%% file: main.tex
\documentclass[pdflatex,sn-nature]{sn-jnl}%

\usepackage{graphicx}%
\usepackage{multirow}%
\usepackage{amsmath,amssymb,amsfonts}%
\usepackage{amsthm}%
\usepackage{mathrsfs}%
\usepackage[title]{appendix}%
\usepackage{xcolor}%
\usepackage{textcomp}%
\usepackage{manyfoot}%
\usepackage{booktabs}%
\usepackage{algorithm}%
\usepackage{algorithmicx}%
\usepackage{algpseudocode}%
\usepackage{listings}%
\usepackage{multicol}% arXiv version: used by the appended Supplementary Information
\usepackage{mathtools}% arXiv version: used by the appended Supplementary Information
\usepackage{enumitem}

\theoremstyle{thmstyleone}%

\theoremstyle{thmstyletwo}%

\theoremstyle{thmstylethree}%

\unnumbered% journal style: no section numbers
\begin{document}

\title{On Improving Graph Neural Networks for QSAR by Pre-training on Extended-Connectivity Fingerprints}

\author[1]{\fnm{Sam} \sur{Money-Kyrle}}
\author[2]{\fnm{Markus} \sur{Dablander}}
\author[3]{\fnm{Thierry} \sur{Hanser}}
\author[3]{\fnm{Stephane} \sur{Werner}}
\author[1]{\fnm{Charlotte M.} \sur{Deane}}
\author*[1]{\fnm{Garrett M.} \sur{Morris}}\email{morris@stats.ox.ac.uk}

\affil[1]{\orgdiv{Department of Statistics}, \orgname{University of Oxford}, \orgaddress{\street{24-29 St Giles'}, \city{Oxford}, \postcode{OX1 3LB}, \country{United Kingdom}}}

\affil[2]{\orgdiv{Mathematical Institute}, \orgname{University of Oxford},  \orgaddress{\street{Andrew Wiles Building, Woodstock Road}, \city{Oxford}, \postcode{OX2 6GG}, \country{United Kingdom}}}

\affil[3]{\orgdiv{Molecular Informatics and AI}, \orgname{Lhasa Limited}, \orgaddress{\street{Granary Wharf House, 2 Canal Wharf}, \city{Leeds}, \postcode{LS11 5PS}, \country{United Kingdom}}}

\abstract{
    Molecular graph neural networks (GNNs) are increasingly common in drug discovery, 
    particularly for Quantitative Structure-Activity Relationship (QSAR) studies; 
    yet their superiority compared to fixed molecular featurisation 
    approaches is disputed. We report a general strategy for improving
    GNNs for QSAR model development by pre-training to predict Extended-Connectivity Fingerprints (ECFP).
    We validate our approach with statistical tests, effect sizes,
    and challenging out-of-distribution (OOD) splits. 
    For five of six Biogen ADMET benchmarks, we observe a statistically significant 
    improvement in performance metrics over evaluated fixed fingerprint baselines when using 
    ECFP pre-trained GNNs. However, we find that pre-trained GNNs outperform randomly initialised GNNs
    on three of the six Biogen tasks.
    We additionally identify that the choice of downstream model substantially 
    changes the relative performance of pre-trained embeddings and fixed fingerprints.
    Importantly, we investigate the impact of increasing molecular similarity to benchmark data during
    pre-training on downstream performance. Our findings show that ECFP-based pre-training can enhance 
    downstream OOD performance on a diverse set of QSAR tasks;
    however, improvements over baselines are contingent on the choice of downstream model architecture.
}

\keywords{Extended-Connectivity Fingerprints, molecular property prediction, graph neural networks, pre-training, QSAR, out-of-distribution generalisation, molecular substructures}

\maketitle

\section{Introduction}\label{sec:intro}

Machine learning (ML) has become a prominent method
in therapeutic drug discovery research~\cite{koohyH-2018-RiseFallMachineLearningMethods}. 
An example of this is Quantitative Structure-Activity Relationship (QSAR) studies, 
where computational models are used to predict molecular properties. 
QSAR studies are useful for profiling and filtering candidate drugs before clinical trials.
In drug discovery, after binding affinity, molecular properties primarily relate to 
Absorption, Distribution, Metabolism, Excretion, and Toxicity (ADMET).
Numerous methods have pushed state-of-the-art performance forward on ADMET benchmarks, e.g.,~\cite{mendez-lucioO-2024-MolE, kearnesS-2016-MolecularGraphConvolutions, yangK-2019-AnalyzingLearnedMolecularRepresentationsProperty, jaegerS-2018-Mol2vec, huW-2020-StrategiesPretrainingGraphNeuralNetworks, houZ-2022-GraphMAE, wangY-2022-MolecularContrastiveLearningRepresentationsGraph, ying2021transformers}.
However, attrition rates in small molecule drug discovery have increased in recent years~\cite{aitkenM-2025-GlobalTrendsRD2025Progress}, 
suggesting benchmark performance does not reflect real-world impact. 
This is partly due to a key limitation in drug discovery:
experimental datasets for training and benchmarking are typically small, noisy, and heterogeneous, 
leading to biased models, limited domain of applicability, and inaccurate performance estimates~\cite{crusiusD-2025-AreWeFittingDataNoise, kretschmerF-2025-CoverageBiasSmallMoleculeMachine, wallachI-2018-MostLigandBasedClassificationBenchmarksReward}.

To overcome the limitations in QSAR, we need robust molecular representations 
that reduce bias and capture relevant task-specific information.
Fixed representations include topological fingerprints, such as Extended-Connectivity Fingerprints (ECFP)~\cite{rogersD-2010-ExtendedconnectivityFingerprints}, 
where substructure or pharmacophore presence is encoded by bits in a binary vector~\cite{mcgibbonM-2024-IntuitionAI}.
Differentiable representations based on deep learning are a more recent approach that
utilises forward- and back-propagation to extract meaningful features from the 
input data in a latent embedding space.
The ability to learn abstract, non-linear patterns from non-tabular data 
should, in principle, make differentiable approaches particularly useful in complex domains like chemistry.

One such differentiable approach is graph-based representation learning, 
which is a natural choice for molecular featurisation; 
molecules can be easily represented as graphs, with atoms as nodes and bonds as edges.
Graph neural networks (GNNs), which operate on molecular graphs and learn task-specific molecular embeddings, 
have been applied to QSAR, showing promising performance as a featurisation method~\cite{kearnesS-2016-MolecularGraphConvolutions, yangK-2019-AnalyzingLearnedMolecularRepresentationsProperty, houZ-2022-GraphMAE, ying2021transformers},
but other results suggest GNN-based methods struggle to outperform fixed representations~\cite{dablanderM-2023-ExploringQSARModelsActivitycliffPrediction, jiangD-2021-CouldGraphNeuralNetworksLearnBetterMolecularRepresentationDrugDiscovery, xiaJ-2023-UnderstandingLimitationsDeepModelsMolecularPropertyPrediction}. 
Self-supervised pre-training has been proposed as a solution for improving GNN performance
and overcoming dataset constraints~\cite{sypetkowskiM-2024-ScalabilityGNNsMolecularGraphs, mendez-lucioO-2024-MolE, zaidiS-2023-PretrainingDenoisingMolecularPropertyPrediction, burnsJ-2026-DescriptorbasedFoundationModelsMolecularProperty}. 
Implemented strategies include pre-training on context prediction and attribute masking~\cite{huW-2020-StrategiesPretrainingGraphNeuralNetworks},
self-supervised pre-training via contrastive learning~\cite{wangY-2022-MolecularContrastiveLearningRepresentationsGraph},
predicting 3D geometry-based objectives alongside molecular fingerprints, including ECFPs~\cite{fangX-2022-GeometryenhancedMolecularRepresentationLearningProperty},
a disentangled transformer approach pre-trained on topological substructure environments~\cite{mendez-lucioO-2024-MolE},
and pre-training on physicochemical descriptor vectors (PDV)~\cite{burnsJ-2026-DescriptorbasedFoundationModelsMolecularProperty}.
Self-supervised pre-training, in combination with downstream training on labelled data via either fine-tuning or another model trained on extracted pre-trained features,
can be viewed as a form of semi-supervised learning~\cite{vanengelenJE-2020-SurveySemisupervisedLearning}.

Although pre-training GNNs for QSAR---both supervised and self-supervised---has been investigated and shown promise, 
there are limitations with how some methods have been assessed.
Firstly, there is a lack of statistical testing against baselines, such as models trained 
from scratch on benchmark data represented via commonly used static fingerprints~\cite{ashJR-2025-PracticallySignificantMethodComparisonProtocols,wognumC-2024-CallIndustryledInitiativeCriticallyAssess}.
Secondly, benchmark comparisons between learned representations pre-trained on different datasets
do not provide a picture of architectural superiority. 
Thirdly, the interpretability of pre-trained models is rarely explored
beyond nearest-neighbour analyses of learned embeddings.
Finally, more often than not, data leakage between pre-training and benchmark datasets is overlooked,
particularly at the substructural level, although Méndez-Lucio et al.\ excluded molecules that were identical
to benchmark molecules from pre-training data~\cite{mendez-lucioO-2024-MolE}.
More recently, van Tilborg et al.\ used a $0.7$ Tanimoto similarity threshold 
between Bemis-Murcko scaffolds to filter pre-training data~\cite{vantilborgD-2026-MolecularDeepLearningEdgeChemical}.

Data leakage is a well-known issue in ML~\cite{kapoorS-2023-LeakageReproducibilityCrisisMachinelearningbasedScience}.
In the strictest sense, data leakage occurs when information that would be unavailable
to the model at inference is inadvertently included during model construction, such as by using test data in the training process~\cite{kapoorS-2023-LeakageReproducibilityCrisisMachinelearningbasedScience}.
Data leakage in this sense is concerning in supervised pre-training,
where benchmark and pre-training datasets can overlap at both task 
and molecule levels due to limited open-source experimental data.
Here, we use the term in a broader sense to describe overlap in chemical space between
pre-training and benchmark datasets.
Such overlap does not invalidate a model for practical use, but it can inflate benchmark
estimates of out-of-distribution (OOD) performance and distort comparisons between pre-training strategies,
as the model may have already encountered similar molecules during pre-training.
In the limit where pre-training data include the benchmark test molecules,
pre-training and downstream benchmarking are effectively transductive
rather than inductive semi-supervised learning~\cite{vanengelenJE-2020-SurveySemisupervisedLearning}.
Generalisation in the form of robust OOD performance is essential 
for identifying novel areas of chemical space with therapeutic potential~\cite{barkerA-2013-ExpandingMedicinalChemistrySpace, klarnerL-2023-DrugDiscoveryCovariateShiftDomainInformed}. 
Therefore, consideration of potential data leakage between 
pre-training and benchmark datasets is essential in QSAR model development.
We investigate the impact of data leakage on benchmark performance by
applying Tanimoto similarity filtering to pre-training data
at various thresholds, down to $\text{Tanimoto} \leq 0.5$.

In addition to data leakage, the choice of benchmark splits is essential for evaluating OOD performance,
as in-distribution splitting can exacerbate benchmark bias~\cite{guoQ-2024-ScaffoldSplitsOverestimateVirtualScreening, klarnerL-2023-DrugDiscoveryCovariateShiftDomainInformed}.
Since exploring novel chemical space is important for developing new drugs,
ML models should be evaluated for robust OOD performance using 
benchmark splits with minimal molecular overlap. 
OOD splits ideally ensure that test molecules are outside the applicability domain (AD)
of the training data, where AD is the region of chemical space where a QSAR model
is expected to make reliable predictions~\cite{royK-2015-SimpleApproachDeterminingApplicabilityDomain}.
As such, OOD evaluation assesses reliability of out-of-domain predictions
and informs whether models can generalise to unseen chemistry.
Commonly used OOD splitting by molecular scaffolds is vulnerable to information leakage and 
performance overestimation~\cite{guoQ-2024-ScaffoldSplitsOverestimateVirtualScreening, klarnerL-2023-DrugDiscoveryCovariateShiftDomainInformed}.
Hierarchical clustering on UMAP projections of molecular fingerprints produces challenging OOD splits;
however, train-test split sizes tend to vary~\cite{guoQ-2025-UMAPbasedClusteringSplitRigorousEvaluation},
which could increase the variance in performance estimates.
On the other hand, Butina clustering on topological fingerprints is a fast, 
distance-based method~\cite{butinaD-1999-UnsupervisedDataBaseClusteringBasedDaylightsFingerprintTanimotoSimilarity}
that produces evenly sized splits with group k-fold cross-validation (CV).
Here we opt for the latter approach to generate many evenly sized OOD splits.

Despite using a consistent splitting strategy, test set performance is variable between splits;
this volatility can obscure whether differences between ML approaches are meaningful.
To validate whether these differences are statistically significant, 
repeated group k-fold CV combined with statistical testing and
effect size estimation is recommended~\cite{ashJR-2025-PracticallySignificantMethodComparisonProtocols}.
However, robust effect size estimation can be challenging in cases where differences are expected to be subtle, 
or for tasks where performance is highly variable (e.g., highly imbalanced classification tasks). 
We therefore use a high number of repetitions ($200\times5$ CV) to generate 
many OOD train-test splits for robust statistical comparison,
averaging metrics over folds within each repeat for stable effect size estimates.
For ablation experiments, we use a subset of $20\times5$ CV repetitions for evaluation 
to reduce computational cost while still providing a comprehensive estimate of performance.

Here, we explore self-supervised pre-training on topological fingerprints as a 
general method for improving GNN predictive performance in QSAR.
We also include an exploratory analysis of substructure importance in Supplementary Information Figs.~51--56,
facilitated by our multi-radius substructure tokenisation approach.
Finally, we investigate the impact of increasing ECFP$4_{2048}$ Tanimoto similarity 
between pre-training and benchmark datasets, finding that increasing
molecular similarity did not consistently improve performance.

\section{Results}\label{sec:results}

Graph Isomorphism Networks (GINs)~\cite{xuK-2019-HowPowerfulAreGraphNeural}
pre-trained to predict topological fingerprints (PT-GIN) show statistically significant 
improvements over fixed fingerprinting methods on challenging OOD splits in five out of six Biogen regression tasks.
Despite these improvements, when compared to randomly initialised GINs, PT-GINs show mixed performance, with some tasks
showing improvements and others showing underperformance.
We also observe relative underperformance on MoleculeNet tasks and 
ChEMBL binding affinity prediction tasks.
However, on the majority of tasks where PT-GIN underperforms,
we observe an inversion of performance rankings when using different downstream models.
Additionally, we find that data leakage, in the form of increasing Tanimoto similarity,
between pre-training and benchmark datasets has an inconsistent impact on benchmark
performance when using our proposed pre-training strategy.

Pre-training to predict hashed ECFPs (ECFP$_{\mathrm{hashed}}$) was 
performed on a subset of the QMugs dataset (see Methods, \textit{\nameref{sec:datasets_pretraining}}).
Benchmarking was carried out on 40 tasks (11 regression and 29 classification) 
from Biogen~\cite{fangC-2023-ProspectiveValidationMachineLearningAlgorithmsAbsorptionDistributionMetabolismExcretionPrediction},
MoleculeNet~\cite{wuZ-2018-MoleculeNet},
and ChEMBL~\cite{zdrazilB-2024-ChEMBLDatabase2023, dablanderM-2023-ExploringQSARModelsActivitycliffPrediction}
(see Methods, \textit{\nameref{sec:datasets_benchmarking}}).
For our baseline featurisation methods, we compared the predictive performance
of PT-GIN against three fixed molecular fingerprinting techniques:
ECFP$_{\mathrm{hashed}}$, 
hashed Functional-Class Fingerprints 
(FCFP$_{\mathrm{hashed}}$)~\cite{rogersD-2010-ExtendedconnectivityFingerprints}, 
and ECFPs vectorised via Sort \& Slice (S\&S)
(ECFP$_{\mathrm{S\&S}}$)~\cite{dablanderM-2024-SortSlice}.
On Biogen, we additionally benchmarked Scratch GINs as a baseline comparison
for learned representations.
All featurisation methods were evaluated with light gradient-boosted machines 
(LightGBM)~\cite{keG-2017-LightGBM}, except Scratch GIN (end-to-end multi-layer perceptron head) (see Methods, \textit{\nameref{sec:scratch_gin_methods}}).
Performance metrics are reported as 5-fold test-set means for each CV repeat; 
statistical significance tests and effect sizes were computed across repeats.
Throughout this work, we use molecular similarity to refer to ECFP$4_{2048}$ Tanimoto similarity.
See~\nameref{sec:methods} for further details.

\subsection{
    ECFP pre-training improves GIN QSAR performance, with gains dependent on task and downstream model
}

\subsubsection{
    ECFP pre-training outperforms fixed fingerprints on five of six Biogen tasks and random initialisation on three
}

\begin{figure}[!htbp]
    \centering
    \textbf{Performance of PT-GIN, fixed fingerprints, and Scratch GIN on Biogen}
    \includegraphics[height=0.8\textwidth, width=\textwidth,keepaspectratio]{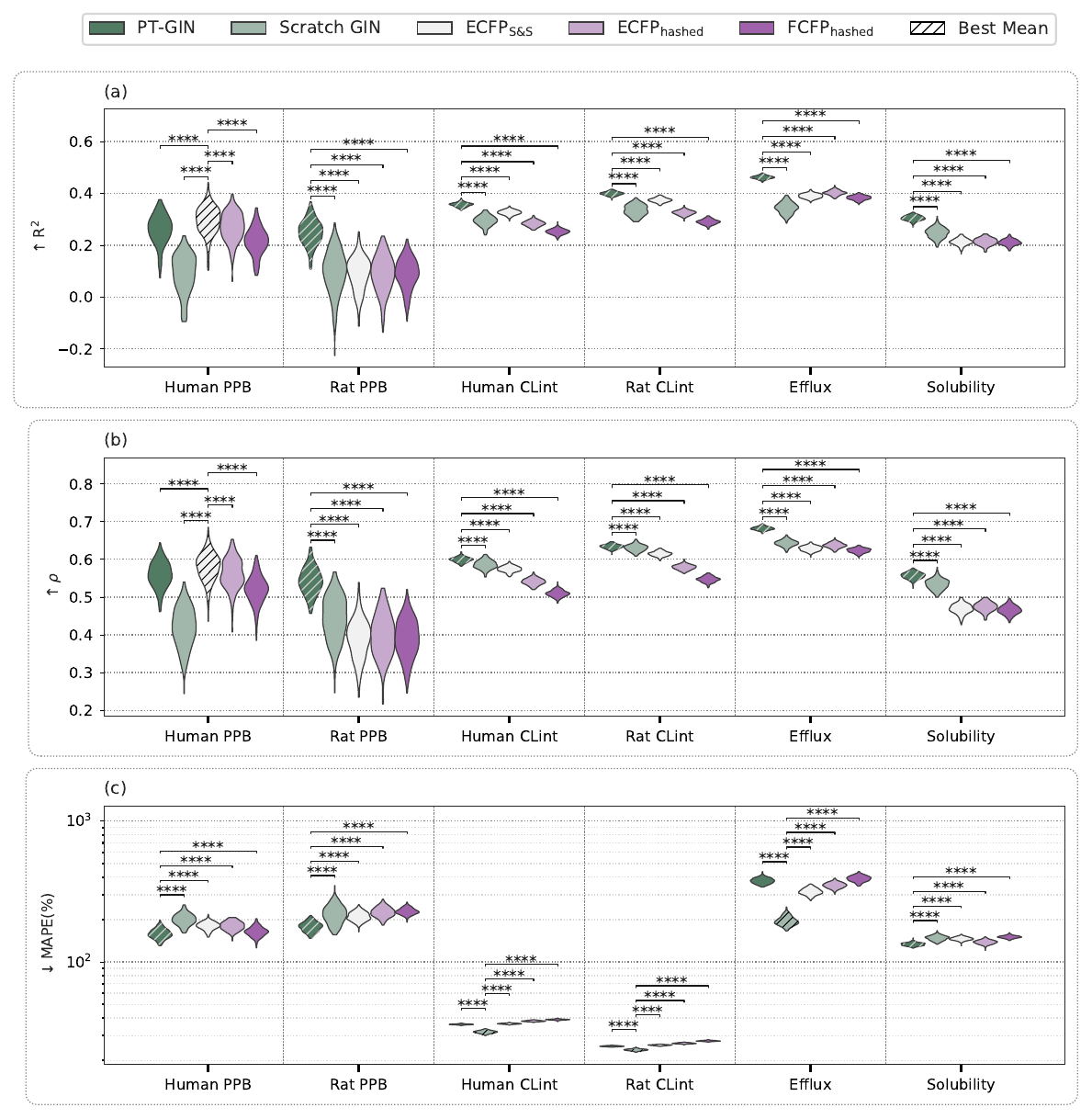}
    \caption{
        Tasks shown are Human Plasma Protein Binding (PPB), Rat PPB, 
        Human Intrinsic Hepatic Clearance (CLint), Rat CLint, Efflux, and Solubility.
        The metrics shown are (a) coefficient of determination ($\uparrow\mathrm{R}^2$), 
        (b) Pearson correlation coefficient ($\uparrow\rho$), and (c) mean absolute percentage error ($\downarrow$MAPE (\%)). 
        Featurisation methods shown are GINs trained from scratch (Scratch GIN), 
        ECFP-pre-trained GINs (PT-GIN),
        hashed Extended-Connectivity Fingerprints (ECFP$_{\mathrm{hashed}}$), 
        hashed Functional-Class Fingerprints (FCFP$_{\mathrm{hashed}}$), 
        and ECFPs vectorised via Sort \& Slice (ECFP$_{\mathrm{S\&S}}$). 
        For each task and metric, the approach with the best mean is hatched.
        Statistical significances, determined by Bonferroni-corrected two-sided Wilcoxon signed-rank tests ($n=199$, $\alpha=0.05$), 
        between the best method by mean and all other approaches are shown with asterisks: 
        ns (not significant) is $p \geq 0.05$;
        * is $0.01 \leq p < 0.05$;
        ** is $0.001 \leq p < 0.01$;
        *** is $0.0001 \leq p < 0.001$;
        and **** is $p < 0.0001$.
        Violins are normalised such that each violin has the same maximum width.
    }
    \label{fig:biogen_violin}
\end{figure}

\begin{figure}[!htbp]
    \centering
    \textbf{Pairwise comparison between PT-GIN and baselines on Biogen}
    \includegraphics[height=0.8\textwidth, width=\textwidth,keepaspectratio]{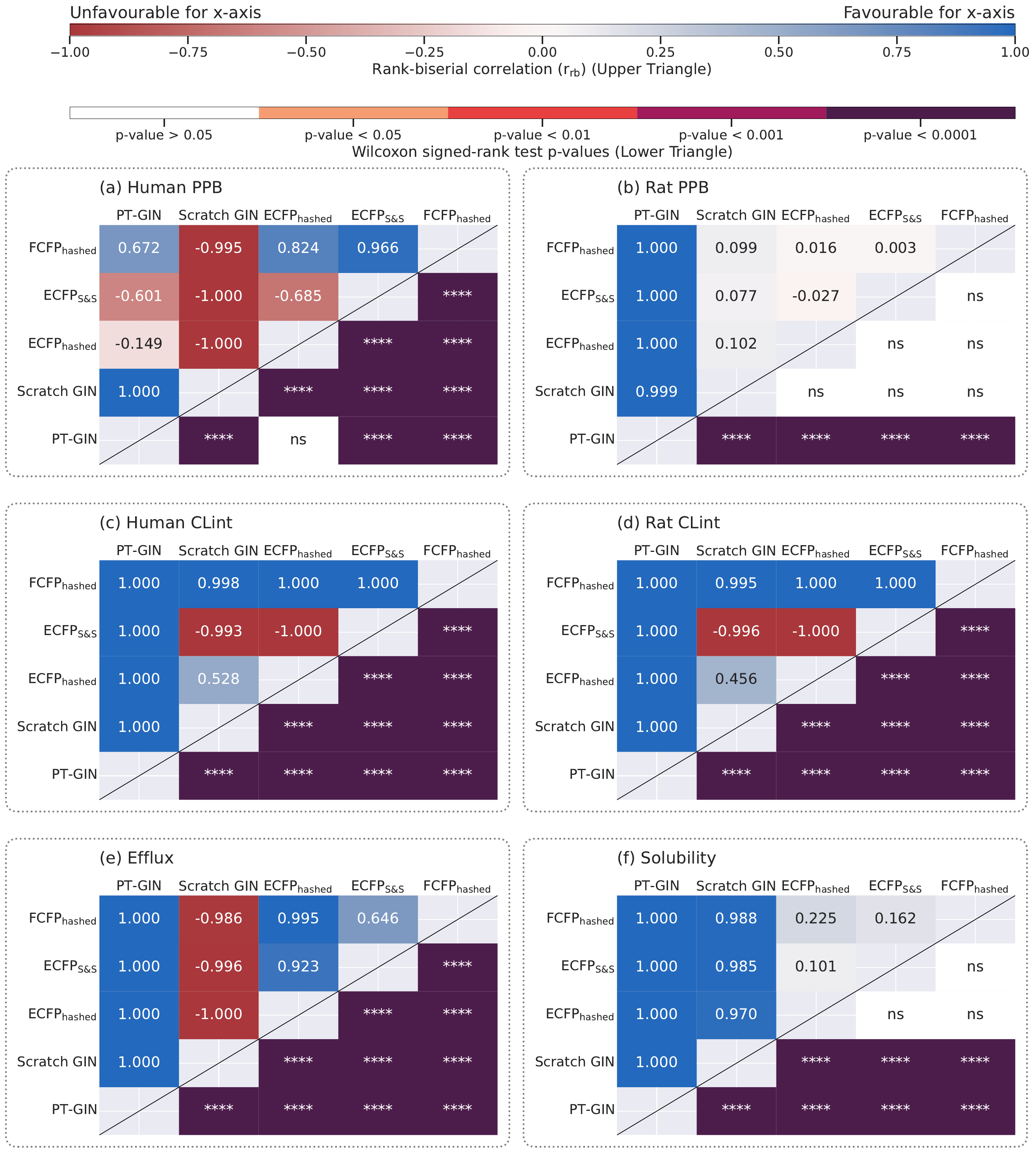}
    \caption{
        Rank-biserial correlation ($r_{rb}$) effect sizes (upper triangle)
        and Bonferroni-corrected two-sided Wilcoxon signed-rank test \textit{p}-values (lower triangle)
        comparison of $\mathrm{R}^2$ between methods on Biogen datasets.
        Tasks are (a) Human Plasma Protein Binding (PPB), (b) Rat PPB, 
        (c) Human Intrinsic Hepatic Clearance (CLint), (d) Rat CLint, (e) Efflux, (f) Solubility. 
        Featurisation methods shown are GINs trained from scratch (Scratch GIN), 
        ECFP-pre-trained GINs (PT-GIN), hashed Extended-Connectivity Fingerprints (ECFP$_{\mathrm{hashed}}$), 
        hashed Functional-Class Fingerprints (FCFP$_{\mathrm{hashed}}$), and 
        ECFPs vectorised via Sort \& Slice (ECFP$_{\mathrm{S\&S}}$).
        For $r_{rb}$ effect sizes (upper triangle): blue indicates the method on the \textit{x}-axis has a higher $\mathrm{R}^2$ than 
        the method on the \textit{y}-axis; red indicates the opposite. 
        For Wilcoxon signed-rank test \textit{p}-values (lower triangle): 
        white indicates no significant difference and darker indicates a smaller \textit{p}-value ($n=199$, $\alpha=0.05$).
    }
    \label{fig:biogen_heatmap}
\end{figure}

Biogen is a homogeneous dataset of drug-like molecules with six diverse QSAR 
regression tasks. A model with greater correlation and lower error on OOD 
splits for these tasks is more likely to generalise well to novel chemical space.
On five of six Biogen tasks, PT-GIN achieves 
the highest mean coefficient of determination ($\mathrm{R}^2$), with statistically significant 
improvements over baseline approaches 
    (Fig.~\ref{fig:biogen_violin}).
On these five tasks, PT-GIN has the highest mean Pearson correlation coefficient ($\rho$) 
    (Fig.~\ref{fig:biogen_violin}). Moreover, 
the rank-biserial correlation ($r_{rb}$) effect sizes for $\mathrm{R}^2$ between PT-GIN and the
baseline methods are consistently $1.00$, indicating that PT-GIN outperforms 
baselines on every CV repeat for most tasks
    (Fig.~\ref{fig:biogen_heatmap}).
These observations suggest that self-supervised pre-training on topological fingerprints 
can provide a predictive advantage over the baseline methods.
However, the exception is on the Human Plasma Protein Binding (PPB) task, 
where ECFP$_{\mathrm{S\&S}}$ performs best by $\mathrm{R}^2$ and $\rho$. 
On Human PPB there is no significant difference in $\mathrm{R}^2$ between PT-GIN and ECFP$_{\mathrm{hashed}}$, and
PT-GIN still has significantly greater $\mathrm{R}^2$ than FCFP$_{\mathrm{hashed}}$ and Scratch GIN, 
with an $r_{rb}$ of $1.00$ between PT-GIN and Scratch GIN
    (Fig.~\ref{fig:biogen_heatmap}).

Scratch GIN models achieve the lowest mean absolute percentage error (MAPE) 
on both intrinsic hepatic clearance (CLint) tasks and the Efflux task, 
with a significant difference from all other approaches 
    (Fig.~\ref{fig:biogen_violin}), 
despite significantly lower $\mathrm{R}^2$ than PT-GIN on these three tasks.
In addition, PT-GIN has the highest $\rho$ on the same three tasks, with a significant difference 
from Scratch GIN on Human CLint and Efflux.
However, by mean absolute error (MAE), PT-GIN outperforms Scratch GIN on Rat CLint and Efflux,
and the difference between the two on Human CLint is reduced
    (Supplementary Information Fig.~26).
The label distributions for these three tasks are right-skewed,
with most compounds having low-magnitude values
    (Supplementary Information Fig.~80).
As MAPE scales each error by the inverse of the true value,
it is dominated by these low-magnitude compounds.
Lower MAPE for Scratch GIN therefore likely reflects smaller errors on the most common, low-magnitude values.
Combined with lower $\mathrm{R}^2$ and $\rho$, this suggests that Scratch GIN is less accurate on
higher-magnitude compounds, whereas PT-GIN better captures variation across the full range of target values, 
consistent with stronger generalisation to these OOD test sets.

\begin{figure}[!htbp]
    \centering
    \textbf{Performance of PT-GIN and randomly initialised GINs on Biogen tasks}
    \includegraphics[height=0.8\textwidth, width=\textwidth,keepaspectratio]{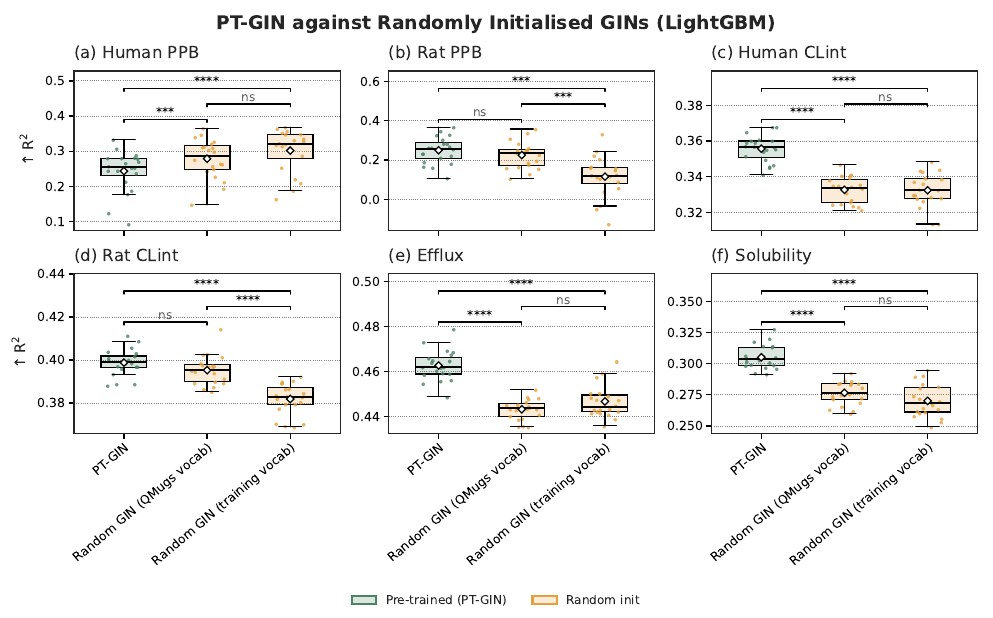}
    \caption{
        Coefficient of determination ($\uparrow\mathrm{R}^2$)
        for PT-GIN (green) and randomly initialised GINs (orange) on Biogen tasks. 
        Tasks are (a) Human Plasma Protein Binding (PPB), (b) Rat PPB, 
        (c) Human Intrinsic Hepatic Clearance (CLint), (d) Rat CLint, (e) Efflux, (f) Solubility. 
        Featurisation methods shown are ECFP-pre-trained GINs (PT-GIN),
        and randomly initialised GINs (Random GIN) with the same architecture as PT-GIN;
        results for the Random GIN using vocabularies derived from QMugs
        and the training split are both shown.
        Statistical significances, determined by Bonferroni-corrected
        two-sided Wilcoxon signed-rank tests ($n=20$, $\alpha=0.05$),
        between all method pairs are shown with asterisks: 
        ns (not significant) is $p \geq 0.05$;
        * is $0.01 \leq p < 0.05$;
        ** is $0.001 \leq p < 0.01$;
        *** is $0.0001 \leq p < 0.001$;
        and **** is $p < 0.0001$.
    }
    \label{fig:random_gin}
\end{figure}

We additionally performed an ablation, comparing PT-GIN to randomly initialised GINs on Biogen,
using a subset of $20\times5$ CV repeats for evaluation.
 (see Methods, \textit{\nameref{sec:random_init_gins}}).
We implemented two versions of Random GINs, one with the same vocabulary as the PT-GIN based on
the QMugs dataset (Random GIN$_{\text{QMugs}}$), and one with vocabularies based on
benchmark training sets (Random GIN$_{\text{Training}}$).
On three of the benchmark tasks, Human CLint, Efflux, and Solubility,
PT-GINs outperformed Random GINs with the same architecture by coefficient of determination 
    ($\mathrm{R}^2$) (Fig.~\ref{fig:random_gin}).
On the Rat CLint and Rat PPB tasks, differences in performance between PT-GINs and Random GIN$_{\text{QMugs}}$
were marginal by $\mathrm{R}^2$, while Random GIN$_{\text{Training}}$ significantly underperformed,
suggesting that the QMugs vocabulary of substructures, rather than pre-training,
is important for performance on these tasks.
For Human PPB, Random GINs were more competitive, with PT-GIN underperforming against 
both Random GINs by $\mathrm{R}^2$, indicating that pre-training a GIN to
predict ECFP$4_{2048}$ fingerprints can be detrimental to performance on certain benchmark tasks.

\subsubsection{
    Relative performance varies across MoleculeNet and ChEMBL tasks
}

\begin{figure}[!htbp]
    \centering
    \textbf{Performance of PT-GIN and fixed fingerprints on MoleculeNet tasks}
    \includegraphics[height=0.8\textwidth, width=\textwidth,keepaspectratio]{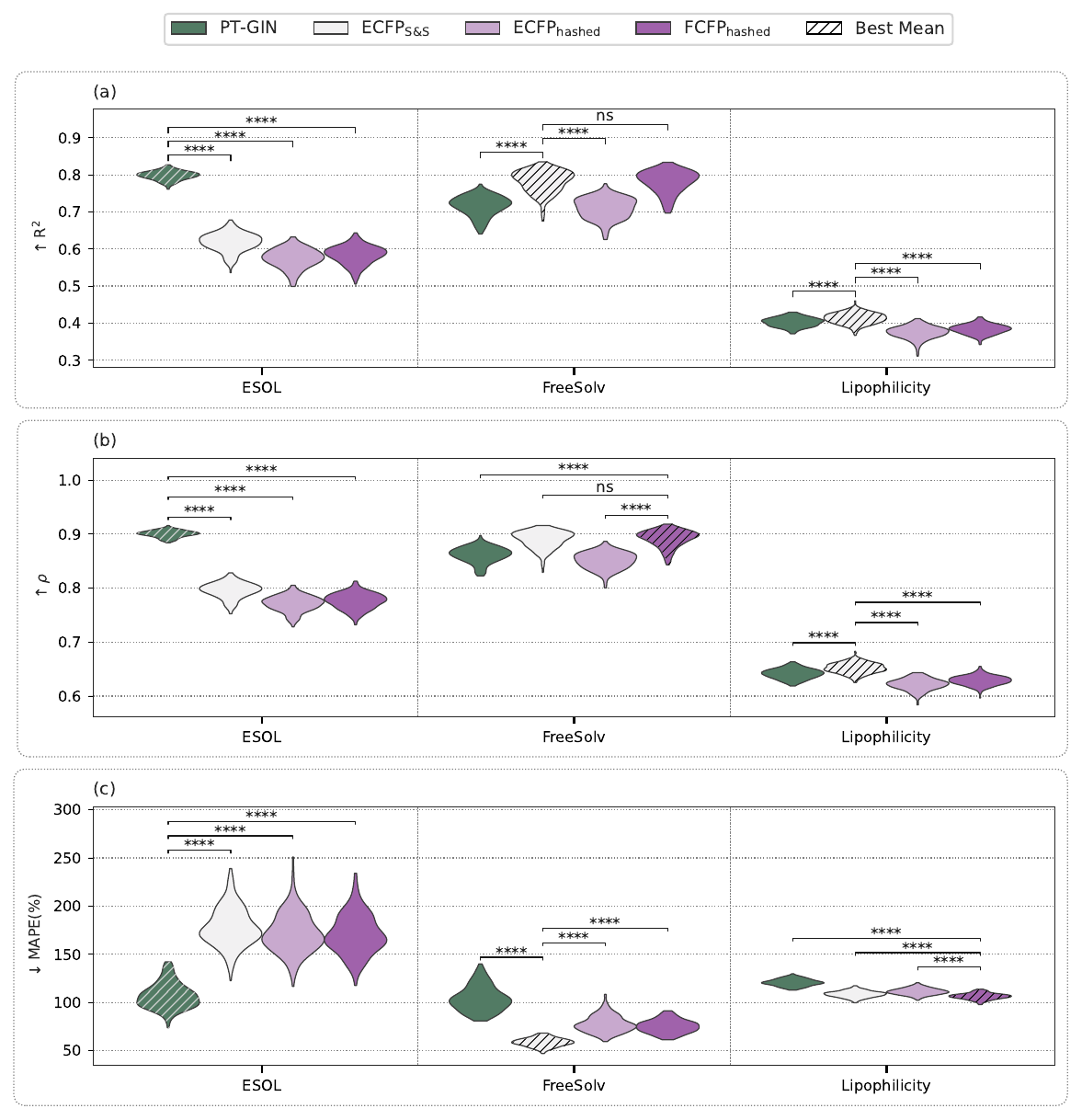}
    \caption{
        Datasets shown are ESOL, FreeSolv, and Lipophilicity.
        The metrics shown are (a) coefficient of determination ($\uparrow\mathrm{R}^2$), 
        (b) Pearson correlation coefficient ($\uparrow\rho$), and (c) mean absolute percentage error ($\downarrow\mathrm{MAPE}$) (\%). 
        Featurisation methods shown are ECFP-pre-trained GINs (PT-GIN),
        hashed Extended-Connectivity Fingerprints (ECFP$_{\mathrm{hashed}}$), 
        hashed Functional-Class Fingerprints (FCFP$_{\mathrm{hashed}}$), 
        and ECFPs vectorised via Sort \& Slice (ECFP$_{\mathrm{S\&S}}$). 
        For each task and metric, the approach with the best mean is hatched.
        Statistical significances, determined by Bonferroni-corrected two-sided Wilcoxon signed-rank tests ($n=199$, $\alpha=0.05$),
        between the best method by mean and all other approaches are shown with asterisks:
        ns (not significant) is $p \geq 0.05$;
        * is $0.01 \leq p < 0.05$;
        ** is $0.001 \leq p < 0.01$;
        *** is $0.0001 \leq p < 0.001$;
        and **** is $p < 0.0001$.
        Violins are normalised such that each violin has the same maximum width.
    }
    \label{fig:molnet_metrics}
\end{figure}

On the three MoleculeNet regression tasks and target-specific ChEMBL affinity prediction,
the performance gains from pre-training on hashed ECFPs are not as clear.
PT-GIN only attains the highest mean $\mathrm{R}^2$ on ESOL, underperforming compared to ECFP$_{\mathrm{S\&S}}$ on
FreeSolv and Lipophilicity
    (Fig.~\ref{fig:molnet_metrics}).
Additionally, PT-GIN has a significantly lower $\mathrm{R}^2$ on FreeSolv compared to 
FCFP$_{\mathrm{hashed}}$
(Supplementary Information Fig.~13).    
PT-GIN attains a significantly greater $\mathrm{R}^2$ than ECFP$_{\mathrm{hashed}}$ on all three tasks,
suggesting pre-training on hashed ECFPs is beneficial over the fixed equivalent; 
however, the effect size of $r_{rb}=0.301$ between PT-GIN and ECFP$_{\mathrm{hashed}}$ 
on FreeSolv suggests that performance gains are not ubiquitous
(Supplementary Information Fig.~13).
The $\mathrm{R}^2$ and $\rho$ scores of all methods decline on ESOL when only considering test set molecules 
with a drug-like solubility, \mbox{$0 < \mathrm{log_{10}(S (\mu M))} < 3$}.
$\mathrm{R}^2$ is negative across all featurisation approaches despite positive $\rho$ values, indicating 
a systematic failure of all methods to predict drug-like solubilities, such as offset predictions
(Supplementary Information Fig.~1).
Similarly, for target-specific binding affinity with LightGBM, PT-GIN underperforms compared to baselines
(Supplementary Information Fig.~2).
PT-GIN has a significantly lower $\mathrm{R}^2$ score than all fixed approaches for DRD2 and Factor Xa.

The MUV dataset presents a challenging task: target-specific virtual screening (VS).
On MUV tasks, models using PT-GIN and baseline methods for featurisation
often predict only negatives, 
leading to a high proportion of Matthews correlation coefficient (MCC)
scores being $\mathrm{MCC} = 0$
(Supplementary Information Figs.~3 and~72). 
This highlights the challenge of VS in an OOD setting,
where 2D ligand topology is likely insufficient for accurately predicting 
3D protein-ligand interactions, and class imbalance is expected.

For VS, models can be useful for ranking ligands
even with low recall at a $0.5$ threshold.
Enrichment factor ($\mathrm{EF}$) is a metric for determining the quality of ligand ranking by a VS model.
$\mathrm{EF}_{x\%}$ measures the ratio of the fraction of active compounds in the
top $x\%$ ranked compared to the fraction of active compounds in the entire test set (see Methods, \textit{\nameref{sec:metrics}}).
At all thresholds, there is no clear advantage for any particular method across the different targets;
for example, PT-GIN achieves the highest mean $\mathrm{EF}_{10\%}$ on only $24\%$ of targets
(Supplementary Information Fig.~4).    
This inconsistency is corroborated by area under the precision-recall curve (AUCPR)
scores, with no method achieving the greatest AUCPR on the majority of targets
(Supplementary Information Fig.~3).    
Friedman $\chi^2$ tests~\cite{demsarJ-2006-StatisticalComparisonsClassifiersMultipleData}
for each metric suggesting that there is insufficient evidence to separate the methods 
across MUV tasks, with no significant differences across all metrics (Supplementary Information Table~1).

As all four approaches are based on 2D molecular topology, 
could not be statistically distinguished across VS tasks,
and all struggle to meaningfully separate active and inactive compounds, we hypothesise 
that 3D information could improve the performance of VS models;
however, this requires further investigation to be validated.

Improvements resulting from our pre-training strategy are clearer for toxicity classification.
In this case, PT-GIN exhibits the highest AUCPR with a significant difference 
from all baseline approaches on $7$ out of $12$ tasks
(Supplementary Information Fig.~5).    
Across all Tox21 tasks, the effect sizes ($r_{rb}$) for AUCPR between PT-GIN and baseline methods exceed
$0.5$ in favour of PT-GIN for $72\%$ of paired comparisons, indicating a strong 
and frequently observed improvement from pre-training over baseline methods
(Supplementary Information Fig.~24).
Low MCC scores occur for all approaches, with $\mathrm{mean(MCC)} = 0$ on several tasks 
and $\mathrm{mean(MCC)} < 0.2$ for $60\%$ of scores
(Supplementary Information Fig.~5),
suggesting poor performance at a $0.5$ classification threshold,
although this is not as prevalent as on the MUV dataset.

Looking at whether PT-GIN consistently outranks other methods on Tox21 tasks,
we performed Friedman $\chi^2$~\cite{demsarJ-2006-StatisticalComparisonsClassifiersMultipleData}
and Nemenyi post-hoc tests~\cite{nemenyiPB-1963-DistributionFreeMultipleComparisons} on the rankings of each method across Tox21 tasks.
Friedman $\chi^2$ tests show a significant difference between method rankings across
Tox21 tasks for area under the receiver operating characteristic curve (AUROC) and AUCPR, but not for MCC
(Supplementary Information Table~2).
Nemenyi post-hoc tests show that PT-GIN significantly outranks ECFP$_{\mathrm{hashed}}$
by AUROC, and outranks FCFP$_{\mathrm{hashed}}$ by AUROC and AUCPR (Supplementary Information Table~3).

\subsubsection{
    Benefits of ECFP pre-training depend on the downstream model
}

Our ablation experiment comparing the performance of LightGBM against Kernel Ridge regression (KR)
with $20\times5$ CV repeats (see Methods, \textit{\nameref{sec:methods_downstream_models}})
suggests that the discrepancy in performance between Biogen and other
tasks is partly attributable to the choice of downstream model architecture.

\begin{figure}[!htbp]
    \centering
    \textbf{Downstream model performance on MoleculeNet and ChEMBL tasks}
    \includegraphics[width=\textwidth, height=0.8\textheight, keepaspectratio]{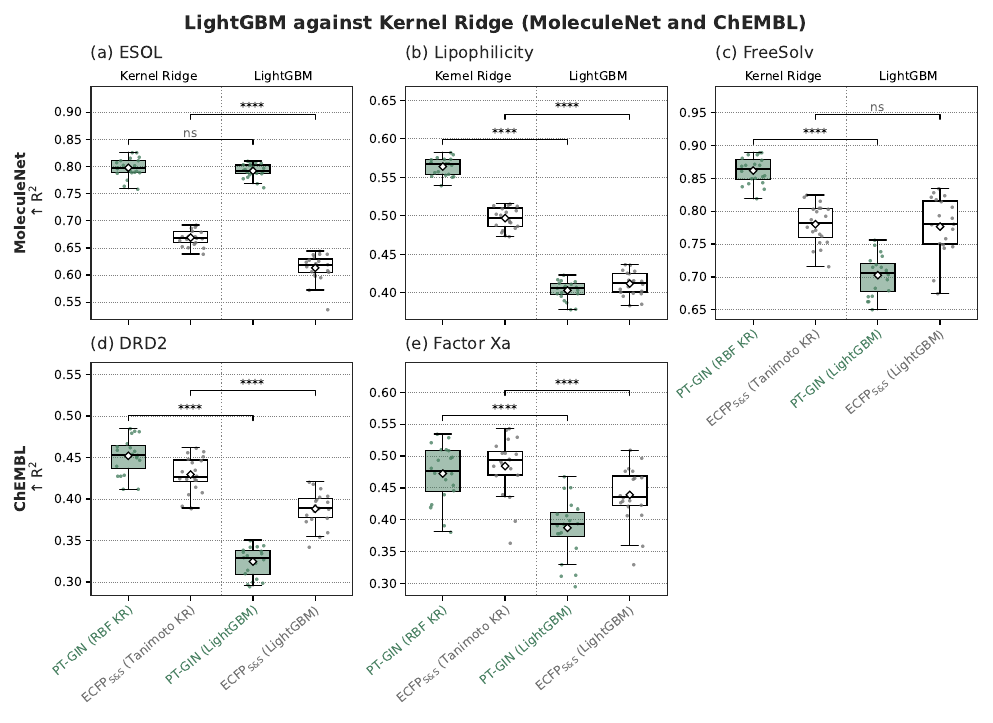}
    \caption{
        Comparison of downstream model architectures on regression tasks
        with coefficient of determination ($\uparrow\mathrm{R}^2$).
        The top row tasks are MoleculeNet regression tasks and
        the bottom row tasks are ChEMBL affinity regression tasks:
        (a) ESOL, (b) Lipophilicity, (c) FreeSolv, (d) DRD2, and (e) Factor Xa.
        Downstream models are Kernel Ridge regression (KR), shown on the left of each subplot,
        and LightGBM, shown on the right.
        The input representations are PT-GIN (green) and ECFP$_{\mathrm{S\&S}}$ (grey).
        For KR, PT-GIN was used with a radial basis function (RBF) kernel and ECFP$_{\mathrm{S\&S}}$ was used
        with a Tanimoto kernel.
        Metrics are calculated over $20$ repeats of $5$-fold cross-validation;
        each point represents the average performance across
        a single 5-fold CV repeat.
        Diamonds indicate the mean performance; the horizontal line
        within each box represents the median performance.
        Significances between downstream models
        are calculated using a two-sided Wilcoxon signed-rank test ($n=20$, $\alpha=0.05$)
        with Bonferroni correction for multiple comparisons:
        ns (not significant) is $p \geq 0.05$;
        * is $0.01 \leq p < 0.05$;
        ** is $0.001 \leq p < 0.01$;
        *** is $0.0001 \leq p < 0.001$;
        and **** is $p < 0.0001$.
    }\label{fig:downstream_model_ablation_regression}
\end{figure}

On the Lipophilicity and FreeSolv tasks, where PT-GIN
underperforms compared to ECFP$_{\mathrm{S\&S}}$ with LightGBM,
PT-GIN with KR significantly outperforms ECFP$_{\mathrm{S\&S}}$
by $\mathrm{R}^2$
(Fig.~\ref{fig:downstream_model_ablation_regression} and Supplementary Information Table~4).
On the ChEMBL binding affinity tasks, PT-GIN with KR significantly outperforms ECFP$_{\mathrm{S\&S}}$
by $\mathrm{R}^2$ on DRD2, with no significant difference on Factor Xa
(Fig.~\ref{fig:downstream_model_ablation_regression} and Supplementary Information Table~4).
Going further, on all regression tasks where PT-GIN underperforms ECFP$_{\mathrm{S\&S}}$ by $\mathrm{R}^2$, $\rho$, and $\mathrm{MAE}$
with LightGBM, PT-GIN either significantly outperforms ECFP$_{\mathrm{S\&S}}$ or, in the case of Factor Xa,
is inseparable from ECFP$_{\mathrm{S\&S}}$ with KR (Supplementary Information Table~4).

Comparing LightGBM and KR models, 
PT-GIN with KR significantly outperforms PT-GIN with LightGBM 
by $\mathrm{R}^2$ on all regression tasks
except for Solubility and ESOL, where there is no significant difference
(Fig.~\ref{fig:downstream_model_ablation_regression} and Supplementary Information Figs.~43--45).
For ECFP$_{\mathrm{S\&S}}$, KR is significantly superior to LightGBM on the majority of regression tasks 
by $\mathrm{R}^2$ ($91\%$), $\rho$ ($91\%$), and $\mathrm{MAE}$ ($73\%$)
(Fig.~\ref{fig:downstream_model_ablation_regression}, Supplementary Information Figs.~43--45 and Table~4).
MAPE is more variable across tasks, likely due to upweighting of errors on low-magnitude targets;
for example, for either representation
KR models are significantly better than LightGBM by all other metrics on Rat PPB, 
but are not significantly different by MAPE (Supplementary Information Fig.~43).

For classification tasks, KR ablation $\mathrm{AUCPR}$ results also suggest that using
KR improves the performance of PT-GIN.
On MUV, PT-GIN with KR significantly outperforms PT-GIN with LightGBM by $\mathrm{AUCPR}$
on 13 out of 17 tasks, compared to only 1 significant result in favour of LightGBM
(Supplementary Information Table~4).
However, $\mathrm{AUROC}$ on MUV
suggests mixed performance across tasks;
PT-GIN with KR achieves significantly better performance than
PT-GIN with LightGBM on only 6 out of 17 tasks, and 
LightGBM significantly outperforms KR on 7 of the tasks.
Compared to ECFP$_{\mathrm{S\&S}}$, the number of significant results in favour of PT-GIN
increases from 3 with LightGBM to 7 with KR by $\mathrm{AUCPR}$.
For Tox21, PT-GIN with KR achieves significantly better performance than
PT-GIN with LightGBM by $\mathrm{AUCPR}$
on 9 out of 12 tasks, compared to no significant results in favour of LightGBM.
By $\mathrm{AUROC}$, PT-GIN with KR significantly outperforms PT-GIN with LightGBM on 7 out of 12 tasks;
3 results are significantly in favour of LightGBM.
For Tox21, the number of significant results in favour of PT-GIN
compared to ECFP$_{\mathrm{S\&S}}$ increases from 9 with LightGBM to 10 with KR by $\mathrm{AUCPR}$.

Overall, these findings suggest that while pre-training on ECFPs can enhance model performance
in an out-of-distribution setting, improvements are often dependent on the data, task, and downstream model.
Pre-training on ECFPs can lead to more robust GNN models,
highlighted by consistently favourable effect sizes for PT-GIN
on five out of six Biogen tasks with LightGBM.
However, this performance benefit varies between tasks with LightGBM;
PT-GIN struggles with LightGBM on FreeSolv, Lipophilicity, 
and ChEMBL binding affinity tasks.
Switching to KR from LightGBM enables 
PT-GIN to outperform ECFP$_{\mathrm{S\&S}}$
on 3 out of 4 regression tasks where PT-GIN underperforms with LightGBM,
and move from significantly worse to not significantly different on the last task.
Conventional tree-based approaches, such as LightGBM, split features based on
axis-parallel decision boundaries; this makes oblique decisions more challenging
for these models to learn~\cite{blaserR-2016-RandomRotationEnsembles}.
As such, we hypothesise that on certain tasks LightGBM may struggle to approximate
oblique decision boundaries in pre-trained embeddings that are relevant to the target label,
whereas KR is able to capture these relationships.

\subsection{On data leakage effects in self-supervised pre-training for QSAR}

As data leakage is a known issue in ML for drug discovery~\cite{bernettJ-2024-GuidingQuestionsAvoidDataLeakage},
we investigated the potential effects of pre-training on molecules similar to those in the benchmark data. 
We generated six QMugs pre-training datasets with $462,189$ molecules each, using
varying Tanimoto similarity thresholds to exclude molecules similar to the benchmark data,
the most stringent being a $0.5$ threshold and the most relaxed being a $1.0$ threshold
    (see Methods, \textit{\nameref{sec:datasets_pretraining}} and \textit{\nameref{sec:methods_splitting_filtering}}).
PT-GIN models were pre-trained on the filtered datasets and benchmarked on Biogen.

\begin{figure}[!htbp]
    \centering
    \textbf{PT-GIN performance with varying Tanimoto similarity filter thresholds}
    \includegraphics[height=0.8\textwidth, width=\textwidth,keepaspectratio]{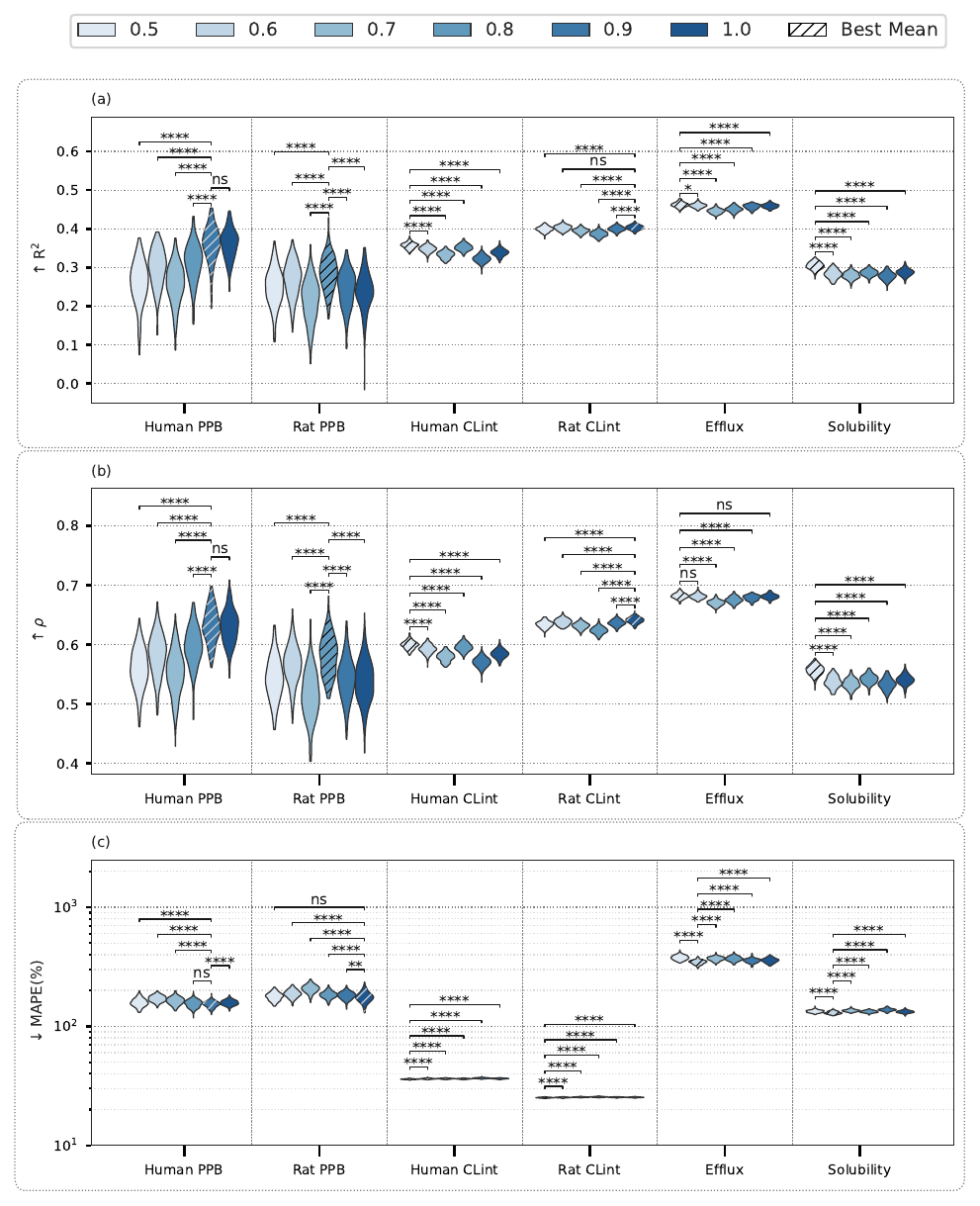}
    \caption{
        Tasks are (a) Human Plasma Protein Binding (PPB), (b) Rat PPB, 
        (c) Human Intrinsic Hepatic Clearance (CLint), (d) Rat CLint, (e) Efflux, (f) Solubility.
        Dissimilarity is greatest between the pre-training and benchmark molecules at the $0.5$ threshold;
        at $1.0$, the pre-training data includes molecules that appear in the benchmark data.
        The metrics shown are coefficient of determination
        ($\uparrow\mathrm{R}^2$), Pearson correlation coefficient ($\uparrow\rho$), and mean absolute percentage error ($\downarrow\mathrm{MAPE}$) (\%).
        For each task and metric, the approach with the best mean is hatched.
        Statistical significances, determined by Bonferroni-corrected two-sided Wilcoxon signed-rank tests ($n=199$, $\alpha=0.05$), 
        between the best method by mean and all other approaches are shown with asterisks: 
        ns (not significant) is $p \geq 0.05$;
        * is $0.01 \leq p < 0.05$;
        ** is $0.001 \leq p < 0.01$;
        *** is $0.0001 \leq p < 0.001$;
        and **** is $p < 0.0001$.
        Violins are normalised such that each violin has the same maximum width. 
    }
    \label{fig:sim_violin}
\end{figure}

\begin{figure}[!htbp]
    \centering
    \textbf{Pairwise comparison between Tanimoto similarity-filtered PT-GINs}
    \includegraphics[height=0.8\textwidth, width=\textwidth,keepaspectratio]{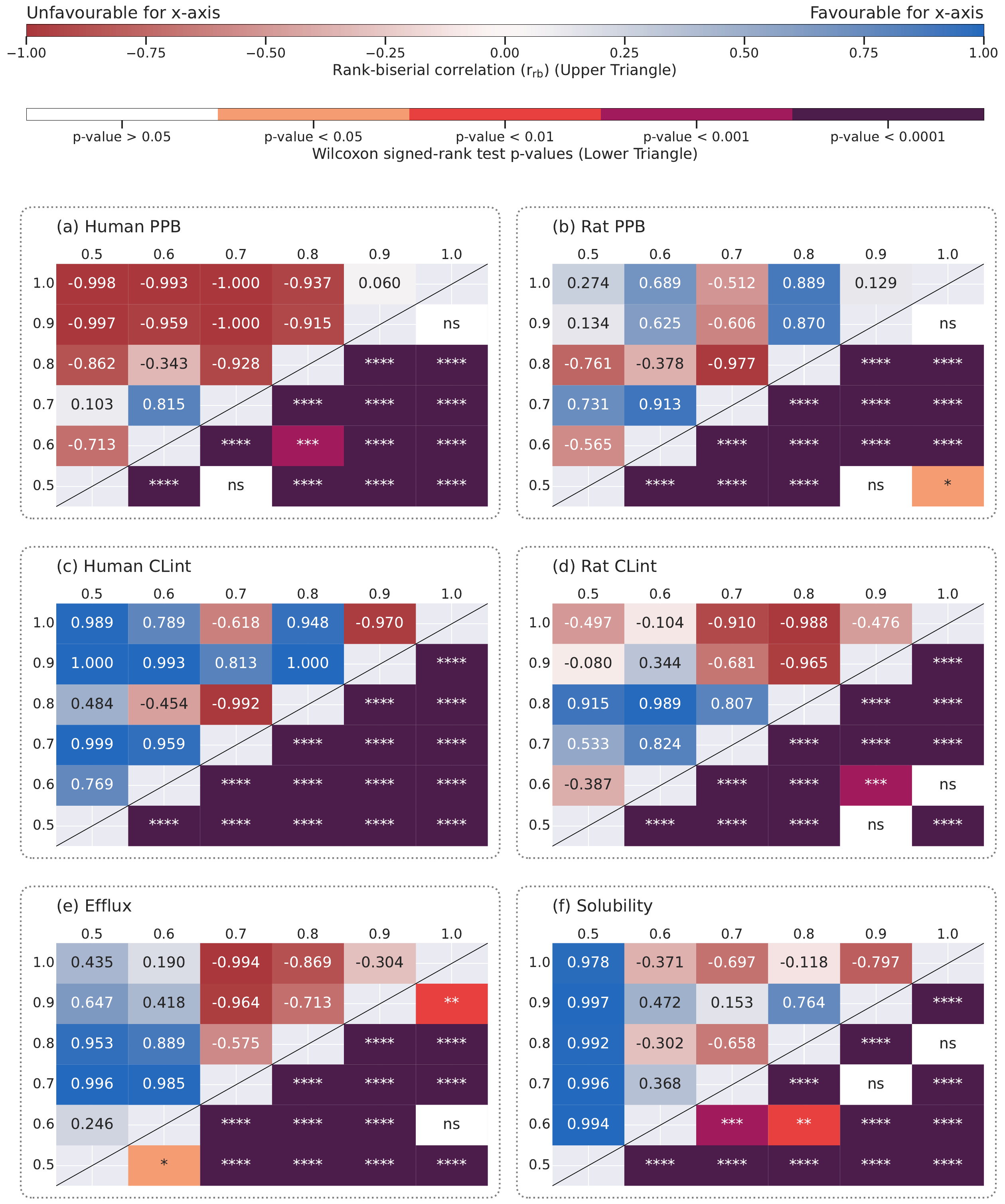}
    \caption{
        Rank-biserial correlation ($r_{rb}$) effect sizes (upper triangle)
        and Bonferroni-corrected two-sided Wilcoxon signed-rank test \textit{p}-values (lower triangle)
        comparison of $\mathrm{R}^2$ between PT-GIN models pre-trained on
        Tanimoto similarity-filtered QMugs data with varying thresholds.
        Tasks are (a) Human Plasma Protein Binding (PPB), (b) Rat PPB, 
        (c) Human Intrinsic Hepatic Clearance (CLint), (d) Rat CLint, (e) Efflux, (f) Solubility.
        Methods shown are PT-GIN models pre-trained on filtered QMugs data at varying similarity filter thresholds;
        $0.5$ indicates that no molecule with Tanimoto similarity $>0.5$ to 
        any molecule in the benchmark data was included in the pre-training data. 
        For $r_{rb}$ effect sizes (upper triangle): blue indicates the method on the \textit{x}-axis has a higher $\mathrm{R}^2$ than 
        the method on the \textit{y}-axis; red indicates the opposite. 
        For Wilcoxon signed-rank test \textit{p}-values (lower triangle): 
        white indicates no significant difference and darker indicates a smaller \textit{p}-value ($n=199$, $\alpha=0.05$).
        See Methods, \textit{\nameref{sec:methods_splitting_filtering}}, for details.
    }
    \label{fig:tanimoto_heatmap}
\end{figure}

On five out of six tasks, there is no clear improvement in performance
as data leakage increases.
At the greatest dissimilarity, $0.5$, PT-GIN exhibits the highest mean $\mathrm{R}^2$ 
on three tasks, Human CLint, Efflux, and Solubility
    (Fig.~\ref{fig:sim_violin}). 
The PT-GIN embeddings that achieve both the highest correlation ($\rho$) and lowest error (MAPE) are the same only for two tasks, Human PPB and Human CLint
    (Fig.~\ref{fig:sim_violin}).
For Human CLint, the general trend indicates a decrease in $\mathrm{R}^2$ 
with increased similarity between pre-training and benchmark data
    (Fig.~\ref{fig:tanimoto_heatmap}). 
In contrast, for Human PPB, increasing data leakage is associated with higher performance, 
with models pre-trained at higher similarity thresholds ($0.8$--$1.0$) consistently outperforming
those pre-trained at lower thresholds ($0.5$--$0.7$) across all three metrics (Fig.~\ref{fig:sim_violin}).
The smallest benchmarks exhibit a far greater maximum percentage difference in $\mathrm{R}^2$ between the best and worst performing models,
$41\%$ for Human PPB and $34\%$ for Rat PPB, compared to tasks with more data; 
the next highest is Human CLint at $11\%$.

The inconsistent, at times negative, effects of increased molecular similarity on 
Biogen ADMET performance scores are counterintuitive, as data leakage is typically associated with overly optimistic model performance~\cite{bernettJ-2024-GuidingQuestionsAvoidDataLeakage}.
The absence of increased performance with increasing similarity suggests 
that the impact of data leakage from overall molecular similarity
is small relative to other differences between pre-trained models,
such as altered loss-optimisation landscapes from pre-training on different molecular structures.
Given that the pre-training objective is to predict bits in a topological fingerprint,
the presence and absence of specific substructures in the pre-training data may
be more important than overall molecular similarity.

\begin{figure}[!htbp]
    \centering
    \textbf{QMugs substructure content and coverage of Biogen substructures}
    \includegraphics[width=\textwidth,keepaspectratio]{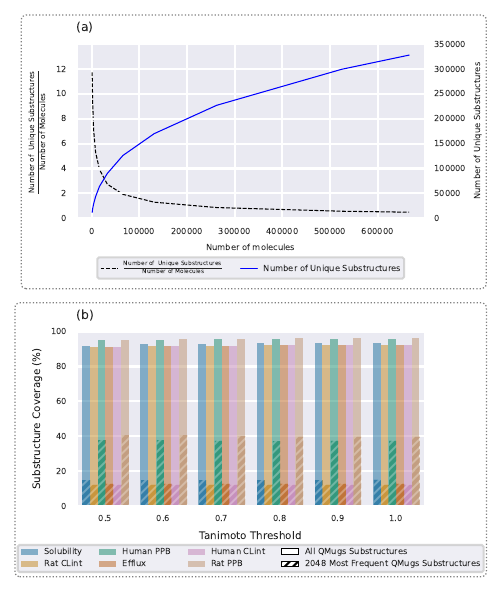}
    \caption{
        (a) The number of molecules in randomly selected subsets of QMugs and the full dataset ($665$k molecules) (\textit{x}-axis)
        versus the ratio of the number of unique substructures to the number of molecules 
        (black, dashed, left \textit{y}-axis) and the total number of unique substructures (blue, right \textit{y}-axis).
        As the number of molecules increases, the ratio of the number of unique substructures to 
        the number of molecules decreases.
        (b) The proportion of substructures in the Biogen subsets that appear in the Tanimoto-filtered QMugs
        pre-training subsets. Substructure coverage is expressed as a percentage. Coverage using the entirety
        of QMugs (no hatching) and only the $2048$ most frequent substructures in the QMugs subset (hatched)
        are shown. The coverage of Biogen substructures does not substantially change with increased data leakage.
        Substructures were determined using RDKit with a radius $r = 2$, chirality set to True, and Daylight atomic invariants.
    }
    \label{fig:substructure_coverage}
\end{figure}

Examining the substructures in the pre-training dataset, QMugs, 
the ratio of unique substructures to the total number of molecules decreases asymptotically with more molecules 
    (Fig.~\ref{fig:substructure_coverage}a).
In Biogen benchmark data, the proportion of substructures that appear in 
QMugs pre-training subsets does not substantially change as the Tanimoto similarity filter becomes more relaxed
    (Fig.~\ref{fig:substructure_coverage}b),
i.e., the relative change in substructure content between QMugs subsets is very small.
At $462,189$ molecules, each QMugs subset is likely to contain over $80\%$ of the
total substructures that appear in QMugs
    (Fig.~\ref{fig:substructure_coverage}a).
Our filtered QMugs subsets are therefore not OOD relative to the benchmark data with respect to individual substructures, 
but rather with respect to their combinations.

Pre-training objectives focused on substructure topology can potentially 
generalise to novel combinations of familiar substructures,
as the majority of the underlying substructure content is already 
observed in the most stringently filtered QMugs subset.
Additionally, the robustness of topological pre-training would likely
increase as pre-training dataset size, and thus substructure coverage, increases.
However, this assumption is unlikely to hold for more extreme similarity thresholds, or more robust similarity measures, 
where the pre-training data may be more OOD with respect to the benchmark data at the individual substructure level.
The expected effect of increased performance with greater data leakage is only observed on Human PPB,
the only Biogen task where PT-GIN fails to outperform fixed approaches,
suggesting that similarity overlap can still lead to overly optimistic performance,
but that this is not always the case.
Decreasing $\mathrm{R}^2$ with greater data leakage in some instances could be due to 
increased variance in molecular topology within the pre-training data,
without a corresponding increase in the number of samples. 

Our findings suggest that in addition to performance improvements on Biogen ADMET tasks,
pre-training on molecular substructures could enhance GNN robustness to benchmark data 
that contains molecules with OOD combinations of substructures.
Increasing similarity between pre-training and benchmark datasets 
may not consistently improve performance
on models pre-trained to predict molecular topology 
due to the substructure corpus being well-represented by a dissimilar set of molecules.
While the filtered subsets may not be OOD 
with respect to individual substructures, they remain OOD in terms of molecular composition.
This suggests that our pre-training strategy effectively learns to generalise 
from substructure components to entire molecules,
achieving robustness that would likely increase with even larger pre-training datasets,
although this requires further validation.

\section{Discussion}\label{sec:discussion}

Graph neural networks have historically struggled to outperform   
fixed molecular featurisation methods for QSAR~\cite{dablanderM-2023-ExploringQSARModelsActivitycliffPrediction, jiangD-2021-CouldGraphNeuralNetworksLearnBetterMolecularRepresentationDrugDiscovery, xiaJ-2023-UnderstandingLimitationsDeepModelsMolecularPropertyPrediction}.
Our work demonstrates that self-supervised pre-training on Extended-Connectivity Fingerprints
is a simple and effective strategy for improving the OOD 
performance of GNNs on key QSAR tasks. By leveraging the 
structural information encoded in ECFPs, ECFP-pre-trained GINs
achieve significantly higher $\mathrm{R}^2$ than GINs trained from scratch and 
fixed fingerprinting methods on five of the six
ADMET tasks in the Biogen dataset,
validated through rigorous statistical testing.
As such, enhancement of GNNs for drug discovery can be achieved 
without increasing model complexity or utilising large, noisy, labelled pre-training
datasets.
Furthermore, our approach requires limited computational resources, as:
ECFPs are computationally efficient and fast to generate;
the pre-training corpus used in this study is relatively small ($<$0.5M molecules);
and we demonstrate downstream use with frozen embeddings and LightGBM, without
fine-tuning.
However, our results with randomly initialised GINs suggest that ECFP pre-training is not always beneficial 
and that performance varies depending on the alignment of substructure vocabularies 
between pre-training and benchmark tasks.

The benefits of ECFP pre-training were task-dependent. 
This result is consistent with prior work demonstrating that QSAR performance of
molecular featurisation methods varies between tasks~\cite{dablanderM-2023-ExploringQSARModelsActivitycliffPrediction, jiangD-2021-CouldGraphNeuralNetworksLearnBetterMolecularRepresentationDrugDiscovery},
including for pre-trained representations~\cite{dengJ-2023-SystematicStudyKeyElementsUnderlying, liZ-2026-SystematicSurveyBenchmarkDeepLearning, sunR-2022-DoesGNNPretrainingHelpMolecular}.
Our comparison of LightGBM and KR suggests that this task-dependent outperformance
of baseline fixed fingerprints is influenced by the downstream model,
with PT-GIN performance improving substantially compared to ECFP$_{\mathrm{S\&S}}$
across tasks when paired with KR.
Task dependence could arise from the pre-training target itself;
for example, we hypothesise that the absence of 3D information in ECFPs
limits what pre-training can transfer to tasks like MUV,
where all evaluated 2D representations performed poorly.
If so, it may be preferable to select self-supervised targets
based on the specific characteristics of the downstream task,
such as E3FPs~\cite{axenSD-2017-SimpleRepresentationThreeDimensionalMolecularStructure}
for tasks that depend on 3D molecular structure.
Jointly pre-training on multiple targets, such as ECFPs and E3FPs, could also be beneficial; however,
this may not entirely mitigate an erroneous relationship between one or more pre-training targets 
and downstream performance.

In addition, we analysed the impact of pre-training data leakage, 
in the form of topological similarity (ECFP$4_{2048}$ Tanimoto similarity), on model performance. 
We found that data leakage does not systematically improve pre-trained model performance, with the exception of Human PPB.
On the remaining Biogen tasks,
greater data leakage showed an association with reduced performance or no association at all.
This is counterintuitive, as data leakage typically leads to overly optimistic performance. 
This suggests that as long as the pre-training corpus is large enough to contain a diverse 
vocabulary of chemical substructures, a GNN could learn to generalise to novel substructure combinations.
The focus for self-supervised pre-training may not need to be on finding perfectly matched pre-training domains, 
but rather on maximising the diversity of chemical motifs.
However, this may not hold for more extreme similarity thresholds, or more robust similarity measures, 
where the pre-training data may be more OOD with respect to the benchmark data at the substructure level.
In a related context, Truong et al.\ found that, for protein-ligand docked pose scoring with logistic regression,
increasing ECFP$4_{2048}$ Tanimoto similarity overlap between train and test sets did not improve docking power,
even at more stringent similarity thresholds~\cite{truongCM-2026-NextTopDocker}.

To summarise:
\begin{enumerate}
  \item ECFP pre-training is a straightforward, inexpensive method for improving GINs,
        outperforming fixed fingerprints on five of six Biogen QSAR tasks and
        randomly initialised GINs on three of the six tasks.
  \item The relative performance between PT-GIN embeddings and fixed fingerprints depends
        on the downstream model; PT-GIN outperforms ECFP$_{\mathrm{S\&S}}$ with KR 
        on the majority of regression tasks, and matches performance on the remaining tasks,
        but underperforms with LightGBM on the Lipophilicity, FreeSolv, and ChEMBL tasks.
  \item Increased data leakage during pre-training, in the form of topological similarity,
        did not systematically improve pre-trained model performance,
        with the exception of Human PPB, suggesting that individual substructure coverage in the
        pre-training corpus may matter more than excluding similar molecules.
\end{enumerate}

Looking forward, our topological pre-training approach 
can be adopted as a simple method to boost GNN 
performance on relevant ADMET tasks in an OOD setting, even with limited pre-training data.
Our findings suggest that pre-training on topological fingerprints
can enhance GNN robustness in regimes with lower data overlap between pre-training and benchmark datasets.
Additionally, our substructure-based tokenisation allows for interpretable model insights, via permutation importance
(Supplementary Information Figs.~51--56).
Furthermore, multi-target self-supervised pre-training strategies should be investigated, such as 
pre-training on topological fingerprints and PDVs simultaneously,
although this would require more data than used in this study.
Moreover, pre-training strategies should be devised and implemented with consideration for the specific
characteristics of the benchmark task or tasks.
Finally, our work focuses on baseline evaluation of ECFP pre-training
and does not establish an advantage over other pre-training approaches;
further comparisons with other pre-training strategies,
such as pre-training on physicochemical descriptors and 3D fingerprints,
are needed on standardised benchmark and pre-training datasets.
The community would benefit greatly from standardised pre-training protocols,
including curated and consistent pre-training datasets that consider benchmark tasks.
Such standards would clarify the strengths and limitations of different pre-training approaches
for molecular property prediction, and are vital for fairly assessing performance differences between pre-training methods.

\section{Methods}\label{sec:methods}

For package versions used here, see Supplementary Information Table~6.

\subsection{Datasets}\label{sec:datasets}

All datasets used in this study were curated and published in prior work;
we applied only the preprocessing steps described below (see Methods, \textit{\nameref{sec:methods_data_preparation}}).

\subsubsection{QMugs}\label{sec:datasets_pretraining}

Self-supervised pre-training of GINs was performed on QMugs~\cite{isertC-2022-QMugsQuantumMechanicalPropertiesDruglike}, 
a chemical dataset consisting of $665$k molecules. QMugs contains drug-like molecules and 
prior work has shown QMugs has potential for pre-training~\cite{rajaA-2024-EffectivenessQuantumChemistryPretrainingPharmacological},
and the dataset is large enough, even with similarity filtering, for pre-training, 
whilst being small enough for extensive ablation studies.

We identified approximately $10$k duplicate SMILES strings in QMugs; 
on inspection, many of the duplicates are enantiomers with missing chirality flags.
To avoid loss of chirality and erroneous duplicates, we used the ChEMBL IDs saved to QMugs to retrieve SMILES
strings directly from ChEMBL v.27~\cite{zdrazilB-2024-ChEMBLDatabase2023}.

\subsubsection{Benchmarking datasets}\label{sec:datasets_benchmarking}

The primary benchmarking dataset used in this study was Biogen~\cite{fangC-2023-ProspectiveValidationMachineLearningAlgorithmsAbsorptionDistributionMetabolismExcretionPrediction},
a dataset of $3521$ molecules with six ADMET tasks. 
Its prediction tasks are plasma protein binding (PPB) in human and rat blood plasma, 
intrinsic hepatic clearance (CLint) in human and rat liver,
a measure of active transport by multidrug resistance protein 1 as an efflux ratio,
and aqueous solubility. 
Biogen was selected for its high degree of experimental consistency, 
with all assays conducted under uniform conditions, ensuring data homogeneity~\cite{fangC-2023-ProspectiveValidationMachineLearningAlgorithmsAbsorptionDistributionMetabolismExcretionPrediction}.

We also used a subset of MoleculeNet~\cite{wuZ-2018-MoleculeNet} 
for benchmarking, including three regression datasets, ESOL~\cite{delaneyJS-2004-ESOL}, 
FreeSolv~\cite{mobleyDL-2014-FreeSolv},
and Lipophilicity~\cite{zdrazilB-2024-ChEMBLDatabase2023}, 
and two binary classification datasets, MUV~\cite{rohrerSG-2009-MaximumUnbiasedValidationMUVData}, 
and Tox21~\cite{richardAM-2021-Tox2110KCompoundLibrary}. 
ESOL is a solubility task, although solubility values extend outside 
the typical range for drug-like molecules~\cite{ryttingE-2005-AqueousCosolventSolubilityDataDruglike}.
FreeSolv is a benchmark for free energies of solvation.
MUV contains 17 target-specific, imbalanced virtual screening (VS) tasks.
Tox21 contains 12 toxicity tasks from high-throughput \textit{in vitro} assays.
Whilst MoleculeNet benchmarks have been subject to critique~\cite{waltersWP-2023-WeNeedBetterBenchmarksMachine, kretschmerF-2025-CoverageBiasSmallMoleculeMachine, wognumC-2024-CallIndustryledInitiativeCriticallyAssess, wallachI-2018-MostLigandBasedClassificationBenchmarksReward}, 
they are still widely used for QSAR benchmarking and can highlight potential model weaknesses, hence their inclusion.

We also benchmarked the featurisation methods on the challenging task of 
target-specific binding affinity prediction.
We employed two datasets derived from ChEMBL~\cite{zdrazilB-2024-ChEMBLDatabase2023} 
that were previously curated and used in another systematic QSAR study~\cite{dablanderM-2023-ExploringQSARModelsActivitycliffPrediction}.
These datasets are heterogeneous, with data from multiple assay deposits in 
ChEMBL, and are therefore likely to contain noise.
The target associated with the first dataset is Dopamine Receptor D2 (DRD2), 
a regulatory signalling protein implicated in neurological disorders~\cite{panX-2019-DopamineDopamineReceptorsAlzheimersDisease},
and the second dataset is Factor Xa, 
a pharmaceutical protein target for anticoagulation therapies~\cite{alexanderJH-2005-InhibitionFactorXa}.

\subsection{Data preparation and splitting}\label{sec:methods_data_preparation}

All Tanimoto similarities were determined using hashed ECFPs, with chirality set to True, 
$\mathrm{radius}=2$, and $\mathrm{fpsize}=2048$.

\subsubsection{Molecular standardisation}\label{sec:methods_standardisation}

Molecular standardisation was performed to ensure that all molecules were in a 
consistent format across the pre-training and benchmark tasks,
and to remove molecules with invalid SMILES strings.
Standardisation across tasks is important to ensure that both learned
embeddings and fingerprints are reliable across tasks;
this additionally ensures that similarity measures between
molecules from different tasks are robust.
All pre-training and benchmark molecules were standardised using RDKit (2024.09.03)~\cite{landrumG-2024-RDKitOpensourceCheminformatics}:

\begin{enumerate}\raggedright
    \item Molecules were checked for incorrect chemistry by \texttt{Chem.SanitizeMol}, including running kekulisation, checking valences are valid, setting aromaticity, conjugation, and hybridisation.
    \item Further checks were done using \texttt{Cleanup} from \texttt{rdMolStandardize}, including hydrogen removal, removal of disconnected metal atoms, functional group normalisation, checking stereogroups, and removing incorrect chiral flags.
    \item Disconnected fragments were removed using \texttt{FragmentParent} from \texttt{rdMolStandardize}.
    \item Molecules were neutralised using \texttt{Uncharger} from \texttt{rdMolStandardize}.
    \item The canonical tautomer was determined using \texttt{TautomerEnumerator} from \texttt{rdMolStandardize}. 
    Tautomerisation was performed without sp$^{3}$ stereochemistry removal to retain chirality.
\end{enumerate}
Any molecule that failed to pass standardisation was excluded.
Excluded SMILES strings, and their failure modes, are listed in the Supplementary Information.

\subsubsection{Tanimoto similarity filtering of pre-training data}\label{sec:methods_splitting_filtering}

QMugs was filtered into different subsets for pre-training
to investigate the effects of data leakage, in the form of topological similarity, on model performance. 
Tanimoto similarities were determined between the molecules in QMugs,
and the molecules in every benchmark dataset used (Biogen, MUV, Tox21, etc.). 
Any molecule in the QMugs dataset with $>0.5$ similarity
to any benchmark molecule was excluded,
leaving an initial pre-training dataset of $462,189$ molecules.
For data leakage analysis, the same process was repeated 
with similarity thresholds of $0.6$, $0.7$, $0.8$, $0.9$, and $1.0$.
To ensure that the same number of molecules was used for pre-training at each similarity threshold,
random uniform sampling without replacement was performed on the filtered QMugs subsets to yield $462,189$ molecules
for each threshold.

\subsubsection{Repeated cross-validation for OOD testing}\label{sec:repeated_cv}

Repeated grouped CV was chosen for generating OOD train-test splits for each benchmark dataset.
CV allows for multiple estimations of model performance, subsequently enabling statistical tests and effect size measurements~\cite{ashJR-2025-PracticallySignificantMethodComparisonProtocols}.
All methods were evaluated on the identical splits to ensure fair comparisons.
An alternative approach would be to use a single hold-out test set for each benchmark dataset
as a measure of OOD performance.
However, this approach is limited in settings where high split-to-split performance variance is expected, 
such as with small datasets, as model comparisons based on a single split can be unreliable~\cite{bouthillierX-2021-AccountingVarianceMachineLearningBenchmarks}.
In drug discovery, datasets are typically small and noisy~\cite{crusiusD-2025-AreWeFittingDataNoise, landrumGA-2024-CombiningIC50KiValuesDifferent, waltersWP-2023-WeNeedBetterBenchmarksMachine} and are therefore prone to data split variance.
Given a greater amount of high-quality data, a single hold-out test set would be an appropriate complementary approach.

Pairwise Tanimoto similarities between molecules were calculated for each benchmark dataset.
Distances between molecules were calculated as \mbox{$\mathrm{dist} = 1 - \mathrm{Tanimoto\ similarity}$}.
Benchmark datasets were individually clustered using Butina clustering on Tanimoto distances~\cite{butinaD-1999-UnsupervisedDataBaseClusteringBasedDaylightsFingerprintTanimotoSimilarity} 
with a cutoff of $0.65$: the default cutoff in Useful RDKit Utils~\cite{waltersWP-2024-UsefulRDKitUtils}.
We chose Butina for its speed, ease of implementation in RDKit, 
and the evenly sized clusters it produces.
Here we assume that molecules within the same cluster are more similar to each other than to molecules in other clusters,
enabling us to split the data by clusters for OOD train-test sets.
Datasets were repeatedly resampled into k-folds, similarly to Ash et al.~\cite{ashJR-2025-PracticallySignificantMethodComparisonProtocols},
for grouped CV; we used \texttt{GroupKFold} for regression tasks and \texttt{StratifiedGroupKFold} for classification tasks~\cite{pedregosaF-2011-Sklearn}.
We used $200$ repetitions for primary evaluation, each time splitting into five folds of held-out clusters with a new random seed,
producing $1000$ train-test splits for each dataset.
Distributions of maximum Tanimoto similarity between the train and test sets compared to random splits 
are shown in Supplementary Information Figs.~75--79.
A high number of repetitions was chosen to ensure stable estimates for differences in effect size between methods.

One repetition was used for hyperparameter tuning; the remaining $199$ were used for model evaluation.
Even with stratifying by target label, the MUV dataset occasionally produced folds with no positive labels 
due to extreme class imbalance; test metrics over these folds were ignored (see Supplementary Information Table~8). 

For ablation evaluations, we used a subset of $20$ repetitions for each dataset, producing $100$ train-test splits;
hyperparameter tuning was performed on the same splits as the primary evaluations.

\subsubsection{Units}
For binding affinity tasks, DRD2 and Factor Xa, 
affinity was converted from $\mathrm{nM}$ to $\mathrm{M}$ before calculating $\mathrm{pK_i}$. 
For ESOL, $\mathrm{log(S)}$ was converted from $\mathrm{log(M)}$ to $\mathrm{log(\mu M)}$.

\subsection{Model architecture and training strategy}

\subsubsection{Graph neural networks}

Graph neural networks were implemented using PyTorch (2.5.1) and PyTorch Geometric (2.6.1)~\cite{paszkeA-2019-PyTorch,feyM-2019-FastGraphRepresentationLearningPyTorch}. 
In this work, we used Graph Isomorphism Networks (GINs)~\cite{xuK-2019-HowPowerfulAreGraphNeural} 
as our GNN architecture of choice.
Our study was motivated by the fact that previous work has suggested that GINs without 
pre-training struggle to outperform fixed featurisation methods for QSAR~\cite{dablanderM-2023-ExploringQSARModelsActivitycliffPrediction}.

\subsubsection{Substructure tokenisation}\label{sec:tokenisation}

\begin{figure}[!htbp]
    \centering
    \textbf{PT-GIN architecture and benchmarking pipeline}
    \includegraphics[width=\textwidth,keepaspectratio]{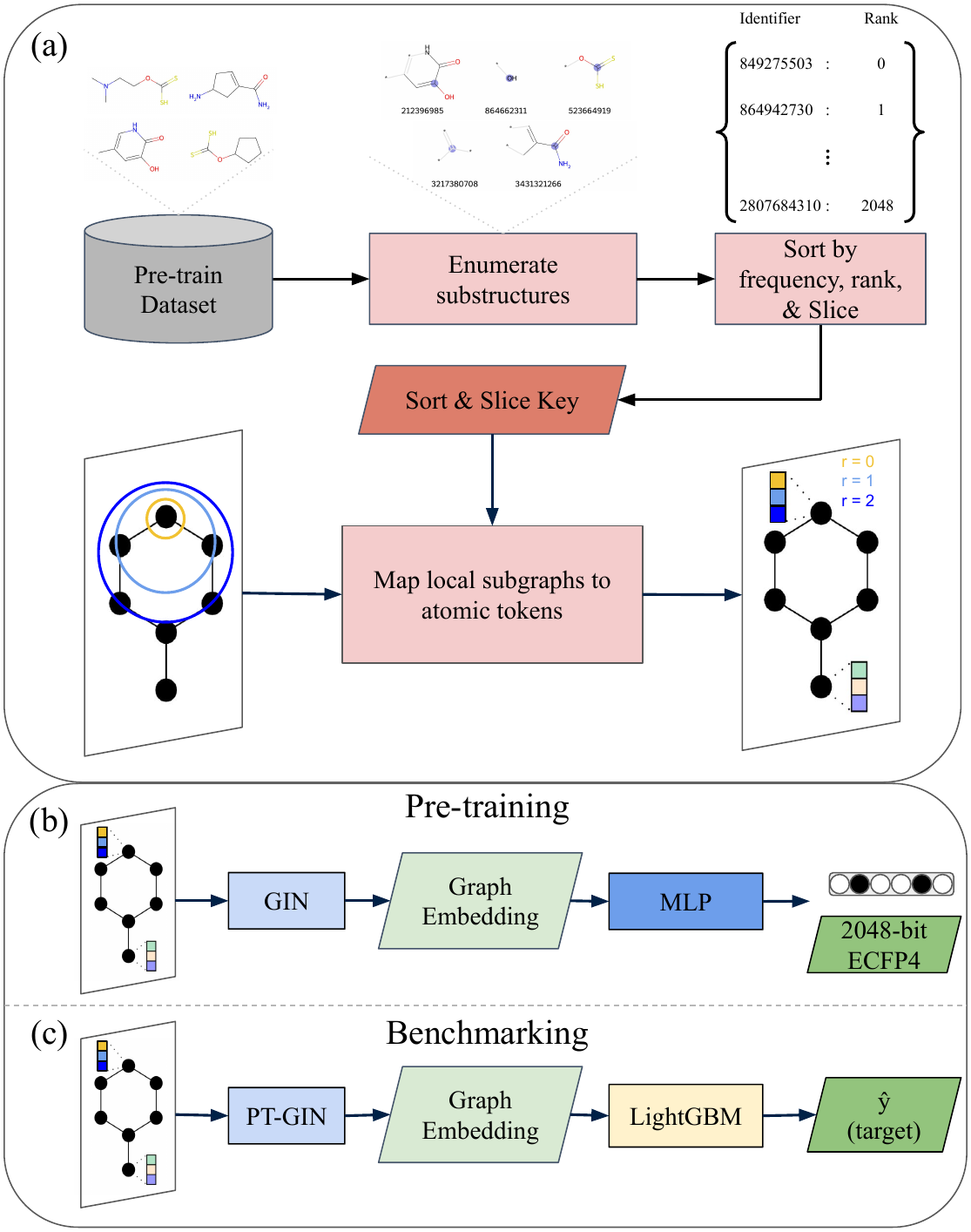}
    \caption{
        (a) Substructure tokenisation of molecular graphs through Sort \& Slice (S\&S)~\cite{dablanderM-2024-SortSlice}.
        (b) Self-supervised, topological pre-training pipeline. 
        GINs were pre-trained to predict hashed ECFPs~\cite{rogersD-2010-ExtendedconnectivityFingerprints} from substructure-tokenised molecular graphs.
        (c) Benchmarking pipeline with frozen ECFP-pre-trained GINs and LightGBM~\cite{keG-2017-LightGBM}.
    }
    \label{fig:architecture}
\end{figure}

Molecular graph featurisation is commonly performed with curated atom-level descriptors~\cite{gilmerJ-2017-NeuralMessagePassingQuantumChemistry, dablanderM-2023-ExploringQSARModelsActivitycliffPrediction, zaidiS-2023-PretrainingDenoisingMolecularPropertyPrediction}. 
An alternative approach is to embed the molecule based on atom types.
For MolE, M\'endez-Lucio et al.~\cite{mendez-lucioO-2024-MolE} 
introduced encoding via Daylight atomic invariants, 
i.e., via Morgan identifiers with radius $r=0$~\cite{rogersD-2010-ExtendedconnectivityFingerprints}. 
We explored taking this further by expanding this strategy 
to include local substructures beyond a radius of $r=0$ 
and assigning multiple tokens to each atom, one token for each radius
    (Fig.~\ref{fig:architecture}a). 
We refer to this process as substructure tokenisation. 
The vocabulary is generated by ranking Morgan identifiers 
based on their frequency in the pre-training corpus, 
a process analogous to the creation of corpus-dependent vocabularies in natural language processing.

Circular substructure vocabularies for pre-training datasets were determined by applying 
Sort \& Slice (S\&S)~\cite{dablanderM-2024-SortSlice} to enumerated ECFP substructures
(Supplementary Information Algorithm~1).
S\&S is an easy-to-implement and bit-collision-free alternative to hash-based folding
for the vectorisation of the set of ECFP-substructures detected in an input molecule. S\&S first sorts
ECFP substructures according to their relative prevalence in a given set of training compounds and
then incorporates only the most frequent substructures into the final binary fingerprint. This
vectorisation technique has been shown to robustly outperform hash-based folding at a diverse set
of molecular property prediction tasks. Here, we apply S\&S in a new context;
rather than mapping substructures to positions in a binary fingerprint,
we map substructures to tokens for vectorial embedding.

Substructure enumeration was performed using the 
Morgan Fingerprint generator with chirality set to True in RDKit~\cite{landrumG-2024-RDKitOpensourceCheminformatics}.
By applying the S\&S algorithm to substructure vocabulary construction,
we were able to rank the circular substructure identifiers by frequency and control the vocabulary size.
We used S\&S to
\begin{enumerate}
    \item sort Morgan circular substructure identifiers by the number of molecules they appear in,
    \item map each identifier to its rank by frequency, and
    \item slice the vocabulary to a specific size, keeping the top-N most frequent substructures in the dataset.
\end{enumerate}

Each identifier rank in the vocabulary was mapped to a learnable embedding using \texttt{torch.nn.Embedding}.
On graph featurisation, each atom in the molecule was assigned a token for each radius up to the maximum radius
based on its local atomic environment identifiers and their ranks in the vocabulary.
Graphs were then embedded using the corresponding embedding weights,
producing a node embedding shape of $(i, j, k)$, where:
\begin{itemize}
    \item $i$ is the number of heavy atoms in the molecule,
    \item $j$ is the number of radii, one token per radius ($j = r + 1$), e.g., $j=2$ for a maximum radius of $r=1$ 
    with a zero radius and a one radius token, and 
    \item $k$ is the embedding size.
\end{itemize}
Multi-token node embeddings were concatenated to produce one unified vectorial representation per atom
with shape $(i, jk)$. Thus, the atomic embedding shape varied depending on the maximum radius of 
the tokenisation strategy (Supplementary Information Table~12).

The maximum radius of circular substructures and vocabulary size were treated as hyperparameters. 
For PT-GIN, vocabulary sizes used were $1024$, $2048$, $4096$, $8192$, and $16384$. 
Vocabularies beyond $16384$ were not considered due to increasingly high parameter counts 
relative to the amount of pre-training data.
Maximum radii of $r=0$, $r=1$, and $r=2$ were explored.

To assess the contribution of substructure tokenisation, we performed an ablation study 
comparing the performance of PT-GINs with substructure tokenisation
to PT-GINs with conventional atom-level descriptors on Biogen (Supplementary Information Figs.~57--60).
The ablation demonstrated that relative performance varies across different benchmark tasks:
by $\mathrm{R}^2$, the two approaches were comparable on three of six tasks,
with atom-level descriptors performing better on two (Human PPB and Human CLint)
and substructure tokenisation on one (Solubility).

For Scratch GIN, the hyperparameter search space was constrained to smaller embedding sizes, vocabulary sizes, and maximum radii
(see Methods, \textit{\nameref{sec:benchmark_tuning}}).
We initially explored Scratch GIN with different graph representations, including using atomic features 
and tokens based only on atom type; however, performance was notably improved by tokenising
based on local atomic environments.

The resulting substructure vocabularies were used to featurise molecular graphs
by assigning substructure tokens to each atom for all radii up to the
pre-defined maximum (Supplementary Information Algorithm~2).
Substructures absent from the vocabulary were assigned an \texttt{UNK} token,
including those that were sliced out during the S\&S process.

\subsubsection{Self-supervised pre-training}\label{sec:pretrain_method}

GINs were pre-trained (PT-GIN) on hashed radius-two, $2048$-bit ECFPs (ECFP$4_{2048}$), with chirality set to True.
This approach is malleable as the label can be substituted
for different topological targets, such as FCFPs.
The ECFP$4_{2048}$ bit-vectors were used as a multi-label binary classification target 
for self-supervised pre-training
    (Fig.~\ref{fig:architecture}b). 
While hash-based vectorisation is prone to information loss resulting from bit collisions~\cite{dablanderM-2024-SortSlice}, 
it allows compression of the constituent circular substructures in a molecule into 
a smaller vectorial representation. Prediction of the entire substructure corpus without hash-based 
folding would require substantially more trainable parameters within our neural architecture.
We did not use S\&S to fold ECFPs for pre-training, as S\&S would have removed rare substructures 
entirely. Loss was calculated using binary cross-entropy with balanced class weighting~\cite{pedregosaF-2011-Sklearn}. 

Pre-training GIN hyperparameters were explored using Optuna (4.2.1)~\cite{akibaT-2019-Optuna} 
with $50$ trials and $5$-fold CV on the $0.5$ similarity-filtered QMugs subset,
aiming to maximise mean test fold AUROC on ECFP$4_{2048}$ predictions.
Splitting was performed via \texttt{KFold}~\cite{pedregosaF-2011-Sklearn}.
For pre-training hyperparameters, see Supplementary Information Tables~10 and~11.
The initial learning rate was calculated as $lr = c\cdot p_{model}^{-0.5}$,
where $p_{model}$ is the number of model parameters and $c$ is the learning rate scaling factor;
a value of $c = 0.5$ and a half-life of $5$ epochs were used.
After tuning, the best hyperparameters were used to pre-train models on the entirety of each pre-training subset.

During pre-training, the output from the last GIN layer was the input for the prediction head. 
Multiple models with varying vocabulary sizes and maximum radii were pre-trained;
for benchmarking, each pre-trained model was used for featurisation in 
individual LightGBM hyperparameter tuning runs on the benchmark data (see Methods, \textit{\nameref{sec:benchmark_tuning}}).
The ECFP-pre-trained GIN with the best performance during the benchmark hyperparameter
tuning was selected for testing on the remaining $199$ repeats,
using the model with the lowest $\mathrm{MAE}$ for regression tasks and highest AUCPR for classification tasks.
For each task, models pre-trained for investigating threshold similarity used the same radius and vocabulary
size as the PT-GIN pre-trained on the $0.5$ filtered dataset with the lowest $\mathrm{MAE}$ in hyperparameter tuning.

\subsection{Benchmarking setup}

\subsubsection{Pre-trained embedding representations}

In downstream benchmarking, frozen embeddings from pre-trained models
were used as molecular representations
    (Fig.~\ref{fig:architecture}c);
PT-GIN outputs after each layer were pooled and concatenated to form a global molecular embedding.
The graph pooling method (mean, max, etc.) was a hyperparameter (Supplementary Information Table~10).
Fine-tuning was explored in preliminary experiments; however,
end-to-end fine-tuning was computationally expensive relative to using frozen embeddings 
with light gradient-boosted machines (LightGBM)~\cite{keG-2017-LightGBM}, and
the benefits were often negligible. In some cases fine-tuning was even detrimental. 
This was most likely due to the small size of the downstream
benchmark datasets relative to the parameter count in the pre-trained model,
even when only fine-tuning a subset of model layers. 
As such, we used frozen pre-trained embeddings as inputs for
LightGBM models in downstream benchmarking.
The use of pre-trained embeddings with LightGBM additionally
allowed for a more direct comparison with baseline molecular fingerprint
representations, such as hashed ECFPs with LightGBM,
as differences in performance could not be attributed to
differences between architectures.

As an initial baseline, we conducted preliminary experiments 
pre-training an autoencoder,
equivalent in parameter count to PT-GIN, to reconstruct ECFP bit vectors. 
However, this approach underperformed compared to models using graph-based representations.

To validate that message passing in pre-training was necessary for downstream performance,
we performed an ablation study comparing PT-GIN with and without message passing, replacing 
the message passing mechanism with a multi-layer perceptron that takes pooled graph representations as input
(Supplementary Information Figs.~66--69).
The ablation study demonstrated that message passing is necessary for downstream performance,
as PT-GIN without message passing underperformed on all Biogen tasks.

\subsubsection{Baseline fixed featurisation methods}

For our baseline comparisons, we used static circular topological fingerprints, 
namely fingerprinting methods vectorised via hash-based folding, including 
Extended-Connectivity Fingerprints (ECFP$_{\mathrm{hashed}}$)~\cite{rogersD-2010-ExtendedconnectivityFingerprints}
and Functional-Class Fingerprints (FCFP$_{\mathrm{hashed}}$). 
Furthermore, we employed ECFPs vectorised via Sort \& Slice (ECFP$_{\mathrm{S\&S}}$)~\cite{dablanderM-2024-SortSlice}.
The S\&S vectorisation technique for ECFPs has been shown to lead to robustly superior 
predictive performance over hash-based folding on various molecular property prediction tasks~\cite{dablanderM-2024-SortSlice}.

For fixed fingerprinting methods, the radius and bit-vector lengths were 
treated as hyperparameters. We used radii of $r=1$ and $r=2$, 
and bit-vector lengths of $1024$ and $2048$.
For each train-test split, features with zero variance in the training split were excluded.

\subsubsection{Light gradient-boosted machines}\label{sec:methods_downstream_models}

\begin{figure}[!htbp]
    \centering
    \textbf{Baseline pipelines for fixed fingerprinting methods and Scratch GIN}
    \includegraphics[width=\textwidth,keepaspectratio]{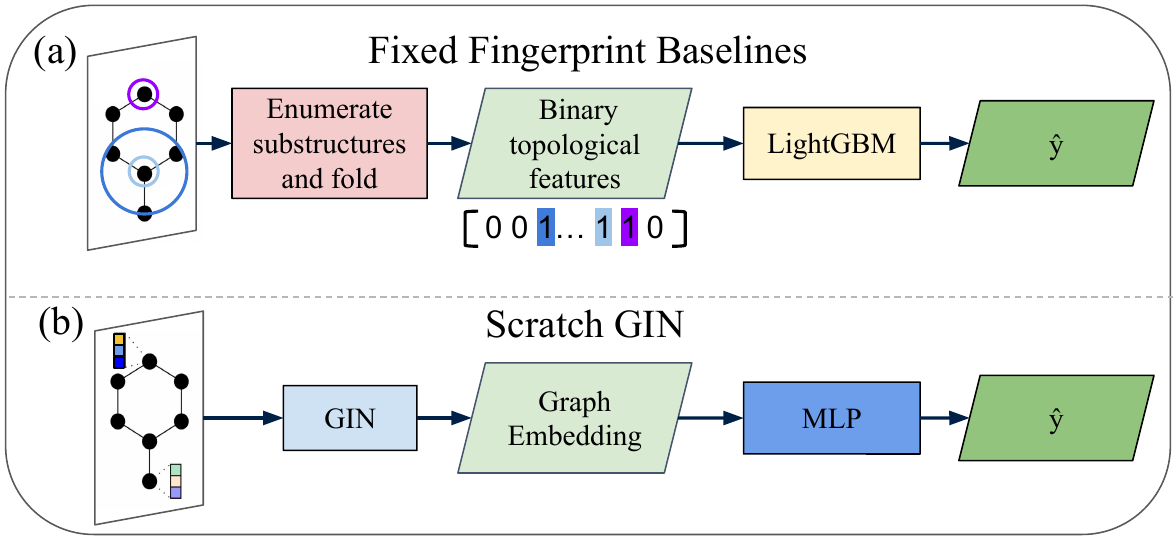}
    \caption{
        (a) Fixed fingerprinting methods.
        Circular substructures are enumerated from the molecular graph and vectorised 
        into a fixed-length feature vector, by hash-based folding for 
        ECFP$_{\mathrm{hashed}}$, FCFP$_{\mathrm{hashed}}$ or by 
        Sort \& Slice for ECFP$_{\mathrm{S\&S}}$. The feature vector is passed to LightGBM 
        to predict the target. The representation is fixed, so only the downstream model 
        is fitted to benchmark labels.
        (b) Scratch GIN. A randomly initialised GIN is trained end-to-end on benchmark labels, 
        pooling node embeddings into a global graph embedding that a multi-layer perceptron 
        head maps to the target label. The representation is learned from scratch, 
        on the benchmark data, so the entire model is fitted to benchmark labels without
        any pre-training.
    }
    \label{fig:baseline_pipelines}
\end{figure}

LightGBM models were used in benchmarking for their training speed and efficiency~\cite{keG-2017-LightGBM}. 
Frozen ECFP-pre-trained GIN embeddings were used as inputs with LightGBM models 
to evaluate PT-GIN as a featurisation method (Fig.~\ref{fig:architecture}c).
All fixed baseline molecular representations were fed into a LightGBM model as well
for downstream prediction (Fig.~\ref{fig:baseline_pipelines}a).
LightGBM models were implemented using the Python package \texttt{lightgbm} (4.6.0)~\cite{keG-2017-LightGBM}.

To evaluate the impact of downstream model architecture on performance,
we compared the performance of Kernel Ridge regression (KR)~\cite{saundersC-1998-RidgeRegressionLearningAlgorithmDual}
and LightGBM~\cite{keG-2017-LightGBM} with $20$ repeats of $5$-fold CV for each task.
We used PT-GIN and ECFP$_{\mathrm{S\&S}}$ as input representations for both downstream models.
For KR, PT-GIN embeddings were standardised and used with a radial basis function (RBF) kernel;
ECFP$_{\mathrm{S\&S}}$ was used with a Tanimoto kernel.
KR hyperparameters were optimised for each task with grid search using the same
single $5$-fold CV repeat used for LightGBM (Supplementary Information Table~5).
For classification tasks (MUV and Tox21), we used class weighting to account for class imbalance.
For MUV, we used Nystr\"om approximation with $2000$ landmark points
to reduce memory requirements~\cite{williamsC-2001-UsingNystroemMethodSpeedKernel};
all training set actives were included as landmarks
and decoys were randomly sampled.
Further results for KR are provided in Supplementary Information Figs.~43--50 and Table~4.

\subsubsection{Scratch GIN}\label{sec:scratch_gin_methods}

Scratch GINs were trained with a standard multi-layer perceptron prediction head (Fig.~\ref{fig:baseline_pipelines}b). 
Regression tasks were trained using mean squared error (MSE) loss. 
On binary classification tasks, models were trained using binary cross-entropy loss with balanced class weighting.
Learning rate was defined similarly to PT-GIN, with learning rate scaling and half-life as hyperparameters (see Supplementary Information Table~14).
We explored retraining LightGBM models on Scratch GIN embeddings
for a consistent pipeline across all molecular representations; however, there was
no noticeable difference in performance between end-to-end Scratch GIN models 
and LightGBMs trained on the same embeddings.

As Scratch GIN models were computationally expensive to repeatedly train compared to the 
LightGBM used with the other approaches, comparisons to Scratch GINs were limited to Biogen.

\subsubsection{Benchmark hyperparameter tuning}\label{sec:benchmark_tuning}
Hyperparameter tuning for benchmarking was performed on a held-out $5$-fold CV using Optuna~\cite{akibaT-2019-Optuna} 
with 50 trials.
Ideally, nested CV would have been used for tuning and evaluation.
However, this would have required approximately $250$k
additional model fits ($50$ trials by $5$ folds per trial by $1000$ train-test splits)
per baseline per dataset,
which was computationally prohibitive.
Tuning was instead performed on one 5-fold set produced from repeated cross-validated splitting; 
these folds were excluded from the final evaluation.
Hyperparameters were selected based on MAE for regression tasks and AUCPR for binary classification tasks.

For LightGBM search spaces, see Supplementary Information Table~13.
The search space for Scratch GIN was constrained such that 
the number of parameters would be approximately within $10^0$--$10^1$ of the dataset size (Supplementary Information Tables~14--18). 
This was done to mitigate overfitting and improve generalisation in an OOD setting, as well as to reduce the computational cost of training.
Additionally, initial testing suggested that over-parameterisation did not lead to performance improvements of Scratch GIN,
likely due to the small size of the benchmark datasets.

We validated that at matched hyperparameters, the PT-GIN approach outperforms the Scratch GIN by performing an ablation study on Biogen, comparing Scratch GIN
and PT-GIN of equivalent capacity, vocabulary size, and radius (Supplementary Information Fig.~70).
The ablation study demonstrates that PT-GINs outperform Scratch GINs with equivalent capacity on all Biogen tasks
by $\mathrm{R}^2$, $\rho$, and $\mathrm{MAE}$.

\subsection{Model evaluation and statistical testing}

Ash et al.\ suggested performing multiple comparisons over repeated k-fold CV splits~\cite{ashJR-2025-PracticallySignificantMethodComparisonProtocols}. 
We adopted this approach for our benchmarking (see Methods, \textit{\nameref{sec:repeated_cv}}).
We averaged the performance metrics across all folds for each repeat 
as repeat-level averages provide more stable performance estimates than individual folds, 
which can be noisy for small datasets.
The $5$-fold split was used for hyperparameter tuning and excluded from model evaluations, 
resulting in $199$ samples for each statistical test.
For ablations, we used a subset of $20$ from the $199$ available repeats.

\subsubsection{Statistical testing and effect size measurement}

Tukey's Honest Significant Difference Test (Tukey's HSD),
a statistical test for evaluating whether performance differences between ML models are significant, 
and Cohen's \textit{d}, an effect size measure, are commonly appropriate for comparing models~\cite{ashJR-2025-PracticallySignificantMethodComparisonProtocols}.
Both of these approaches are parametric, assuming sample sets are normally distributed 
and that the variance within each group is comparable.
However, we found that metric distributions can be non-normally distributed,
particularly for performance scores with values close to the upper or lower bounds of the metric,
such as AUCPR
    (Supplementary Information Fig.~72).
Additionally, Tukey's HSD assumes independent observations; 
however, our results are paired, since the performance of a model on a CV repeat can be directly matched 
to the performance of a different model on the same CV repeat.
Therefore, we used two-sided Wilcoxon signed-rank tests~\cite{wilcoxonF-1945-IndividualComparisonsRankingMethods}, 
a non-parametric test for determining differences between paired samples by evaluating their ranks. 
To account for multiple comparisons, we applied the Bonferroni correction method to adjust \textit{p}-values
and reduce the occurrence of type I error~\cite{armstrongRA-2014-WhenUseBonferroniCorrection}.
Wilcoxon with Bonferroni correction was computed using scikit-posthocs (0.11.4)~\cite{maksimterpilovskii-2025-Scikit-posthocs}.
Even when normality held, we still used non-parametric tests and effect sizes
for consistency across tasks and metrics.

As we used many train-test splits in our benchmarking,
it is expected that even small differences in performance between models will be statistically significant;
to address this, we focused on effect sizes rather than just statistical significance.
Additionally, we plotted the \textit{p}-values and effect sizes for PT-GIN versus
all other approaches as a function of the number of CV repeats used in the statistical tests
to demonstrate the consistency of the effect sizes and the statistical significance,
using 100 repeated subsamples to obtain interquartile ranges.
The results for Biogen are shown in Supplementary Information Figs.~37--42.

To evaluate the effect size between models, we used rank-biserial correlation ($r_{rb}$)~\cite{kerbyDS-2014-SimpleDifferenceFormula, curetonEE-1956-RankbiserialCorrelation},
a paired non-parametric measure of effect size.
Given two models, $A$ and $B$, evaluated on the same CV repeats, $r_{rb}$ is the
difference between the proportion of the total rank sum that favours $A$
and the proportion that favours $B$:
\[r_{rb} = \frac{W^{+} - W^{-}}{W^{+} + W^{-}}\]
where $W^{+}$ is the sum of ranks for CV repeats where model $A$ has a greater value than model $B$,
and $W^{-}$ is the sum of ranks for CV repeats where model $B$ has a greater value than model $A$.
By ranks, we mean the ranks of the absolute differences between the performance metrics of the two models on each CV repeat, ignoring zero differences.
$r_{rb}$ ranges from $-1$ to $1$, where $-1$ indicates that model $B$ has a greater value than model $A$ on all CV repeats
and $1$ indicates that model $A$ has a greater value than model $B$ on all CV repeats.
As $r_{rb} = \frac{4W^{+}}{n(n+1)} - 1$ is a linear transformation of the Wilcoxon signed-rank statistic $W^{+}$~\cite{kerbyDS-2014-SimpleDifferenceFormula},
where $n$ is the number of CV repeats with non-zero differences,
its null distribution follows from the null distribution of $W^{+}$.
For large $n$ and no ties, $r_{rb}$ is approximately normally distributed under the null hypothesis $H_0$ of no difference between models:
\[r_{rb} \sim^{H_0} \mathcal{N}\left(0, \frac{2(2n + 1)}{3n(n + 1)}\right)\]

To clarify whether any approach across MUV and Tox21 tasks consistently outperforms the others,
we used the Friedman $\chi^2$ test~\cite{demsarJ-2006-StatisticalComparisonsClassifiersMultipleData}
for each dataset and each evaluation metric (Supplementary Information Tables~1 and~2).
Additionally, for Tox21 we used the Nemenyi post-hoc test~\cite{nemenyiPB-1963-DistributionFreeMultipleComparisons}
to determine which models were significantly different from each other
(Supplementary Information Table~3).

\subsubsection{Performance metrics}\label{sec:metrics}

Following Ash et al.~\cite{ashJR-2025-PracticallySignificantMethodComparisonProtocols},
we used multiple metrics to evaluate model performance as each has its own strengths and limitations.
For regression tasks, we used
\begin{itemize}
    \item the coefficient of determination ($\mathrm{R}^2$), calculated using \texttt{r2\_score} in scikit-learn (1.6.0)~\cite{pedregosaF-2011-Sklearn}, which measures the proportion of variance explained by the model: $\mathrm{R}^2 = 1 - \frac{\mathrm{MSE}}{\sigma^2(y)}$, where $\mathrm{MSE}$ is the mean squared error between predicted and true values, and $\sigma^2(y)$ is the variance of the true values;
    \item Pearson correlation coefficient ($\rho$), calculated using \texttt{pearsonr} in scipy (1.15.2)~\cite{virtanenP-2020-SciPy10FundamentalAlgorithmsScientific}, which measures the linear correlation between predicted and true values: $\rho = \frac{\mathrm{Cov}(y, \hat{y})}{\sigma_y \sigma_{\hat{y}}}$, where $\mathrm{Cov}(y, \hat{y})$ is the covariance between predicted and true values, and $\sigma_y$ and $\sigma_{\hat{y}}$ are the standard deviations of the true and predicted values, respectively;
    \item mean absolute percentage error ($\mathrm{MAPE}$), calculated using \texttt{mean\_absolute\_percentage\_error} in scikit-learn~\cite{pedregosaF-2011-Sklearn} and converted to a percentage, which measures the average absolute percentage difference between predicted and actual values: $\mathrm{MAPE} = 100 \cdot \frac{1}{n}\sum_{i=1}^{n} \left|\frac{y_i - \hat{y}_i}{y_i}\right|$, where $n$ is the number of samples, $y_i$ is the true value, and $\hat{y}_i$ is the predicted value; and
    \item mean absolute error ($\mathrm{MAE}$), calculated using \texttt{mean\_absolute\_error} in scikit-learn~\cite{pedregosaF-2011-Sklearn}, which measures the average absolute difference between predicted and actual values: $\mathrm{MAE} = \frac{1}{n}\sum_{i=1}^{n} |y_i - \hat{y}_i|$, where $n$ is the number of samples, $y_i$ is the true value, and $\hat{y}_i$ is the predicted value. As MAE is unit-dependent, it was used primarily for Supplementary Information box plots and ablations.
\end{itemize}

MAPE was chosen as the primary error metric for regression tasks as it is unit-agnostic, allowing for plotting of multiple regression tasks on the same axis,
which is not possible for other error metrics, such as MAE, which are unit-dependent.
However, MAPE weights errors on low-value targets more heavily than high-value targets, which can be misleading for some regression tasks.
For MAPE calculations, we excluded test samples with $y=0$,
as MAPE is undefined for these samples due to the division by zero.
This resulted in two samples being excluded from the Biogen Solubility task,
one from ESOL, two from FreeSolv, and three from Lipophilicity.
For $\rho$ calculations on the Biogen Rat PPB task with Scratch GIN,
we excluded three test folds, each in a different CV repeat,
which had constant predictions;
$\rho$ is undefined for these folds due to division by zero.

For classification tasks, we used
\begin{itemize}
    \item the area under the receiver operating characteristic curve (AUROC), calculated using \texttt{roc\_auc\_score} in scikit-learn~\cite{pedregosaF-2011-Sklearn}, which measures the ability of the model to distinguish between classes: $\mathrm{AUROC} = \int_{0}^{1} \mathrm{TPR}(\mathrm{FPR}^{-1}(x)) dx$, where $\mathrm{TPR}$ is the true positive rate and $\mathrm{FPR}$ is the false positive rate;
    \item the area under the precision-recall curve (AUCPR), calculated using \texttt{average\_precision\_score} in scikit-learn~\cite{pedregosaF-2011-Sklearn}, which measures the trade-off between precision and recall: $\mathrm{AUCPR} = \int_{0}^{1} \mathrm{P}(\mathrm{R}^{-1}(x)) dx$, where $\mathrm{P}$ is the precision and $\mathrm{R}$ is the recall;
    \item the Matthews correlation coefficient (MCC) at a threshold of $0.5$, calculated using \texttt{matthews\_corrcoef} in scikit-learn~\cite{pedregosaF-2011-Sklearn}, which measures the quality of binary classifications: $\mathrm{MCC} = \frac{\mathrm{TP} \cdot \mathrm{TN} - \mathrm{FP} \cdot \mathrm{FN}}{\sqrt{(\mathrm{TP} + \mathrm{FP})(\mathrm{TP} + \mathrm{FN})(\mathrm{TN} + \mathrm{FP})(\mathrm{TN} + \mathrm{FN})}}$.
\end{itemize}

For MUV virtual screening tasks, we also report the enrichment factor (EF)~\cite{benderA-2005-DiscussionMeasuresEnrichmentVirtualScreening},
the ratio of the fraction of actives among the top-ranked predictions to the fraction of actives in the whole test set:
\[\mathrm{EF}_{x\%} = \frac{A_{x\%} / N_{x\%}}{A / N}, \qquad N_{x\%} = \left\lfloor \frac{x}{100} N \right\rfloor,\]
where $N$ is the number of test compounds, $A$ is the number of actives among them,
$N_{x\%}$ is the number of compounds ranked in the top $x\%$ by predicted score,
and $A_{x\%}$ is the number of actives among those $N_{x\%}$ compounds.
We report $\mathrm{EF}_{x\%}$ for $x \in \{10, 5, 1, 0.5\}$.

\subsection{Randomly initialised GINs}\label{sec:random_init_gins}

Graph neural networks, such as GINs, have been previously shown to perform well 
on graph and node classification tasks without any training~\cite{sunR-2022-DoesGNNPretrainingHelpMolecular}.
To validate the effectiveness of pre-training GINs to predict ECFP$4_{2048}$, we conducted an ablation study 
comparing the performance of pre-trained GINs (PT-GIN) against a randomly initialised GIN (Random GIN) with the same architecture
on Biogen regression tasks.

We implemented Random GINs with two different substructure vocabularies, each with the same seed:
one with the same Sort \& Slice atomic environment vocabulary as the pre-trained GIN based on
the QMugs dataset (Random GIN$_{\text{QMugs}}$), and one with Sort \& Slice vocabularies based on
benchmark training sets, generated from each training set (Random GIN$_{\text{Training}}$).
Random GINs were initialised with Xavier normal initialisation,
following the same initialisation scheme as the PT-GIN prior to pre-training.
Random GINs used the same readout and pooling approach as PT-GINs:
sum pooling over nodesand concatenating the outputs of each message passing layer.
Radii and vocabulary sizes were
the same as those used for evaluation against fixed fingerprinting methods
and are summarised in Supplementary Information Table~19.
Evaluated PT-GINs were pre-trained on the filtered QMugs subset with a similarity threshold of 0.5.

We used LightGBM for all benchmark tasks.
LightGBM hyperparameter optimisation was performed on a single $5$-fold CV repeat for 
each GIN method and each task, as described above.
We used $20$ $5$-fold CV repeats for evaluation.
The results are shown in Fig.~\ref{fig:random_gin} and Supplementary Information 
Figs.~61--65.

\subsection{Substructure importance}\label{sec:importance}

As S\&S maps substructures to bits and PT-GIN maps substructures to tokens,
it is possible to measure the relative importance of substructures for both methods.

For ECFP$_{\mathrm{S\&S}}$, substructure importance was determined by measuring
change in performance upon permutation as described previously~\cite{breimanL-2001-RandomForests},
using $\mathrm{R}^2$ for regression and $\mathrm{AUCPR}$ for classification.

For PT-GIN models, substructure importance was determined by iteratively permuting 
the embedding for each substructure token (Supplementary Information Algorithms~3 and~4).
To prevent token prevalence within each graph from influencing the measurement,
a fixed number of permutations, equal to the node embedding dimension, 
was distributed across every instance of a token in a graph; 
an alternative approach would be to permute a maximum of one node in each graph. 
By permuting substructure embeddings of molecules in the test set,
we had a heuristic for substructure importance that we could compare directly to S\&S.

Feature importance analysis was performed with LightGBM models over one of the 
$5$-fold cross-validated repeats used in benchmark evaluation, 
and repeated five times on each train-test split, totalling $25$ iterations. 
See Supplementary Information Figs.~51--56 for substructure importance results and analysis.

\subsection{Computational resources}

All experiments were performed on desktop hardware with a single NVIDIA RTX A2000 12GB GPU, an Intel Core i5-13500 (13th generation) CPU, and 32GB RAM.

\backmatter

\bmhead{Supplementary Information}

Additional discussion, figures, tables, and algorithms are provided in the Supplementary Information.

\bmhead{Funding}

This work was supported by the Engineering and Physical Sciences Research Council and Lhasa Ltd [Grant number EP/W524311/1].

\bmhead{Competing interests}

C.M.D. discloses membership of the Scientific Advisory Board of Fusion Antibodies plc and Astex, and is a founder of DaltonTx. All other authors declare no competing interests.

\bmhead{Data availability}

The data used in this work, including source data for all figures and tables, are available on Zenodo at: \url{https://doi.org/10.5281/zenodo.24693872}.

\bmhead{Code availability}

The code for this work is available on GitHub at \url{https://github.com/oxpig/topological-pretraining}.
The version used in this work is archived on Zenodo~\cite{moneykyrleS-2026-ImprovingGNNsQSARPretrainingECFPCode} (\url{https://doi.org/10.5281/zenodo.24707481}).

\bmhead{Author contributions}

S.M-K. and M.D. conceptualised the study.
S.M-K. developed the methodology, performed the experiments, and analysed the results. 
G.M.M., C.M.D., T.H., and S.W. supervised the project.
All authors discussed the results and contributed to the interpretation.
S.M-K. wrote the manuscript with editorial input from all authors.

\clearpage

\bibliography{main_bibliography} %\

\clearpage

\section*{Figures}\label{sec:figures}

\input{supplementary_appendix}

\end{document}

%% file: supplementary_appendix.tex
% Supplementary Information, appended to the main text for the arXiv version.
% Body copied from supplementary.tex; its standalone preamble and separate
% bibliography are replaced by the settings below (references are merged
% into the main reference list).
\clearpage
\numbered
\setcounter{section}{0}
\renewcommand{\thesection}{Supplementary Information \arabic{section}}
\setcounter{figure}{0}
\setcounter{table}{0}
\setcounter{equation}{0}
\setcounter{algorithm}{0}
\renewcommand{\figurename}{Supplementary Information Fig.}
\renewcommand{\tablename}{Supplementary Information Table}
% Distinct hyperlink anchors, since the counters restart from 1
\renewcommand{\theHsection}{SI.\arabic{section}}
\renewcommand{\theHfigure}{SI.\arabic{figure}}
\renewcommand{\theHtable}{SI.\arabic{table}}
\renewcommand{\theHequation}{SI.\arabic{equation}}
\renewcommand{\theHalgorithm}{SI.\arabic{algorithm}}
\algnewcommand{\LineComment}[1]{\State \(\triangleright\) #1}
\algrenewcommand\algorithmicrequire{\textbf{Input:}}
\let\cleardoublepage\clearpage
\begin{center}
    \begin{Large}
    \textbf{Supplementary Information for: On Improving Graph Neural Networks for QSAR by Pre-training on Extended-Connectivity Fingerprints}
    \end{Large}
\end{center}

\newpage
\section{Supplementary results}
\subsection{Metric violin plots}

\begin{figure}[!htbp]
    \centering
    \includegraphics[width=\textwidth, height=0.7\textheight, keepaspectratio]{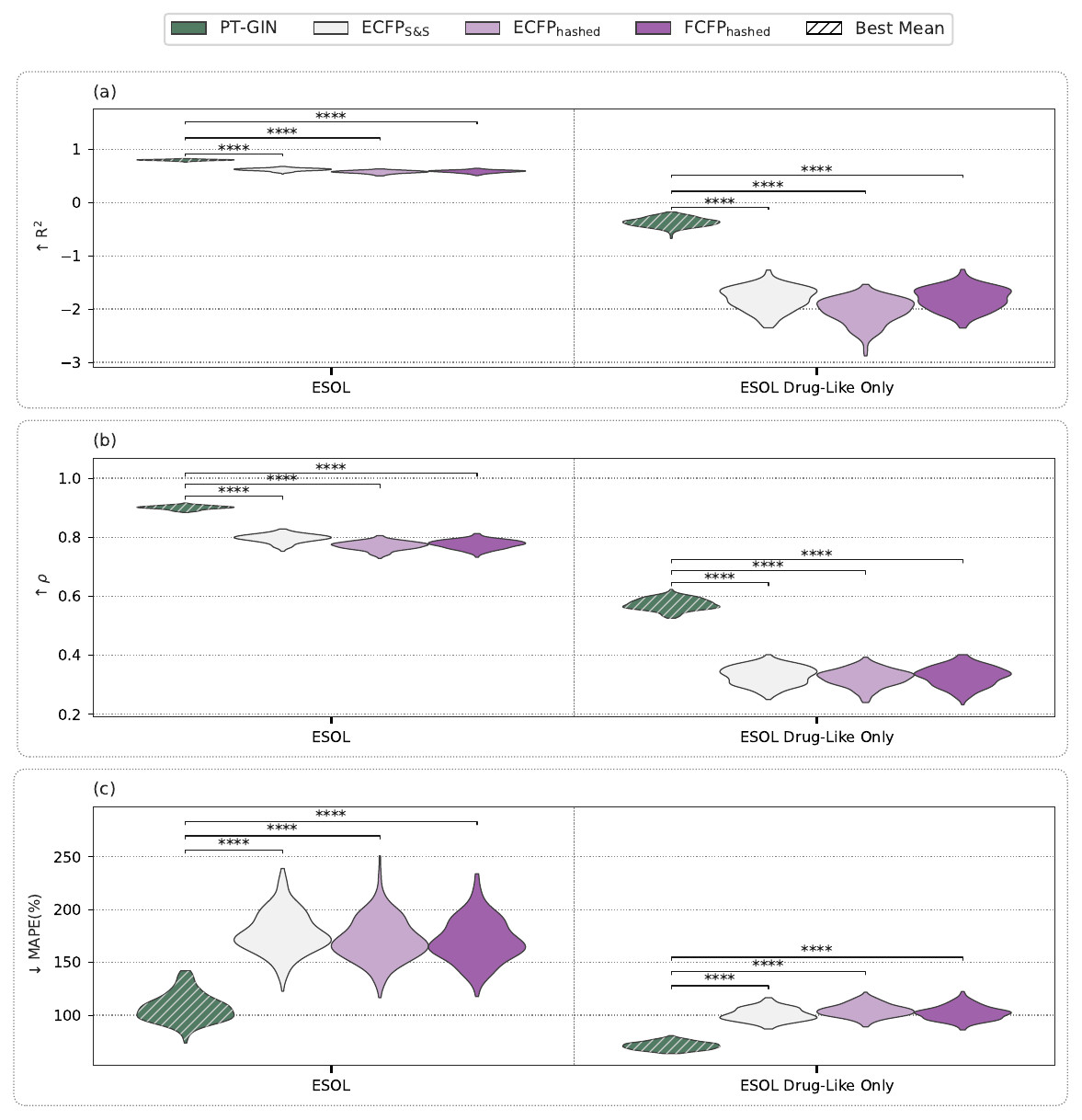}
    \caption{
        Performance metrics on ESOL aqueous solubility with all molecules (ESOL) 
        and only molecules with a drug-like solubility (ESOL Drug-Like Only) (\mbox{$0 < \mathrm{log_{10}(S (\mu M))} < 3$}). 
        The metrics shown are (a) coefficient of determination ($\uparrow\mathrm{R}^2$), 
        (b) Pearson correlation coefficient ($\uparrow\rho$), and (c) mean absolute percentage error ($\downarrow$MAPE (\%)). 
        Featurisation methods shown are ECFP-pre-trained GINs (PT-GIN),
        hashed Extended-Connectivity Fingerprints (ECFP$_{\mathrm{hashed}}$), 
        hashed Functional-Class Fingerprints (FCFP$_{\mathrm{hashed}}$), 
        and ECFPs vectorised via Sort \& Slice (ECFP$_{\mathrm{S\&S}}$). 
        For each task and metric, the approach with the best mean is hatched.
        Statistical significances, determined by Bonferroni-corrected two-sided Wilcoxon signed-rank tests ($n=199$, $\alpha=0.05$), 
        between the best method by mean and all other approaches are shown with asterisks; 
        ns (not significant) is $p \geq 0.05$;
        * is $0.01 \leq p < 0.05$;
        ** is $0.001 \leq p < 0.01$;
        *** is $0.0001 \leq p < 0.001$;
        and **** is $p < 0.0001$. 
        Violins are normalised such that each violin has the same maximum width.
    }
    \label{fig:esol_metrics}
\end{figure}

\begin{figure}[!htbp]
    \centering
    \includegraphics[width=\textwidth,keepaspectratio]{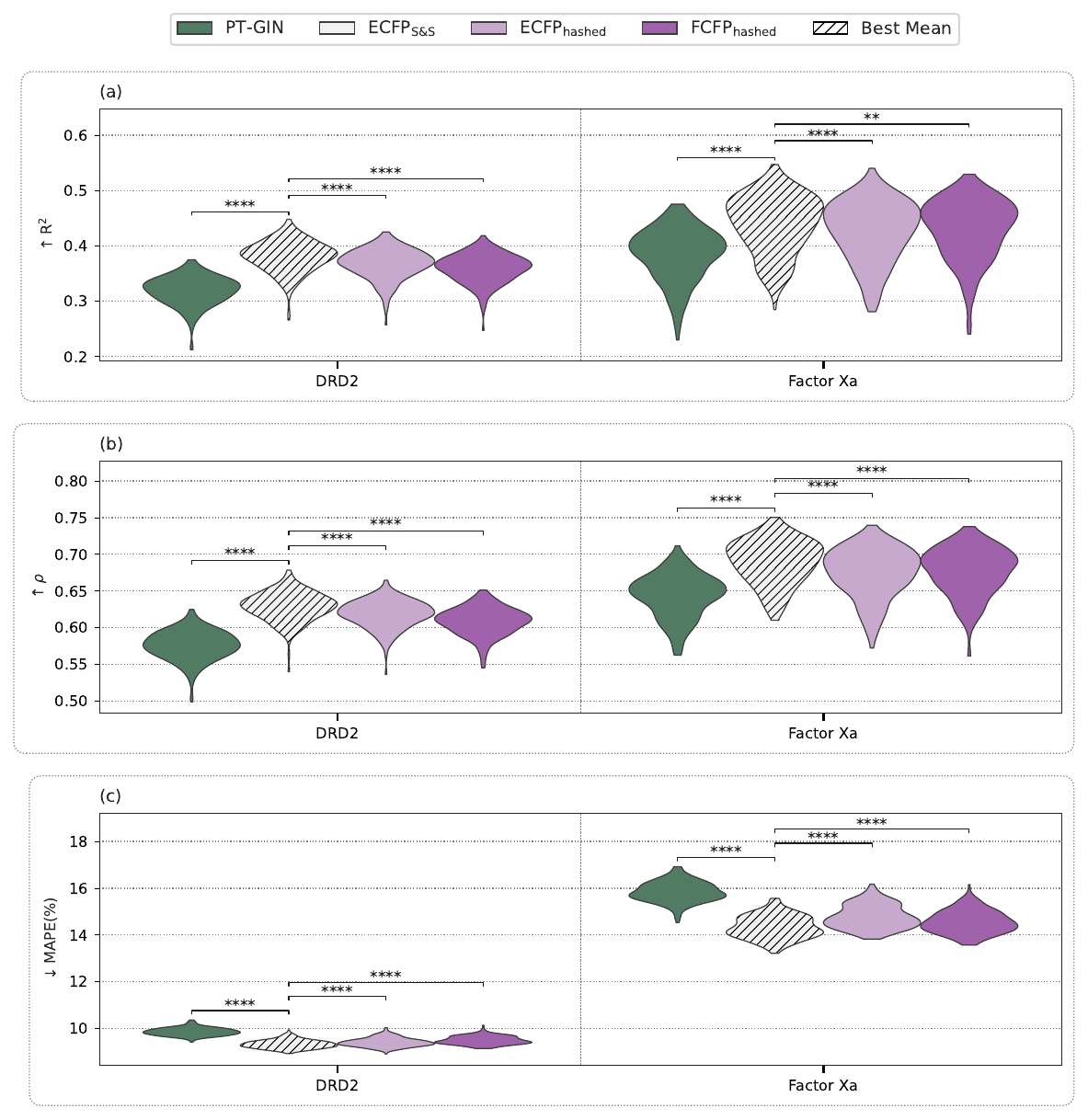}
    \caption{
        Performance metrics on ChEMBL binding affinity tasks. 
        Datasets shown are DRD2 and Factor Xa.
        The metrics shown are (a) coefficient of determination ($\uparrow\mathrm{R}^2$), 
        (b) Pearson correlation coefficient ($\uparrow\rho$), and (c) mean absolute percentage error ($\downarrow$MAPE (\%)). 
        Featurisation methods shown are ECFP-pre-trained GINs (PT-GIN),
        hashed Extended-Connectivity Fingerprints (ECFP$_{\mathrm{hashed}}$), 
        hashed Functional-Class Fingerprints (FCFP$_{\mathrm{hashed}}$), 
        and ECFPs vectorised via Sort \& Slice (ECFP$_{\mathrm{S\&S}}$). 
        For each task and metric, the approach with the best mean is hatched.
        Statistical significances, determined by Bonferroni-corrected two-sided Wilcoxon signed-rank tests ($n=199$, $\alpha=0.05$), 
        between the best method by mean and all other approaches are shown with asterisks; 
        ns (not significant) is $p \geq 0.05$;
        * is $0.01 \leq p < 0.05$;
        ** is $0.001 \leq p < 0.01$;
        *** is $0.0001 \leq p < 0.001$;
        and **** is $p < 0.0001$. 
        Violins are normalised such that each violin has the same maximum width. 
    }
    \label{fig:chembl_metrics}
\end{figure}

\begin{figure}[!htbp]
    \centering
    \includegraphics[width=\textwidth,keepaspectratio]{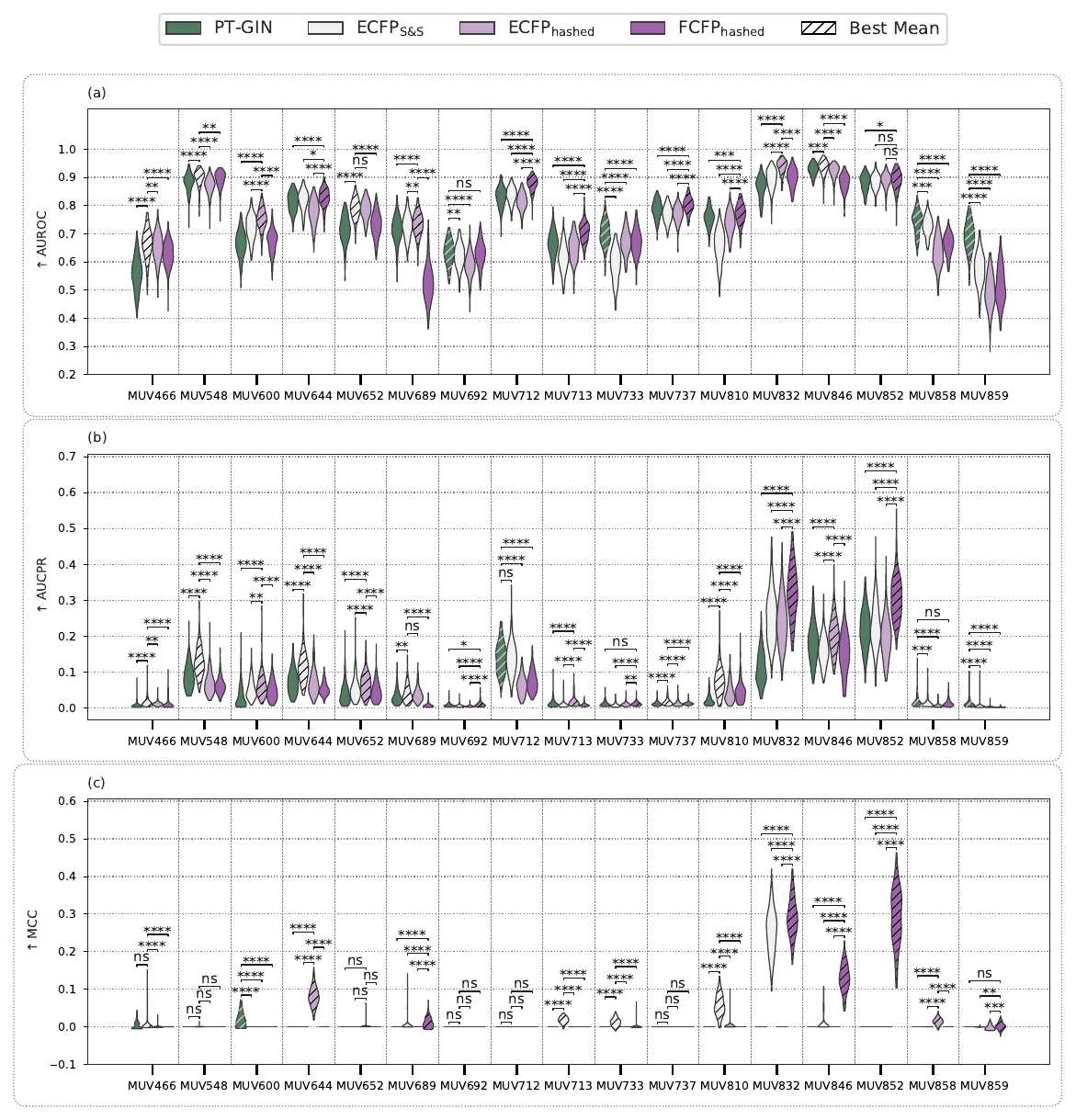}
    \caption{
        Performance metrics on MUV virtual screening tasks. 
        The metrics shown are (a) area under the receiver operating characteristic curve ($\uparrow\mathrm{AUROC}$),
        (b) area under the precision-recall curve ($\uparrow\mathrm{AUCPR}$), 
        and (c) Matthews correlation coefficient ($\uparrow\mathrm{MCC}$).  
        Featurisation methods shown are ECFP-pre-trained GINs (PT-GIN),
        hashed Extended-Connectivity Fingerprints (ECFP$_{\mathrm{hashed}}$), 
        hashed Functional-Class Fingerprints (FCFP$_{\mathrm{hashed}}$), 
        and ECFPs vectorised via Sort \& Slice (ECFP$_{\mathrm{S\&S}}$). 
        For each task and metric, the approach with the best mean is hatched.
        Statistical significances, determined by Bonferroni-corrected two-sided Wilcoxon signed-rank tests ($n=199$, $\alpha=0.05$), 
        between the best method by mean and all other approaches are shown with asterisks; 
        ns (not significant) is $p \geq 0.05$;
        * is $0.01 \leq p < 0.05$;
        ** is $0.001 \leq p < 0.01$;
        *** is $0.0001 \leq p < 0.001$;
        and **** is $p < 0.0001$. 
        Violins are normalised such that each violin has the same maximum width. 
        }
    \label{fig:muv_metrics}
\end{figure}

\begin{figure}[!htbp]
    \centering
    \includegraphics[width=\textwidth,keepaspectratio]{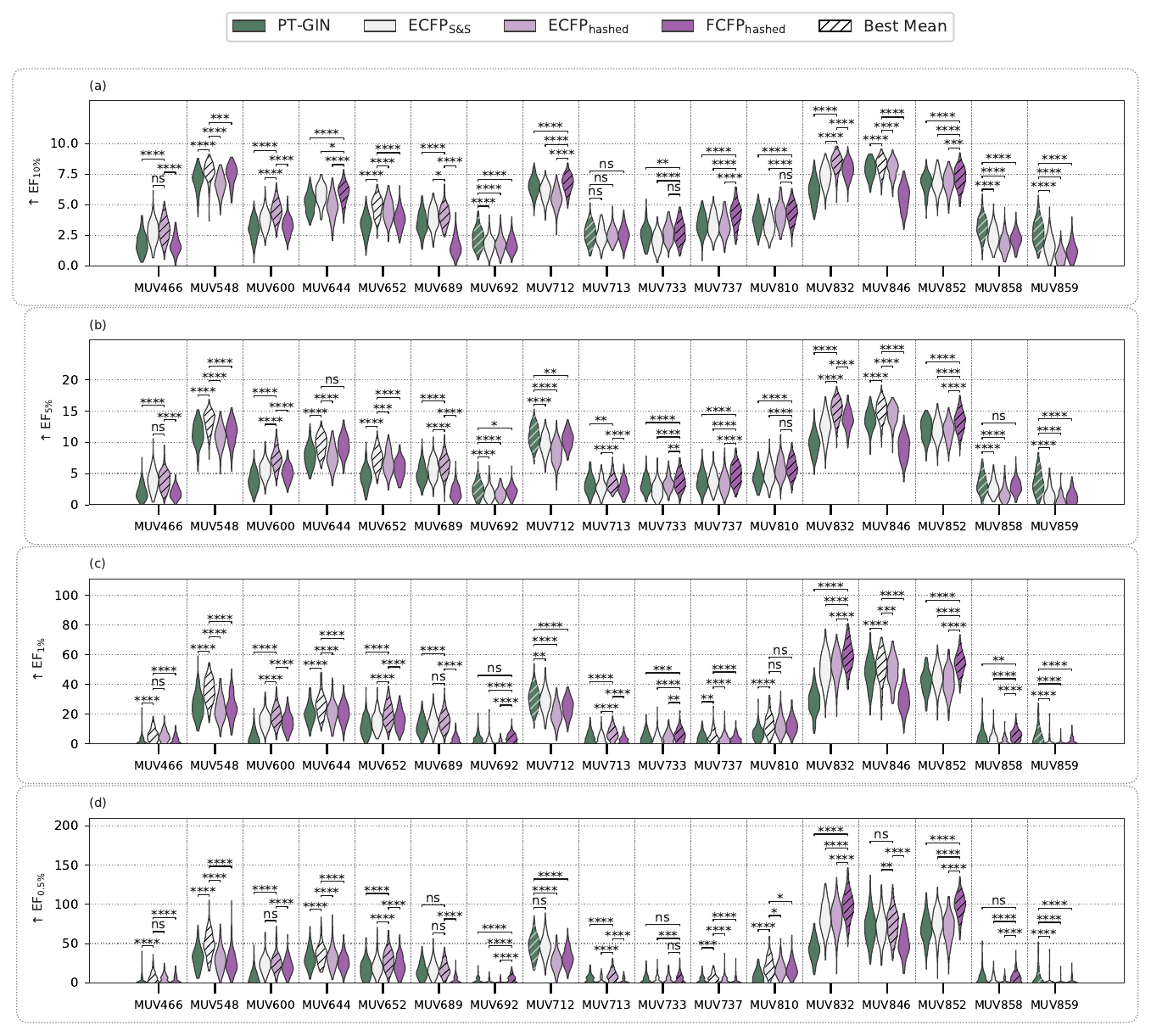}
    \caption{
        Enrichment factors at (a) $10\%$, (b) $5\%$, (c) $1\%$, and (d) $0.5\%$ on MUV virtual screening tasks.  
        Featurisation methods shown are ECFP-pre-trained GINs (PT-GIN),
        hashed Extended-Connectivity Fingerprints (ECFP$_{\mathrm{hashed}}$), 
        hashed Functional-Class Fingerprints (FCFP$_{\mathrm{hashed}}$), 
        and ECFPs vectorised via Sort \& Slice (ECFP$_{\mathrm{S\&S}}$). 
        For each task and metric, the approach with the best mean is hatched.
        Statistical significances, determined by Bonferroni-corrected two-sided Wilcoxon signed-rank tests ($n=199$, $\alpha=0.05$), 
        between the best method by mean and all other approaches are shown with asterisks; 
        ns (not significant) is $p \geq 0.05$;
        * is $0.01 \leq p < 0.05$;
        ** is $0.001 \leq p < 0.01$;
        *** is $0.0001 \leq p < 0.001$;
        and **** is $p < 0.0001$. 
        Violins are normalised such that each violin has the same maximum width. 
    }
    \label{fig:muv_ef}
\end{figure}

\begin{figure}[!htbp]
    \centering
    \includegraphics[width=\textwidth,keepaspectratio]{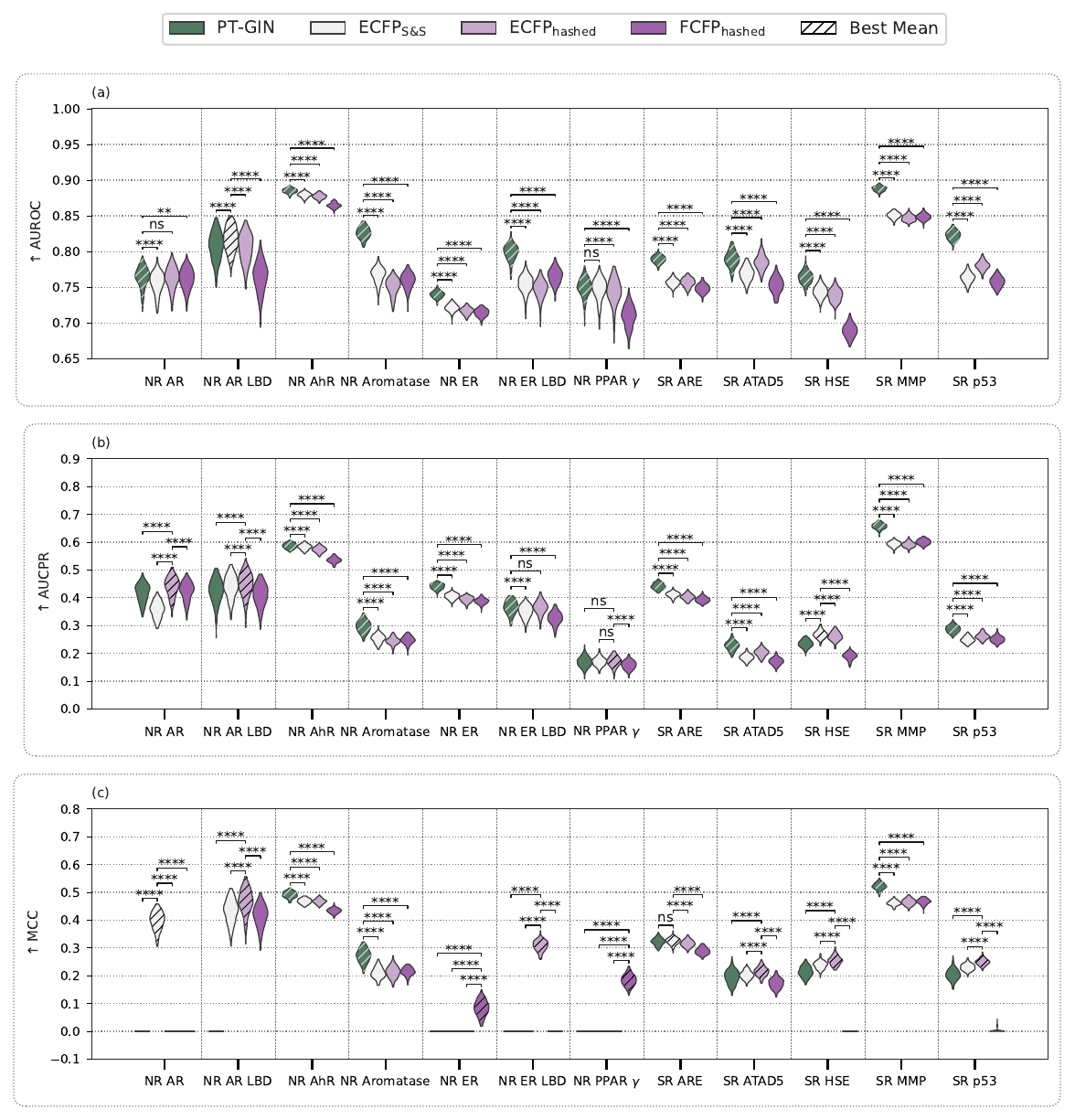}
    \caption{
        Performance metrics on Tox21 binary toxicity tasks. 
        The metrics shown are (a) area under the receiver operating characteristic curve ($\uparrow\mathrm{AUROC}$),
        (b) area under the precision-recall curve ($\uparrow\mathrm{AUCPR}$), 
        and (c) Matthews correlation coefficient ($\uparrow\mathrm{MCC}$).  
        Featurisation methods shown are ECFP-pre-trained GINs (PT-GIN),
        hashed Extended-Connectivity Fingerprints (ECFP$_{\mathrm{hashed}}$), 
        hashed Functional-Class Fingerprints (FCFP$_{\mathrm{hashed}}$), 
        and ECFPs vectorised via Sort \& Slice (ECFP$_{\mathrm{S\&S}}$). 
        For each task and metric, the approach with the best mean is hatched.
        Statistical significances, determined by Bonferroni-corrected two-sided Wilcoxon signed-rank tests ($n=199$, $\alpha=0.05$), 
        between the best method by mean and all other approaches are shown with asterisks; 
        ns (not significant) is $p \geq 0.05$;
        * is $0.01 \leq p < 0.05$;
        ** is $0.001 \leq p < 0.01$;
        *** is $0.0001 \leq p < 0.001$;
        and **** is $p < 0.0001$.
        Violins are normalised such that each violin has the same maximum width. 
    }
    \label{fig:tox21_metrics}
\end{figure}

\clearpage
\newpage

\subsection{Multiple comparisons plots}

\subsubsection{Biogen}

\begin{figure}[!htbp]
    \centering
    \includegraphics[width=\textwidth, height=0.7\textheight, keepaspectratio]{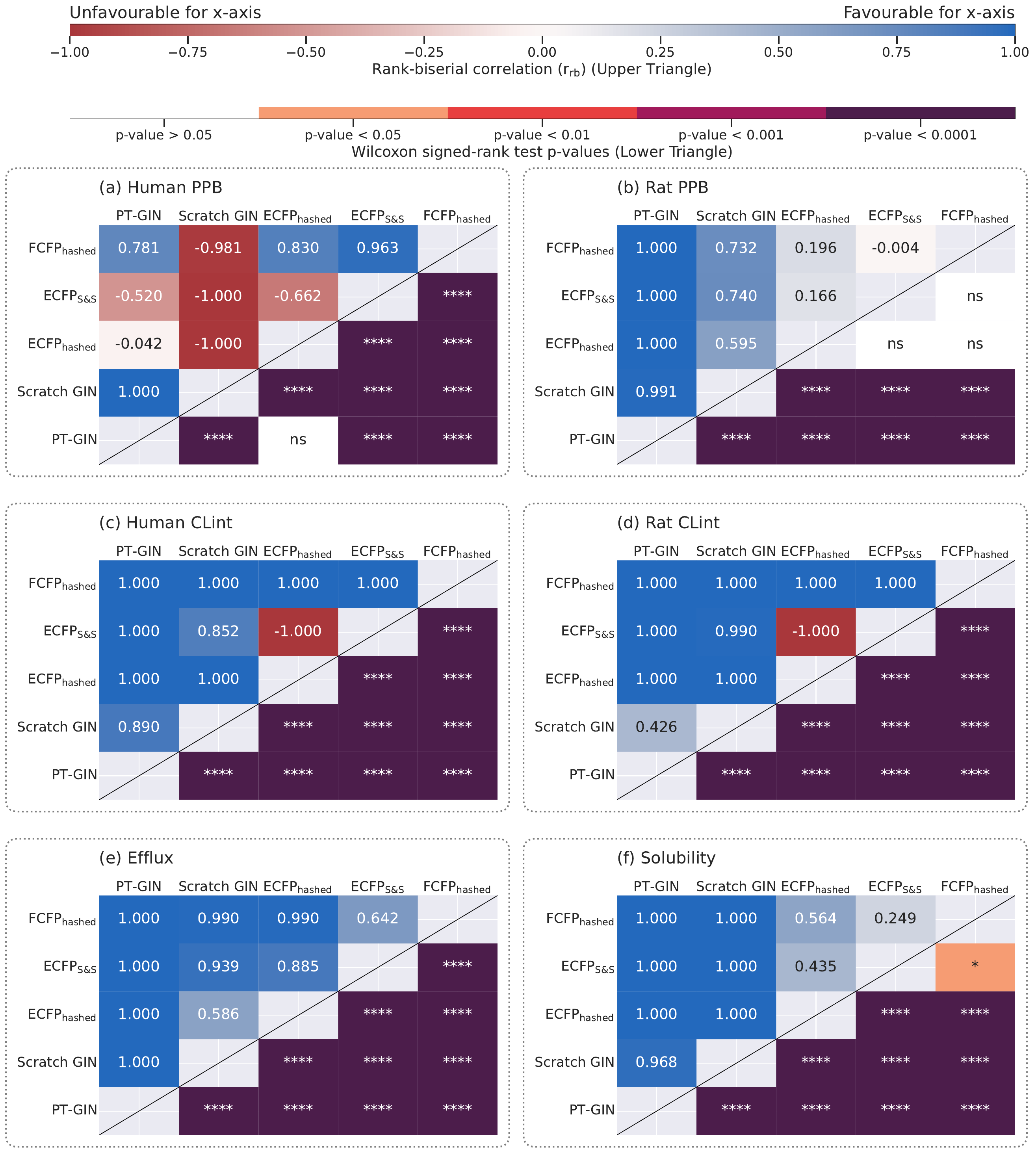}
    \caption{
        Rank-biserial correlation ($r_{rb}$) effect sizes (upper triangle) and Bonferroni-corrected two-sided Wilcoxon signed-rank test \textit{p}-values (lower triangle) of the Pearson correlation coefficient
        ($\uparrow \rho$) between methods on Biogen datasets.
        Tasks are (a) Human Plasma Protein Binding (PPB), (b) Rat PPB, 
        (c) Human Intrinsic Hepatic Clearance (CLint), (d) Rat CLint, (e) Efflux, (f) Solubility. 
        Featurisation methods shown are GINs trained from scratch (Scratch GIN), 
        ECFP-pre-trained GINs (PT-GIN), hashed Extended-Connectivity Fingerprints (ECFP$_{\mathrm{hashed}}$), 
        hashed Functional-Class Fingerprints (FCFP$_{\mathrm{hashed}}$), and 
        ECFPs vectorised via Sort \& Slice (ECFP$_{\mathrm{S\&S}}$).
        For $r_{rb}$ effect sizes (upper triangle): blue indicates the method on the \textit{x}-axis has a higher $\rho$ than 
        the method on the \textit{y}-axis; red indicates the opposite. 
        For Wilcoxon signed-rank test \textit{p}-values (lower triangle): 
        white indicates no significant difference and darker indicates a smaller \textit{p}-value ($n=199$, $\alpha=0.05$).
    }
    \label{fig:biogen_heatmap_pearson}
\end{figure}

\begin{figure}[!htbp]
    \centering
    \includegraphics[width=\textwidth, height=0.7\textheight, keepaspectratio]{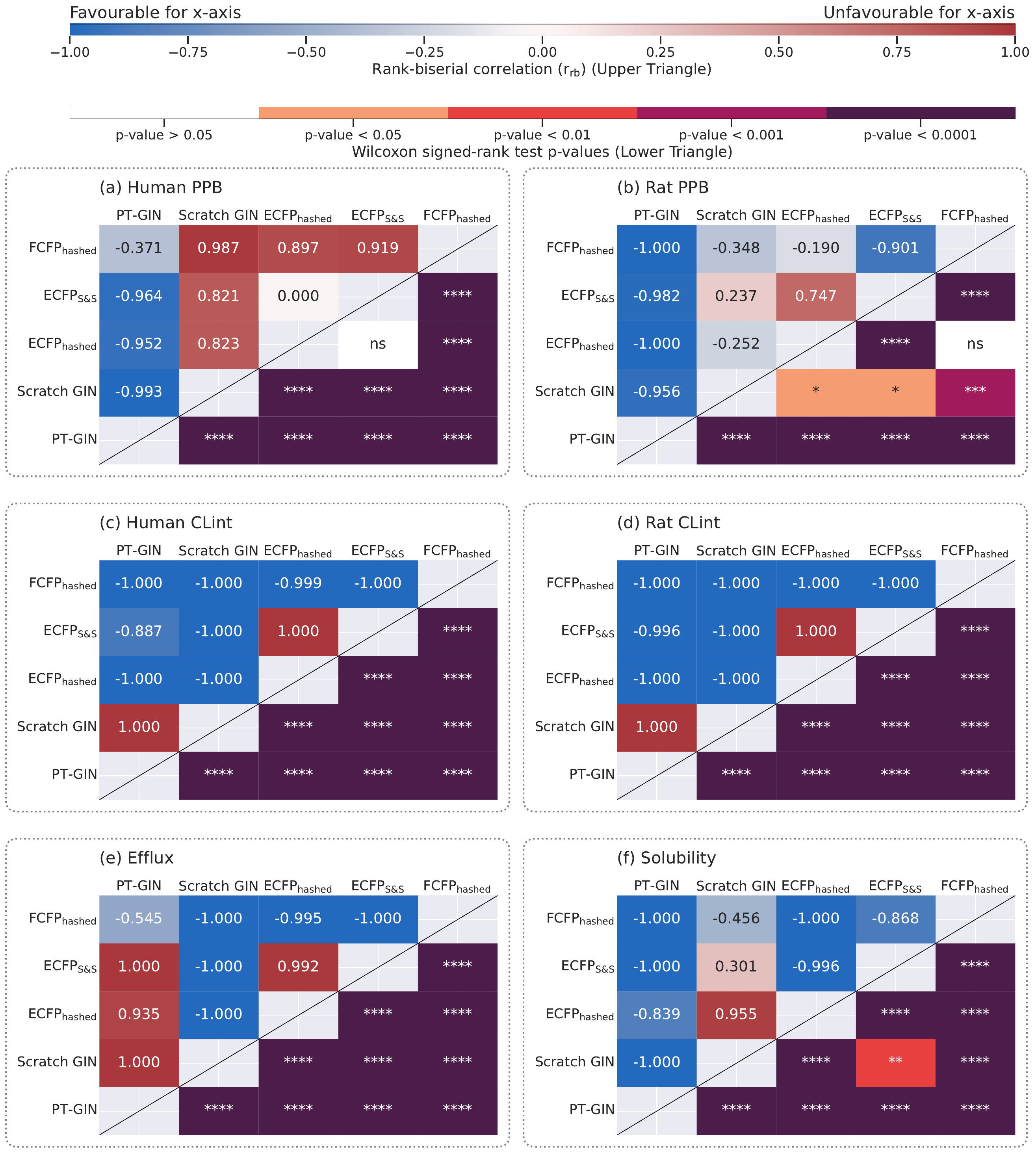}
    \caption{
        Rank-biserial correlation ($r_{rb}$) effect sizes (upper triangle) and Bonferroni-corrected two-sided Wilcoxon signed-rank test \textit{p}-values (lower triangle) of the mean absolute percentage error
        ($\downarrow\mathrm{MAPE}$) between methods on Biogen datasets.
        Tasks are (a) Human Plasma Protein Binding (PPB), (b) Rat PPB, 
        (c) Human Intrinsic Hepatic Clearance (CLint), (d) Rat CLint, (e) Efflux, (f) Solubility. 
        Featurisation methods shown are GINs trained from scratch (Scratch GIN), 
        ECFP-pre-trained GINs (PT-GIN), hashed Extended-Connectivity Fingerprints (ECFP$_{\mathrm{hashed}}$), 
        hashed Functional-Class Fingerprints (FCFP$_{\mathrm{hashed}}$), and 
        ECFPs vectorised via Sort \& Slice (ECFP$_{\mathrm{S\&S}}$).
        For $r_{rb}$ effect sizes (upper triangle): blue indicates the method on the \textit{x}-axis has a lower $\mathrm{MAPE}$ than 
        the method on the \textit{y}-axis; red indicates the opposite. 
        For Wilcoxon signed-rank test \textit{p}-values (lower triangle): 
        white indicates no significant difference and darker indicates a smaller \textit{p}-value ($n=199$, $\alpha=0.05$).
    }
    \label{fig:biogen_heatmap_mape}
\end{figure}

\clearpage
\newpage

\subsubsection{Biogen Tanimoto similarity filter}

\begin{figure}[!htbp]
    \centering
    \includegraphics[width=\textwidth, height=0.7\textheight, keepaspectratio]{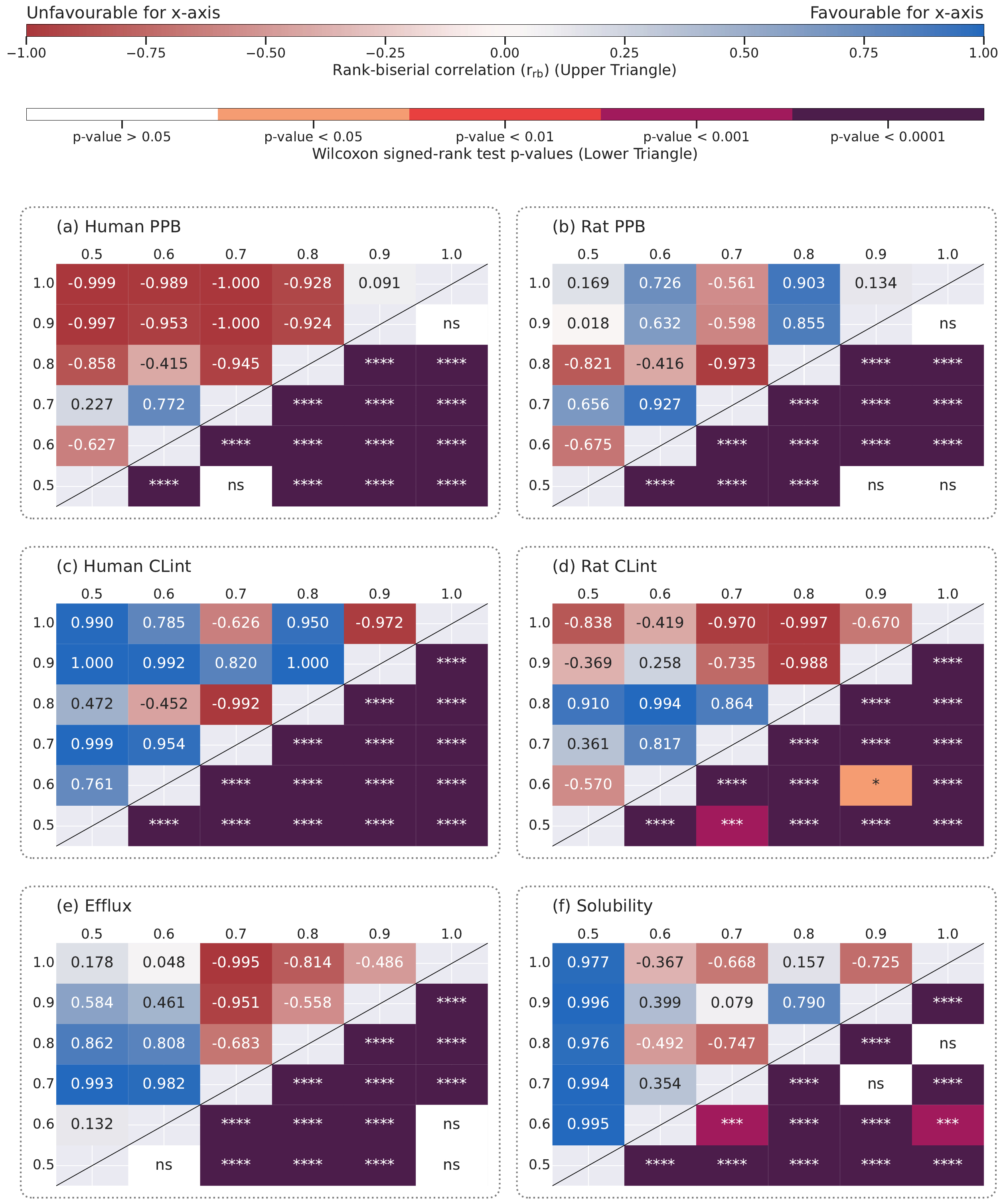}
    \caption{
        Rank-biserial correlation ($r_{rb}$) effect sizes (upper triangle) and Bonferroni-corrected two-sided Wilcoxon signed-rank test \textit{p}-values (lower triangle) of the Pearson correlation coefficient
        ($\uparrow \rho$) of PT-GIN models pre-trained on Tanimoto filtered QMugs subsets.
        Tasks are (a) Human Plasma Protein Binding (PPB), (b) Rat PPB, 
        (c) Human Intrinsic Hepatic Clearance (CLint), (d) Rat CLint, (e) Efflux, (f) Solubility. 
        Featurisation methods shown are GINs pre-trained (PT-GIN) to predict ECFP$4_{2048}$ on 
        QMugs with a Tanimoto filter of $0.5$, $0.6$, $0.7$, $0.8$, $0.9$, and $1.0$.
        All filter subsets were equal in size.
        For $r_{rb}$ effect sizes (upper triangle): blue indicates the method on the \textit{x}-axis has a higher $\rho$ than 
        the method on the \textit{y}-axis; red indicates the opposite. 
        For Wilcoxon signed-rank test \textit{p}-values (lower triangle): 
        white indicates no significant difference and darker indicates a smaller \textit{p}-value ($n=199$, $\alpha=0.05$).
    }
    \label{fig:tanimoto_heatmap_pearson}
\end{figure}

\begin{figure}[!htbp]
    \centering
    \includegraphics[width=\textwidth, height=0.7\textheight, keepaspectratio]{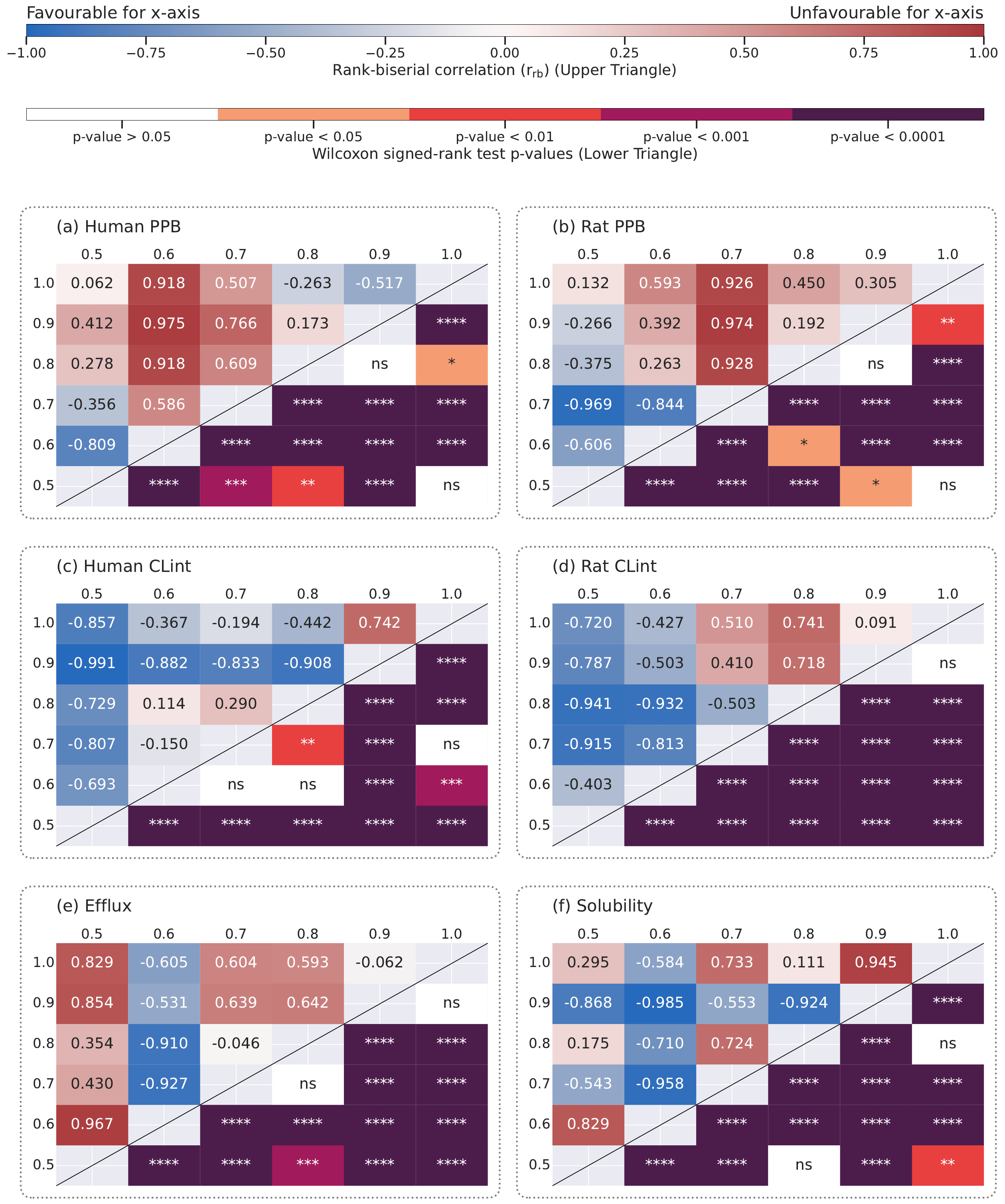}
    \caption{
        Rank-biserial correlation ($r_{rb}$) effect sizes (upper triangle) and Bonferroni-corrected two-sided Wilcoxon signed-rank test \textit{p}-values (lower triangle) of the mean absolute percentage error
        ($\downarrow\mathrm{MAPE}$) of PT-GIN models pre-trained on Tanimoto filtered QMugs subsets.
        Tasks are (a) Human Plasma Protein Binding (PPB), (b) Rat PPB, 
        (c) Human Intrinsic Hepatic Clearance (CLint), (d) Rat CLint, (e) Efflux, (f) Solubility. 
        Featurisation methods shown are GINs pre-trained (PT-GIN) to predict ECFP$4_{2048}$ on 
        QMugs with a Tanimoto filter of $0.5$, $0.6$, $0.7$, $0.8$, $0.9$, and $1.0$.
        All filter subsets were equal in size.
        For $r_{rb}$ effect sizes (upper triangle): blue indicates the method on the \textit{x}-axis has a lower $\mathrm{MAPE}$ than 
        the method on the \textit{y}-axis; red indicates the opposite. 
        For Wilcoxon signed-rank test \textit{p}-values (lower triangle): 
        white indicates no significant difference and darker indicates a smaller \textit{p}-value ($n=199$, $\alpha=0.05$).
    }
    \label{fig:tanimoto_heatmap_mape}
\end{figure}

\clearpage
\newpage
\subsubsection{ChEMBL}

\begin{figure}[!htbp]
    \centering
    \includegraphics[scale=0.32]{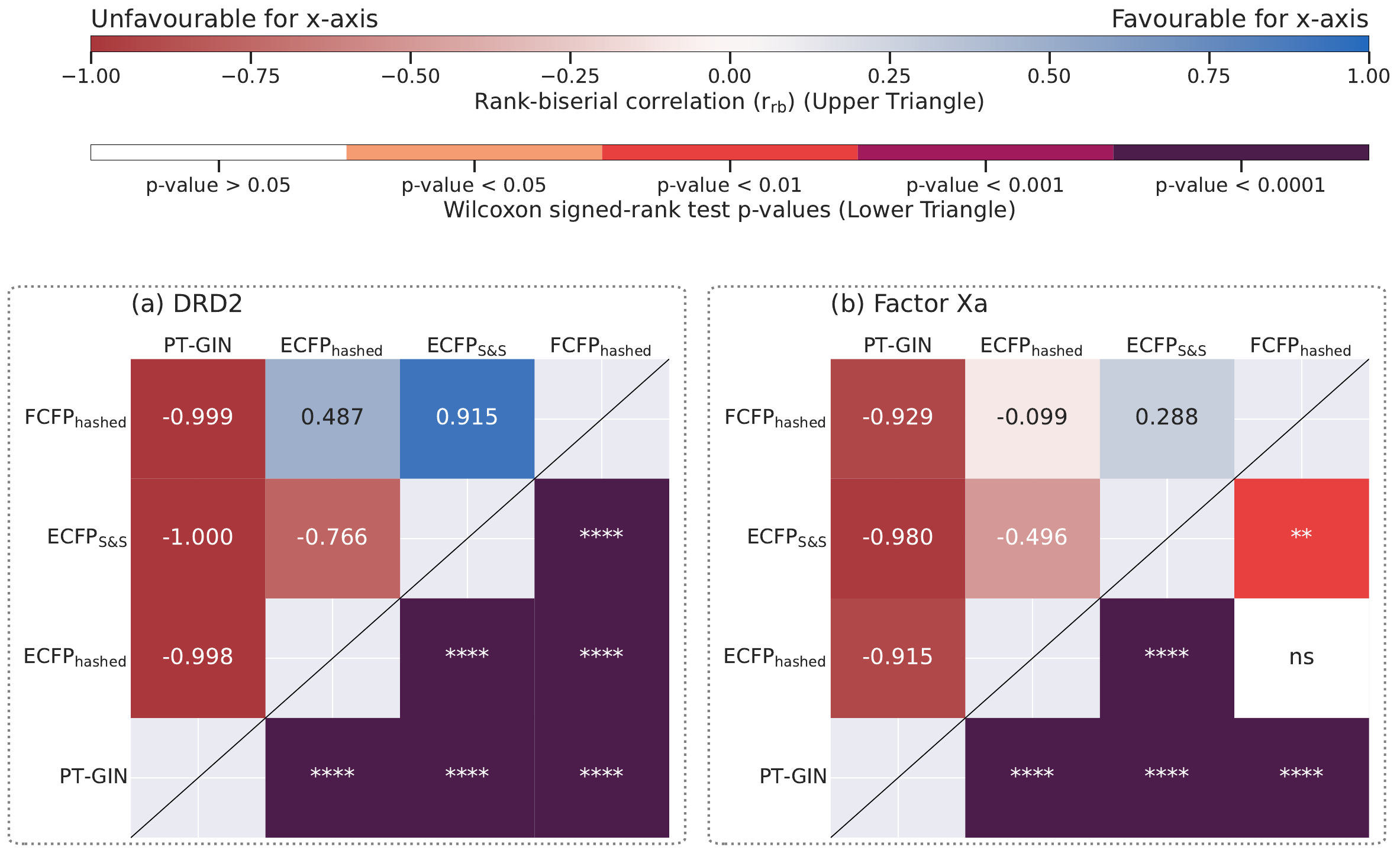}
    \caption{
        Rank-biserial correlation ($r_{rb}$) effect sizes (upper triangle) and Bonferroni-corrected two-sided Wilcoxon signed-rank test \textit{p}-values (lower triangle) of the coefficient of determination 
        ($\uparrow \mathrm{R}^2$) between methods on ChEMBL datasets.
        Tasks are (a) DRD2, (b) Factor Xa binding affinity prediction.
        Featurisation methods shown are 
        ECFP-pre-trained GINs (PT-GIN), hashed Extended-Connectivity Fingerprints (ECFP$_{\mathrm{hashed}}$), 
        hashed Functional-Class Fingerprints (FCFP$_{\mathrm{hashed}}$), and 
        ECFPs vectorised via Sort \& Slice (ECFP$_{\mathrm{S\&S}}$).
        For $r_{rb}$ effect sizes (upper triangle): blue indicates the method on the \textit{x}-axis has a higher $\mathrm{R}^2$ than 
        the method on the \textit{y}-axis; red indicates the opposite. 
        For Wilcoxon signed-rank test \textit{p}-values (lower triangle): 
        white indicates no significant difference and darker indicates a smaller \textit{p}-value ($n=199$, $\alpha=0.05$).
    }
    \label{fig:chembl_heatmap_r2}
\end{figure}

\begin{figure}[!htbp]
    \centering
    \includegraphics[scale=0.32]{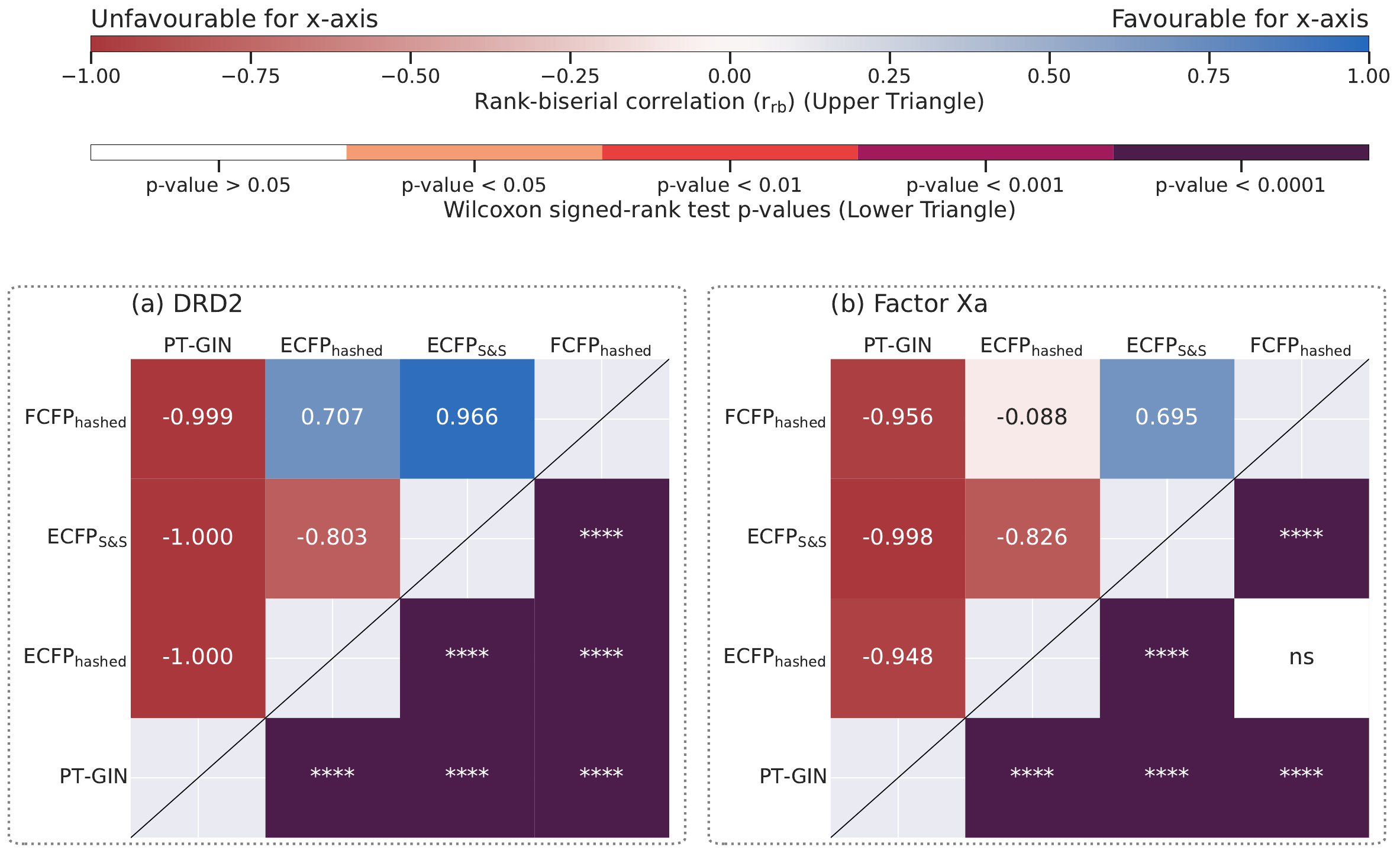}
    \caption{
        Rank-biserial correlation ($r_{rb}$) effect sizes (upper triangle) and Bonferroni-corrected two-sided Wilcoxon signed-rank test \textit{p}-values (lower triangle) of the Pearson correlation coefficient
        ($\uparrow \rho$) between methods on ChEMBL datasets.
        Tasks are (a) DRD2, (b) Factor Xa binding affinity prediction.
        Featurisation methods shown are 
        ECFP-pre-trained GINs (PT-GIN), hashed Extended-Connectivity Fingerprints (ECFP$_{\mathrm{hashed}}$), 
        hashed Functional-Class Fingerprints (FCFP$_{\mathrm{hashed}}$), and 
        ECFPs vectorised via Sort \& Slice (ECFP$_{\mathrm{S\&S}}$).
        For $r_{rb}$ effect sizes (upper triangle): blue indicates the method on the \textit{x}-axis has a higher $\rho$ than 
        the method on the \textit{y}-axis; red indicates the opposite. 
        For Wilcoxon signed-rank test \textit{p}-values (lower triangle): 
        white indicates no significant difference and darker indicates a smaller \textit{p}-value ($n=199$, $\alpha=0.05$).
    }
    \label{fig:chembl_heatmap_pearson}
\end{figure}

\begin{figure}[!htbp]
    \centering
    \includegraphics[scale=0.32]{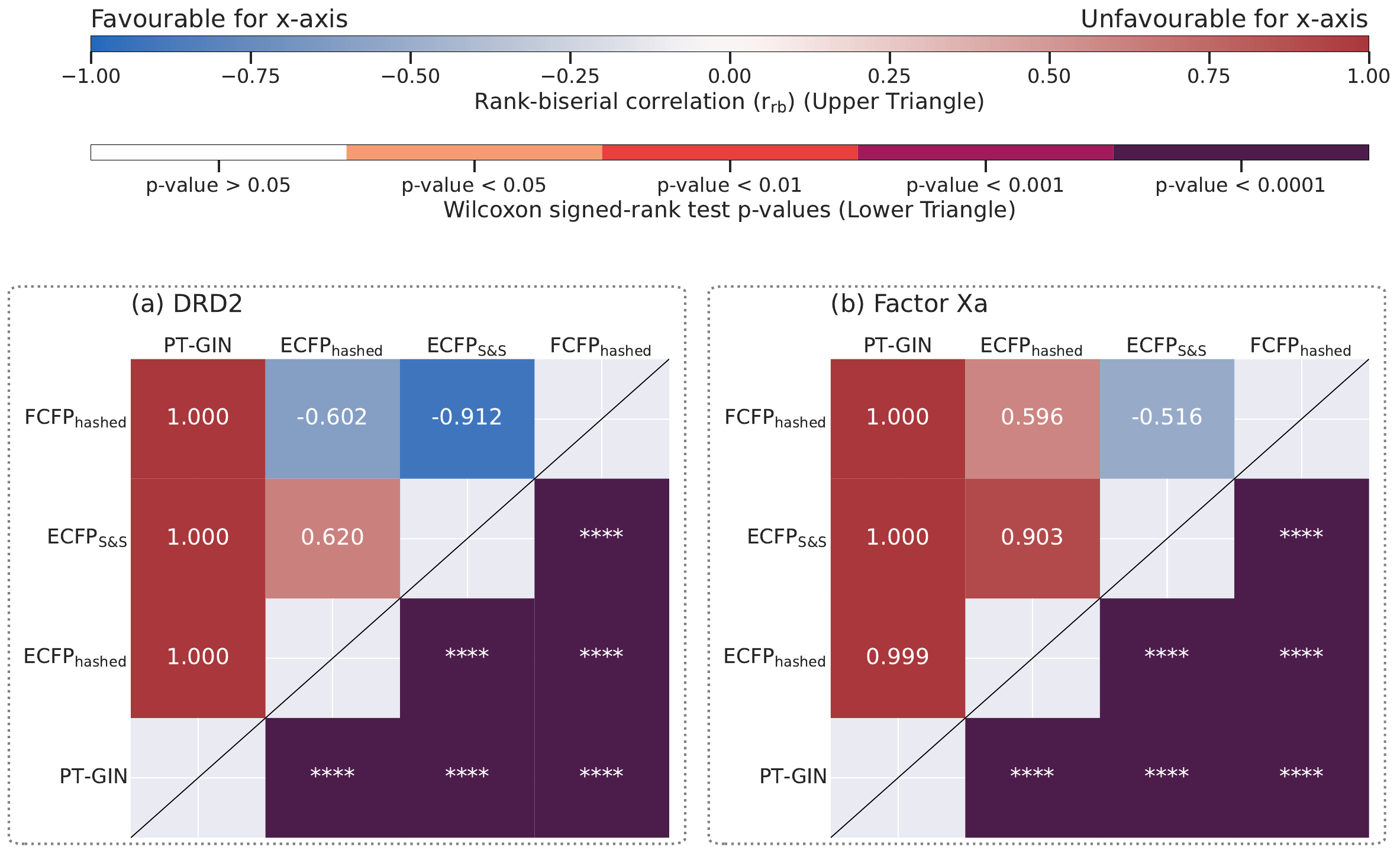}
    \caption{
        Rank-biserial correlation ($r_{rb}$) effect sizes (upper triangle) and Bonferroni-corrected two-sided Wilcoxon signed-rank test \textit{p}-values (lower triangle) of the mean absolute percentage error
        ($\downarrow\mathrm{MAPE}$) between methods on ChEMBL datasets.
        Tasks are (a) DRD2, (b) Factor Xa binding affinity prediction.
        Featurisation methods shown are 
        ECFP-pre-trained GINs (PT-GIN), hashed Extended-Connectivity Fingerprints (ECFP$_{\mathrm{hashed}}$), 
        hashed Functional-Class Fingerprints (FCFP$_{\mathrm{hashed}}$), and 
        ECFPs vectorised via Sort \& Slice (ECFP$_{\mathrm{S\&S}}$).
        For $r_{rb}$ effect sizes (upper triangle): blue indicates the method on the \textit{x}-axis has a lower $\mathrm{MAPE}$ than 
        the method on the \textit{y}-axis; red indicates the opposite. 
        For Wilcoxon signed-rank test \textit{p}-values (lower triangle): 
        white indicates no significant difference and darker indicates a smaller \textit{p}-value ($n=199$, $\alpha=0.05$).
    }
    \label{fig:chembl_heatmap_mape}
\end{figure}

\clearpage
\newpage
\subsubsection{MoleculeNet regression}

\begin{figure}[!htbp]
    \centering
    \includegraphics[width=\textwidth,keepaspectratio]{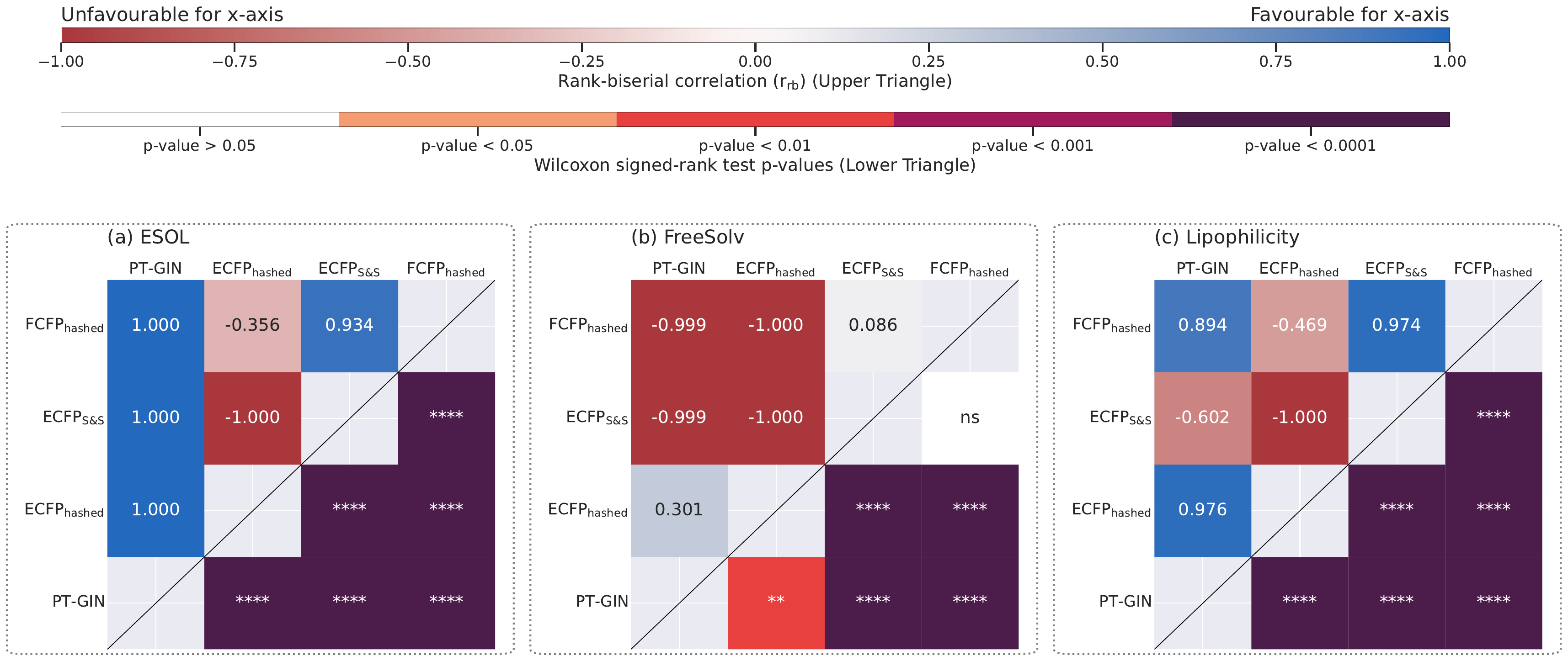}
    \caption{
        Rank-biserial correlation ($r_{rb}$) effect sizes (upper triangle)
        and Bonferroni-corrected two-sided Wilcoxon signed-rank test \textit{p}-values (lower triangle)
        comparison of the coefficient of determination ($\mathrm{R}^2$) between methods on MoleculeNet regression datasets.
        Datasets shown are (a) ESOL, (b) FreeSolv, and (c) Lipophilicity.
        Featurisation methods shown are 
        ECFP-pre-trained GINs (PT-GIN), hashed Extended-Connectivity Fingerprints (ECFP$_{\mathrm{hashed}}$), 
        hashed Functional-Class Fingerprints (FCFP$_{\mathrm{hashed}}$), and 
        ECFPs vectorised via Sort \& Slice (ECFP$_{\mathrm{S\&S}}$).
        For $r_{rb}$ effect sizes (upper triangle): blue indicates the method on the \textit{x}-axis has a higher $\mathrm{R}^2$ than 
        the method on the \textit{y}-axis; red indicates the opposite. 
        For Wilcoxon signed-rank test \textit{p}-values (lower triangle): 
        white indicates no significant difference and darker indicates a smaller \textit{p}-value ($n=199$, $\alpha=0.05$).
    }
    \label{fig:molnet_heatmap_r2}
\end{figure}

 \begin{figure}[!htbp]
    \centering
    \includegraphics[width=\textwidth, height=0.7\textheight, keepaspectratio]{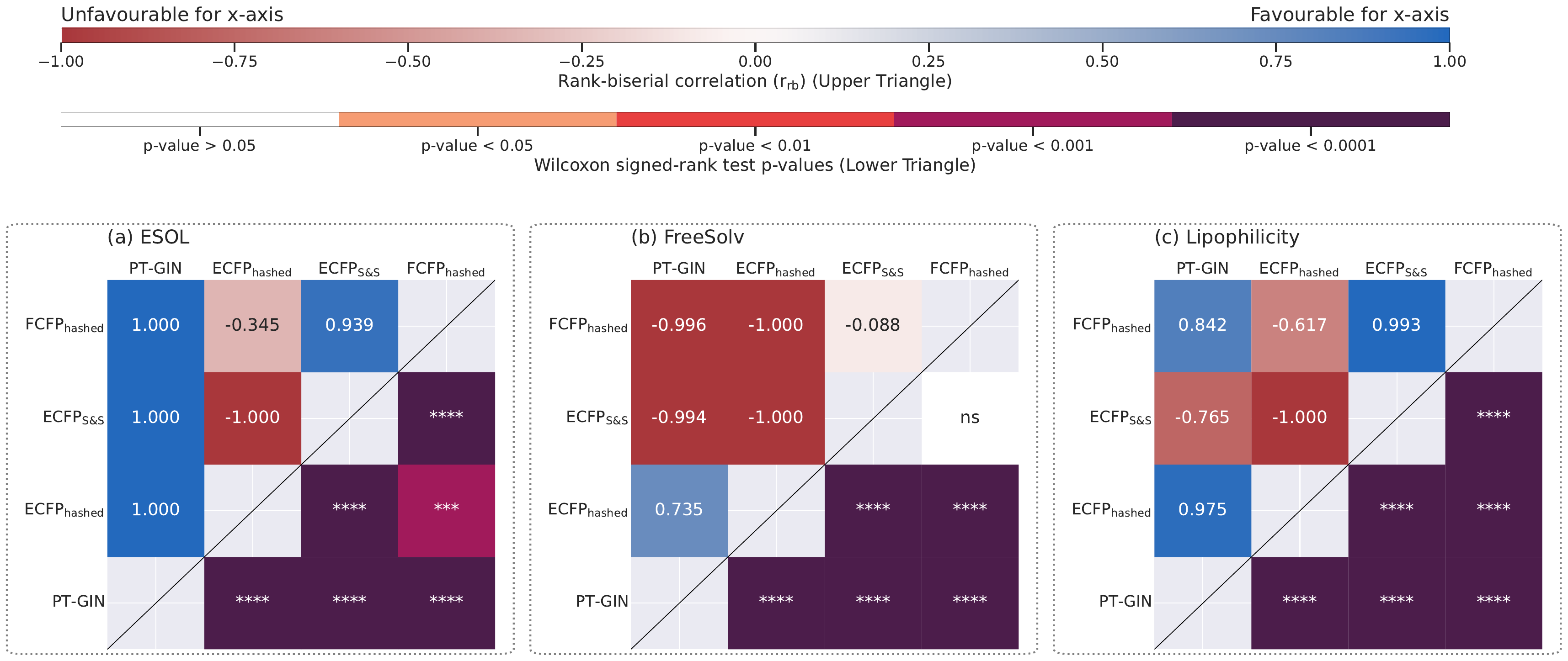}
    \caption{
        Rank-biserial correlation ($r_{rb}$) effect sizes (upper triangle) and Bonferroni-corrected two-sided Wilcoxon signed-rank test \textit{p}-values (lower triangle) of the Pearson correlation coefficient
        ($\uparrow \rho$) between methods on MoleculeNet regression datasets.
        Datasets shown are (a) ESOL, (b) FreeSolv, and (c) Lipophilicity.
        Featurisation methods shown are 
        ECFP-pre-trained GINs (PT-GIN), hashed Extended-Connectivity Fingerprints (ECFP$_{\mathrm{hashed}}$), 
        hashed Functional-Class Fingerprints (FCFP$_{\mathrm{hashed}}$), and 
        ECFPs vectorised via Sort \& Slice (ECFP$_{\mathrm{S\&S}}$).
        For $r_{rb}$ effect sizes (upper triangle): blue indicates the method on the \textit{x}-axis has a higher $\rho$ than 
        the method on the \textit{y}-axis; red indicates the opposite. 
        For Wilcoxon signed-rank test \textit{p}-values (lower triangle): 
        white indicates no significant difference and darker indicates a smaller \textit{p}-value ($n=199$, $\alpha=0.05$).
    }
    \label{fig:molnet_heatmap_pearson}
\end{figure}

\begin{figure}[!htbp]
    \centering
    \includegraphics[width=\textwidth, height=0.7\textheight, keepaspectratio]{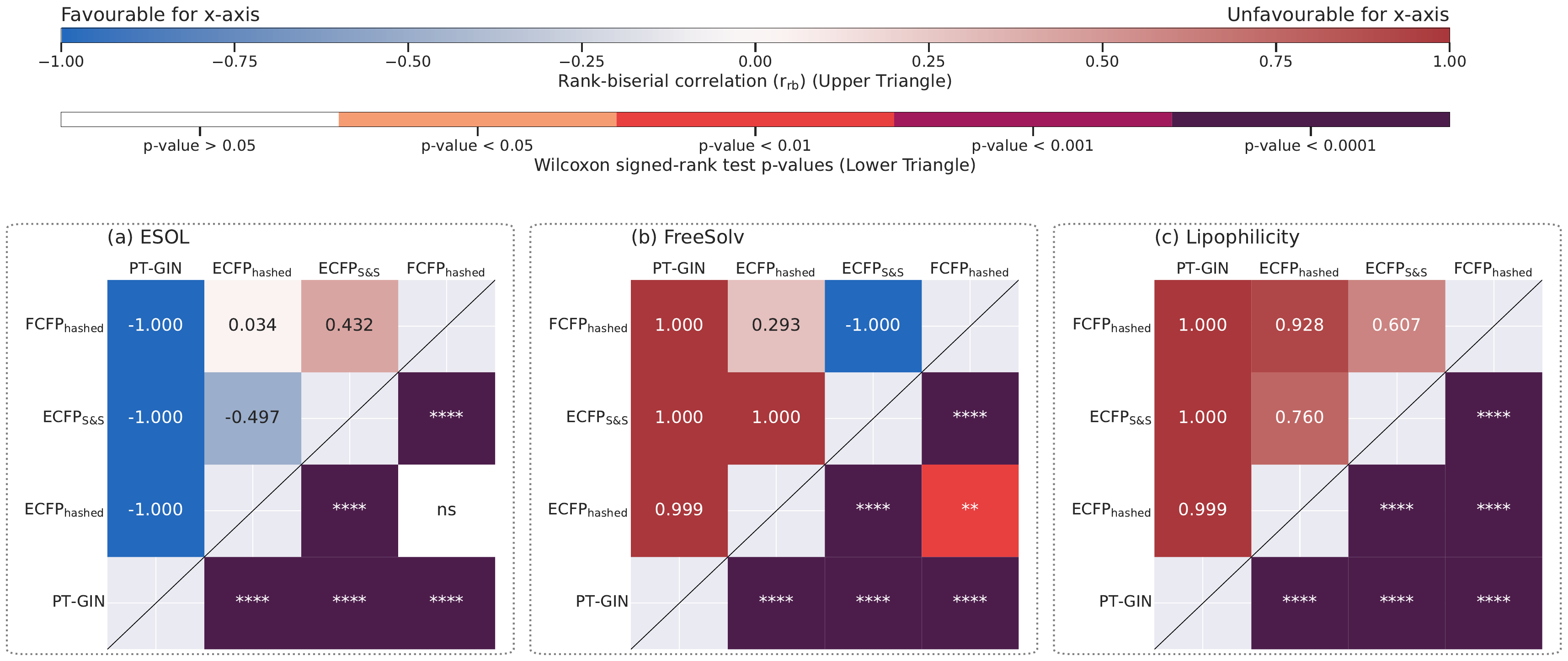}
    \caption{
        Rank-biserial correlation ($r_{rb}$) effect sizes (upper triangle) and Bonferroni-corrected two-sided Wilcoxon signed-rank test \textit{p}-values (lower triangle) of the mean absolute percentage error
        ($\downarrow\mathrm{MAPE}$) between methods on MoleculeNet regression datasets.
        Datasets shown are (a) ESOL, (b) FreeSolv, and (c) Lipophilicity.
        Featurisation methods shown are 
        ECFP-pre-trained GINs (PT-GIN), hashed Extended-Connectivity Fingerprints (ECFP$_{\mathrm{hashed}}$), 
        hashed Functional-Class Fingerprints (FCFP$_{\mathrm{hashed}}$), and 
        ECFPs vectorised via Sort \& Slice (ECFP$_{\mathrm{S\&S}}$).
        For $r_{rb}$ effect sizes (upper triangle): blue indicates the method on the \textit{x}-axis has a lower $\mathrm{MAPE}$ than 
        the method on the \textit{y}-axis; red indicates the opposite. 
        For Wilcoxon signed-rank test \textit{p}-values (lower triangle): 
        white indicates no significant difference and darker indicates a smaller \textit{p}-value ($n=199$, $\alpha=0.05$).
    }
    \label{fig:molnet_heatmap_mape}
\end{figure}

\clearpage
\newpage
\subsubsection{MUV}

\begin{figure}[!htbp]
    \centering
    \includegraphics[width=\textwidth, height=0.7\textheight, keepaspectratio]{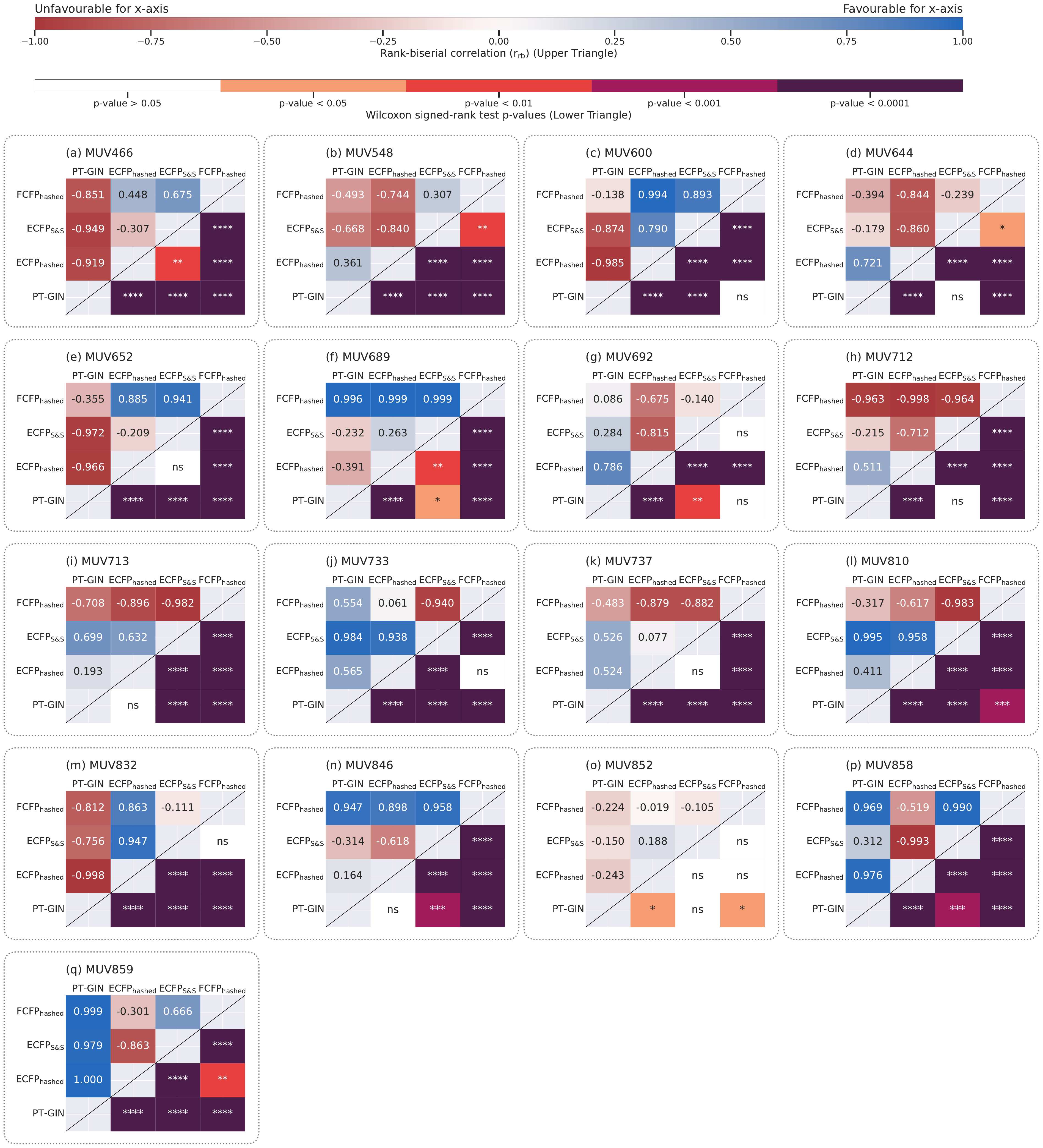}
    \caption{
        Rank-biserial correlation ($r_{rb}$) effect sizes (upper triangle) and Bonferroni-corrected two-sided Wilcoxon signed-rank test \textit{p}-values (lower triangle)
        of the area under the receiver operating characteristic curve
        ($\uparrow \mathrm{AUROC}$) between methods on MUV virtual screening tasks.
        Tasks are (a) MUV466, (b) MUV548, (c) MUV600, (d) MUV644, (e) MUV652, 
        (f) MUV689, (g) MUV692, (h) MUV712, (i) MUV713, (j) MUV733, (k) MUV737, 
        (l) MUV810, (m) MUV832, (n) MUV846, (o) MUV852, (p) MUV858, (q) MUV859.
        Featurisation methods shown are 
        ECFP-pre-trained GINs (PT-GIN), hashed Extended-Connectivity Fingerprints (ECFP$_{\mathrm{hashed}}$), 
        hashed Functional-Class Fingerprints (FCFP$_{\mathrm{hashed}}$), and 
        ECFPs vectorised via Sort \& Slice (ECFP$_{\mathrm{S\&S}}$).
        For $r_{rb}$ effect sizes (upper triangle): blue indicates the method on the \textit{x}-axis has a higher $\mathrm{AUROC}$ than 
        the method on the \textit{y}-axis; red indicates the opposite. 
        For Wilcoxon signed-rank test \textit{p}-values (lower triangle): 
        white indicates no significant difference and darker indicates a smaller \textit{p}-value ($n=199$, $\alpha=0.05$).
    }
    \label{fig:muv_heatmap_auroc}
\end{figure}

\begin{figure}[!htbp]
    \centering
    \includegraphics[width=\textwidth, height=0.7\textheight, keepaspectratio]{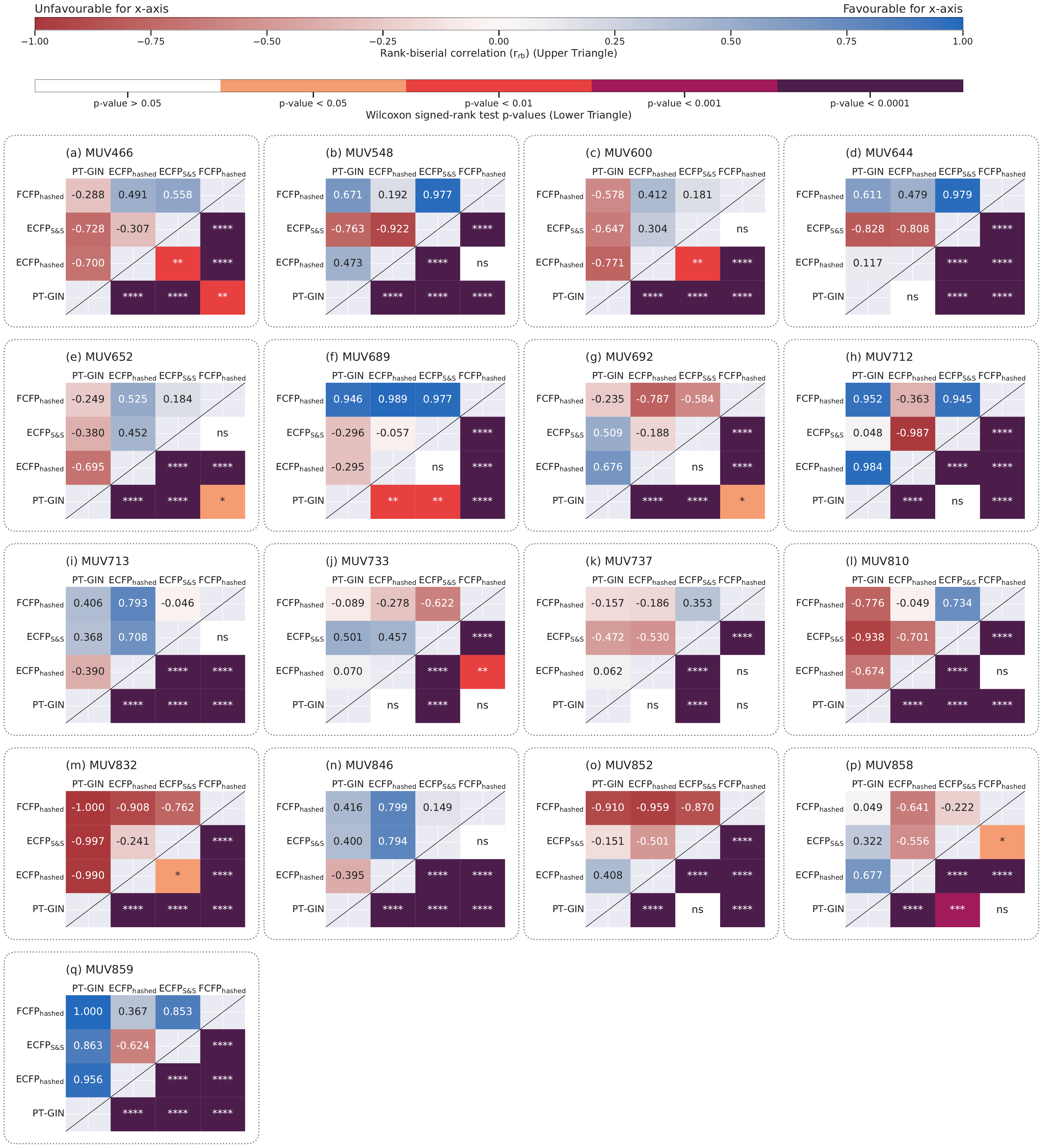}
    \caption{
        Rank-biserial correlation ($r_{rb}$) effect sizes (upper triangle) and Bonferroni-corrected two-sided Wilcoxon signed-rank test \textit{p}-values (lower triangle) of the area under the precision-recall curve
        ($\uparrow \mathrm{AUCPR}$) between methods on MUV virtual screening tasks.
        Tasks are (a) MUV466, (b) MUV548, (c) MUV600, (d) MUV644, (e) MUV652, 
        (f) MUV689, (g) MUV692, (h) MUV712, (i) MUV713, (j) MUV733, (k) MUV737, 
        (l) MUV810, (m) MUV832, (n) MUV846, (o) MUV852, (p) MUV858, (q) MUV859.
        Featurisation methods shown are 
        ECFP-pre-trained GINs (PT-GIN), hashed Extended-Connectivity Fingerprints (ECFP$_{\mathrm{hashed}}$), 
        hashed Functional-Class Fingerprints (FCFP$_{\mathrm{hashed}}$), and 
        ECFPs vectorised via Sort \& Slice (ECFP$_{\mathrm{S\&S}}$).
        For $r_{rb}$ effect sizes (upper triangle): blue indicates the method on the \textit{x}-axis has a higher $\mathrm{AUCPR}$ than 
        the method on the \textit{y}-axis; red indicates the opposite. 
        For Wilcoxon signed-rank test \textit{p}-values (lower triangle): 
        white indicates no significant difference and darker indicates a smaller \textit{p}-value ($n=199$, $\alpha=0.05$).
    }
    \label{fig:muv_heatmap_aucpr}
\end{figure}

\begin{figure}[!htbp]
    \centering
    \includegraphics[width=\textwidth, height=0.7\textheight, keepaspectratio]{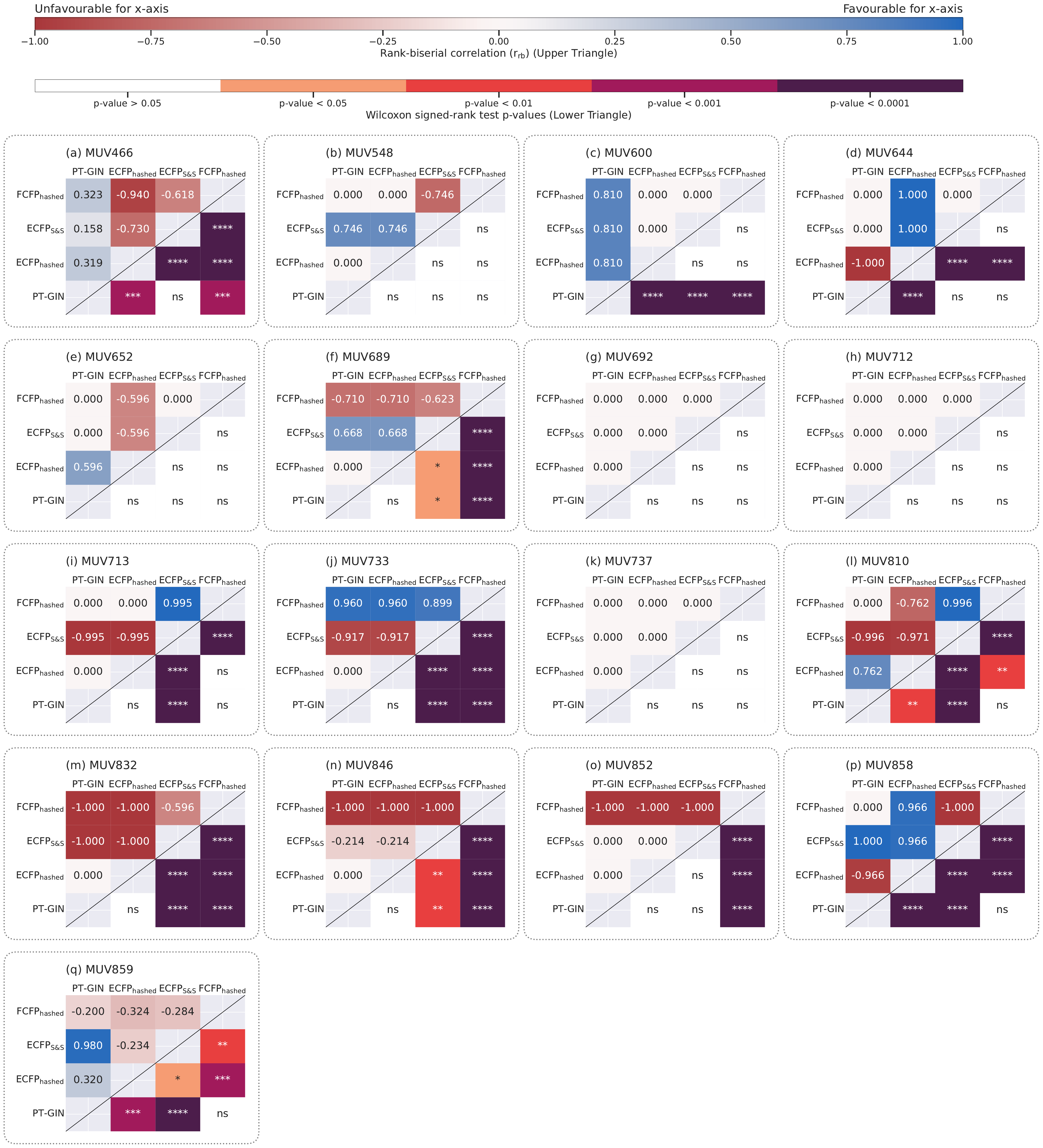}
    \caption{
        Rank-biserial correlation ($r_{rb}$) effect sizes (upper triangle) and Bonferroni-corrected two-sided Wilcoxon signed-rank test \textit{p}-values (lower triangle) of the Matthews correlation coefficient
        ($\uparrow \mathrm{MCC}$) between methods on MUV virtual screening tasks.
        Tasks are (a) MUV466, (b) MUV548, (c) MUV600, (d) MUV644, (e) MUV652, 
        (f) MUV689, (g) MUV692, (h) MUV712, (i) MUV713, (j) MUV733, (k) MUV737, 
        (l) MUV810, (m) MUV832, (n) MUV846, (o) MUV852, (p) MUV858, (q) MUV859.
        Featurisation methods shown are 
        ECFP-pre-trained GINs (PT-GIN), hashed Extended-Connectivity Fingerprints (ECFP$_{\mathrm{hashed}}$), 
        hashed Functional-Class Fingerprints (FCFP$_{\mathrm{hashed}}$), and 
        ECFPs vectorised via Sort \& Slice (ECFP$_{\mathrm{S\&S}}$).
        For $r_{rb}$ effect sizes (upper triangle): blue indicates the method on the \textit{x}-axis has a higher $\mathrm{MCC}$ than 
        the method on the \textit{y}-axis; red indicates the opposite. 
        For Wilcoxon signed-rank test \textit{p}-values (lower triangle): 
        white indicates no significant difference and darker indicates a smaller \textit{p}-value ($n=199$, $\alpha=0.05$).
    }
    \label{fig:muv_heatmap_mcc}
\end{figure}

\begin{figure}[!htbp]
    \centering
    \includegraphics[width=\textwidth, height=0.7\textheight, keepaspectratio]{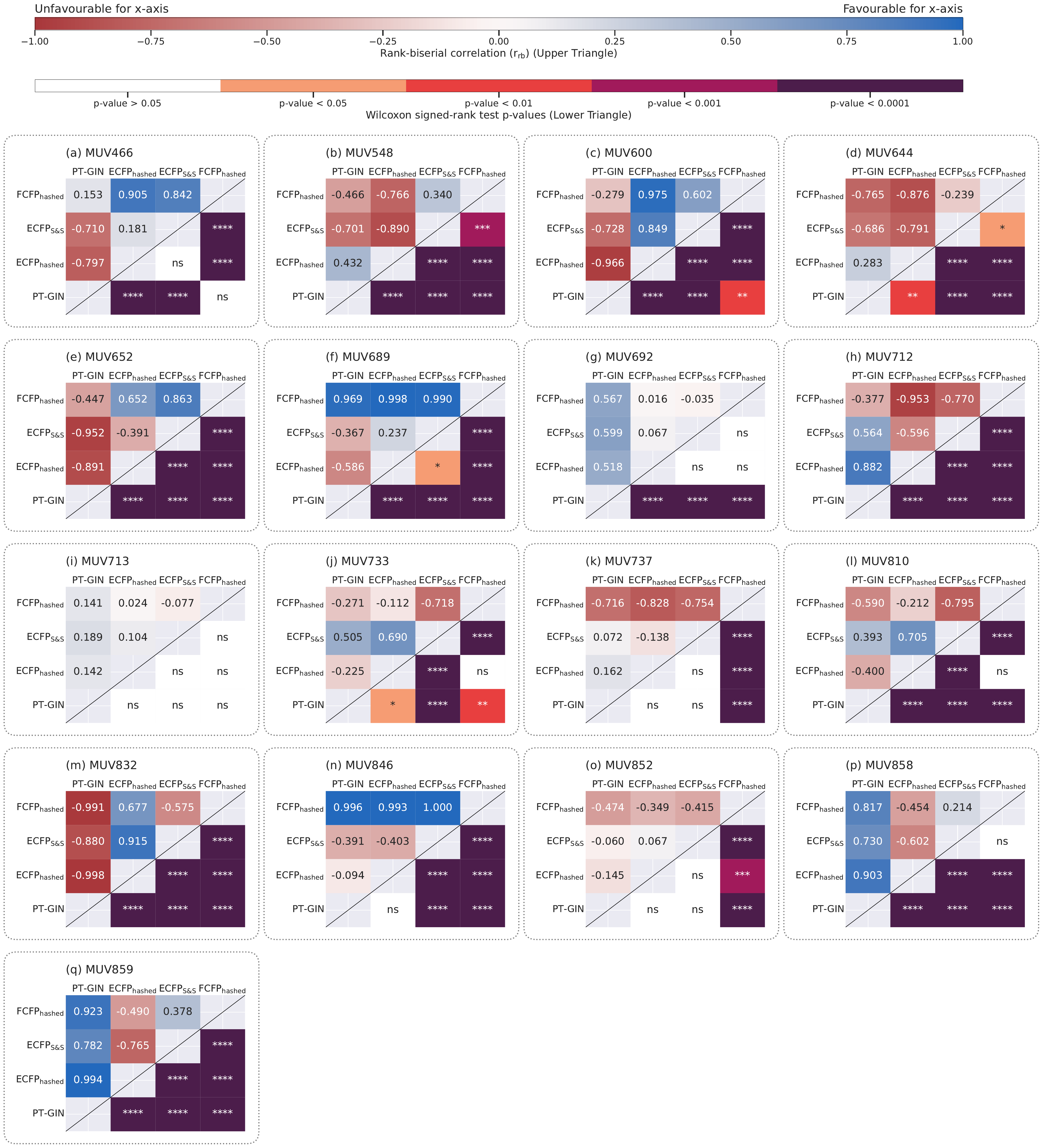}
    \caption{
        Rank-biserial correlation ($r_{rb}$) effect sizes (upper triangle) and Bonferroni-corrected two-sided Wilcoxon signed-rank test \textit{p}-values (lower triangle) of the Enrichment Factor at $10\%$
        ($\uparrow \mathrm{EF_{10\%}}$) between methods on MUV virtual screening tasks.
        Tasks are (a) MUV466, (b) MUV548, (c) MUV600, (d) MUV644, (e) MUV652, 
        (f) MUV689, (g) MUV692, (h) MUV712, (i) MUV713, (j) MUV733, (k) MUV737, 
        (l) MUV810, (m) MUV832, (n) MUV846, (o) MUV852, (p) MUV858, (q) MUV859.
        Featurisation methods shown are 
        ECFP-pre-trained GINs (PT-GIN), hashed Extended-Connectivity Fingerprints (ECFP$_{\mathrm{hashed}}$), 
        hashed Functional-Class Fingerprints (FCFP$_{\mathrm{hashed}}$), and 
        ECFPs vectorised via Sort \& Slice (ECFP$_{\mathrm{S\&S}}$).
        For $r_{rb}$ effect sizes (upper triangle): blue indicates the method on the \textit{x}-axis has a higher $\mathrm{EF}_{10\%}$ than 
        the method on the \textit{y}-axis; red indicates the opposite. 
        For Wilcoxon signed-rank test \textit{p}-values (lower triangle): 
        white indicates no significant difference and darker indicates a smaller \textit{p}-value ($n=199$, $\alpha=0.05$).
    }
    \label{fig:muv_heatmap_ef10}
\end{figure}

\begin{figure}[!htbp]
    \centering
    \includegraphics[width=\textwidth, height=0.7\textheight, keepaspectratio]{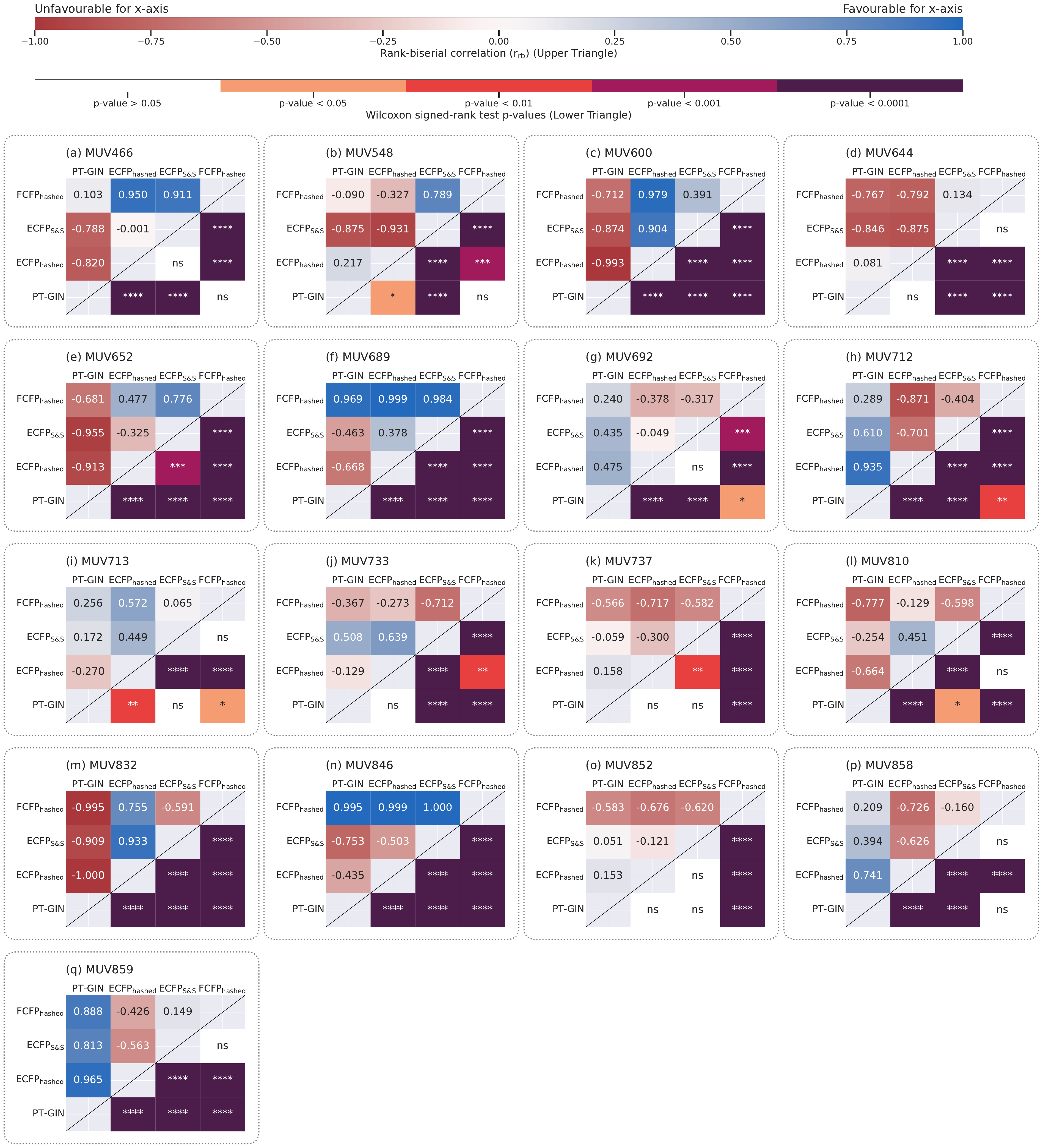}
    \caption{
        Rank-biserial correlation ($r_{rb}$) effect sizes (upper triangle) and Bonferroni-corrected two-sided Wilcoxon signed-rank test \textit{p}-values (lower triangle) of the Enrichment Factor at $5\%$
        ($\uparrow \mathrm{EF_{5\%}}$) between methods on MUV virtual screening tasks.
        Tasks are (a) MUV466, (b) MUV548, (c) MUV600, (d) MUV644, (e) MUV652, 
        (f) MUV689, (g) MUV692, (h) MUV712, (i) MUV713, (j) MUV733, (k) MUV737, 
        (l) MUV810, (m) MUV832, (n) MUV846, (o) MUV852, (p) MUV858, (q) MUV859.
        Featurisation methods shown are 
        ECFP-pre-trained GINs (PT-GIN), hashed Extended-Connectivity Fingerprints (ECFP$_{\mathrm{hashed}}$), 
        hashed Functional-Class Fingerprints (FCFP$_{\mathrm{hashed}}$), and 
        ECFPs vectorised via Sort \& Slice (ECFP$_{\mathrm{S\&S}}$).
        For $r_{rb}$ effect sizes (upper triangle): blue indicates the method on the \textit{x}-axis has a higher $\mathrm{EF}_{5\%}$ than 
        the method on the \textit{y}-axis; red indicates the opposite. 
        For Wilcoxon signed-rank test \textit{p}-values (lower triangle): 
        white indicates no significant difference and darker indicates a smaller \textit{p}-value ($n=199$, $\alpha=0.05$).
    }
    \label{fig:muv_heatmap_ef5}
\end{figure}

\begin{figure}[!htbp]
    \centering
    \includegraphics[width=\textwidth, height=0.7\textheight, keepaspectratio]{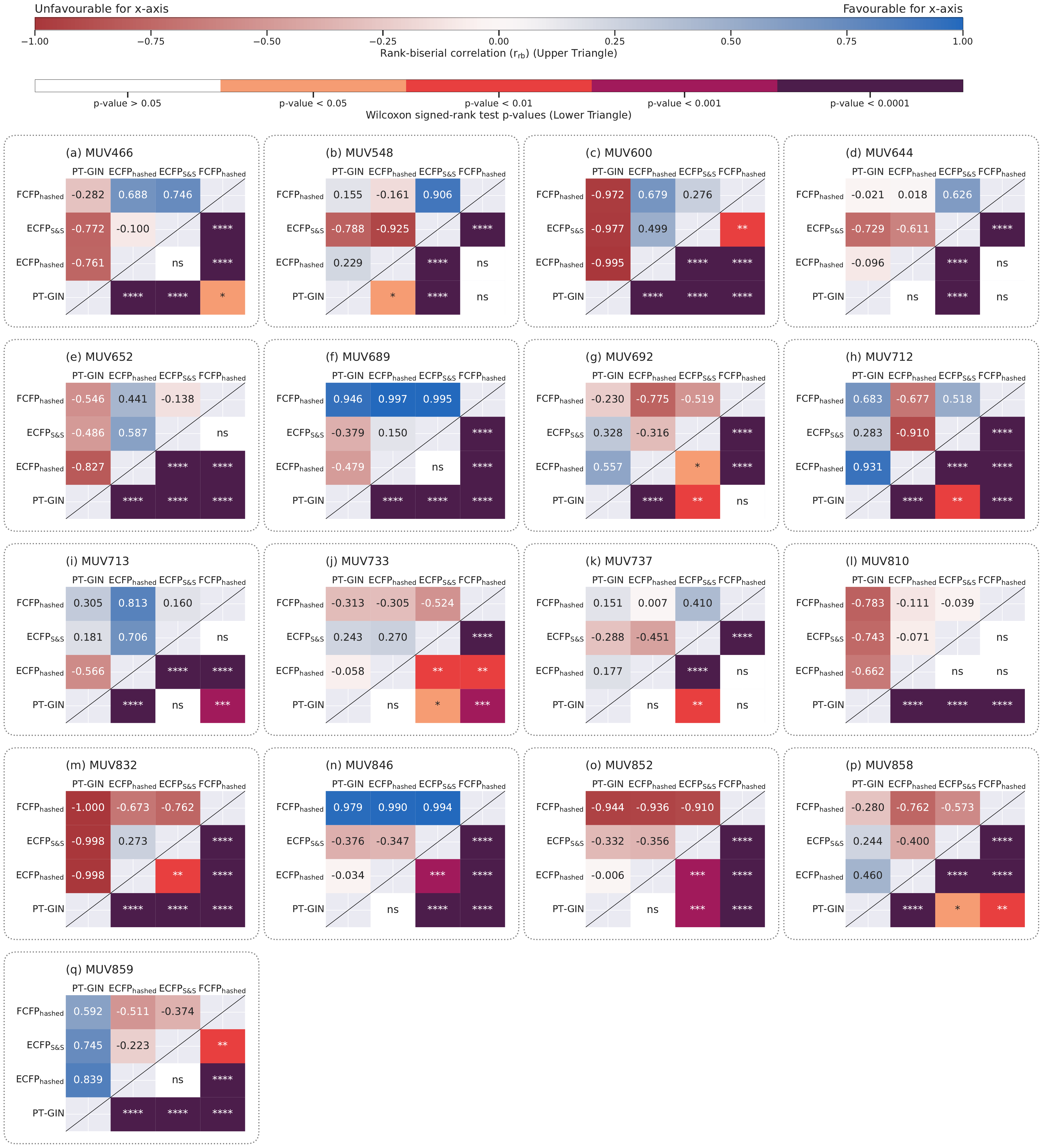}
    \caption{
        Rank-biserial correlation ($r_{rb}$) effect sizes (upper triangle) and Bonferroni-corrected two-sided Wilcoxon signed-rank test \textit{p}-values (lower triangle) of the Enrichment Factor at $1\%$
        ($\uparrow \mathrm{EF_{1\%}}$) between methods on MUV virtual screening tasks.
        Tasks are (a) MUV466, (b) MUV548, (c) MUV600, (d) MUV644, (e) MUV652, 
        (f) MUV689, (g) MUV692, (h) MUV712, (i) MUV713, (j) MUV733, (k) MUV737, 
        (l) MUV810, (m) MUV832, (n) MUV846, (o) MUV852, (p) MUV858, (q) MUV859.
        Featurisation methods shown are 
        ECFP-pre-trained GINs (PT-GIN), hashed Extended-Connectivity Fingerprints (ECFP$_{\mathrm{hashed}}$), 
        hashed Functional-Class Fingerprints (FCFP$_{\mathrm{hashed}}$), and 
        ECFPs vectorised via Sort \& Slice (ECFP$_{\mathrm{S\&S}}$).
        For $r_{rb}$ effect sizes (upper triangle): blue indicates the method on the \textit{x}-axis has a higher $\mathrm{EF}_{1\%}$ than 
        the method on the \textit{y}-axis; red indicates the opposite. 
        For Wilcoxon signed-rank test \textit{p}-values (lower triangle): 
        white indicates no significant difference and darker indicates a smaller \textit{p}-value ($n=199$, $\alpha=0.05$).
    }
    \label{fig:muv_heatmap_ef1}
\end{figure}

\begin{figure}[!htbp]
    \centering
    \includegraphics[width=\textwidth, height=0.7\textheight, keepaspectratio]{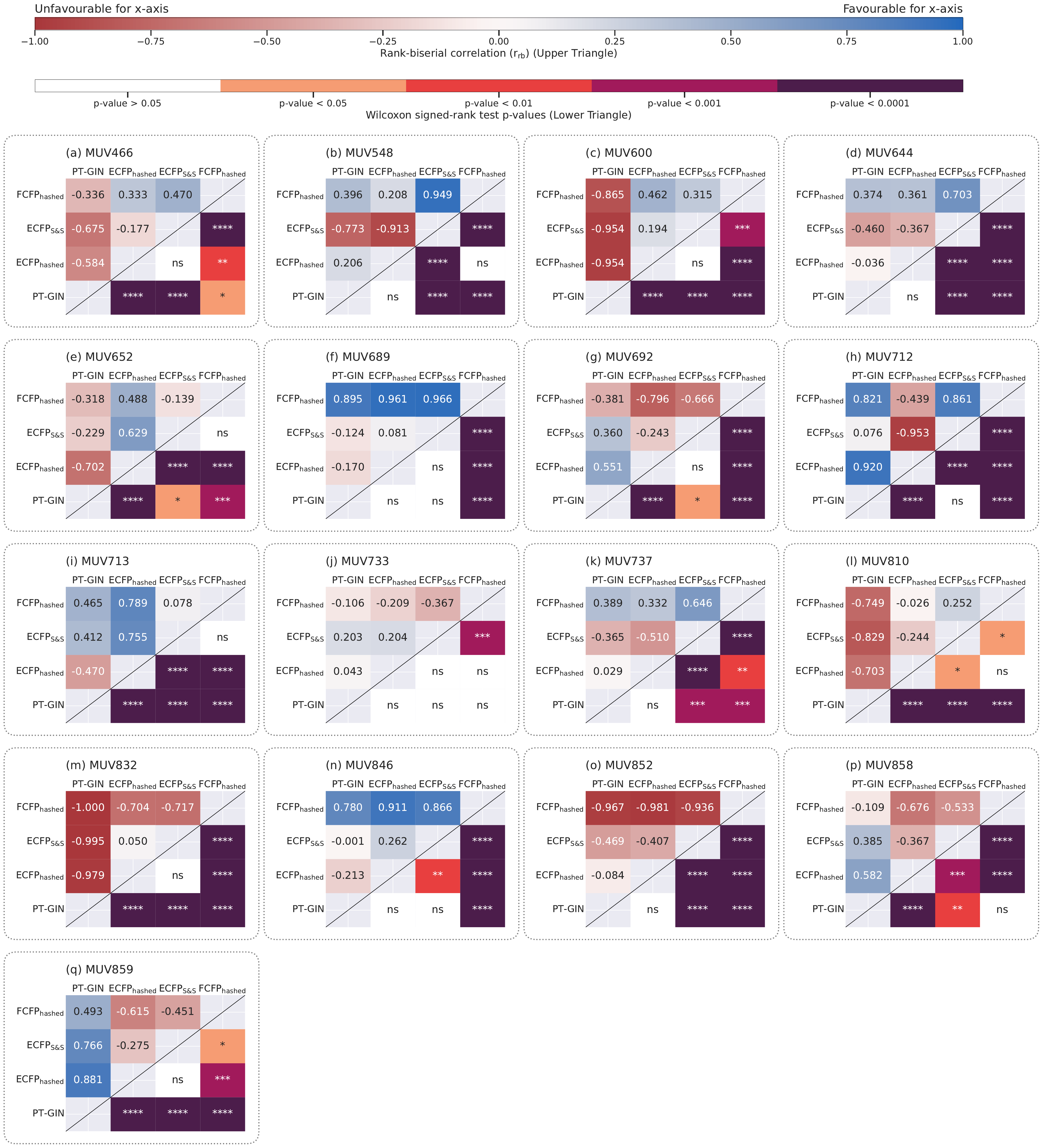}
    \caption{
        Rank-biserial correlation ($r_{rb}$) effect sizes (upper triangle) and Bonferroni-corrected two-sided Wilcoxon signed-rank test \textit{p}-values (lower triangle) of the Enrichment Factor at $0.5\%$
        ($\uparrow \mathrm{EF_{0.5\%}}$) between methods on MUV virtual screening tasks.
        Tasks are (a) MUV466, (b) MUV548, (c) MUV600, (d) MUV644, (e) MUV652, 
        (f) MUV689, (g) MUV692, (h) MUV712, (i) MUV713, (j) MUV733, (k) MUV737, 
        (l) MUV810, (m) MUV832, (n) MUV846, (o) MUV852, (p) MUV858, (q) MUV859.
        Featurisation methods shown are 
        ECFP-pre-trained GINs (PT-GIN), hashed Extended-Connectivity Fingerprints (ECFP$_{\mathrm{hashed}}$), 
        hashed Functional-Class Fingerprints (FCFP$_{\mathrm{hashed}}$), and 
        ECFPs vectorised via Sort \& Slice (ECFP$_{\mathrm{S\&S}}$).
        For $r_{rb}$ effect sizes (upper triangle): blue indicates the method on the \textit{x}-axis has a higher $\mathrm{EF}_{0.5\%}$ than 
        the method on the \textit{y}-axis; red indicates the opposite. 
        For Wilcoxon signed-rank test \textit{p}-values (lower triangle): 
        white indicates no significant difference and darker indicates a smaller \textit{p}-value ($n=199$, $\alpha=0.05$).
    }
    \label{fig:muv_heatmap_ef05}
\end{figure}

\clearpage
\newpage
\subsubsection{Tox21}
\begin{figure}[!htbp]
    \centering
    \includegraphics[width=\textwidth, height=0.7\textheight, keepaspectratio]{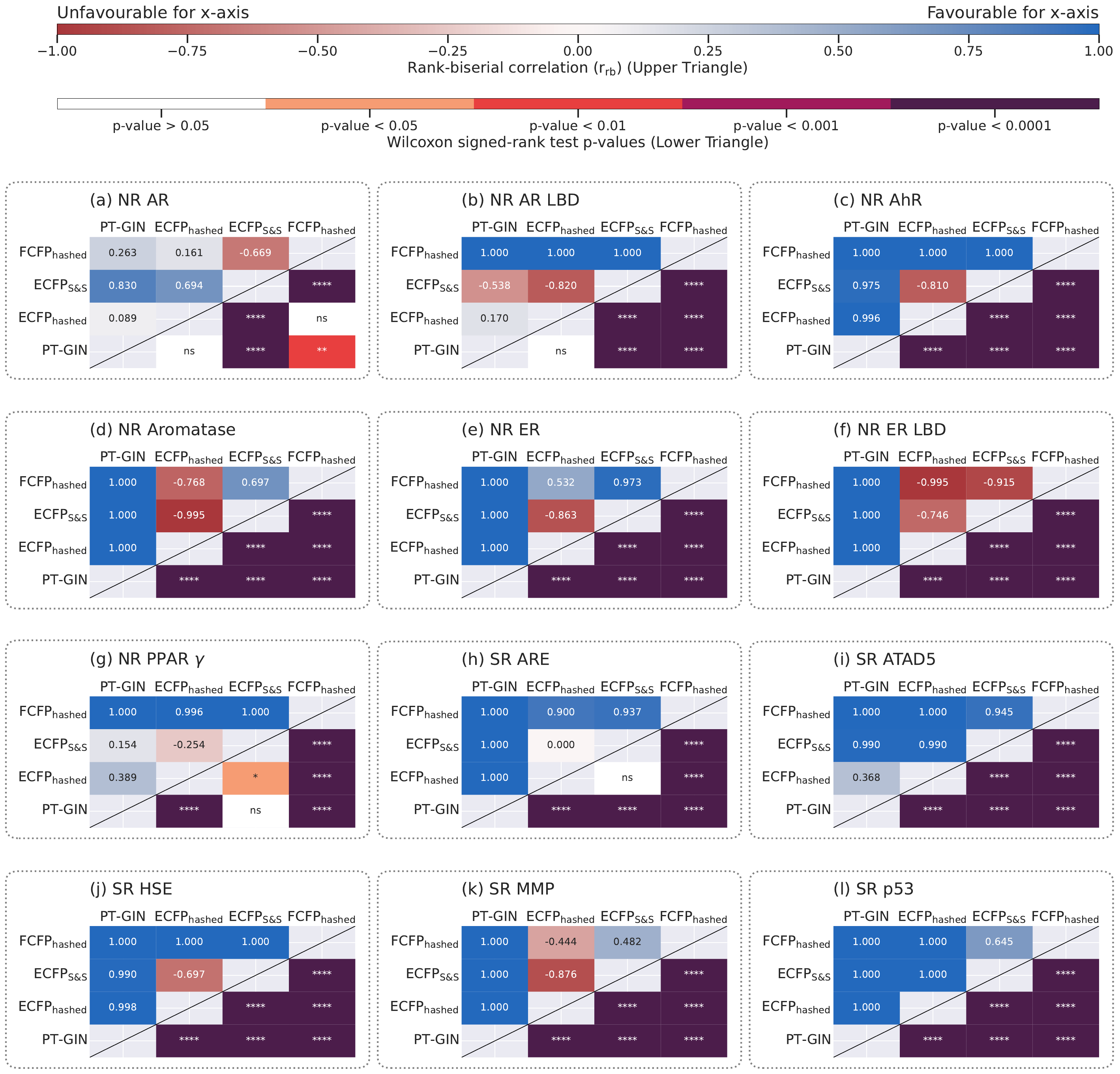}
    \caption{
        Rank-biserial correlation ($r_{rb}$) effect sizes (upper triangle) and Bonferroni-corrected two-sided Wilcoxon signed-rank test \textit{p}-values (lower triangle) of the area under the receiver operating characteristic curve
        ($\uparrow \mathrm{AUROC}$) between methods on Tox21 toxicity tasks.
        For task labels, see Supplementary Information, \textit{\nameref{si:datasets}}.
        Featurisation methods shown are 
        ECFP-pre-trained GINs (PT-GIN), hashed Extended-Connectivity Fingerprints (ECFP$_{\mathrm{hashed}}$), 
        hashed Functional-Class Fingerprints (FCFP$_{\mathrm{hashed}}$), and 
        ECFPs vectorised via Sort \& Slice (ECFP$_{\mathrm{S\&S}}$).
        For $r_{rb}$ effect sizes (upper triangle): blue indicates the method on the \textit{x}-axis has a higher $\mathrm{AUROC}$ than 
        the method on the \textit{y}-axis; red indicates the opposite. 
        For Wilcoxon signed-rank test \textit{p}-values (lower triangle): 
        white indicates no significant difference and darker indicates a smaller \textit{p}-value ($n=199$, $\alpha=0.05$).
    }
    \label{fig:tox21_heatmap_auroc}
\end{figure}

\begin{figure}[!htbp]
    \centering
    \includegraphics[width=\textwidth, height=0.7\textheight, keepaspectratio]{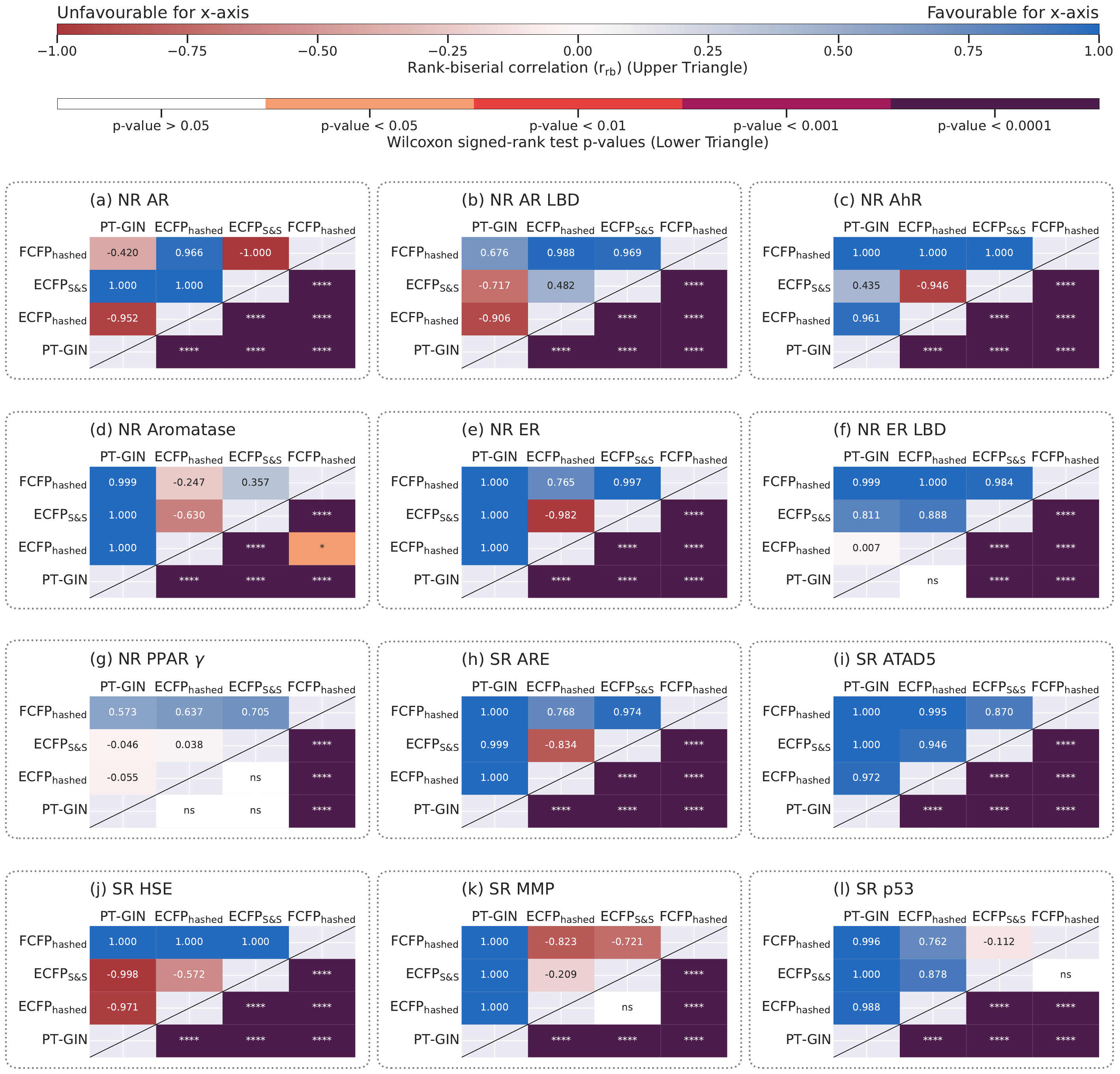}
    \caption{
        Rank-biserial correlation ($r_{rb}$) effect sizes (upper triangle) and Bonferroni-corrected two-sided Wilcoxon signed-rank test \textit{p}-values (lower triangle) of the area under the precision-recall curve
        ($\uparrow \mathrm{AUCPR}$) between methods on Tox21 toxicity tasks.
        For task labels, see Supplementary Information, \textit{\nameref{si:datasets}}.
        Featurisation methods shown are 
        ECFP-pre-trained GINs (PT-GIN), hashed Extended-Connectivity Fingerprints (ECFP$_{\mathrm{hashed}}$), 
        hashed Functional-Class Fingerprints (FCFP$_{\mathrm{hashed}}$), and 
        ECFPs vectorised via Sort \& Slice (ECFP$_{\mathrm{S\&S}}$).
        For $r_{rb}$ effect sizes (upper triangle): blue indicates the method on the \textit{x}-axis has a higher $\mathrm{AUCPR}$ than 
        the method on the \textit{y}-axis; red indicates the opposite. 
        For Wilcoxon signed-rank test \textit{p}-values (lower triangle): 
        white indicates no significant difference and darker indicates a smaller \textit{p}-value ($n=199$, $\alpha=0.05$).
    }
    \label{fig:tox21_heatmap_aucpr}
\end{figure}

\begin{figure}[!htbp]
    \centering
    \includegraphics[width=\textwidth, height=0.7\textheight, keepaspectratio]{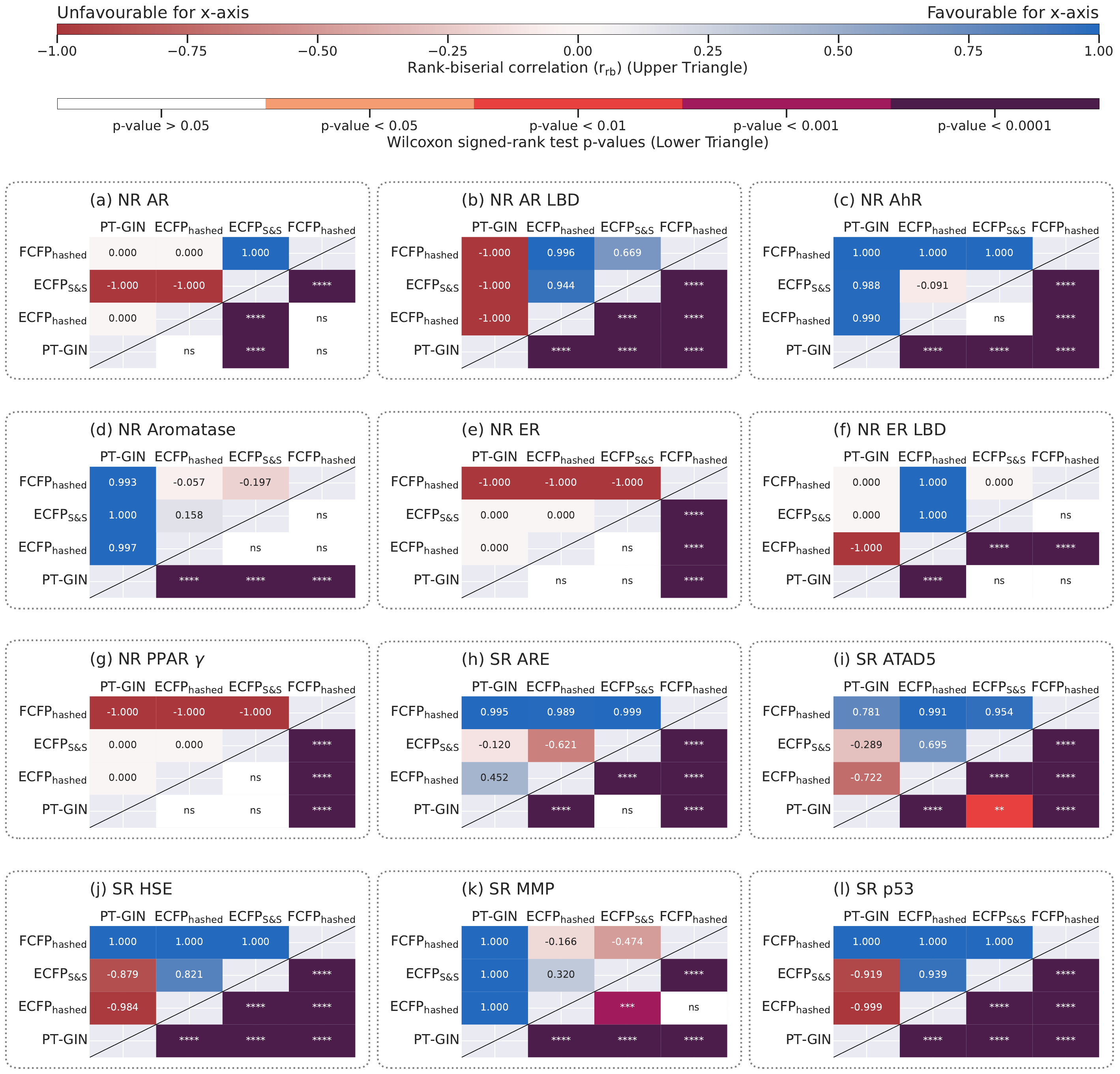}
    \caption{
        Rank-biserial correlation ($r_{rb}$) effect sizes (upper triangle) and Bonferroni-corrected two-sided Wilcoxon signed-rank test \textit{p}-values (lower triangle) of the Matthews correlation coefficient
        ($\uparrow \mathrm{MCC}$) between methods on Tox21 toxicity tasks.
        For task labels, see Supplementary Information, \textit{\nameref{si:datasets}}.
        Featurisation methods shown are 
        ECFP-pre-trained GINs (PT-GIN), hashed Extended-Connectivity Fingerprints (ECFP$_{\mathrm{hashed}}$), 
        hashed Functional-Class Fingerprints (FCFP$_{\mathrm{hashed}}$), and 
        ECFPs vectorised via Sort \& Slice (ECFP$_{\mathrm{S\&S}}$).
        For $r_{rb}$ effect sizes (upper triangle): blue indicates the method on the \textit{x}-axis has a higher $\mathrm{MCC}$ than 
        the method on the \textit{y}-axis; red indicates the opposite. 
        For Wilcoxon signed-rank test \textit{p}-values (lower triangle): 
        white indicates no significant difference and darker indicates a smaller \textit{p}-value ($n=199$, $\alpha=0.05$).
    }
    \label{fig:tox21_heatmap_mcc}
\end{figure}
\clearpage
\newpage

\subsection{Box plots with point density overlay}\label{si:boxplots}

\subsubsection{Biogen}

\begin{figure}[!htbp]
    \centering
    \includegraphics[width=\textwidth, height=0.8\textheight, keepaspectratio]{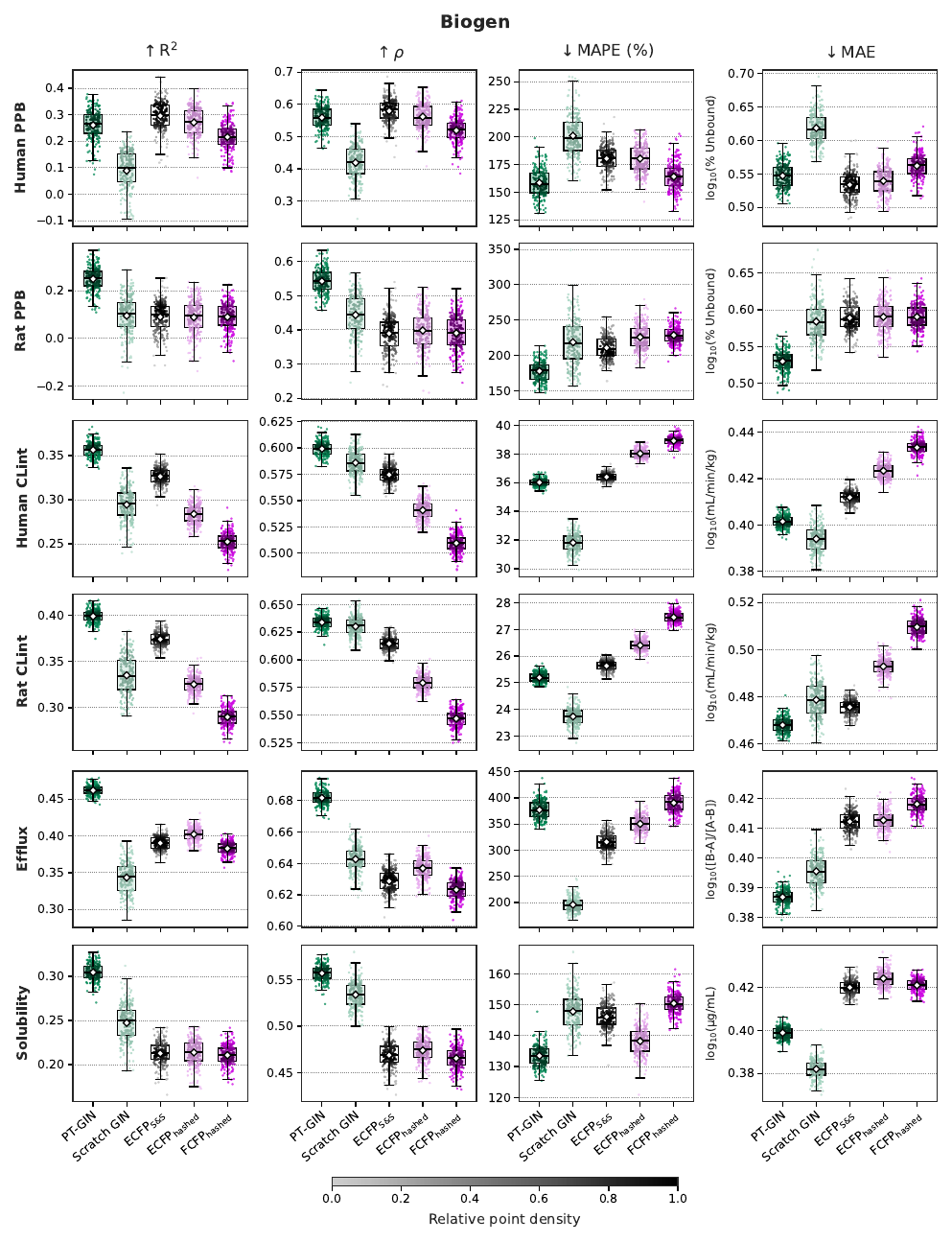}
    \caption{
        Box-and-whisker plots with point density overlay for the Biogen ADME tasks.
        Rows show the Human PPB, Rat PPB, Human CLint, Rat CLint, Efflux, and Solubility tasks.
        Columns show
        coefficient of determination ($\uparrow \mathrm{R}^2$),
        Pearson correlation coefficient ($\uparrow \rho$),
        mean absolute percentage error ($\downarrow\mathrm{MAPE} (\%)$), and
        mean absolute error ($\downarrow\mathrm{MAE}$, in the units of each task's label, given on the y-axis).
        Featurisation methods shown are
        ECFP-pre-trained GINs (PT-GIN),
        GINs trained from scratch (Scratch GIN),
        ECFPs vectorised via Sort \& Slice (ECFP$_{\mathrm{S\&S}}$),
        hashed Extended-Connectivity Fingerprints (ECFP$_{\mathrm{hashed}}$), and
        hashed Functional-Class Fingerprints (FCFP$_{\mathrm{hashed}}$).
        Point density is estimated using a Gaussian kernel density estimate ($n=199$).
        Diamonds are the mean of each distribution.
        Lines are the median of each distribution.
    }
    \label{fig:biogen_boxplot}
\end{figure}

\clearpage
\newpage

\subsubsection{ChEMBL}

\begin{figure}[!htbp]
    \centering
    \includegraphics[width=\textwidth, height=0.8\textheight, keepaspectratio]{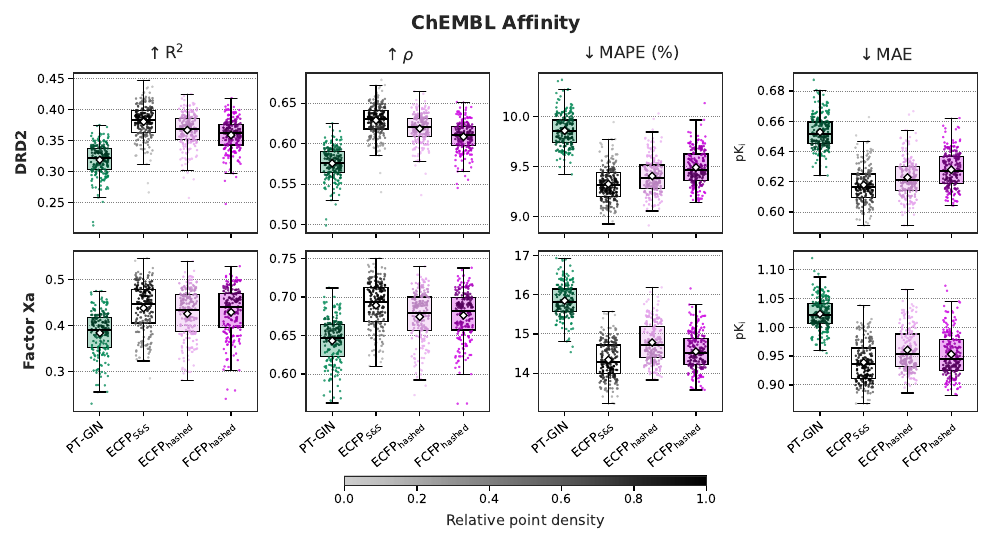}
    \caption{
        Box-and-whisker plots with point density overlay for the ChEMBL binding affinity tasks.
        Rows show the DRD2 and Factor Xa tasks.
        Columns show
        coefficient of determination ($\uparrow \mathrm{R}^2$),
        Pearson correlation coefficient ($\uparrow \rho$),
        mean absolute percentage error ($\downarrow\mathrm{MAPE} (\%)$), and
        mean absolute error ($\downarrow\mathrm{MAE}$, in the units of each task's label, given on the y-axis).
        Featurisation methods shown are
        ECFP-pre-trained GINs (PT-GIN),
        ECFPs vectorised via Sort \& Slice (ECFP$_{\mathrm{S\&S}}$),
        hashed Extended-Connectivity Fingerprints (ECFP$_{\mathrm{hashed}}$), and
        hashed Functional-Class Fingerprints (FCFP$_{\mathrm{hashed}}$).
        Point density is estimated using a Gaussian kernel density estimate ($n=199$).
        Diamonds are the mean of each distribution.
        Lines are the median of each distribution.
    }
    \label{fig:chembl_boxplot}
\end{figure}

\clearpage
\newpage

\subsubsection{MoleculeNet regression}

\begin{figure}[!htbp]
    \centering
    \includegraphics[width=\textwidth, height=0.8\textheight, keepaspectratio]{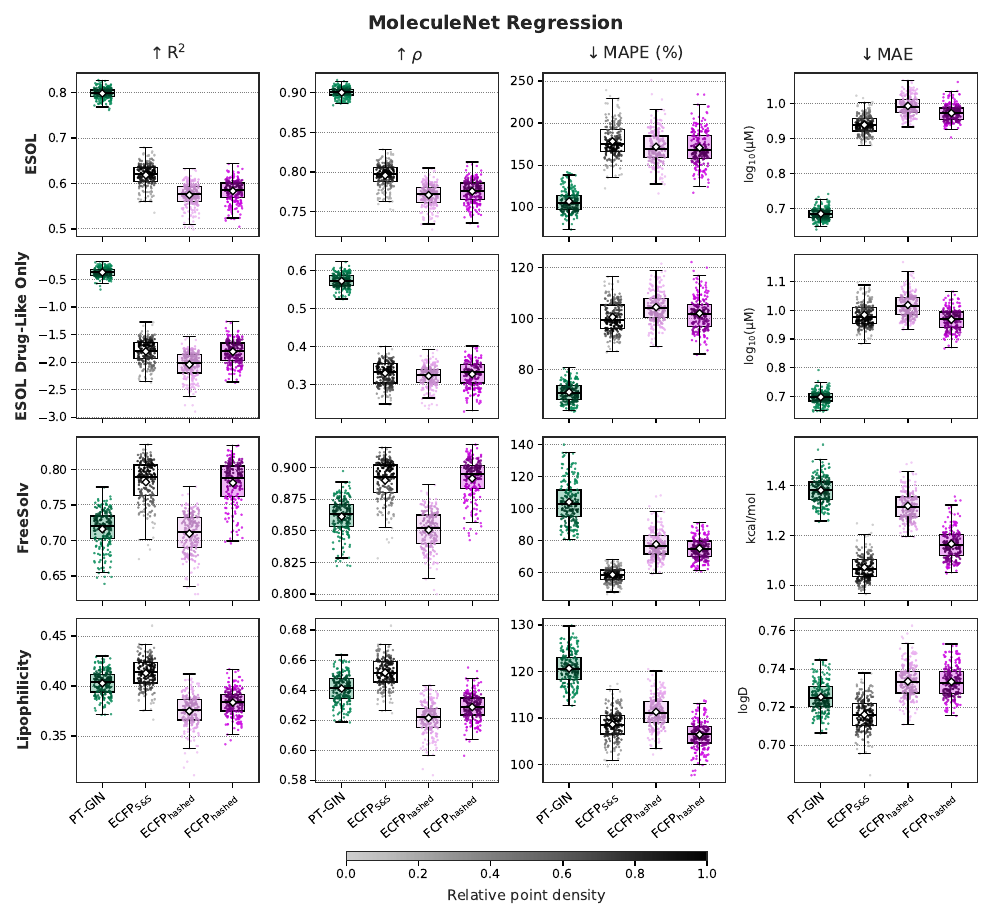}
    \caption{
        Box-and-whisker plots with point density overlay for the MoleculeNet regression tasks.
        Rows show the ESOL aqueous solubility task,
        the ESOL task restricted to drug-like molecules with $0 < \mathrm{log_{10}(S (\mu M))} < 3$ (ESOL Drug-Like Only),
        the FreeSolv free energy of solvation task, and the Lipophilicity task.
        Columns show
        coefficient of determination ($\uparrow \mathrm{R}^2$),
        Pearson correlation coefficient ($\uparrow \rho$),
        mean absolute percentage error ($\downarrow\mathrm{MAPE} (\%)$), and
        mean absolute error ($\downarrow\mathrm{MAE}$, in the units of each task's label, given on the y-axis).
        Featurisation methods shown are
        ECFP-pre-trained GINs (PT-GIN),
        ECFPs vectorised via Sort \& Slice (ECFP$_{\mathrm{S\&S}}$),
        hashed Extended-Connectivity Fingerprints (ECFP$_{\mathrm{hashed}}$), and
        hashed Functional-Class Fingerprints (FCFP$_{\mathrm{hashed}}$).
        Point density is estimated using a Gaussian kernel density estimate ($n=199$).
        Diamonds are the mean of each distribution.
        Lines are the median of each distribution.
    }
    \label{fig:molnet_boxplot}
\end{figure}

\clearpage
\newpage

\subsubsection{MUV}

\begin{figure}[!htbp]
    \centering
    \includegraphics[width=\textwidth, height=0.8\textheight, keepaspectratio]{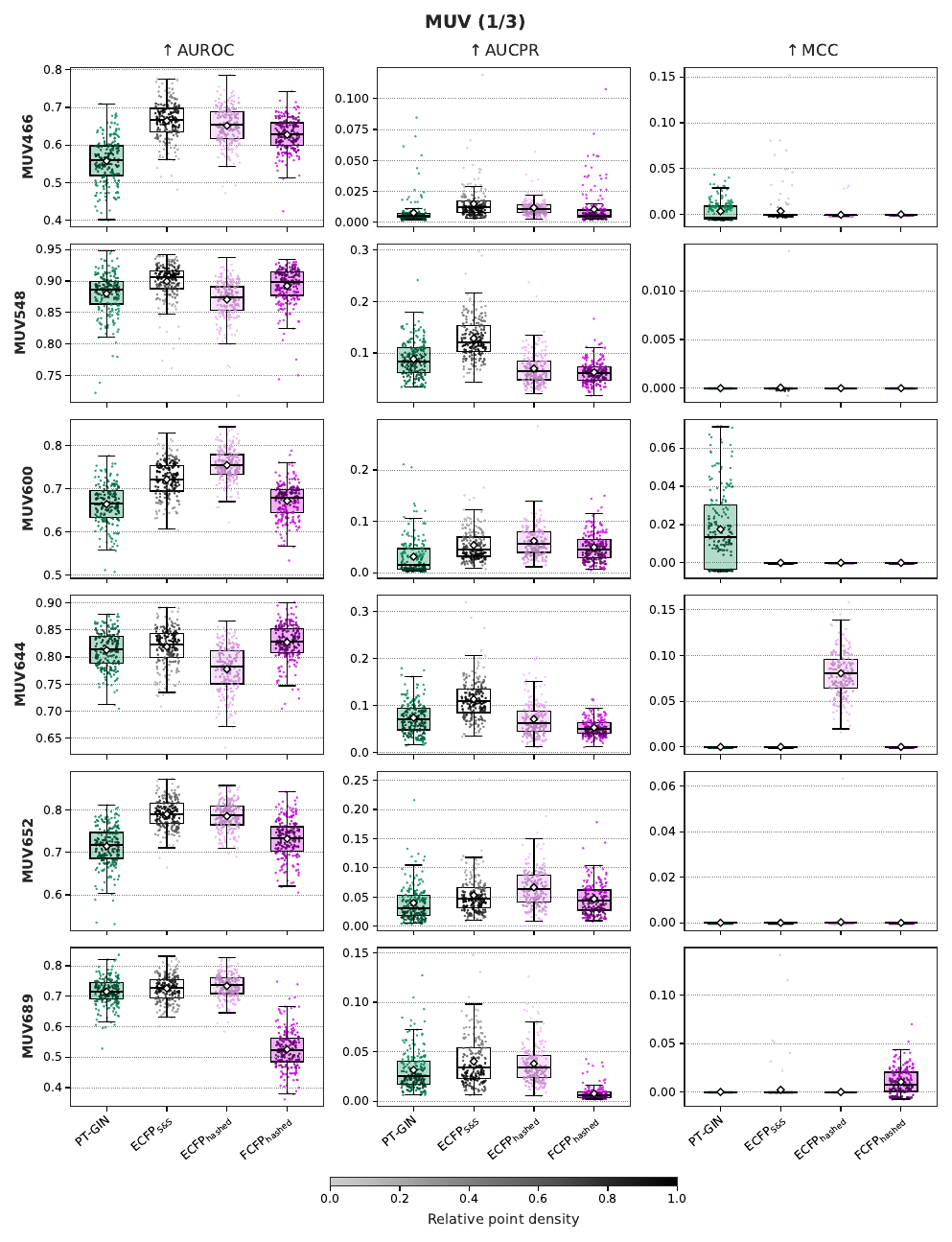}
    \caption{
        Box-and-whisker plots with point density overlay for the MUV virtual screening tasks: classification metrics (part 1 of 3).
        Rows show the MUV466, MUV548, MUV600, MUV644, MUV652, and MUV689 tasks.
        Columns show
        area under the receiver operating characteristic curve ($\uparrow \mathrm{AUROC}$),
        area under the precision-recall curve ($\uparrow \mathrm{AUCPR}$), and
        Matthews correlation coefficient ($\uparrow \mathrm{MCC}$).
        Featurisation methods shown are
        ECFP-pre-trained GINs (PT-GIN),
        ECFPs vectorised via Sort \& Slice (ECFP$_{\mathrm{S\&S}}$),
        hashed Extended-Connectivity Fingerprints (ECFP$_{\mathrm{hashed}}$), and
        hashed Functional-Class Fingerprints (FCFP$_{\mathrm{hashed}}$).
        Point density is estimated using a Gaussian kernel density estimate
        ($n=199$).
        Diamonds are the mean of each distribution.
        Lines are the median of each distribution.
    }
    \label{fig:muv_classification_boxplot_1}
\end{figure}

\begin{figure}[!htbp]
    \centering
    \includegraphics[width=\textwidth, height=0.8\textheight, keepaspectratio]{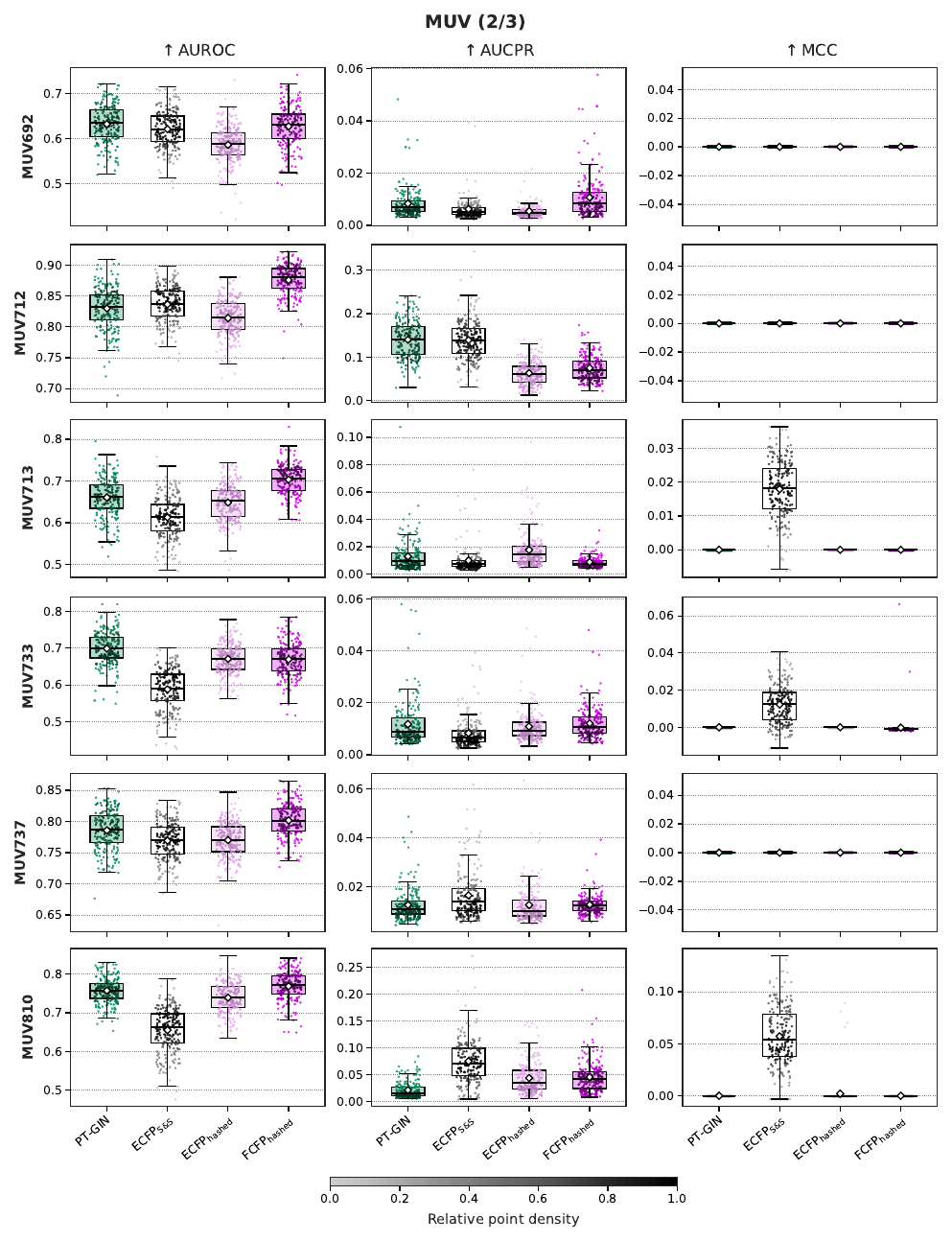}
    \caption{
        Box-and-whisker plots with point density overlay for the MUV virtual screening tasks: classification metrics (part 2 of 3).
        Rows show the MUV692, MUV712, MUV713, MUV733, MUV737, and MUV810 tasks.
        Columns show
        area under the receiver operating characteristic curve ($\uparrow \mathrm{AUROC}$),
        area under the precision-recall curve ($\uparrow \mathrm{AUCPR}$), and
        Matthews correlation coefficient ($\uparrow \mathrm{MCC}$).
        Featurisation methods shown are
        ECFP-pre-trained GINs (PT-GIN),
        ECFPs vectorised via Sort \& Slice (ECFP$_{\mathrm{S\&S}}$),
        hashed Extended-Connectivity Fingerprints (ECFP$_{\mathrm{hashed}}$), and
        hashed Functional-Class Fingerprints (FCFP$_{\mathrm{hashed}}$).
        Point density is estimated using a Gaussian kernel density estimate ($n=199$).
        Diamonds are the mean of each distribution.
        Lines are the median of each distribution.
    }
    \label{fig:muv_classification_boxplot_2}
\end{figure}

\begin{figure}[!htbp]
    \centering
    \includegraphics[width=\textwidth, height=0.8\textheight, keepaspectratio]{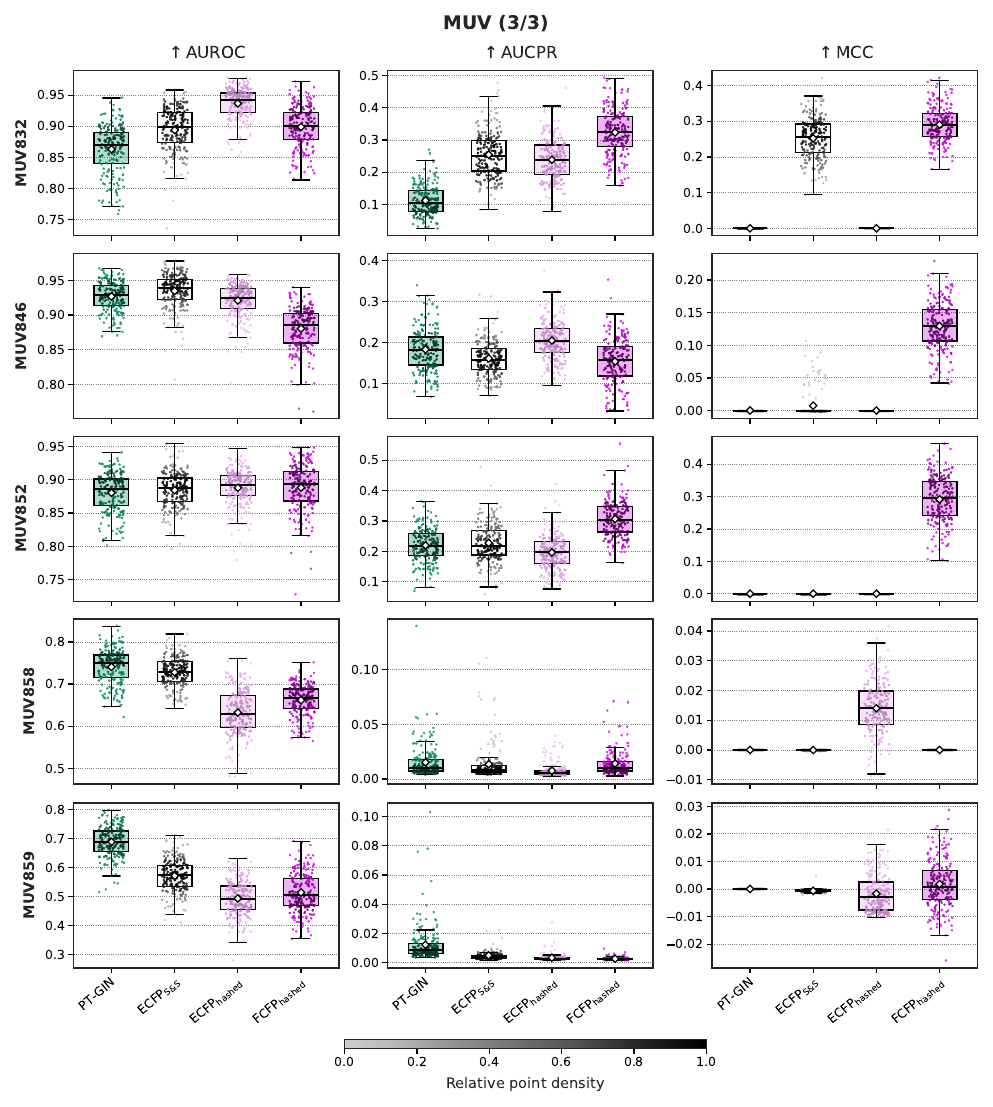}
    \caption{
        Box-and-whisker plots with point density overlay for the MUV virtual screening tasks: classification metrics (part 3 of 3).
        Rows show the MUV832, MUV846, MUV852, MUV858, and MUV859 tasks.
        Columns show
        area under the receiver operating characteristic curve ($\uparrow \mathrm{AUROC}$),
        area under the precision-recall curve ($\uparrow \mathrm{AUCPR}$), and
        Matthews correlation coefficient ($\uparrow \mathrm{MCC}$).
        Featurisation methods shown are
        ECFP-pre-trained GINs (PT-GIN),
        ECFPs vectorised via Sort \& Slice (ECFP$_{\mathrm{S\&S}}$),
        hashed Extended-Connectivity Fingerprints (ECFP$_{\mathrm{hashed}}$), and
        hashed Functional-Class Fingerprints (FCFP$_{\mathrm{hashed}}$).
        Point density is estimated using a Gaussian kernel density estimate ($n=199$).
        Diamonds are the mean of each distribution.
        Lines are the median of each distribution.
    }
    \label{fig:muv_classification_boxplot_3}
\end{figure}

\begin{figure}[!htbp]
    \centering
    \includegraphics[width=\textwidth, height=0.8\textheight, keepaspectratio]{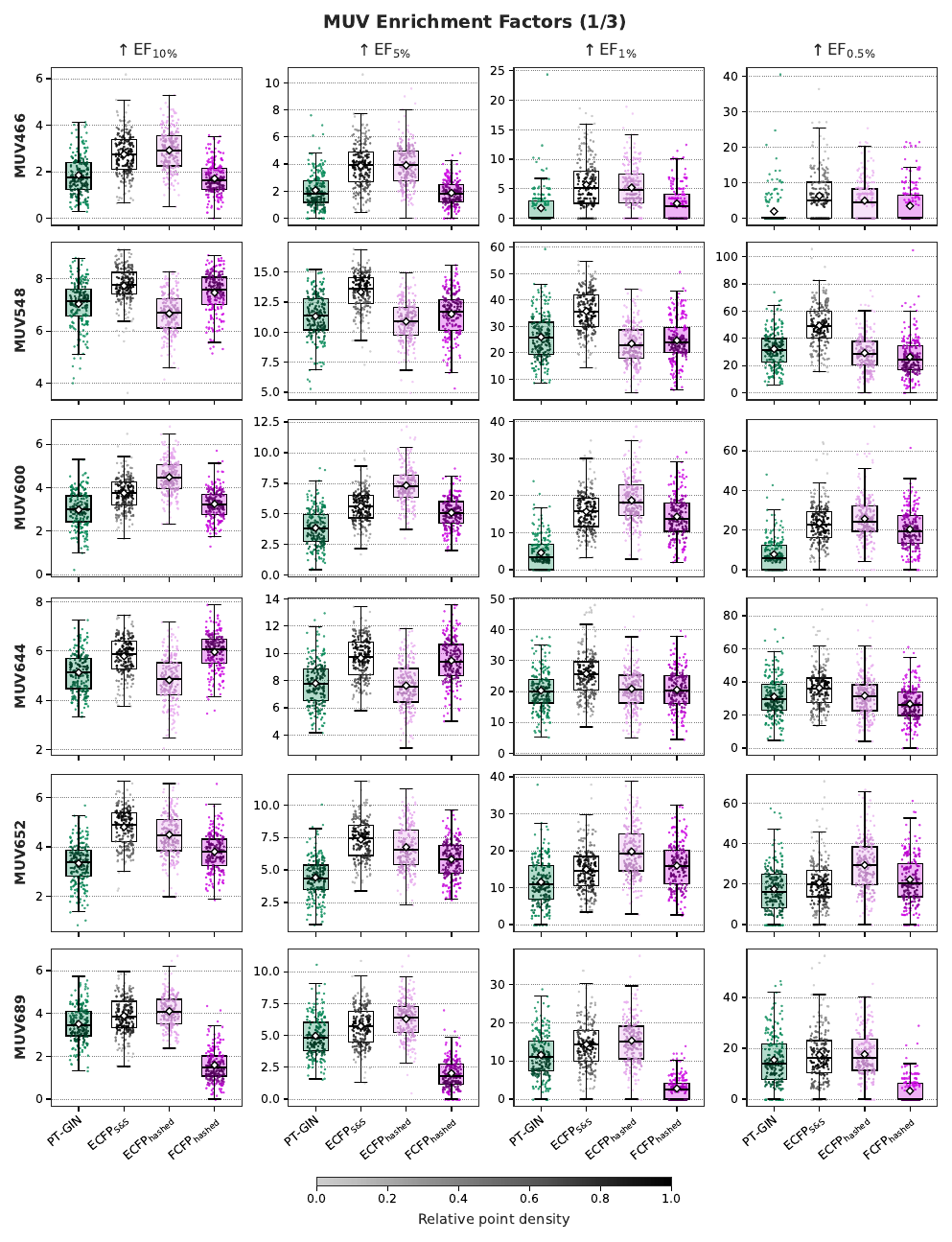}
    \caption{
        Box-and-whisker plots with point density overlay for the MUV virtual screening tasks: enrichment factors (part 1 of 3).
        Rows show the MUV466, MUV548, MUV600, MUV644, MUV652, and MUV689 tasks.
        Columns show
        enrichment factors at $10\%$ ($\uparrow \mathrm{EF_{10\%}}$),
        $5\%$ ($\uparrow \mathrm{EF_{5\%}}$),
        $1\%$ ($\uparrow \mathrm{EF_{1\%}}$), and
        $0.5\%$ ($\uparrow \mathrm{EF_{0.5\%}}$).
        Featurisation methods shown are
        ECFP-pre-trained GINs (PT-GIN),
        ECFPs vectorised via Sort \& Slice (ECFP$_{\mathrm{S\&S}}$),
        hashed Extended-Connectivity Fingerprints (ECFP$_{\mathrm{hashed}}$), and
        hashed Functional-Class Fingerprints (FCFP$_{\mathrm{hashed}}$).
        Point density is estimated using a Gaussian kernel density estimate ($n=199$).
        Diamonds are the mean of each distribution.
        Lines are the median of each distribution.
    }
    \label{fig:muv_enrichment_boxplot_1}
\end{figure}

\begin{figure}[!htbp]
    \centering
    \includegraphics[width=\textwidth, height=0.8\textheight, keepaspectratio]{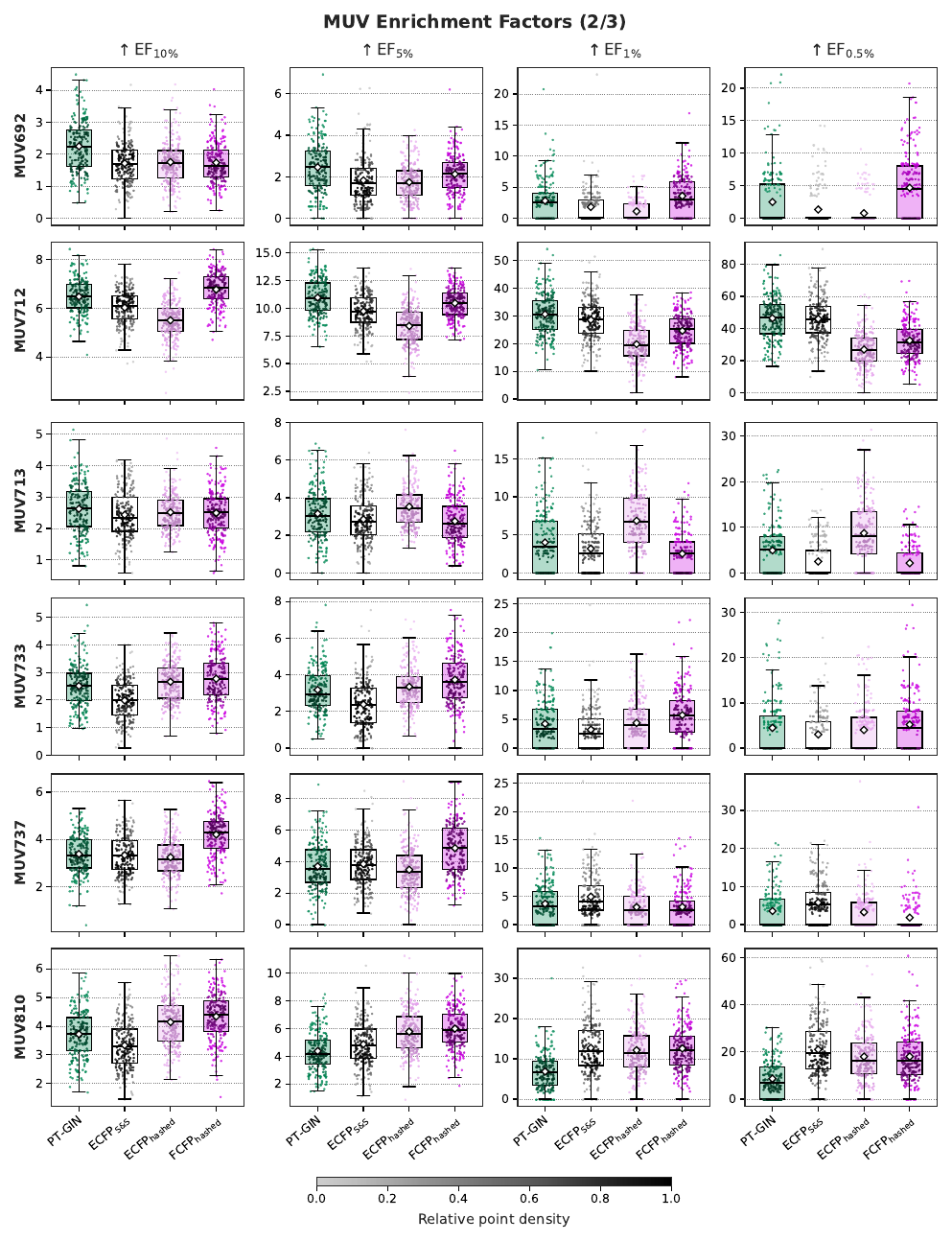}
    \caption{
        Box-and-whisker plots with point density overlay for the MUV virtual screening tasks: enrichment factors (part 2 of 3).
        Rows show the MUV692, MUV712, MUV713, MUV733, MUV737, and MUV810 tasks.
        Columns show
        enrichment factors at $10\%$ ($\uparrow \mathrm{EF_{10\%}}$),
        $5\%$ ($\uparrow \mathrm{EF_{5\%}}$),
        $1\%$ ($\uparrow \mathrm{EF_{1\%}}$), and
        $0.5\%$ ($\uparrow \mathrm{EF_{0.5\%}}$).
        Featurisation methods shown are
        ECFP-pre-trained GINs (PT-GIN),
        ECFPs vectorised via Sort \& Slice (ECFP$_{\mathrm{S\&S}}$),
        hashed Extended-Connectivity Fingerprints (ECFP$_{\mathrm{hashed}}$), and
        hashed Functional-Class Fingerprints (FCFP$_{\mathrm{hashed}}$).
        Point density is estimated using a Gaussian kernel density estimate ($n=199$).
        Diamonds are the mean of each distribution.
        Lines are the median of each distribution.
    }
    \label{fig:muv_enrichment_boxplot_2}
\end{figure}

\begin{figure}[!htbp]
    \centering
    \includegraphics[width=\textwidth, height=0.8\textheight, keepaspectratio]{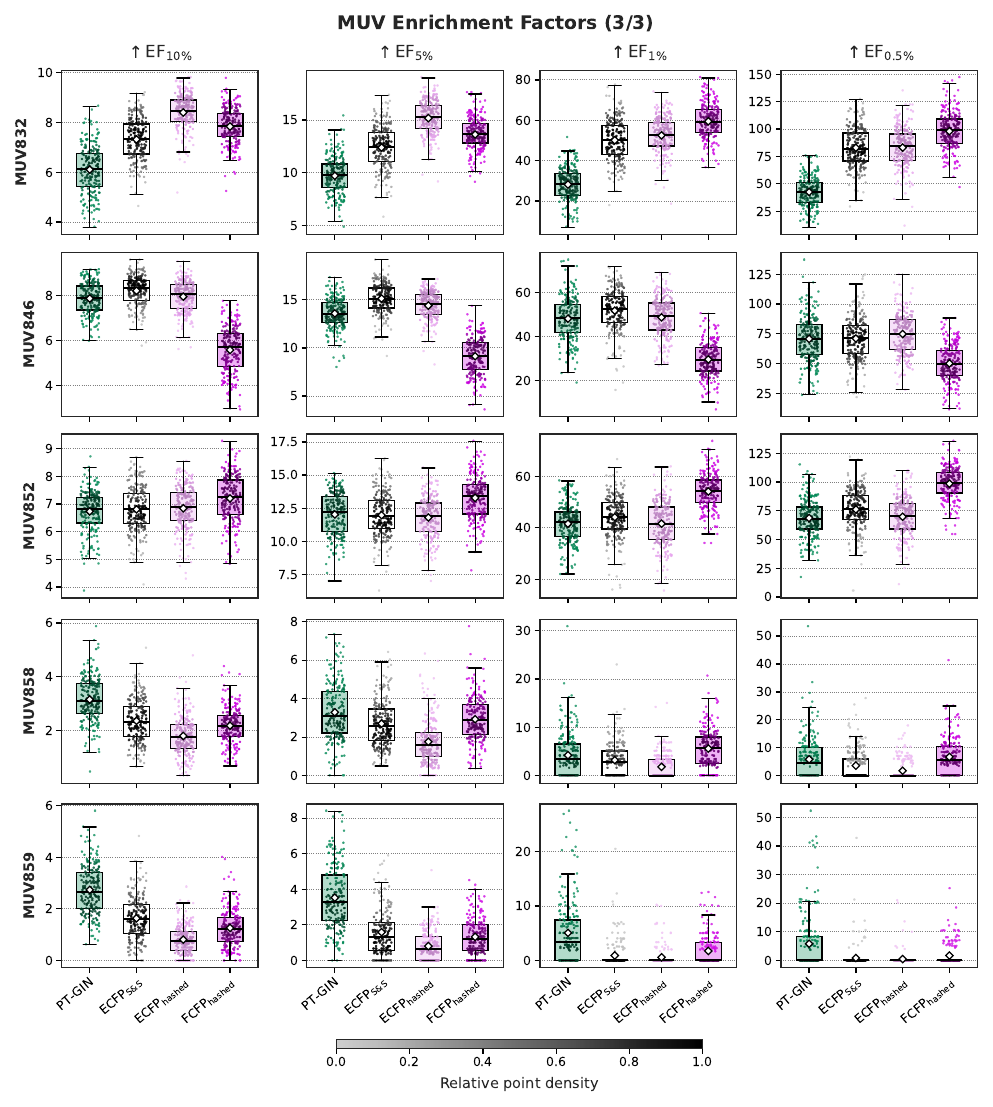}
    \caption{
        Box-and-whisker plots with point density overlay for the MUV virtual screening tasks: enrichment factors (part 3 of 3).
        Rows show the MUV832, MUV846, MUV852, MUV858, and MUV859 tasks.
        Columns show
        enrichment factors at $10\%$ ($\uparrow \mathrm{EF_{10\%}}$),
        $5\%$ ($\uparrow \mathrm{EF_{5\%}}$),
        $1\%$ ($\uparrow \mathrm{EF_{1\%}}$), and
        $0.5\%$ ($\uparrow \mathrm{EF_{0.5\%}}$).
        Featurisation methods shown are
        ECFP-pre-trained GINs (PT-GIN),
        ECFPs vectorised via Sort \& Slice (ECFP$_{\mathrm{S\&S}}$),
        hashed Extended-Connectivity Fingerprints (ECFP$_{\mathrm{hashed}}$), and
        hashed Functional-Class Fingerprints (FCFP$_{\mathrm{hashed}}$).
        Point density is estimated using a Gaussian kernel density estimate ($n=199$).
        Diamonds are the mean of each distribution.
        Lines are the median of each distribution.
    }
    \label{fig:muv_enrichment_boxplot_3}
\end{figure}

\clearpage
\newpage

\subsubsection{Tox21}

\begin{figure}[!htbp]
    \centering
    \includegraphics[width=\textwidth, height=0.75\textheight, keepaspectratio]{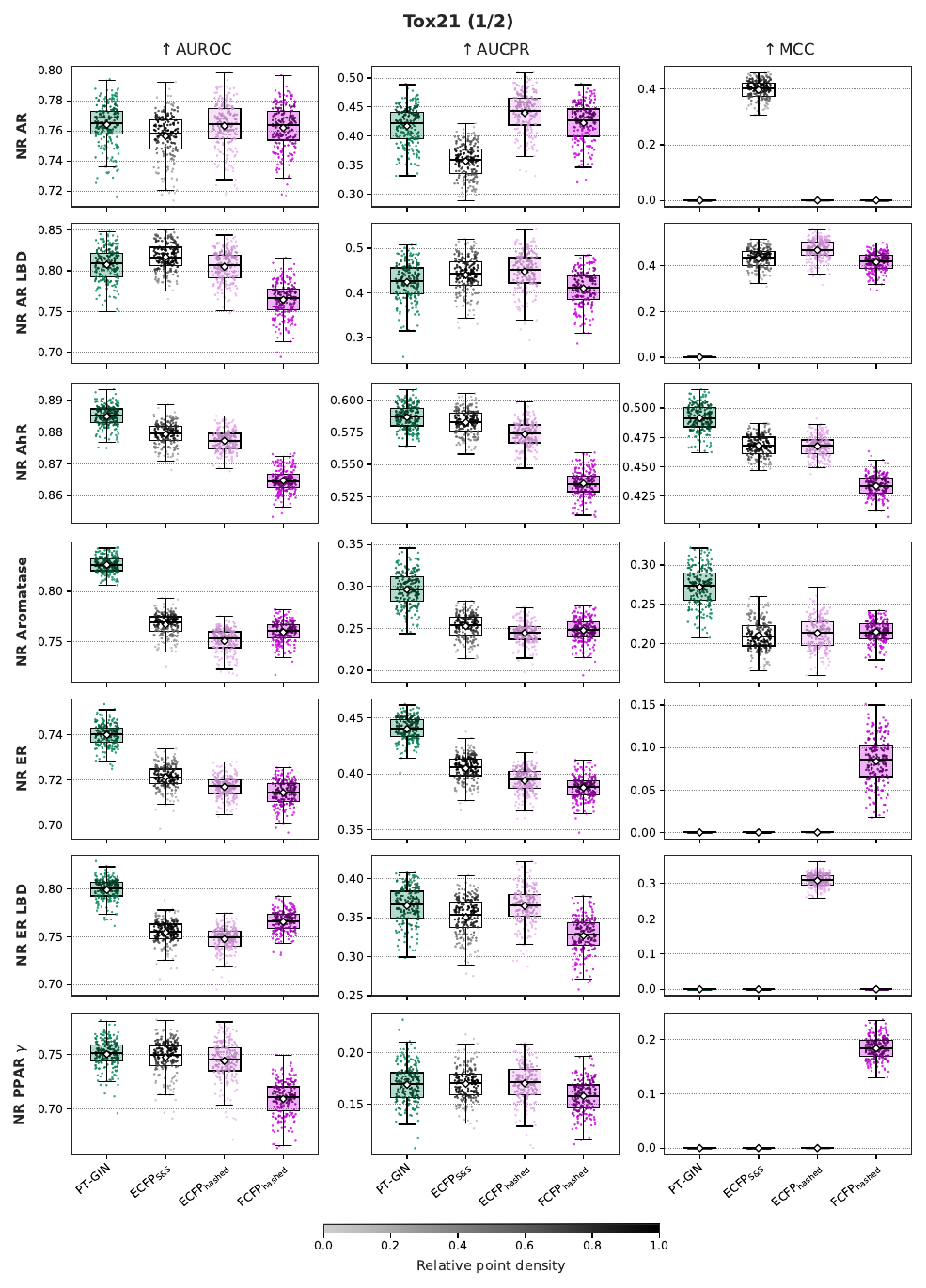}% arXiv: 0.8 in journal SI
    \caption{
        Box-and-whisker plots with point density overlay for the Tox21 nuclear receptor toxicity tasks (part 1 of 2).
        Rows show the
        NR AR (androgen receptor activation), NR AR LBD (androgen receptor ligand binding domain activation),
        NR AhR (aryl hydrocarbon receptor activation), NR Aromatase (aromatase enzyme inhibition),
        NR ER (estrogen receptor alpha activation), NR ER LBD (estrogen receptor alpha ligand binding domain activation), and
        NR PPAR $\gamma$ (peroxisome proliferator-activated receptor $\gamma$ activation) tasks.
        Columns show
        area under the receiver operating characteristic curve ($\uparrow \mathrm{AUROC}$),
        area under the precision-recall curve ($\uparrow \mathrm{AUCPR}$), and
        Matthews correlation coefficient ($\uparrow \mathrm{MCC}$).
        Featurisation methods shown are
        ECFP-pre-trained GINs (PT-GIN),
        ECFPs vectorised via Sort \& Slice (ECFP$_{\mathrm{S\&S}}$),
        hashed Extended-Connectivity Fingerprints (ECFP$_{\mathrm{hashed}}$), and
        hashed Functional-Class Fingerprints (FCFP$_{\mathrm{hashed}}$).
        Point density is estimated using a Gaussian kernel density estimate ($n=199$).
        Diamonds are the mean of each distribution.
        Lines are the median of each distribution.
    }
    \label{fig:tox21_boxplot_1}
\end{figure}

\begin{figure}[!htbp]
    \centering
    \includegraphics[width=\textwidth, height=0.8\textheight, keepaspectratio]{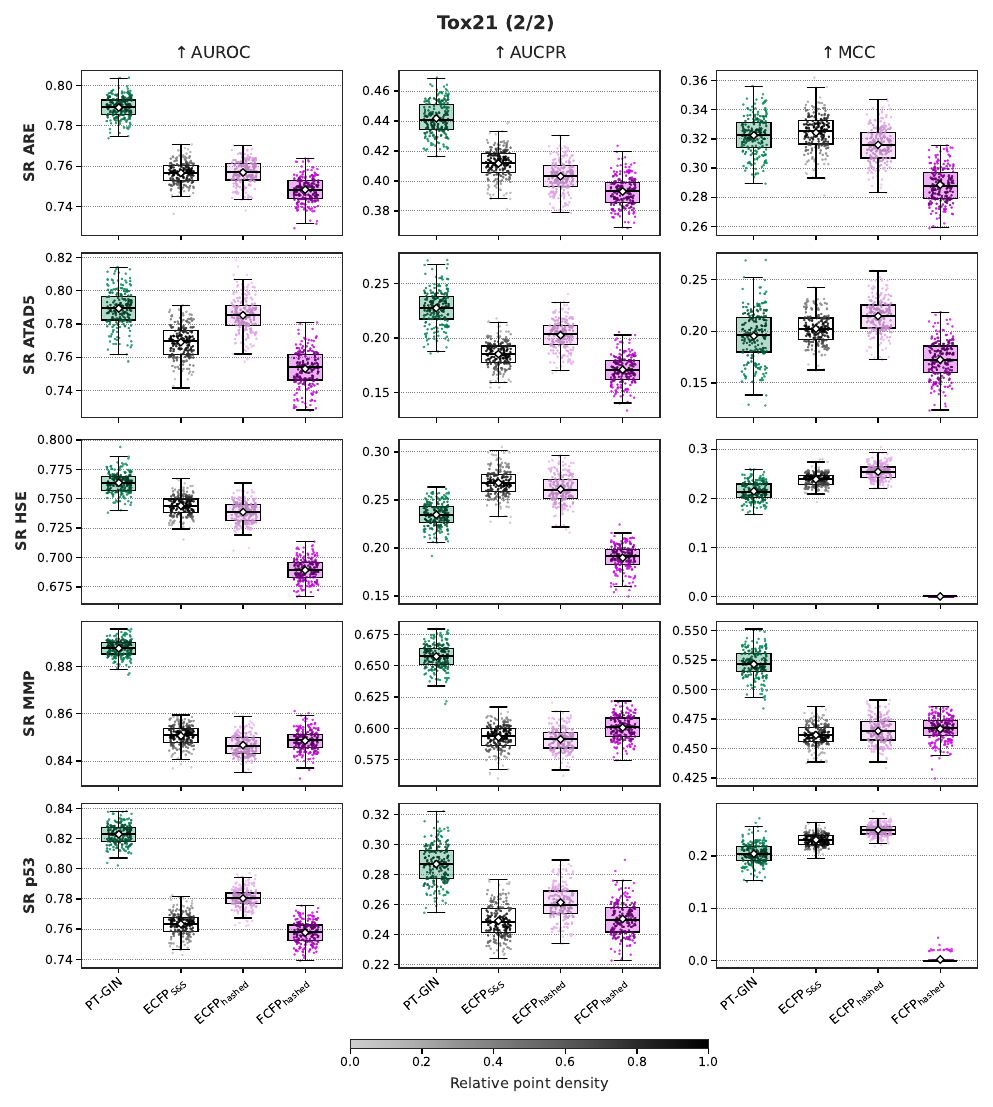}
    \caption{
        Box-and-whisker plots with point density overlay for the Tox21 stress response toxicity tasks (part 2 of 2).
        Rows show the
        SR ARE (antioxidant responsive element signalling pathway activation), SR ATAD5 (ATPase family AAA domain containing 5 expression),
        SR HSE (heat shock element signalling pathway activation), SR MMP (mitochondrial membrane potential disruption), and
        SR p53 (p53 signalling pathway activation) tasks.
        Columns show
        area under the receiver operating characteristic curve ($\uparrow \mathrm{AUROC}$),
        area under the precision-recall curve ($\uparrow \mathrm{AUCPR}$), and
        Matthews correlation coefficient ($\uparrow \mathrm{MCC}$).
        Featurisation methods shown are
        ECFP-pre-trained GINs (PT-GIN),
        ECFPs vectorised via Sort \& Slice (ECFP$_{\mathrm{S\&S}}$),
        hashed Extended-Connectivity Fingerprints (ECFP$_{\mathrm{hashed}}$), and
        hashed Functional-Class Fingerprints (FCFP$_{\mathrm{hashed}}$).
        Point density is estimated using a Gaussian kernel density estimate ($n=199$).
        Diamonds are the mean of each distribution.
        Lines are the median of each distribution.
    }
    \label{fig:tox21_boxplot_2}
\end{figure}

\clearpage
\newpage

\subsection{PT-GIN \textit{p}-values and effect sizes as a function of the number of CV repeats for the Biogen tasks}\label{si:statistical_tests}

\begin{figure}[!htbp]
    \centering
    \includegraphics[width=\textwidth, height=0.7\textheight, keepaspectratio]{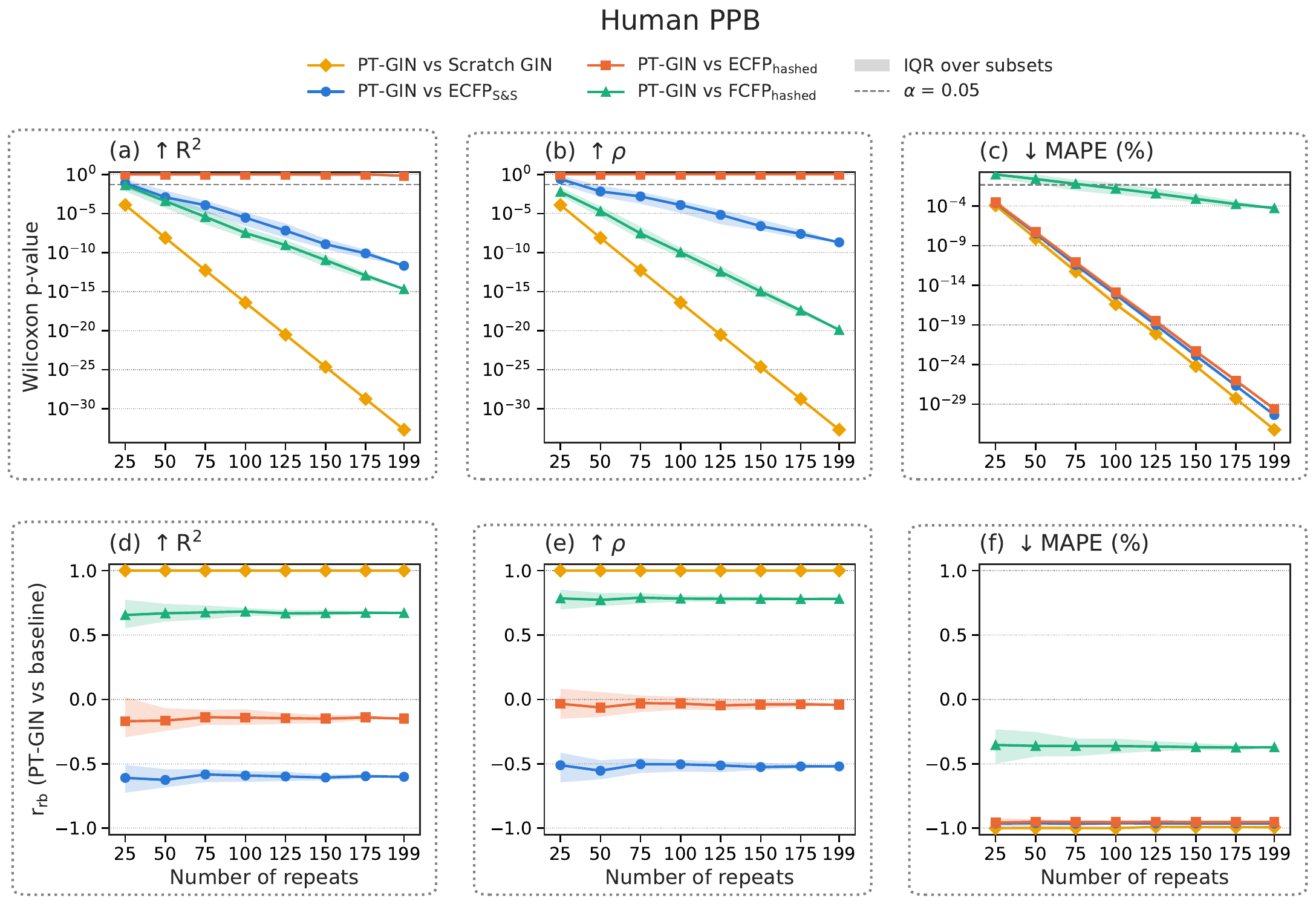}
    \caption{
        Bonferroni-corrected two-sided Wilcoxon signed-rank tests (a, b, c) and rank-biserial correlation ($r_{rb}$) effect sizes (d, e, f)
        of PT-GINs versus Scratch GIN (yellow),
        ECFP$_{\mathrm{S\&S}}$ (blue),
        ECFP$_{\mathrm{hashed}}$ (orange), and
        FCFP$_{\mathrm{hashed}}$ (green),
        as a function of the number of cross-validation
        repeats for the Biogen Human PPB task.
        Metrics shown are (a, d) coefficient of determination ($\uparrow \mathrm{R}^2$),
        (b, e) Pearson correlation coefficient ($\uparrow \rho$), and
        (c, f) mean absolute percentage error ($\downarrow\mathrm{MAPE}$ (\%)).
        CV repeats were randomly subsampled without replacement
        100 times.
        Shaded regions show the interquartile range (IQR) over the 100 subsamples;
        the dashed line marks $\alpha = 0.05$.
    }
    \label{fig:biogen_human_ppb_stats_tests}
\end{figure}

\begin{figure}[!htbp]
    \centering
    \includegraphics[width=\textwidth, height=0.7\textheight, keepaspectratio]{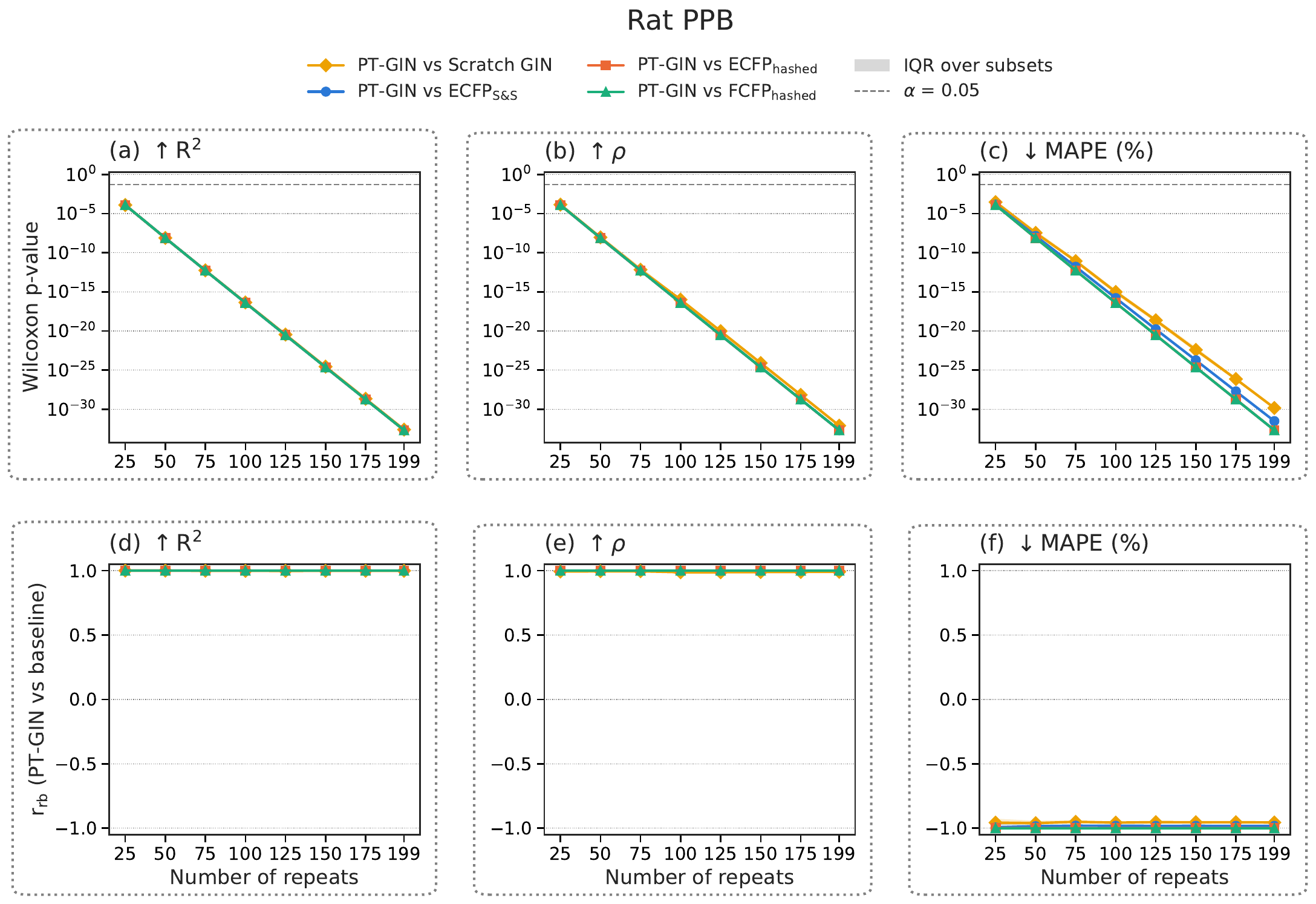}
    \caption{
        Bonferroni-corrected two-sided Wilcoxon signed-rank tests (a, b, c) and rank-biserial correlation ($r_{rb}$) effect sizes (d, e, f)
        of PT-GINs versus Scratch GIN (yellow),
        ECFP$_{\mathrm{S\&S}}$ (blue),
        ECFP$_{\mathrm{hashed}}$ (orange), and
        FCFP$_{\mathrm{hashed}}$ (green),
        as a function of the number of cross-validation
        repeats for the Biogen Rat PPB task.
        Metrics shown are (a, d) coefficient of determination ($\uparrow \mathrm{R}^2$),
        (b, e) Pearson correlation coefficient ($\uparrow \rho$), and
        (c, f) mean absolute percentage error ($\downarrow\mathrm{MAPE}$ (\%)).
        CV repeats were randomly subsampled without replacement
        100 times.
        Shaded regions show the interquartile range (IQR) over the 100 subsamples;
        the dashed line marks $\alpha = 0.05$.
    }
    \label{fig:biogen_rat_ppb_stats_tests}
\end{figure}

\begin{figure}[!htbp]
    \centering
    \includegraphics[width=\textwidth, height=0.7\textheight, keepaspectratio]{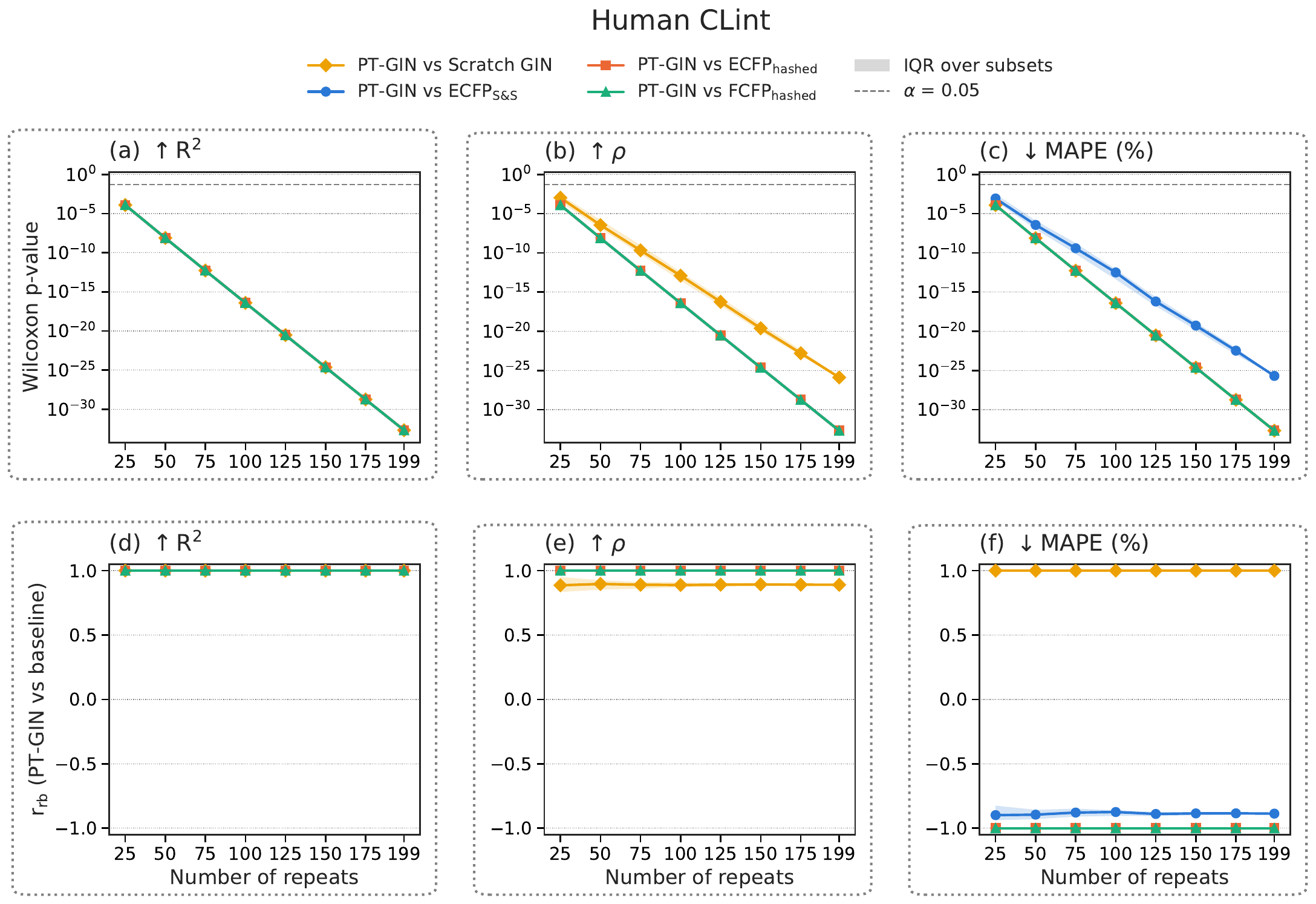}
    \caption{
        Bonferroni-corrected two-sided Wilcoxon signed-rank tests (a, b, c) and rank-biserial correlation ($r_{rb}$) effect sizes (d, e, f)
        of PT-GINs versus Scratch GIN (yellow),
        ECFP$_{\mathrm{S\&S}}$ (blue),
        ECFP$_{\mathrm{hashed}}$ (orange), and
        FCFP$_{\mathrm{hashed}}$ (green),
        as a function of the number of cross-validation
        repeats for the Biogen Human CLint task.
        Metrics shown are (a, d) coefficient of determination ($\uparrow \mathrm{R}^2$),
        (b, e) Pearson correlation coefficient ($\uparrow \rho$), and
        (c, f) mean absolute percentage error ($\downarrow\mathrm{MAPE}$ (\%)).
        CV repeats were randomly subsampled without replacement
        100 times.
        Shaded regions show the interquartile range (IQR) over the 100 subsamples;
        the dashed line marks $\alpha = 0.05$.
    }
    \label{fig:biogen_human_clint_stats_tests}
\end{figure}

\begin{figure}[!htbp]
    \centering
    \includegraphics[width=\textwidth, height=0.7\textheight, keepaspectratio]{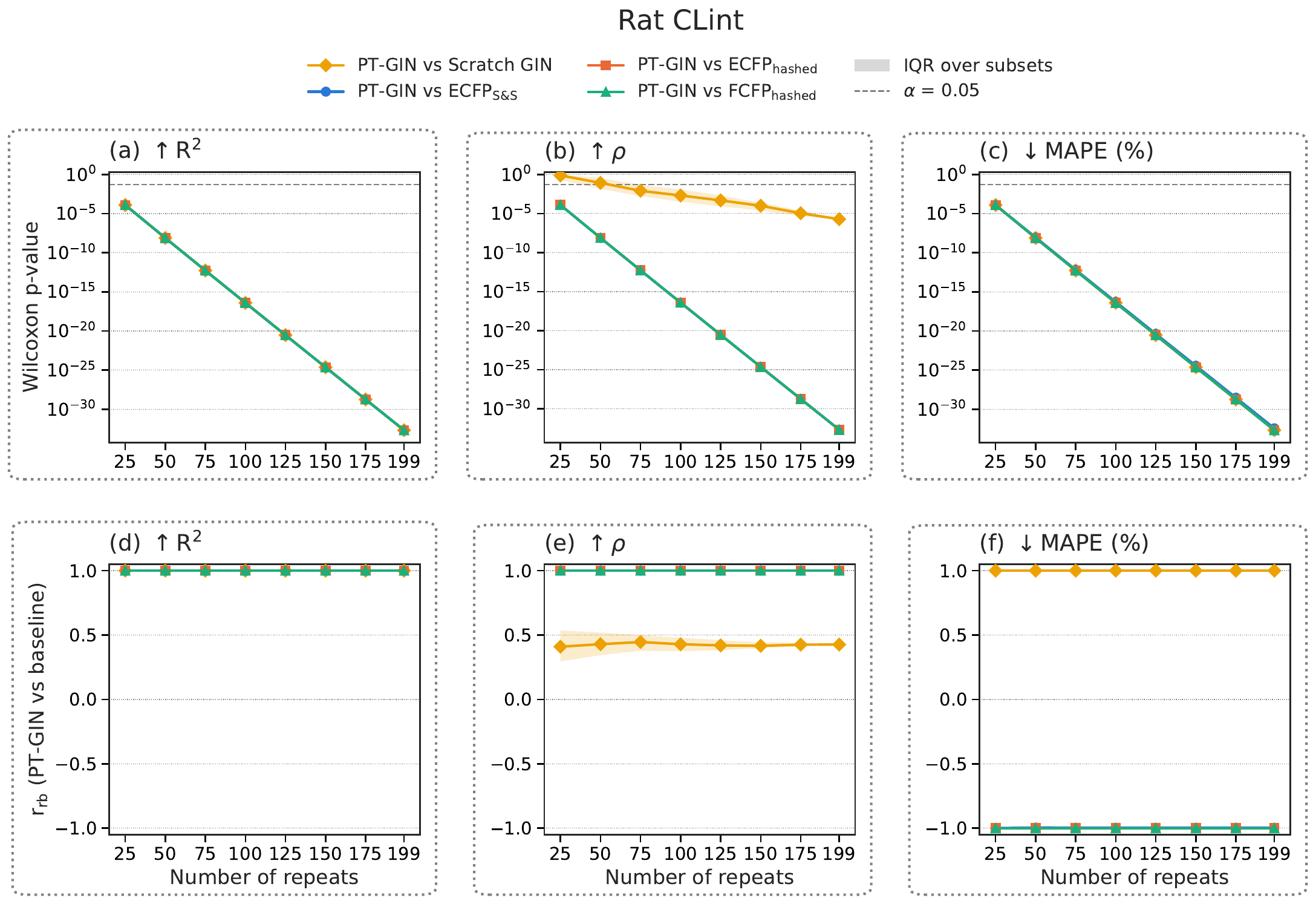}
    \caption{
        Bonferroni-corrected two-sided Wilcoxon signed-rank tests (a, b, c) and rank-biserial correlation ($r_{rb}$) effect sizes (d, e, f)
        of PT-GINs versus Scratch GIN (yellow),
        ECFP$_{\mathrm{S\&S}}$ (blue),
        ECFP$_{\mathrm{hashed}}$ (orange), and
        FCFP$_{\mathrm{hashed}}$ (green),
        as a function of the number of cross-validation
        repeats for the Biogen Rat CLint task.
        Metrics shown are (a, d) coefficient of determination ($\uparrow \mathrm{R}^2$),
        (b, e) Pearson correlation coefficient ($\uparrow \rho$), and
        (c, f) mean absolute percentage error ($\downarrow\mathrm{MAPE}$ (\%)).
        CV repeats were randomly subsampled without replacement
        100 times.
        Shaded regions show the interquartile range (IQR) over the 100 subsamples;
        the dashed line marks $\alpha = 0.05$.
    }
    \label{fig:biogen_rat_clint_stats_tests}
\end{figure}

\begin{figure}[!htbp]
    \centering
    \includegraphics[width=\textwidth, height=0.7\textheight, keepaspectratio]{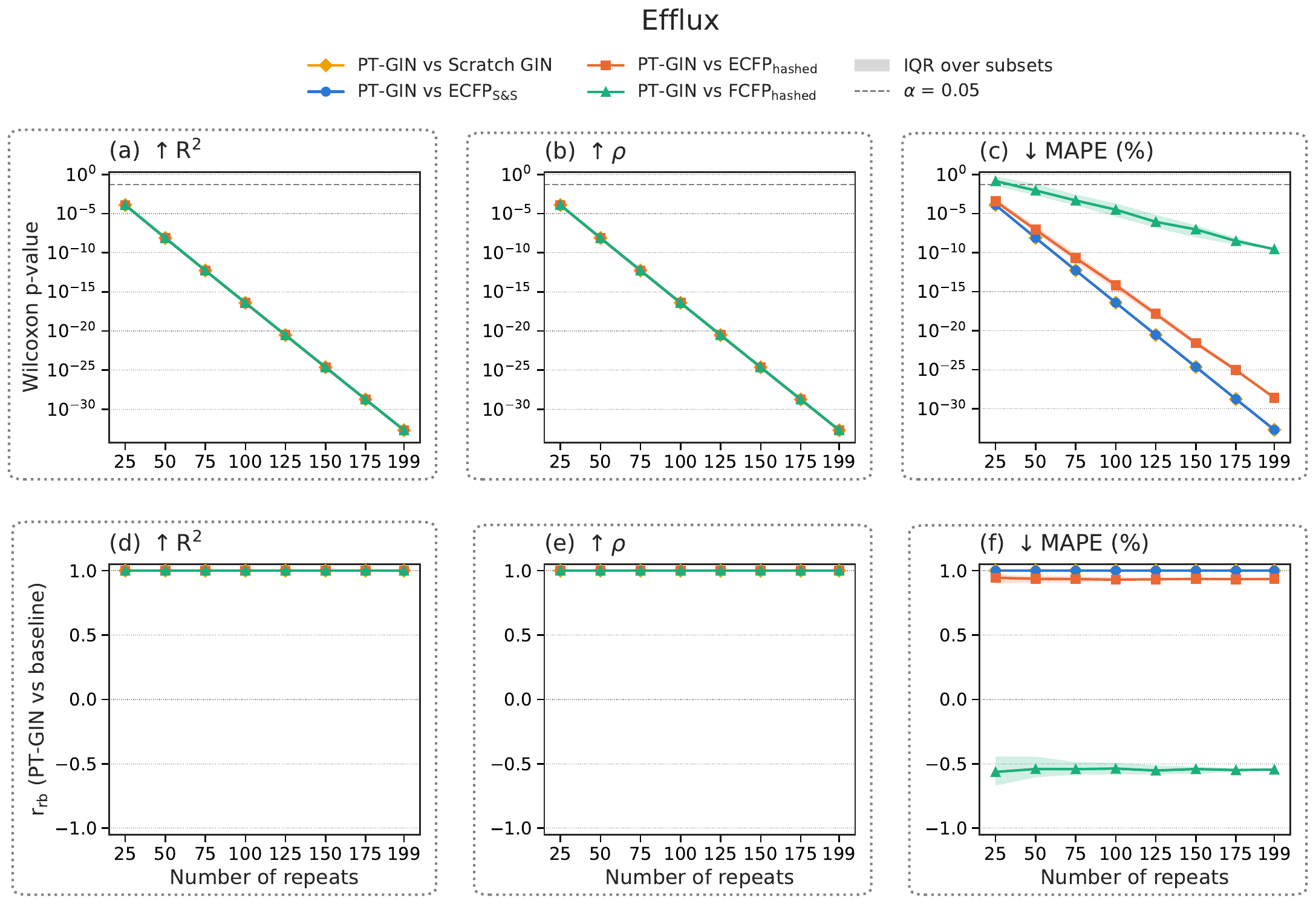}
    \caption{
        Bonferroni-corrected two-sided Wilcoxon signed-rank tests (a, b, c) and rank-biserial correlation ($r_{rb}$) effect sizes (d, e, f)
        of PT-GINs versus Scratch GIN (yellow),
        ECFP$_{\mathrm{S\&S}}$ (blue),
        ECFP$_{\mathrm{hashed}}$ (orange), and
        FCFP$_{\mathrm{hashed}}$ (green),
        as a function of the number of cross-validation
        repeats for the Biogen Efflux task.
        Metrics shown are (a, d) coefficient of determination ($\uparrow \mathrm{R}^2$),
        (b, e) Pearson correlation coefficient ($\uparrow \rho$), and
        (c, f) mean absolute percentage error ($\downarrow\mathrm{MAPE}$ (\%)).
        CV repeats were randomly subsampled without replacement
        100 times.
        Shaded regions show the interquartile range (IQR) over the 100 subsamples;
        the dashed line marks $\alpha = 0.05$.
    }
    \label{fig:biogen_efflux_stats_tests}
\end{figure}

\begin{figure}[!htbp]
    \centering
    \includegraphics[width=\textwidth, height=0.7\textheight, keepaspectratio]{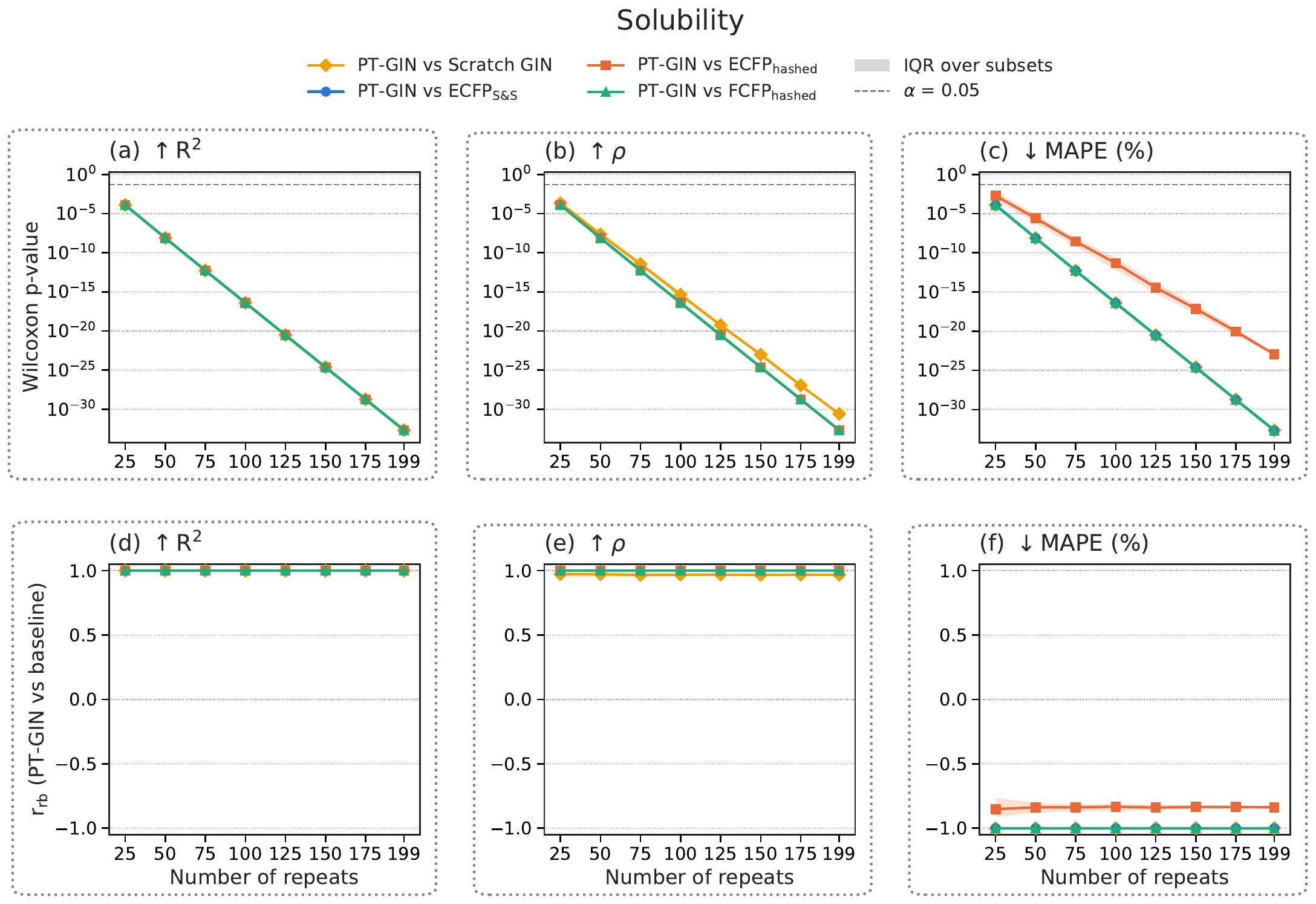}
    \caption{
        Bonferroni-corrected two-sided Wilcoxon signed-rank tests (a, b, c) and rank-biserial correlation ($r_{rb}$) effect sizes (d, e, f)
        of PT-GINs versus Scratch GIN (yellow),
        ECFP$_{\mathrm{S\&S}}$ (blue),
        ECFP$_{\mathrm{hashed}}$ (orange), and
        FCFP$_{\mathrm{hashed}}$ (green),
        as a function of the number of cross-validation
        repeats for the Biogen Solubility task.
        Metrics shown are (a, d) coefficient of determination ($\uparrow \mathrm{R}^2$),
        (b, e) Pearson correlation coefficient ($\uparrow \rho$), and
        (c, f) mean absolute percentage error ($\downarrow\mathrm{MAPE}$ (\%)).
        CV repeats were randomly subsampled without replacement
        100 times.
        Shaded regions show the interquartile range (IQR) over the 100 subsamples;
        the dashed line marks $\alpha = 0.05$.
    }
    \label{fig:biogen_solubility_stats_tests}
\end{figure}

\clearpage
\newpage

\subsection{MUV and Tox21 Friedman test and Nemenyi post-hoc test}\label{si:friedman_test}

As representations were more challenging to separate on MUV and Tox21 tasks, we performed Friedman tests and Nemenyi post-hoc tests
to determine whether there were significant differences between methods.
Nemenyi post-hoc tests were only performed for Tox21 tasks, as no significant differences were found between methods for any metric on MUV tasks.

\begin{table}[ht]
\centering
    \caption{
        Friedman test over the 17 MUV tasks
        between PT-GIN and fixed baselines, all with LightGBM.
        $\chi^2_F$: Friedman statistic ($\mathrm{df}=3$; $n=17$);
        \textit{p}: \textit{p}-value of the Friedman test ($p < 0.05$ in bold);
        $W$: Kendall's coefficient of concordance; the remaining columns are the mean rank of each method over the tasks (1 = best, best in bold).
        No significant differences were found between methods for any metric ($p > 0.05$).
    }
    \label{tab:muv_friedman}
    \begin{tabular}{lrrrrrrr}
    \toprule
    Metric & $\chi^2_F$ & \textit{p} & $W$ & PT-GIN & ECFP$_{\mathrm{S\&S}}$ & ECFP$_{\mathrm{hashed}}$ & FCFP$_{\mathrm{hashed}}$ \\
    \midrule
    $\uparrow\mathrm{AUROC}$ & 1.59 & 0.662 & 0.031 & 2.59 & 2.41 & 2.76 & \textbf{2.24} \\
    $\uparrow\mathrm{AUCPR}$ & 3.49 & 0.322 & 0.069 & 2.59 & \textbf{2.00} & 2.71 & 2.71 \\
    $\uparrow\mathrm{MCC}$ & 2.95 & 0.400 & 0.058 & 2.74 & \textbf{2.21} & 2.68 & 2.38 \\
    $\uparrow\mathrm{EF_{10\%}}$ & 0.46 & 0.928 & 0.009 & 2.65 & 2.53 & 2.47 & \textbf{2.35} \\
    $\uparrow\mathrm{EF_{5\%}}$ & 0.95 & 0.813 & 0.019 & 2.65 & \textbf{2.29} & 2.65 & 2.41 \\
    $\uparrow\mathrm{EF_{1\%}}$ & 3.14 & 0.370 & 0.062 & 2.88 & \textbf{2.12} & 2.59 & 2.41 \\
    $\uparrow\mathrm{EF_{0.5\%}}$ & 1.87 & 0.600 & 0.037 & 2.76 & \textbf{2.18} & 2.47 & 2.59 \\
    \bottomrule
    \end{tabular}
\end{table}

\begin{table}[ht]
    \centering
    \caption{Friedman test over the 12 Tox21 tasks between PT-GIN and fixed baselines, all with LightGBM.
    $\chi^2_F$: Friedman statistic ($\mathrm{df}=3$; $n=12$);
    \textit{p}: \textit{p}-value of the Friedman test ($p < 0.05$ in bold);
    $W$: Kendall's coefficient of concordance;
    the remaining columns are the mean rank of each method over the tasks (1 = best, best in bold).}
    \label{tab:tox21_friedman}
    \begin{tabular}{lrrrrrrr}
    \toprule
    Metric & $\chi^2_F$ & \textit{p} & $W$ & PT-GIN & ECFP$_{\mathrm{S\&S}}$ & ECFP$_{\mathrm{hashed}}$ & FCFP$_{\mathrm{hashed}}$ \\
    \midrule
    $\uparrow\mathrm{AUROC}$ & 24.20 & \textbf{<0.001} & 0.672 & \textbf{1.08} & 2.42 & 2.92 & 3.58 \\
    $\uparrow\mathrm{AUCPR}$ & 12.40 & \textbf{0.006} & 0.344 & \textbf{1.67} & 2.50 & 2.33 & 3.50 \\
    $\uparrow\mathrm{MCC}$ & 2.42 & 0.489 & 0.067 & 2.50 & 2.42 & \textbf{2.17} & 2.92 \\
    \bottomrule
    \end{tabular}
\end{table}

\begin{table}[ht]
    \centering
    \caption{Nemenyi post-hoc test \textit{p}-values for every pair of methods over the 12 Tox21 tasks ($p < 0.05$ in bold).}
    \label{tab:tox21_nemenyi}
    \begin{tabular}{lrrr}
    \toprule
    Comparison & $\uparrow\mathrm{AUROC}$ & $\uparrow\mathrm{AUCPR}$ & $\uparrow\mathrm{MCC}$ \\
    \midrule
    PT-GIN vs ECFP$_{\mathrm{S\&S}}$ & 0.055 & 0.389 & 0.999 \\
    PT-GIN vs ECFP$_{\mathrm{hashed}}$ & \textbf{0.003} & 0.585 & 0.922 \\
    PT-GIN vs FCFP$_{\mathrm{hashed}}$ & \textbf{<0.001} & \textbf{0.003} & 0.859 \\
    ECFP$_{\mathrm{S\&S}}$ vs ECFP$_{\mathrm{hashed}}$ & 0.778 & 0.989 & 0.965 \\
    ECFP$_{\mathrm{S\&S}}$ vs FCFP$_{\mathrm{hashed}}$ & 0.119 & 0.229 & 0.778 \\
    ECFP$_{\mathrm{hashed}}$ vs FCFP$_{\mathrm{hashed}}$ & 0.585 & 0.119 & 0.485 \\
    \bottomrule
    \end{tabular}
\end{table}

\clearpage
\newpage

\subsection{Impact of downstream model architecture}\label{si:downstream_model_ablation}

This section contains additional results for the downstream model experiment comparing Kernel Ridge regression (KR) and LightGBM,
as well as the hyperparameter search space for KR.
For KR, PT-GIN was used with a radial basis function (RBF) kernel,
and ECFP$_{\mathrm{S\&S}}$ was used with a Tanimoto kernel.

\begin{figure}[!htbp]
    \centering
    \includegraphics[width=\textwidth, height=0.7\textheight, keepaspectratio]{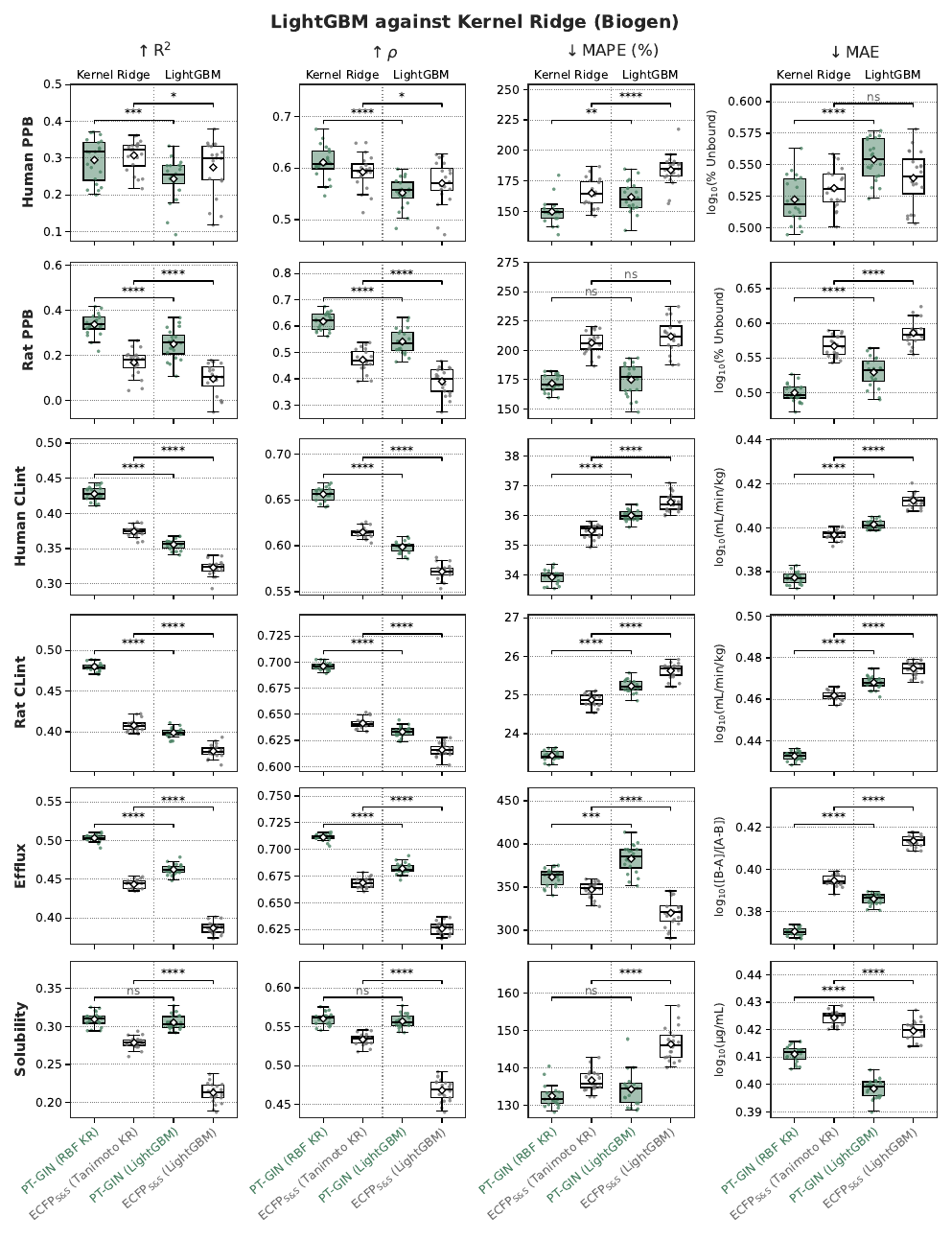}
    \caption{
        Comparison of downstream model architectures on Biogen regression tasks.
        Downstream models are Kernel Ridge regression (KR) and LightGBM.
        The input representations are PT-GIN (green) and ECFP$_{\mathrm{S\&S}}$ (grey).
        For KR, PT-GIN is used with an RBF kernel and ECFP$_{\mathrm{S\&S}}$ is used
        with a Tanimoto kernel.
        Metrics are calculated over 20 5-fold cross-validation repeats;
        each point represents the average performance across
        a single 5-fold CV repeat.
        Diamonds indicate the mean performance; the horizontal line
        within each box represents the median performance.
        Each row is a different task: Human Plasma Protein Binding (PPB), Rat PPB,
        Human Intrinsic Hepatic Clearance (CLint), Rat CLint, Efflux, and Solubility.
        Each column is a different metric: coefficient of determination ($\uparrow\mathrm{R}^2$),
        Pearson correlation coefficient ($\uparrow\rho$), mean absolute percentage error ($\downarrow$MAPE (\%)),
        and mean absolute error ($\downarrow$MAE).
        Significances are calculated using a two-sided Wilcoxon signed-rank test ($n=20$, $\alpha=0.05$)
        with Bonferroni correction for multiple comparisons.
    }\label{fig:downstream_model_ablation_biogen}
\end{figure}

\begin{figure}[!htbp]
    \centering
    \includegraphics[width=\textwidth, height=0.7\textheight, keepaspectratio]{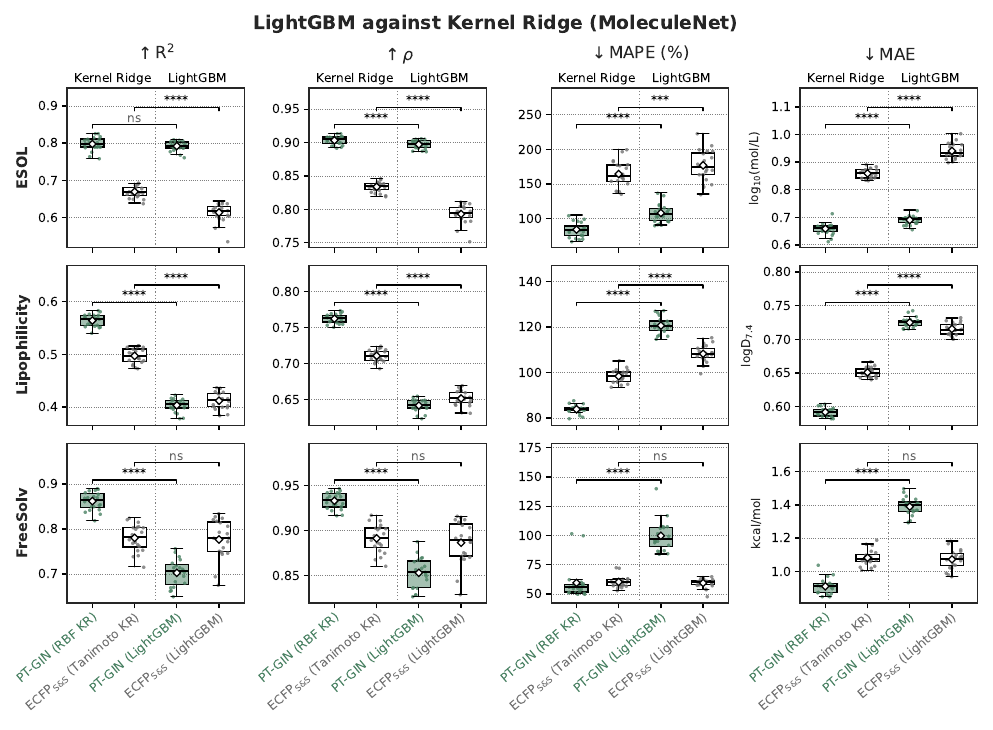}
    \caption{
        Comparison of downstream model architectures on MoleculeNet regression tasks.
        Downstream models are Kernel Ridge regression (KR) and LightGBM.
        The input representations are PT-GIN (green) and ECFP$_{\mathrm{S\&S}}$ (grey).
        For KR, PT-GIN is used with an RBF kernel and ECFP$_{\mathrm{S\&S}}$ is used
        with a Tanimoto kernel.
        Metrics are calculated over 20 5-fold cross-validation repeats;
        each point represents the average performance across
        a single 5-fold CV repeat.
        Diamonds indicate the mean performance; the horizontal line
        within each box represents the median performance.
        Each row is a different task: ESOL aqueous solubility, Lipophilicity,
        and FreeSolv free energy of solvation.
        Each column is a different metric: coefficient of determination ($\uparrow\mathrm{R}^2$),
        Pearson correlation coefficient ($\uparrow\rho$), mean absolute percentage error ($\downarrow$MAPE (\%)),
        and mean absolute error ($\downarrow$MAE).
        Significances are calculated using a two-sided Wilcoxon signed-rank test ($n=20$, $\alpha=0.05$)
        with Bonferroni correction for multiple comparisons.
    }\label{fig:downstream_model_ablation_moleculenet}
\end{figure}

\begin{figure}[!htbp]
    \centering
    \includegraphics[width=\textwidth, height=0.7\textheight, keepaspectratio]{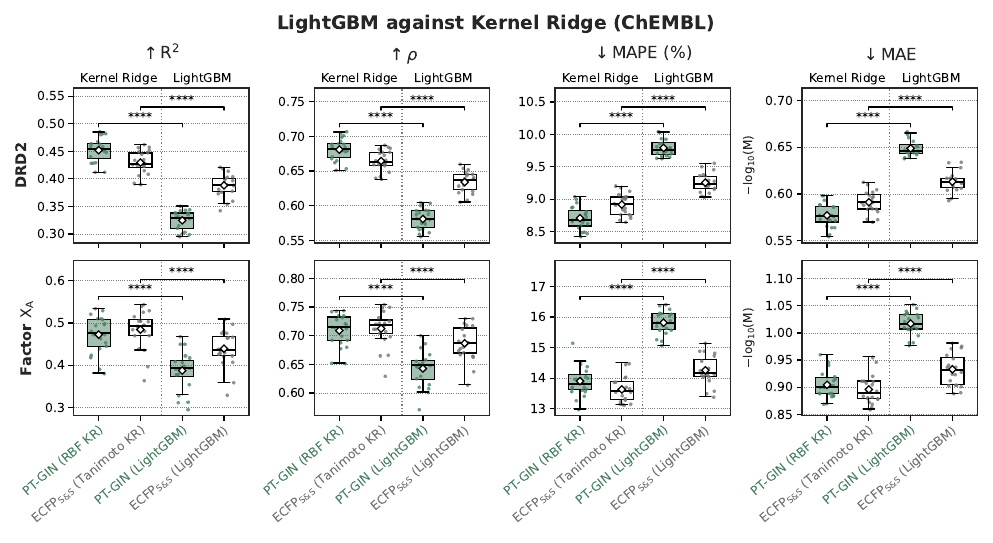}
    \caption{
        Comparison of downstream model architectures on ChEMBL binding affinity tasks.
        Downstream models are Kernel Ridge regression (KR) and LightGBM.
        The input representations are PT-GIN (green) and ECFP$_{\mathrm{S\&S}}$ (grey).
        For KR, PT-GIN is used with an RBF kernel and ECFP$_{\mathrm{S\&S}}$ is used
        with a Tanimoto kernel.
        Metrics are calculated over 20 5-fold cross-validation repeats;
        each point represents the average performance across
        a single 5-fold CV repeat.
        Diamonds indicate the mean performance; the horizontal line
        within each box represents the median performance.
        Each row is a different task: DRD2 and Factor Xa.
        Each column is a different metric: coefficient of determination ($\uparrow\mathrm{R}^2$),
        Pearson correlation coefficient ($\uparrow\rho$), mean absolute percentage error ($\downarrow$MAPE (\%)),
        and mean absolute error ($\downarrow$MAE).
        Significances are calculated using a two-sided Wilcoxon signed-rank test ($n=20$, $\alpha=0.05$)
        with Bonferroni correction for multiple comparisons.
    }\label{fig:downstream_model_ablation_chembl}
\end{figure}

\begin{figure}[!htbp]
    \centering
    \includegraphics[width=\textwidth, height=0.7\textheight, keepaspectratio]{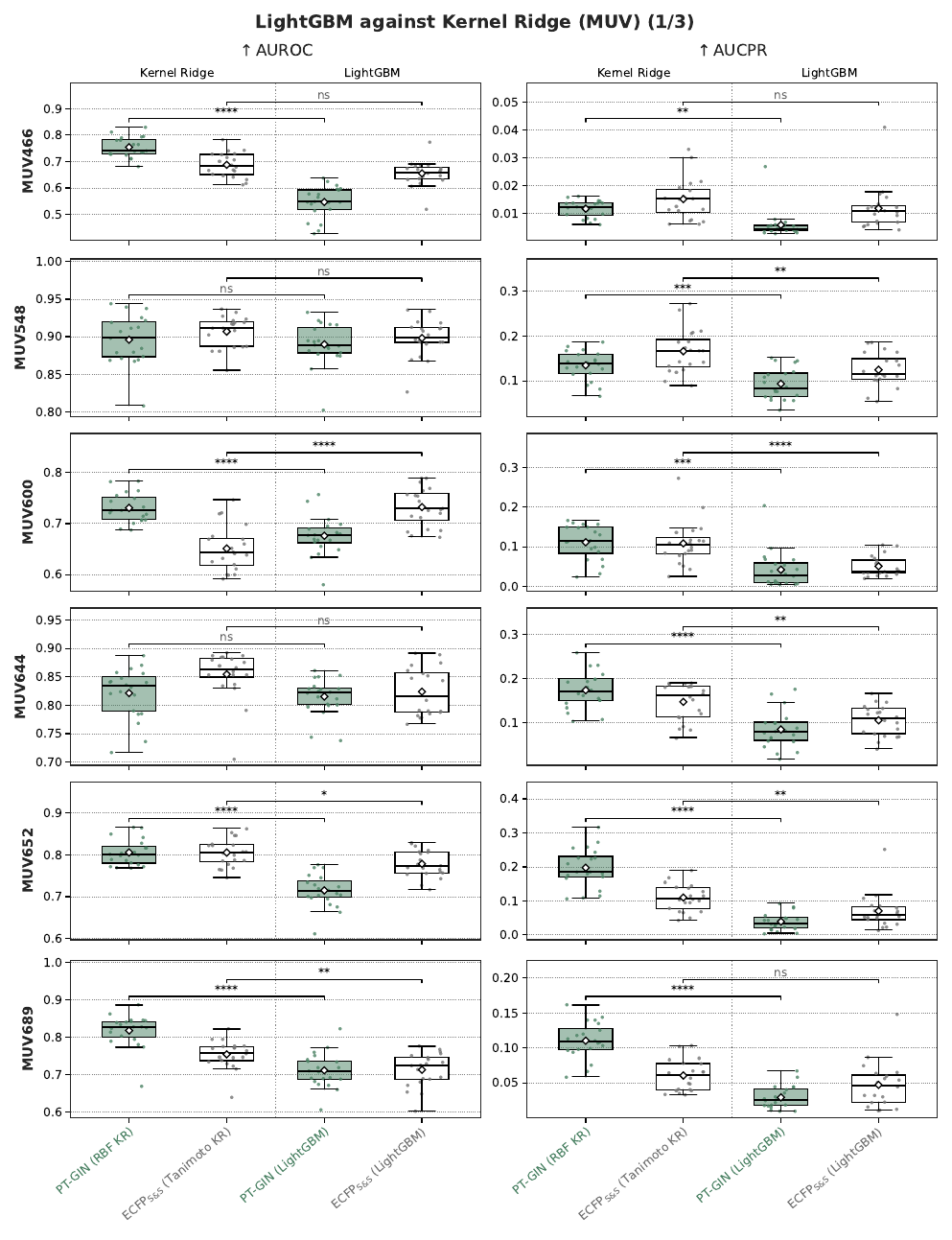}
    \caption{
        Comparison of downstream model architectures on MUV virtual screening tasks (part 1).
        Downstream models are Kernel Ridge regression (KR) and LightGBM.
        The input representations are PT-GIN (green) and ECFP$_{\mathrm{S\&S}}$ (grey).
        For KR, PT-GIN is used with an RBF kernel and ECFP$_{\mathrm{S\&S}}$ is used
        with a Tanimoto kernel.
        Metrics are calculated over 20 5-fold cross-validation repeats;
        each point represents the average performance across
        a single 5-fold CV repeat.
        Diamonds indicate the mean performance; the horizontal line
        within each box represents the median performance.
        Each row is a different MUV task and each column is a different metric ($\mathrm{AUROC}$, $\mathrm{AUCPR}$).
        Significances are calculated using a two-sided Wilcoxon signed-rank test ($n=20$, $\alpha=0.05$)
        with Bonferroni correction for multiple comparisons.
    }\label{fig:downstream_model_ablation_muv_1}
\end{figure}

\begin{figure}[!htbp]
    \centering
    \includegraphics[width=\textwidth, height=0.7\textheight, keepaspectratio]{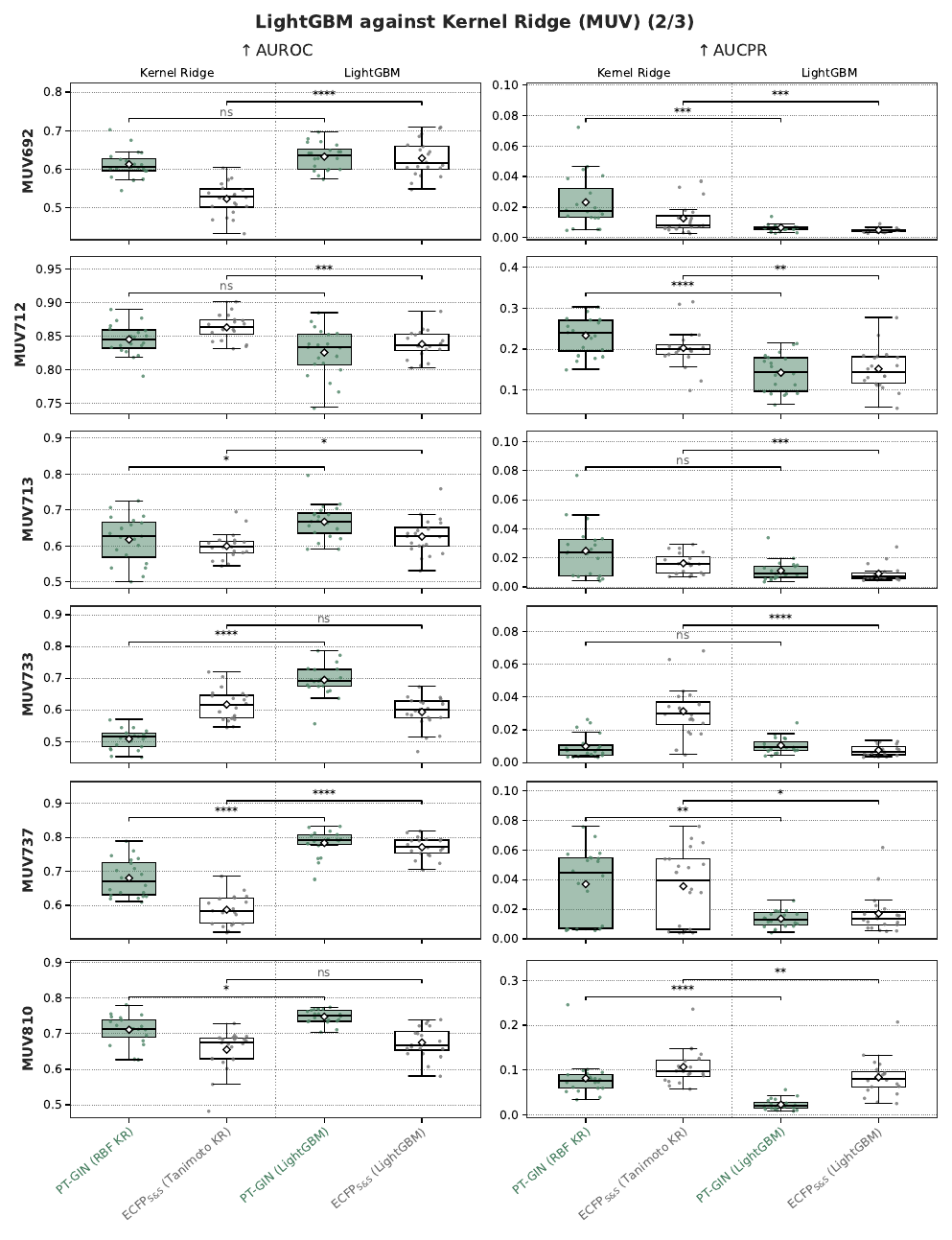}
    \caption{
        Comparison of downstream model architectures on MUV virtual screening tasks (part 2).
        Downstream models are Kernel Ridge regression (KR) and LightGBM.
        The input representations are PT-GIN (green) and ECFP$_{\mathrm{S\&S}}$ (grey).
        For KR, PT-GIN is used with an RBF kernel and ECFP$_{\mathrm{S\&S}}$ is used
        with a Tanimoto kernel.
        Metrics are calculated over 20 5-fold cross-validation repeats;
        each point represents the average performance across
        a single 5-fold CV repeat.
        Diamonds indicate the mean performance; the horizontal line
        within each box represents the median performance.
        Each row is a different MUV task and each column is a different metric ($\mathrm{AUROC}$, $\mathrm{AUCPR}$).
        Significances are calculated using a two-sided Wilcoxon signed-rank test ($n=20$, $\alpha=0.05$)
        with Bonferroni correction for multiple comparisons.
    }\label{fig:downstream_model_ablation_muv_2}
\end{figure}

\begin{figure}[!htbp]
    \centering
    \includegraphics[width=\textwidth, height=0.7\textheight, keepaspectratio]{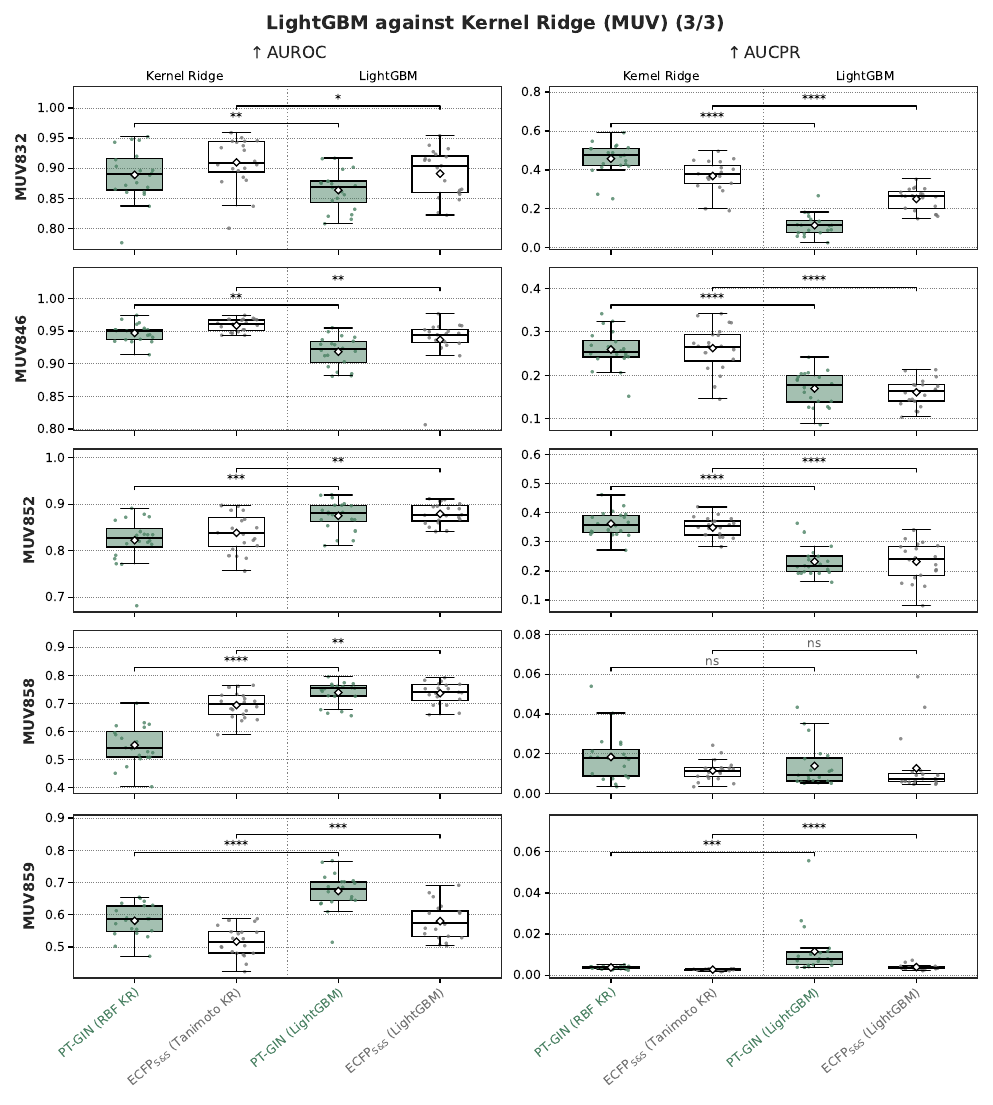}
    \caption{
        Comparison of downstream model architectures on MUV virtual screening tasks (part 3).
        Downstream models are Kernel Ridge regression (KR) and LightGBM.
        The input representations are PT-GIN (green) and ECFP$_{\mathrm{S\&S}}$ (grey).
        For KR, PT-GIN is used with an RBF kernel and ECFP$_{\mathrm{S\&S}}$ is used
        with a Tanimoto kernel.
        Metrics are calculated over 20 5-fold cross-validation repeats;
        each point represents the average performance across
        a single 5-fold CV repeat.
        Diamonds indicate the mean performance; the horizontal line
        within each box represents the median performance.
        Each row is a different MUV task and each column is a different metric ($\mathrm{AUROC}$, $\mathrm{AUCPR}$).
        Significances are calculated using a two-sided Wilcoxon signed-rank test ($n=20$, $\alpha=0.05$)
        with Bonferroni correction for multiple comparisons.
    }\label{fig:downstream_model_ablation_muv_3}
\end{figure}

\begin{figure}[!htbp]
    \centering
    \includegraphics[width=\textwidth, height=0.7\textheight, keepaspectratio]{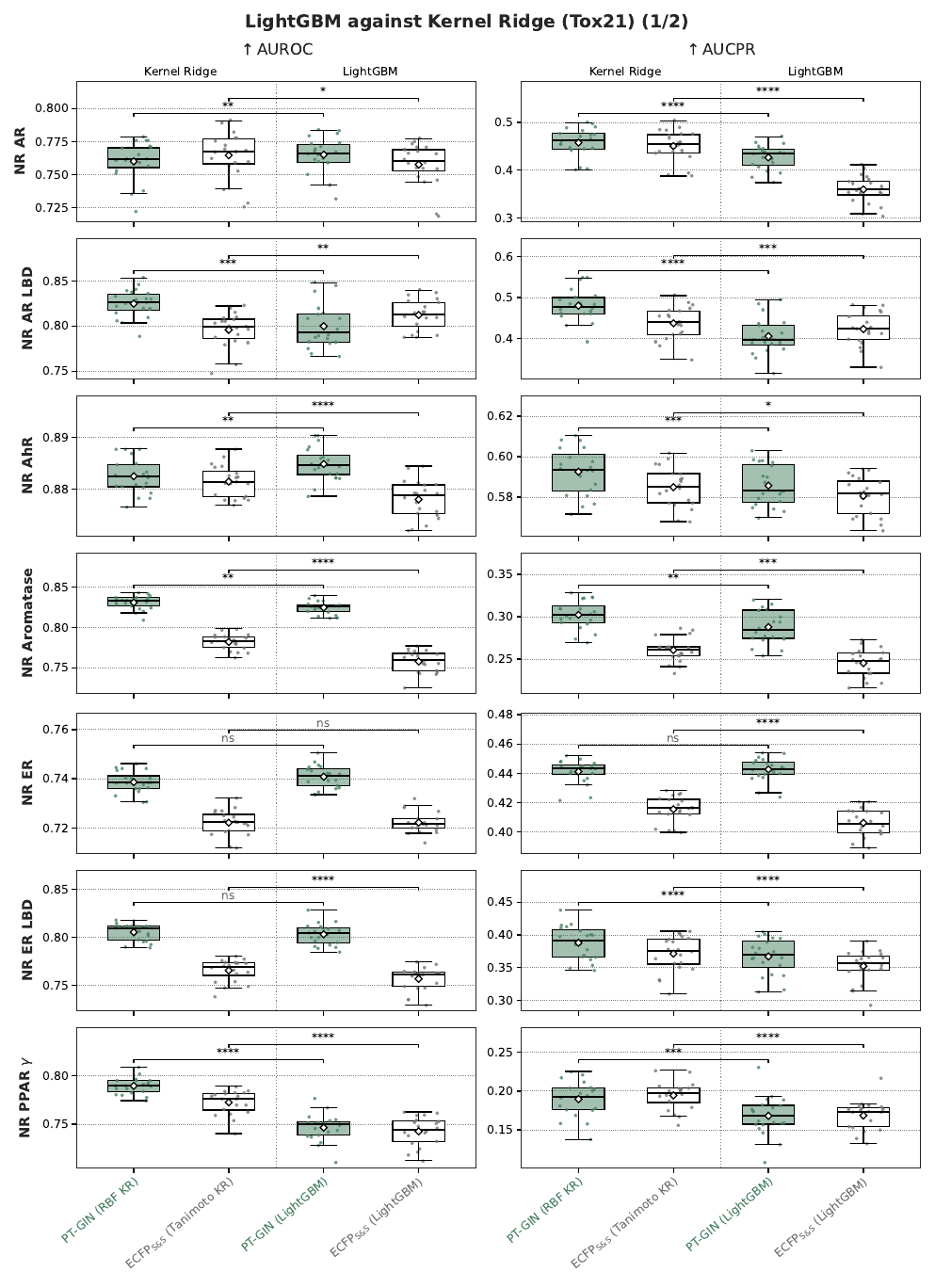}
    \caption{
        Comparison of downstream model architectures on Tox21 toxicity tasks (part 1).
        Downstream models are Kernel Ridge regression (KR) and LightGBM.
        The input representations are PT-GIN (green) and ECFP$_{\mathrm{S\&S}}$ (grey).
        For KR, PT-GIN is used with an RBF kernel and ECFP$_{\mathrm{S\&S}}$ is used
        with a Tanimoto kernel.
        Metrics are calculated over 20 5-fold cross-validation repeats;
        each point represents the average performance across
        a single 5-fold CV repeat.
        Diamonds indicate the mean performance; the horizontal line
        within each box represents the median performance.
        Each row is a different Tox21 task and each column is a different metric ($\mathrm{AUROC}$, $\mathrm{AUCPR}$).
        Significances are calculated using a two-sided Wilcoxon signed-rank test ($n=20$, $\alpha=0.05$)
        with Bonferroni correction for multiple comparisons.
    }\label{fig:downstream_model_ablation_tox21_1}
\end{figure}

\begin{figure}[!htbp]
    \centering
    \includegraphics[width=\textwidth, height=0.7\textheight, keepaspectratio]{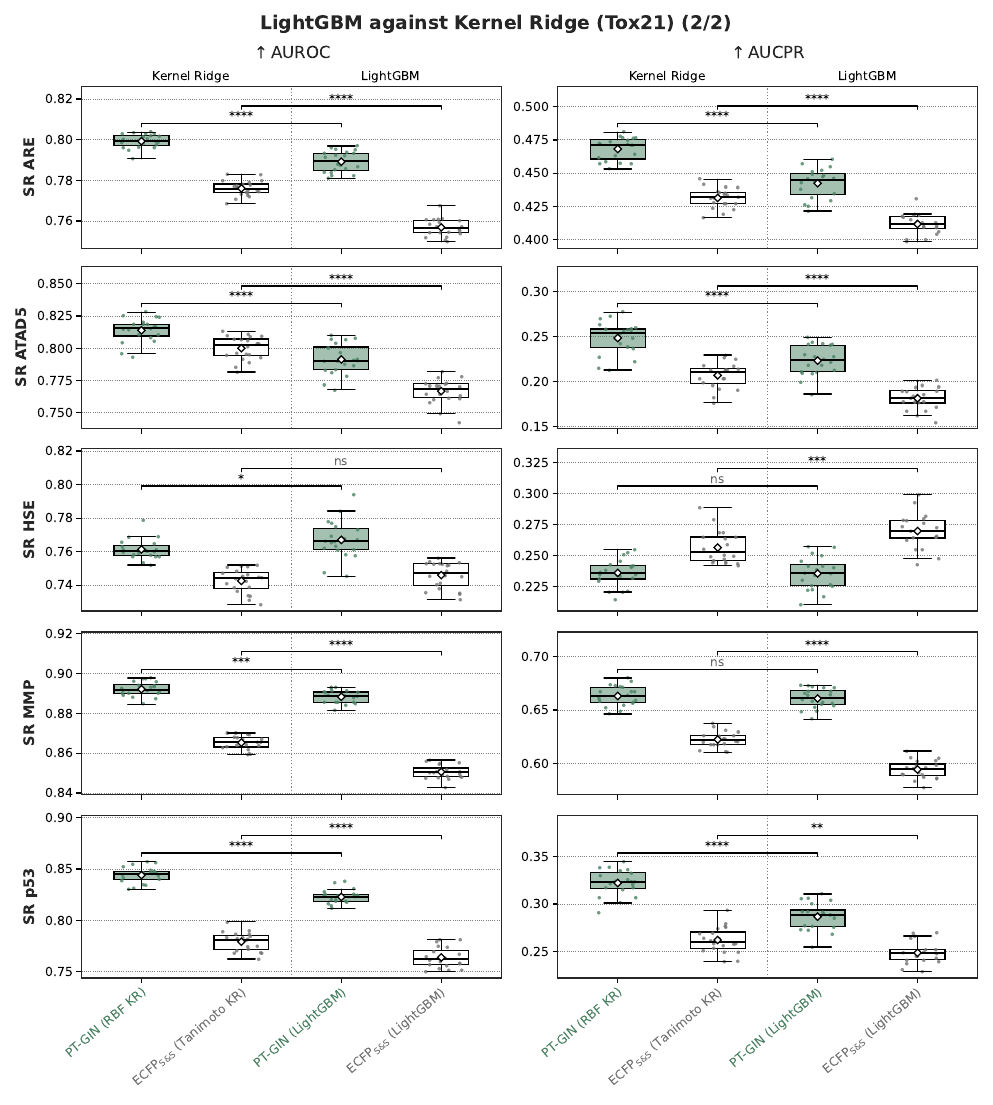}
    \caption{
        Comparison of downstream model architectures on Tox21 toxicity tasks (part 2).
        Downstream models are Kernel Ridge regression (KR) and LightGBM.
        The input representations are PT-GIN (green) and ECFP$_{\mathrm{S\&S}}$ (grey).
        For KR, PT-GIN is used with an RBF kernel and ECFP$_{\mathrm{S\&S}}$ is used
        with a Tanimoto kernel.
        Metrics are calculated over 20 5-fold cross-validation repeats;
        each point represents the average performance across
        a single 5-fold CV repeat.
        Diamonds indicate the mean performance; the horizontal line
        within each box represents the median performance.
        Each row is a different Tox21 task and each column is a different metric ($\mathrm{AUROC}$, $\mathrm{AUCPR}$).
        Significances are calculated using a two-sided Wilcoxon signed-rank test ($n=20$, $\alpha=0.05$)
        with Bonferroni correction for multiple comparisons.
    }\label{fig:downstream_model_ablation_tox21_2}
\end{figure}

\begin{table}[htbp]
\centering
\caption{
    Comparison of Kernel Ridge regression (KR) and LightGBM on PT-GIN embeddings and ECFP$_{\mathrm{S\&S}}$ fingerprints.
    Statistical significance is determined by Bonferroni-corrected two-sided Wilcoxon signed-rank tests on the 20 repeat means
    ($\alpha=0.05$).
    Each cell gives the number of tasks on which the first method of the comparison is 
    significantly better / not significantly different / significantly worse.
    KR uses an RBF kernel on the standardised PT-GIN embeddings and a Tanimoto kernel on ECFP$_{\mathrm{S\&S}}$;
    for MUV the kernels are approximated with the Nystr\"om method.
    Classification KR is class-weighted.
    }\label{tab:lightgbm_vs_kernel_ridge_wins}
    \small
    \begin{tabular}{lcccc}
        \toprule
        & \multicolumn{2}{c}{KR vs LightGBM} & \multicolumn{2}{c}{PT-GIN vs ECFP$_{\mathrm{S\&S}}$} \\
        \cmidrule(lr){2-3} \cmidrule(lr){4-5}
        Metric & PT-GIN & ECFP$_{\mathrm{S\&S}}$ & KR & LightGBM \\
        \midrule
        \multicolumn{5}{l}{\textit{Biogen (6 tasks)}} \\
        $\mathrm{R}^2$ & 5 / 1 / 0 & 6 / 0 / 0 & 5 / 1 / 0 & 5 / 1 / 0 \\
        $\rho$ & 5 / 1 / 0 & 6 / 0 / 0 & 6 / 0 / 0 & 5 / 1 / 0 \\
        MAPE (\%) & 4 / 2 / 0 & 4 / 1 / 1 & 5 / 0 / 1 & 5 / 0 / 1 \\
        MAE & 5 / 0 / 1 & 4 / 1 / 1 & 5 / 1 / 0 & 5 / 1 / 0 \\
        \midrule
        \multicolumn{5}{l}{\textit{MoleculeNet (3 tasks)}} \\
        $\mathrm{R}^2$ & 2 / 1 / 0 & 2 / 1 / 0 & 3 / 0 / 0 & 1 / 1 / 1 \\
        $\rho$ & 3 / 0 / 0 & 2 / 1 / 0 & 3 / 0 / 0 & 1 / 0 / 2 \\
        MAPE (\%) & 3 / 0 / 0 & 2 / 1 / 0 & 2 / 1 / 0 & 1 / 0 / 2 \\
        MAE & 3 / 0 / 0 & 2 / 1 / 0 & 3 / 0 / 0 & 1 / 0 / 2 \\
        \midrule
        \multicolumn{5}{l}{\textit{ChEMBL (2 tasks)}} \\
        $\mathrm{R}^2$ & 2 / 0 / 0 & 2 / 0 / 0 & 1 / 1 / 0 & 0 / 0 / 2 \\
        $\rho$ & 2 / 0 / 0 & 2 / 0 / 0 & 1 / 1 / 0 & 0 / 0 / 2 \\
        MAPE (\%) & 2 / 0 / 0 & 2 / 0 / 0 & 1 / 0 / 1 & 0 / 0 / 2 \\
        MAE & 2 / 0 / 0 & 2 / 0 / 0 & 1 / 1 / 0 & 0 / 0 / 2 \\
        \midrule
        \multicolumn{5}{l}{\textit{MUV (17 tasks)}} \\
        AUROC & 6 / 4 / 7 & 5 / 5 / 7 & 7 / 5 / 5 & 4 / 8 / 5 \\
        AUCPR & 13 / 3 / 1 & 13 / 3 / 1 & 7 / 7 / 3 & 3 / 11 / 3 \\
        \midrule
        \multicolumn{5}{l}{\textit{Tox21 (12 tasks)}} \\
        AUROC & 7 / 2 / 3 & 9 / 2 / 1 & 10 / 1 / 1 & 10 / 2 / 0 \\
        AUCPR & 9 / 3 / 0 & 11 / 0 / 1 & 10 / 1 / 1 & 9 / 1 / 2 \\
        \bottomrule
    \end{tabular}
\end{table}

\begin{table}[htbp]
    \centering
    \caption{
        Hyperparameter search space for KR with RBF and Tanimoto kernels.
        $\alpha$ is the regularisation parameter and $\gamma$ is the kernel coefficient for RBF.
    }
    \begin{tabular}{lcc}
        \toprule
        Kernel & Hyperparameter & Search Space \\
        \midrule
        RBF & $\alpha$ & $[10^{-3}, 10^{-2}, 10^{-1}, 1, 10]$ \\
        RBF & $\gamma$ & $[10^{-1}, 1, 10]$ \\
        Tanimoto & $\alpha$ & $[10^{-5}, 10^{-4}, 10^{-3}, 10^{-2}, 10^{-1}, 1, 10]$ \\
        \bottomrule
    \end{tabular}
    \label{tab:kr_hyperparameter_search_space}
\end{table}

\clearpage
\newpage

\subsection{Determining substructure importance via permutation}\label{si:substructure_importance}

As both ECFPs vectorised via Sort \& Slice (ECFP$_{\mathrm{S\&S}}$)~\cite{dablanderM-2024-SortSlice} 
and PT-GIN leverage substructure representations---ECFP$_{\mathrm{S\&S}}$ maps substructures to bits and PT-GIN maps substructures to tokens in molecular graphs---the relative importance of substructures can be compared.
We investigated substructure importance for LightGBM models trained on 
ECFP$_{\mathrm{S\&S}}$ and PT-GIN with the FreeSolv dataset~\cite{mobleyDL-2014-FreeSolv},
along with additional examples on other datasets.
We implemented permutation importance~\cite{breimanL-2001-RandomForests}
for ECFP$_{\mathrm{S\&S}}$, and we implemented embedding permutation on substructure tokens 
prior to message passing for PT-GIN
    (see main text, Methods, \textit{Substructure importance}).

\begin{figure}[!htbp]
    \centering
    \includegraphics[height=0.8\textwidth, width=\textwidth,keepaspectratio]{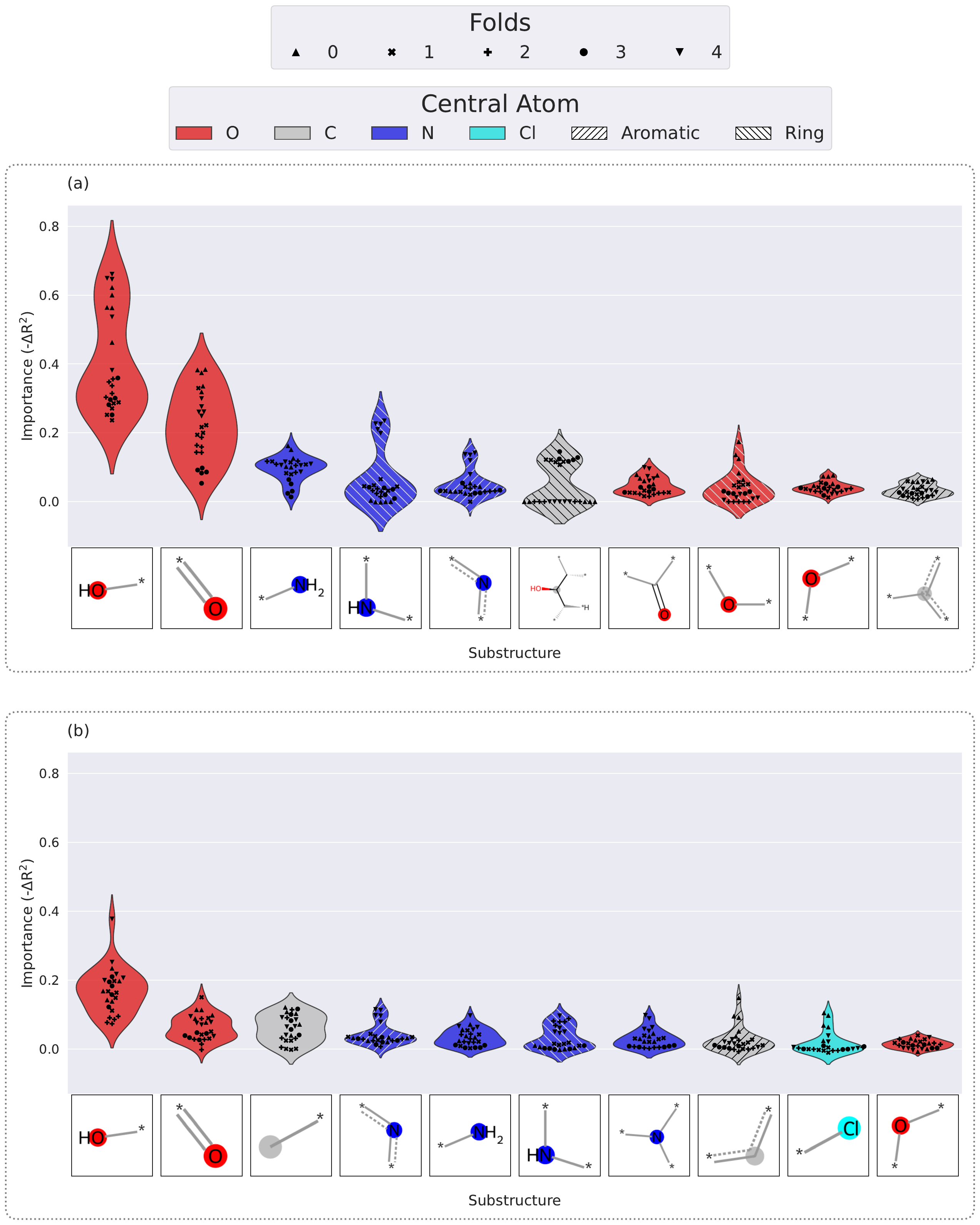}
    \caption{
        Substructure importances on the FreeSolv dataset. 
        (a) Permutation feature importances for ECFP$_{\mathrm{S\&S}}$, mapped to substructures. 
        (b) Token embedding importances for PT-GIN, mapped to substructures. 
        For both (a) and (b), importances were measured as $-\Delta\mathrm{R}^2$ 
        on the test set upon permutation. 
        Only the top ten ranked substructures by mean importance are shown. 
        Violin plots are coloured by central atom element type; 
        hatches from the bottom left to the top right indicate the central atom is aromatic;
        hatches from the bottom right to the top left indicate the central atom is in a non-aromatic ring. 
        Individual scores are shown as a swarm plot; marker shape indicates the test set fold.
    }
    \label{fig:feat-import}
\end{figure}

We found that pharmacophoric features commonly associated with a high magnitude of free energy of solvation
(such as polar groups and hydrogen bond donors) are present in high-ranking substructures for both approaches. 
Hydroxyls are the top-ranked substructure by importance
when predicting free energy of solvation ($\Delta \mathrm{G_{solv}}$)
for both ECFP$_{\mathrm{S\&S}}$ and PT-GIN
    (Fig.~\ref{fig:feat-import}).
For ECFP$_{\mathrm{S\&S}}$, five other oxygen-containing substructures rank in the 
top ten, including esters, ketones, and cycloalkanols
    (Fig.~\ref{fig:feat-import}a). 
Ketones and esters also rank highly for PT-GIN
    (Fig.~\ref{fig:feat-import}b).
For both featurisation approaches, models also consider nitrogenous substructures important, 
with three substructures containing a central nitrogen atom appearing for ECFP$_{\mathrm{S\&S}}$, and four for PT-GIN. 
PT-GIN additionally ranks a haloalkane
    (Fig.~\ref{fig:feat-import}b). 
The overt presence of oxygen and nitrogen is reasonable; 
both elements are indicative of dipole moments and zwitterions, 
which enable a molecule to form electrostatic interactions with water, 
reducing $\Delta \mathrm{G_{solv}}$. 
One would expect molecules with functional groups containing 
the majority of the top-ranked substructures to have a lower 
$\Delta \mathrm{G_{solv}}$.
Collectively, this suggests that both featurisation approaches effectively 
capture relevant features for predicting $\Delta \mathrm{G_{solv}}$.

\begin{figure}[!htbp]
    \centering
    \includegraphics[width=0.7\textwidth,keepaspectratio]{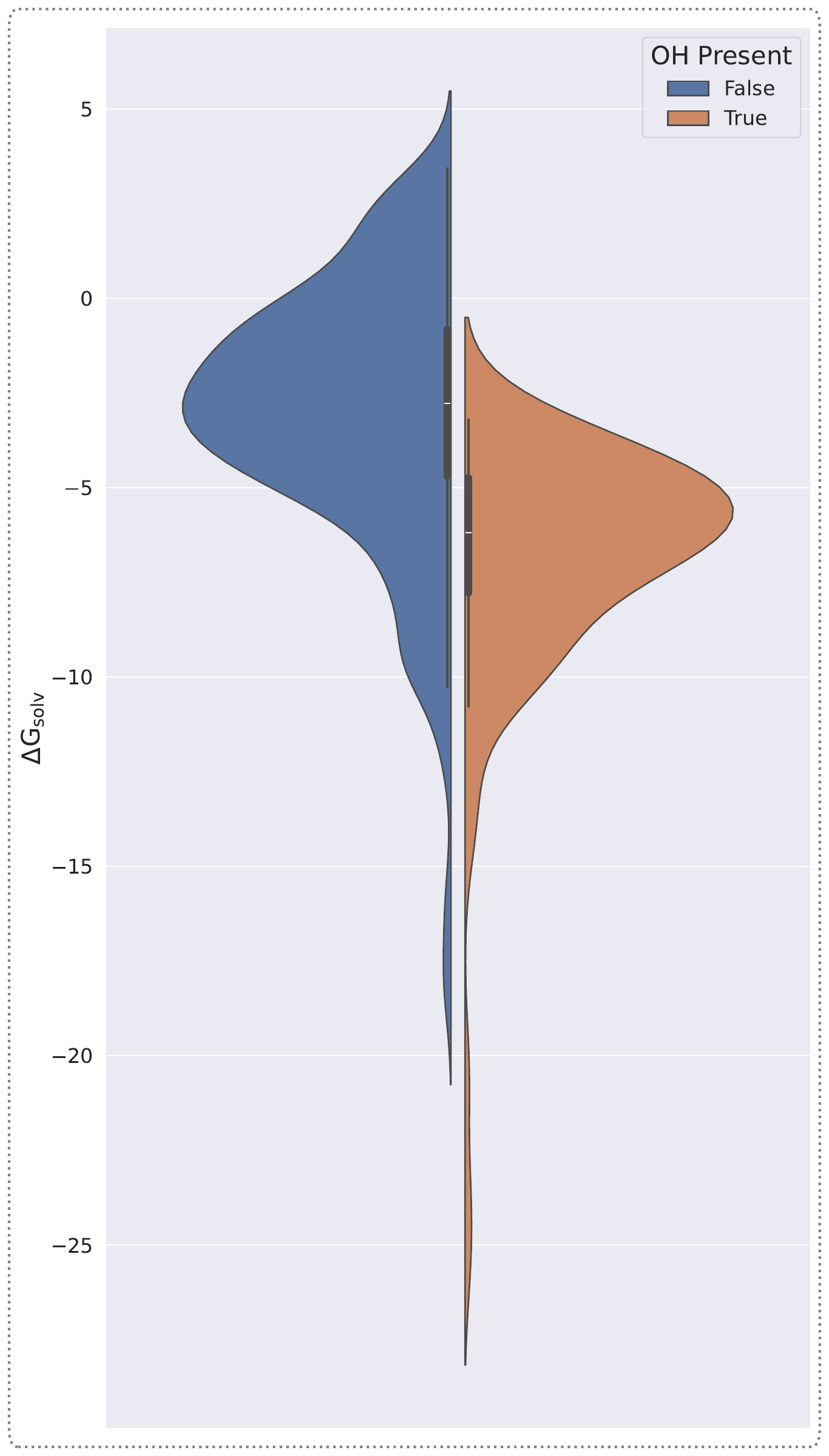}
    \caption{Distribution of $\Delta \mathrm{G_{solv}}$ in the FreeSolv dataset for molecules with and without R-OH.}
    \label{fig:freesolv_oh}
\end{figure}

The models differ noticeably in their reliance on individual substructures. 
Models trained on ECFP$_{\mathrm{S\&S}}$ exhibit a stronger reliance on a specific substructure compared to 
those trained on PT-GIN.
There is a clear skew in importance towards the top-ranked hydroxyl substructure using ECFP$_{\mathrm{S\&S}}$
    (Fig.~\ref{fig:feat-import}a);
$\mathrm{R}^2$ reduces by over $0.6$ in four instances with ECFP$_{\mathrm{S\&S}}$.
For folds $0$ and $4$ with ECFP$_{\mathrm{S\&S}}$, 
every importance measurement except one exceeds a 
$0.4$ decrease in $\mathrm{R}^2$.
Hydroxyls are also the top-ranked substructure for PT-GIN,
with R-OH being the only substructure where the majority of measurements
have a $-\Delta \mathrm{R}^2>0.1$; however, the skew is not as extreme
    (Fig.~\ref{fig:feat-import}b).
Models, especially those trained on ECFP$_{\mathrm{S\&S}}$, 
are potentially relying on a single substructure as a proxy 
for predicting $\Delta \mathrm{G_{solv}}$. 
Comparing $\Delta \mathrm{G_{solv}}$ values for molecules with and 
without R-OH reveals a distribution shift
    (Fig.~\ref{fig:freesolv_oh}).
For FreeSolv, a model trained on a representation that conveys hydroxyl presence may be
prone to reliance on a single feature for a complex task.
Predicting $\Delta \mathrm{G_{solv}}$ only on molecules with R-OH substructures 
might be a more accurate measure of model robustness.
The decrease in reliance on hydroxyl presence of PT-GIN compared to ECFP$_{\mathrm{S\&S}}$ suggests 
that PT-GIN relies less on a single substructure in this instance; 
this may indicate a loss of atom-level information from graph pooling.

Our novel graph tokenisation approach based on local circular substructures
enabled analysis of substructure importance in PT-GINs, with direct comparison to ECFP$_{\mathrm{S\&S}}$,
providing interpretable insight into model behaviour and chemical understanding.
Our case study on the FreeSolv benchmark revealed that ECFP$_{\mathrm{S\&S}}$ may
depend heavily on a single, dominant chemical feature (e.g., hydroxyls),
a shortcut that PT-GINs were less prone to.
This capability allows researchers to move beyond simple performance metrics
and ask why a model is making its predictions on a dataset-wide level, enabling the diagnosis of
model weaknesses and the identification of biases in benchmark datasets themselves.

\begin{figure}[!htbp]
    \centering
    \includegraphics[width=\textwidth, height=0.8\textheight, keepaspectratio]{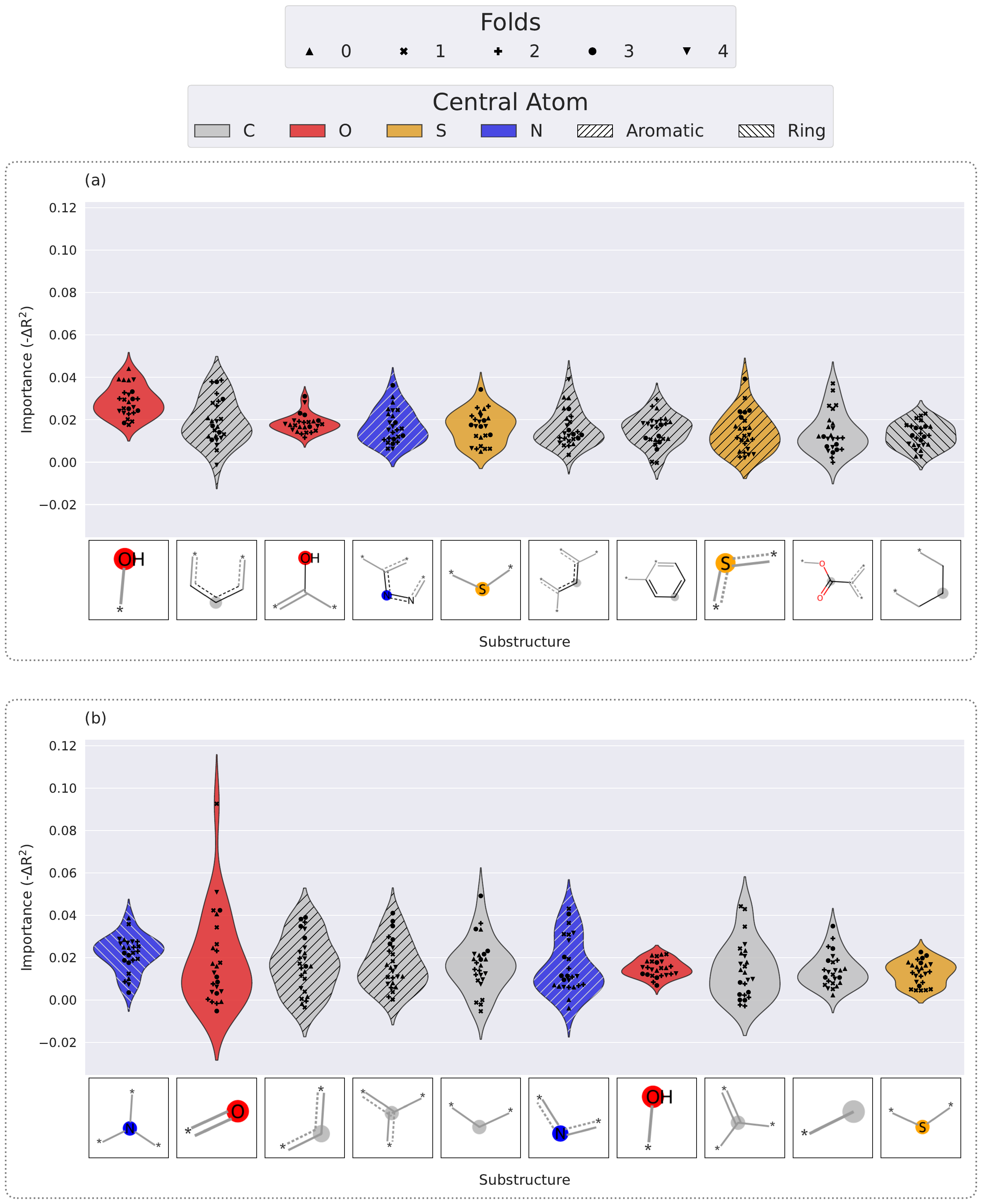}
    \caption{
        Substructure importances on the Human CLint dataset~\cite{fangC-2023-ProspectiveValidationMachineLearningAlgorithmsAbsorptionDistributionMetabolismExcretionPrediction}. 
        (a) Permutation feature importances for ECFP$_{\mathrm{S\&S}}$, mapped to substructures. 
        (b) Token embedding importances for PT-GIN, mapped to substructures. 
        For both (a) and (b), importances were measured as $-\Delta \mathrm{R}^2$ 
        on the test set upon permutation. 
        Only the top ten ranked substructures by mean importance are shown. 
        Violin plots are coloured by central atom element type; 
        hatches from the bottom left to the top right indicate the central atom is aromatic;
        hatches from the bottom right to the top left indicate the central atom is in a non-aromatic ring. 
        Individual scores are shown as a swarm plot; marker shape indicates the test set fold.
    }
    \label{fig:feat-import-clint}
\end{figure}

\begin{figure}[!htbp]
    \centering
    \includegraphics[width=\textwidth, height=0.8\textheight, keepaspectratio]{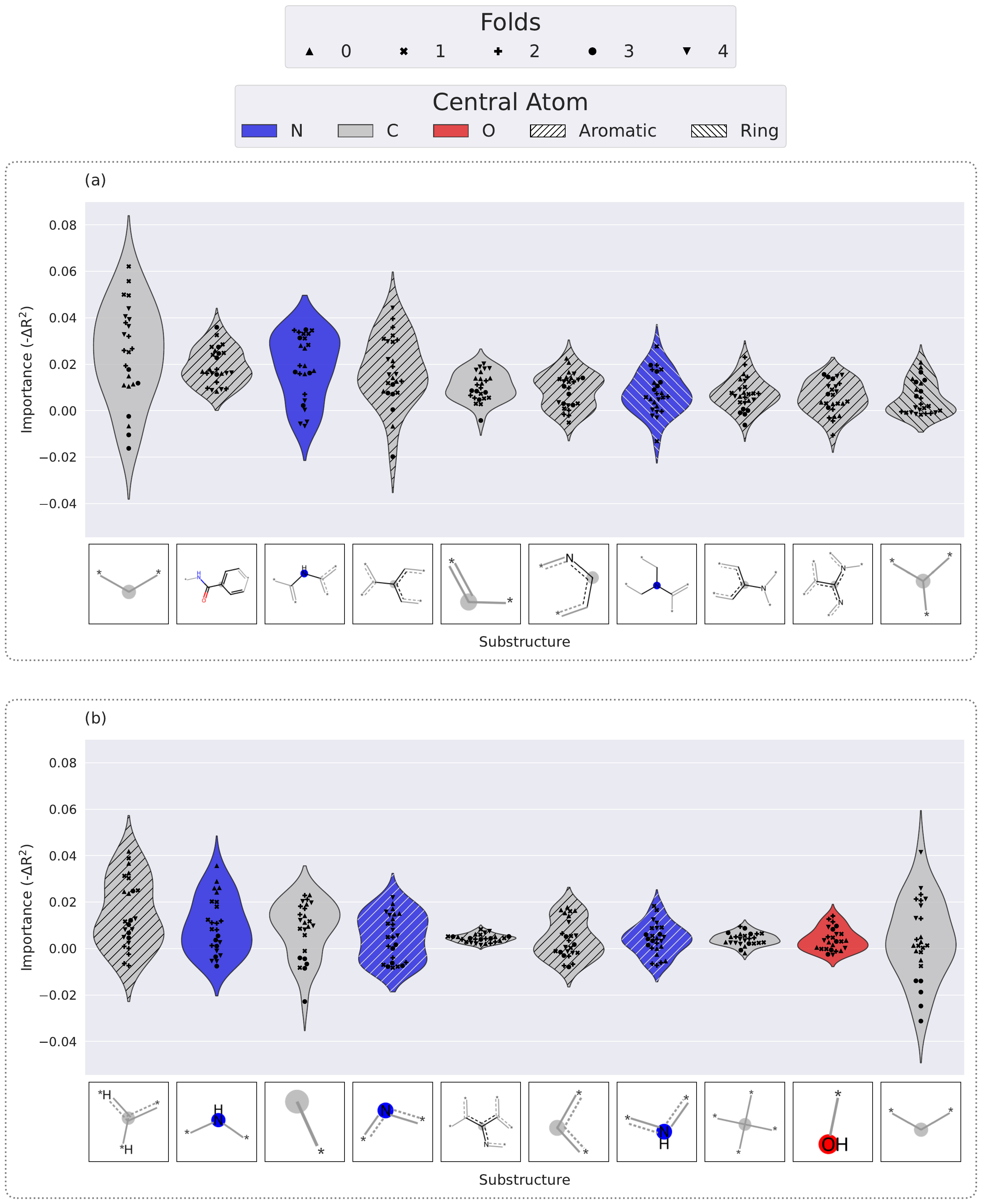}
    \caption{
        Substructure importances on the Biogen Solubility task~\cite{fangC-2023-ProspectiveValidationMachineLearningAlgorithmsAbsorptionDistributionMetabolismExcretionPrediction}. 
        (a) Permutation feature importances for ECFP$_{\mathrm{S\&S}}$, mapped to substructures. 
        (b) Token embedding importances for PT-GIN, mapped to substructures. 
        For both (a) and (b), importances were measured as $-\Delta \mathrm{R}^2$ 
        on the test set upon permutation. 
        Only the top ten ranked substructures by mean importance are shown. 
        Violin plots are coloured by central atom element type; 
        hatches from the bottom left to the top right indicate the central atom is aromatic;
        hatches from the bottom right to the top left indicate the central atom is in a non-aromatic ring. 
        Individual scores are shown as a swarm plot; marker shape indicates the test set fold.
    }
    \label{fig:feat-import-solu}
\end{figure}

\begin{figure}[!htbp]
    \centering
    \includegraphics[width=\textwidth, height=0.8\textheight, keepaspectratio]{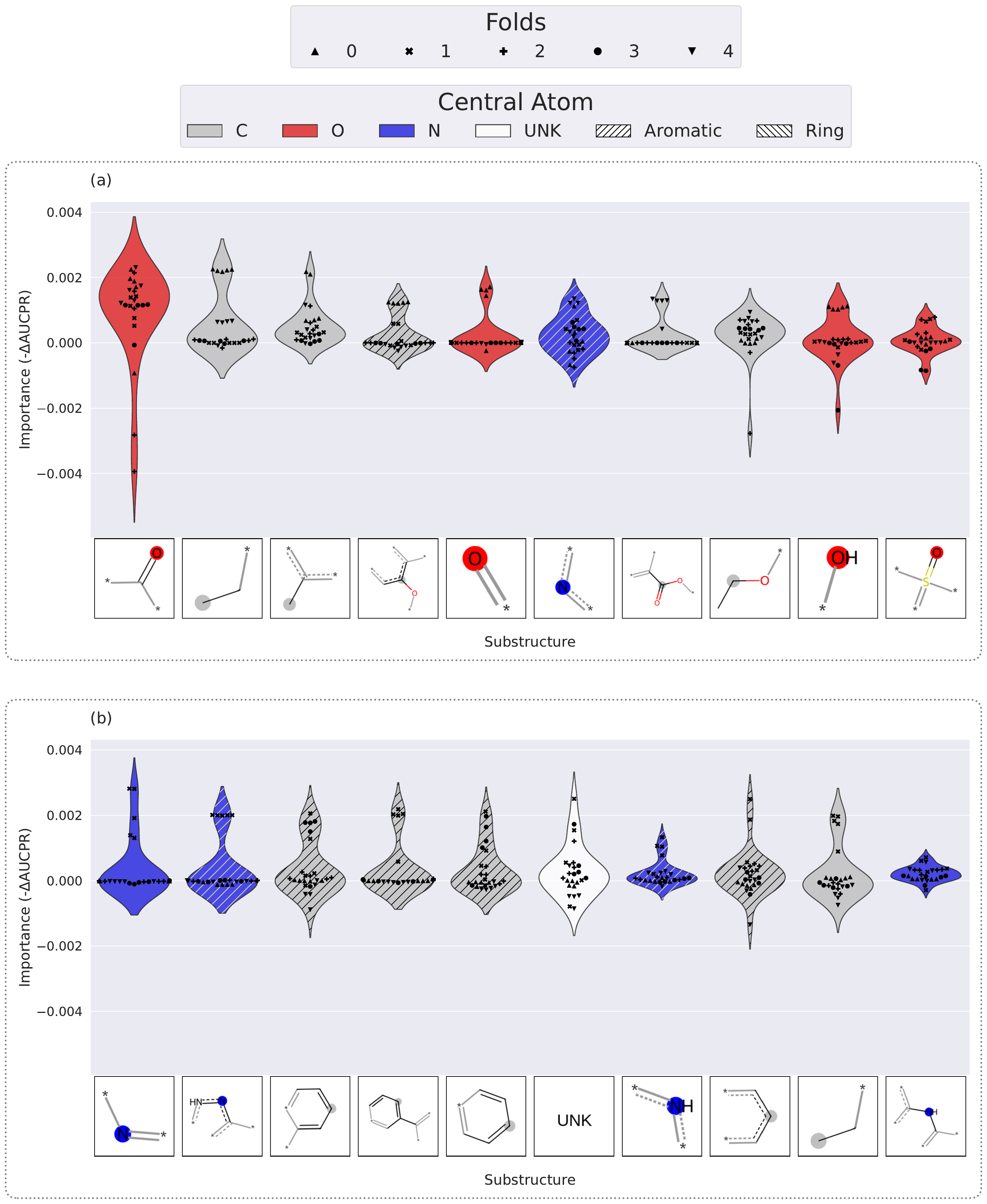}
    \caption{
        Substructure importances on the MUV858 task~\cite{rohrerSG-2009-MaximumUnbiasedValidationMUVData}. 
        (a) Permutation feature importances for ECFP$_{\mathrm{S\&S}}$, mapped to substructures. 
        (b) Token embedding importances for PT-GIN, mapped to substructures. 
        For both (a) and (b), importances were measured as $-\mathrm{\Delta AUCPR}$ 
        on the test set upon permutation. 
        Only the top ten ranked substructures by mean importance are shown. 
        Violin plots are coloured by central atom element type; 
        hatches from the bottom left to the top right indicate the central atom is aromatic;
        hatches from the bottom right to the top left indicate the central atom is in a non-aromatic ring. 
        Individual scores are shown as a swarm plot; marker shape indicates the test set fold.
    }
    \label{fig:feat-import-muv}
\end{figure}

\begin{figure}[!htbp]
    \centering
    \includegraphics[width=\textwidth, height=0.8\textheight, keepaspectratio]{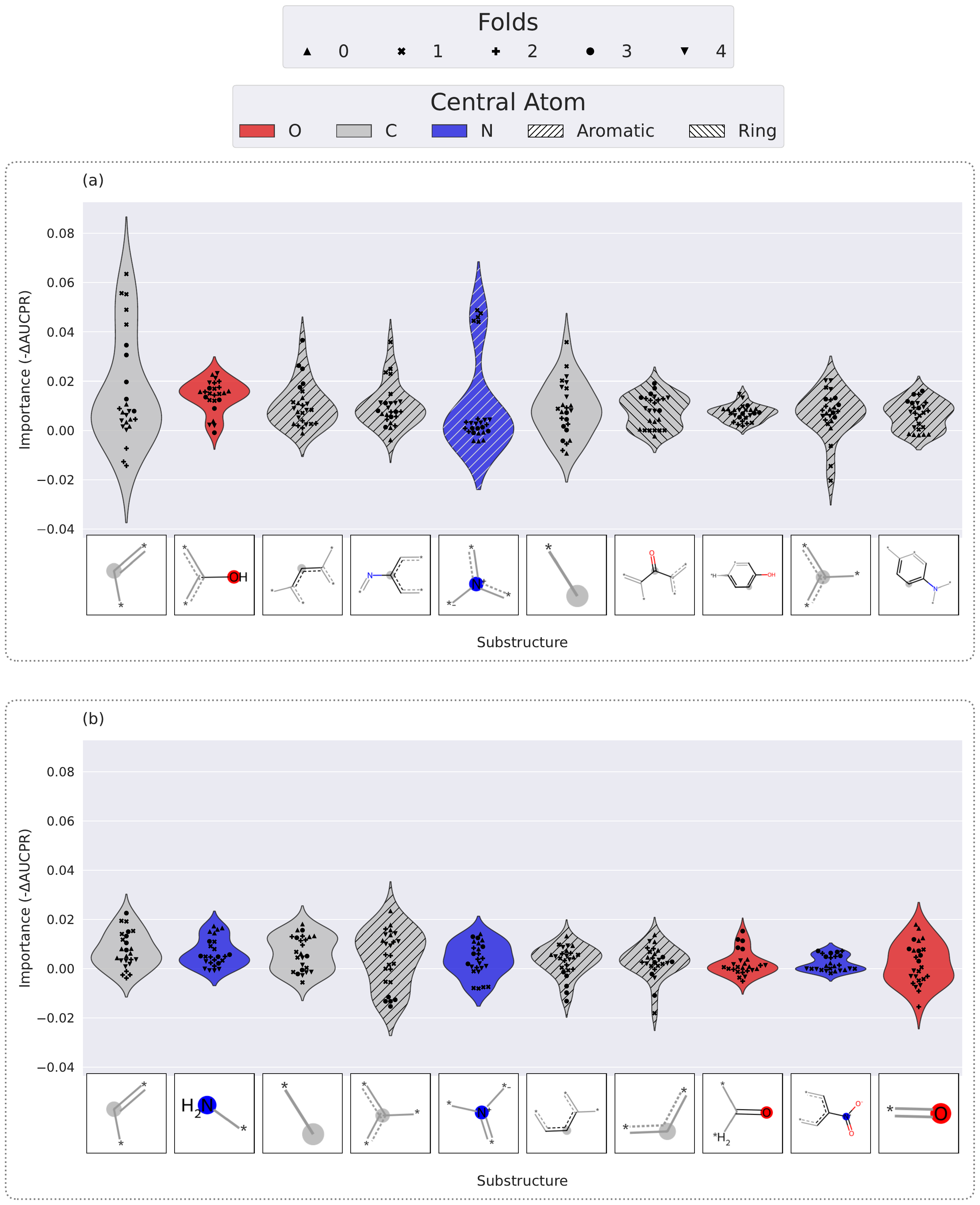}
    \caption{
        Substructure importances on the Tox21 SR ARE task~\cite{richardAM-2021-Tox2110KCompoundLibrary}. 
        (a) Permutation feature importances for ECFP$_{\mathrm{S\&S}}$, mapped to substructures. 
        (b) Token embedding importances for PT-GIN, mapped to substructures. 
        For both (a) and (b), importances were measured as $-\mathrm{\Delta AUCPR}$ 
        on the test set upon permutation. 
        Only the top ten ranked substructures by mean importance are shown. 
        Violin plots are coloured by central atom element type; 
        hatches from the bottom left to the top right indicate the central atom is aromatic;
        hatches from the bottom right to the top left indicate the central atom is in a non-aromatic ring. 
        Individual scores are shown as a swarm plot; marker shape indicates the test set fold.
    }
    \label{fig:feat-import-tox21}
\end{figure}
\clearpage
\newpage

\subsection{Ablation study: Morgan-based tokenisation versus conventional atom features}\label{si:atom_feature_ablation}

Throughout this work, we have used a novel graph tokenisation approach based on local circular substructures (see main text, Methods, \textit{Substructure tokenisation})
to featurise molecular graphs for PT-GINs. Our approach allowed us to perform additional investigations into substructure importance via
embedding permutation (see \textit{\nameref{si:substructure_importance}}). Embedding graph nodes using a lookup table with learnable embeddings 
based on atomic features, such as atom type, has been previously used in GNNs, e.g.,~\cite{huW-2020-StrategiesPretrainingGraphNeuralNetworks}; extending this approach to include 
substructure-level information may lead to enriched representations and improved model performance. To validate the effectiveness of our tokenisation approach,
we compared PT-GINs with our tokenisation approach to PT-GINs with engineered conventional atomic features without learnable embeddings.
We evaluated performance on Biogen regression tasks, using LightGBM for all benchmark tasks.
Rather than using all $199$ $5$-fold cross-validation repeats for evaluation, we used the same single $5$-fold cross-validation split
for hyperparameter tuning of LightGBMs (see main text, Methods, \textit{Benchmark hyperparameter tuning}), and a subset of $20$ $5$-fold
cross-validation repeats for evaluation.
Atomic features were based on the features used in~\cite{dablanderM-2023-ExploringQSARModelsActivitycliffPrediction}, 
including one-hot encodings for: 
\begin{itemize}
    \item atom type (C, N, O, S, F, Si, P, Cl, Br, Mg, Na, Ca, Fe, As, Al, I, B, V, K, Tl, Yb, Sb, Sn, Ag, Pd, Co, Se, Ti, Zn, Li, Ge, Cu, Au, Ni, Cd, In, Mn, Zr, Cr, Pt and Pb),
    \item the number of neighbouring heavy atoms $\in \{0, 1, 2, 3, 4\}$,
    \item formal charge $\in \{-2, -1, 0, +1, +2\}$,
    \item hybridisation $\in \{\text{sp, sp}^2, \text{sp}^3, \text{sp}^3\text{d}, \text{sp}^3\text{d}^2\}$,
    \item chirality $\in \{\text{R, S, None}\}$,
    \item the number of neighbouring hydrogen atoms $\in \{0, 1, 2, 3, 4\}$,
\end{itemize}
along with binary indicators for
\begin{itemize}
    \item ring status $\in \{0, 1\}$,
    \item aromaticity $\in \{0, 1\}$,
\end{itemize}
and three scaled continuous features:
\begin{itemize}
    \item atomic mass,
    \item van der Waals radius,
    \item covalent radius.
\end{itemize}

Our results suggest that the performance of PT-GINs with our tokenisation approach varies across different Biogen regression tasks.
On Solubility, learnable embeddings based on our tokenisation approach outperformed conventional atomic features by
coefficient of determination ($\mathrm{R}^2$) and Pearson correlation coefficient ($\rho$). However, on Human Plasma Protein Binding (PPB) and
Human Intrinsic Hepatic Clearance (CLint), conventional atomic features outperformed our tokenisation approach by $\mathrm{R}^2$ and $\rho$.
On the remaining tasks, there were no significant differences between the two approaches by $\mathrm{R}^2$ and $\rho$. By mean absolute percentage error (MAPE), 
our tokenisation approach underperformed compared to conventional atomic features on all tasks except for Solubility and Rat PPB, where there were no significant differences.

Discrepancies between $\mathrm{MAPE}$ and the other metrics likely reflect the sensitivity of $\mathrm{MAPE}$ to targets close to zero. 
Taken together, these results suggest that our tokenisation approach achieves broadly comparable predictive performance 
to conventional atomic features, with task-dependent advantages and disadvantages,
while additionally enabling substructure-level interpretability.
Given a greater amount of pre-training data, and greater model capacity, our tokenisation approach
could outperform conventional atomic features across all tasks due to the greater flexibility of learnable embeddings
and the ability to capture greater local substructure-level information.

\begin{figure}[!htbp]
    \centering
    \includegraphics[width=\textwidth, height=0.7\textheight, keepaspectratio]{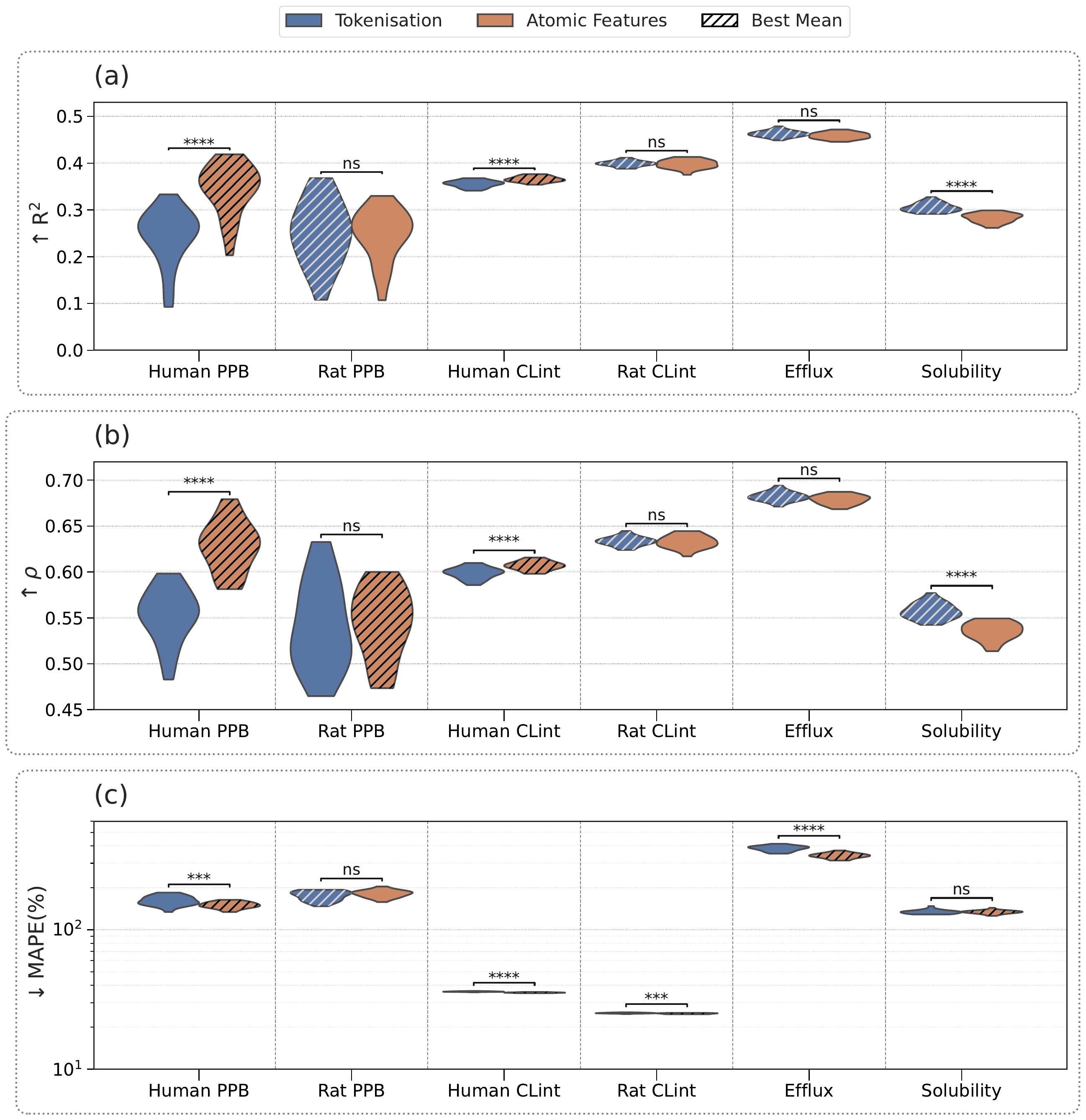}
    \caption{
        Performance metrics on Biogen regression tasks for PT-GIN with different graph featurisation methods:
        Morgan algorithm-based tokenisation (blue) (see main text, Methods, \textit{Substructure tokenisation}) and conventional atomic features (orange) (see \textit{\nameref{si:atom_feature_ablation}}).
        Tasks shown are Human Plasma Protein Binding (PPB), Rat PPB, 
        Human Intrinsic Hepatic Clearance (CLint), Rat CLint, Efflux, and Solubility.
        The metrics shown are (a) coefficient of determination ($\uparrow\mathrm{R}^2$), 
        (b) Pearson correlation coefficient ($\uparrow\rho$), and (c) mean absolute percentage error ($\downarrow$MAPE (\%)).
        For each task and metric, the approach with the best mean is hatched.
        Statistical significances, determined by the two-sided Wilcoxon signed-rank test ($n=20$, $\alpha=0.05$), 
        between the best method by mean and all other approaches are shown with asterisks; 
        ns (not significant) is $p \geq 0.05$;
        * is $0.01 \leq p < 0.05$;
        ** is $0.001 \leq p < 0.01$;
        *** is $0.0001 \leq p < 0.001$;
        and **** is $p < 0.0001$. 
        Violins are normalised such that each violin has the same maximum width.
    }
    \label{fig:atom_features_ablation_violin}
\end{figure}

\begin{figure}[!htbp]
    \centering
    \includegraphics[width=\textwidth, height=0.7\textheight, keepaspectratio]{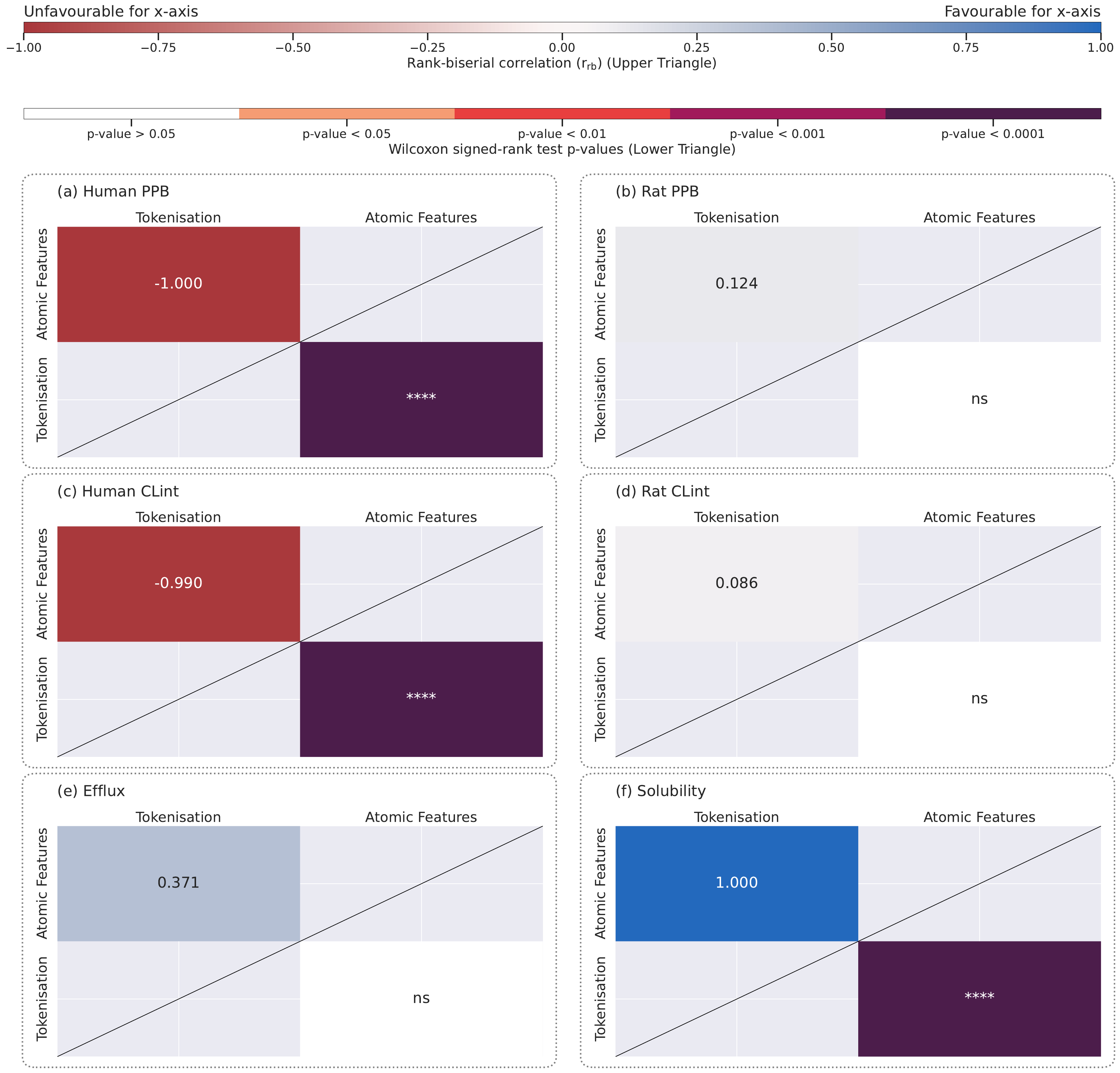}
    \caption{
        Rank-biserial correlation ($r_{rb}$) effect sizes (upper triangle) and two-sided Wilcoxon signed-rank test \textit{p}-values (lower triangle) of the coefficient of determination 
        ($\uparrow \mathrm{R}^2$) between PT-GINs with different atom featurisation methods on Biogen tasks.
        Graph featurisation methods are Tokenisation (see main text, Methods, \textit{Substructure tokenisation}) and Atomic features (see \textit{\nameref{si:atom_feature_ablation}}).
        Tasks are (a) Human Plasma Protein Binding (PPB), (b) Rat PPB, 
        (c) Human Intrinsic Hepatic Clearance (CLint), (d) Rat CLint, (e) Efflux, (f) Solubility.
        For $r_{rb}$ effect sizes (upper triangle): blue indicates the method on the \textit{x}-axis has a higher $\mathrm{R}^2$ than 
        the method on the \textit{y}-axis; red indicates the opposite. 
        For Wilcoxon signed-rank test \textit{p}-values (lower triangle): 
        white indicates no significant difference and darker indicates a smaller \textit{p}-value ($n=20$, $\alpha=0.05$).
    }
    \label{fig:atom_features_ablation_r2}
\end{figure}

\begin{figure}[!htbp]
    \centering
    \includegraphics[width=\textwidth, height=0.7\textheight, keepaspectratio]{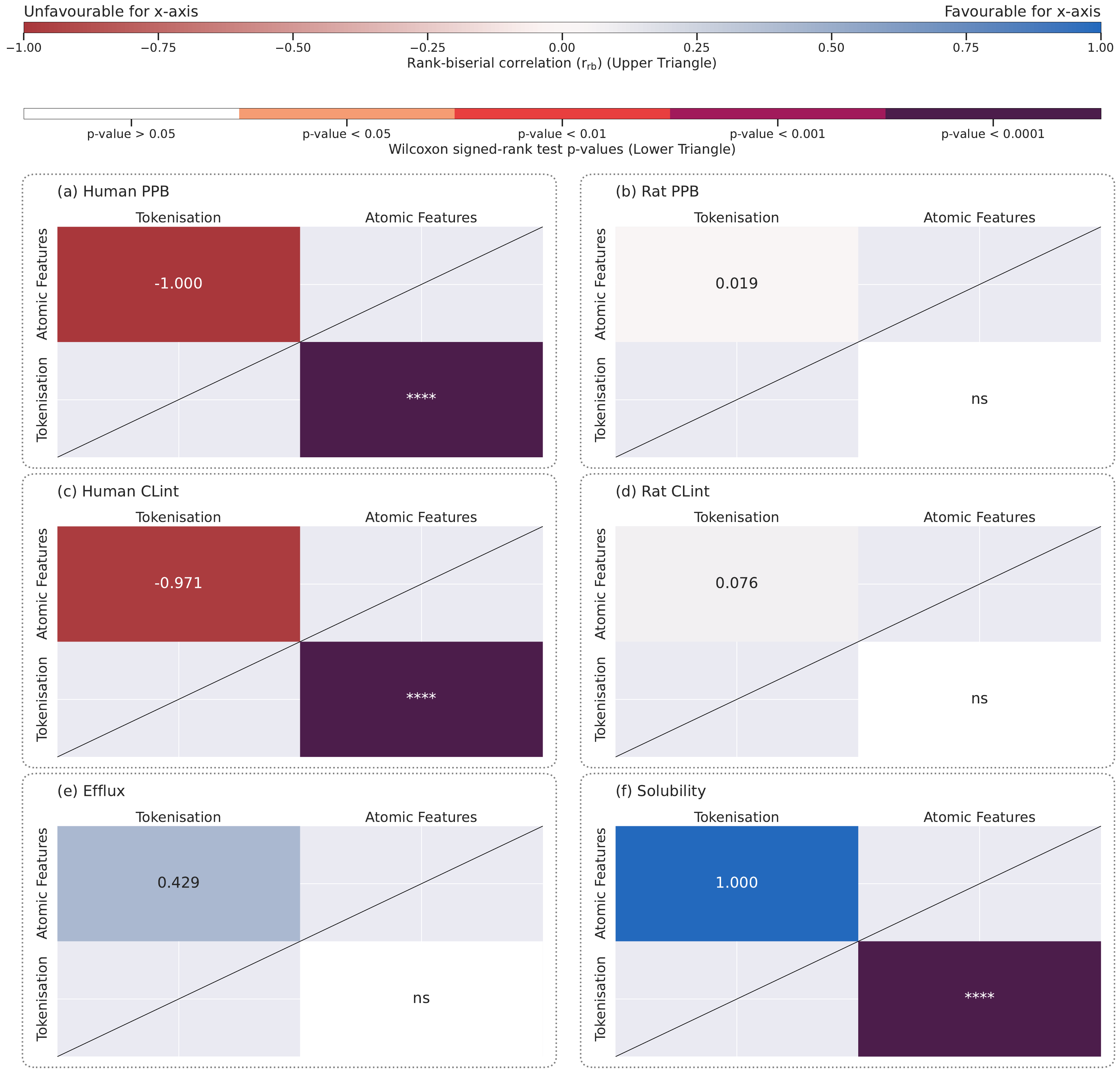}
    \caption{
        Rank-biserial correlation ($r_{rb}$) effect sizes (upper triangle) and two-sided Wilcoxon signed-rank test \textit{p}-values (lower triangle) of the Pearson correlation coefficient
        ($\uparrow \rho$) between PT-GINs with different atom featurisation methods on Biogen tasks.
        Graph featurisation methods are Tokenisation (see main text, Methods, \textit{Substructure tokenisation}) and Atomic features (see \textit{\nameref{si:atom_feature_ablation}}).
        Tasks are (a) Human Plasma Protein Binding (PPB), (b) Rat PPB, 
        (c) Human Intrinsic Hepatic Clearance (CLint), (d) Rat CLint, (e) Efflux, (f) Solubility.
        For $r_{rb}$ effect sizes (upper triangle): blue indicates the method on the \textit{x}-axis has a higher $\rho$ than 
        the method on the \textit{y}-axis; red indicates the opposite. 
        For Wilcoxon signed-rank test \textit{p}-values (lower triangle): 
        white indicates no significant difference and darker indicates a smaller \textit{p}-value ($n=20$, $\alpha=0.05$).
    }
    \label{fig:atom_features_ablation_pearson}
\end{figure}

\begin{figure}[!htbp]
    \centering
    \includegraphics[width=\textwidth, height=0.7\textheight, keepaspectratio]{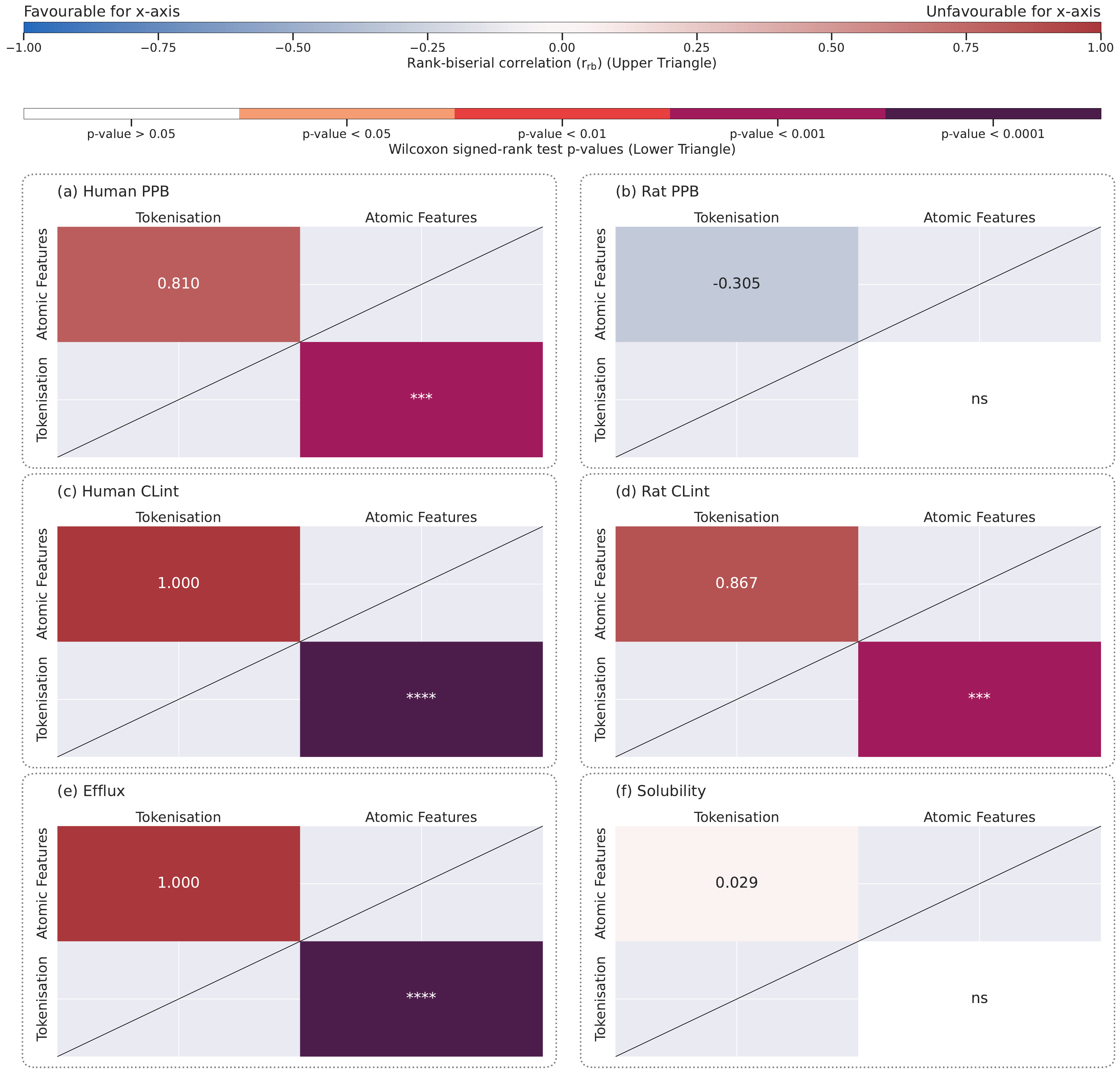}
    \caption{
        Rank-biserial correlation ($r_{rb}$) effect sizes (upper triangle) and two-sided Wilcoxon signed-rank test \textit{p}-values (lower triangle) of the mean absolute percentage error
        ($\downarrow\mathrm{MAPE}$) between PT-GINs with different atom featurisation methods on Biogen tasks.
        Graph featurisation methods are Tokenisation (see main text, Methods, \textit{Substructure tokenisation}) and Atomic features (see \textit{\nameref{si:atom_feature_ablation}}).
        Tasks are (a) Human Plasma Protein Binding (PPB), (b) Rat PPB, 
        (c) Human Intrinsic Hepatic Clearance (CLint), (d) Rat CLint, (e) Efflux, (f) Solubility.
        For $r_{rb}$ effect sizes (upper triangle): blue indicates the method on the \textit{x}-axis has a lower $\mathrm{MAPE}$ than 
        the method on the \textit{y}-axis; red indicates the opposite. 
        For Wilcoxon signed-rank test \textit{p}-values (lower triangle): 
        white indicates no significant difference and darker indicates a smaller \textit{p}-value ($n=20$, $\alpha=0.05$).
    }
    \label{fig:atom_features_ablation_mape}
\end{figure}

\clearpage
\newpage

\subsection{Ablation study: pre-training versus a randomly initialised GIN}\label{si:random_init_ablation}

\begin{figure}[!htbp]
    \centering
    \includegraphics[width=\textwidth, height=0.7\textheight, keepaspectratio]{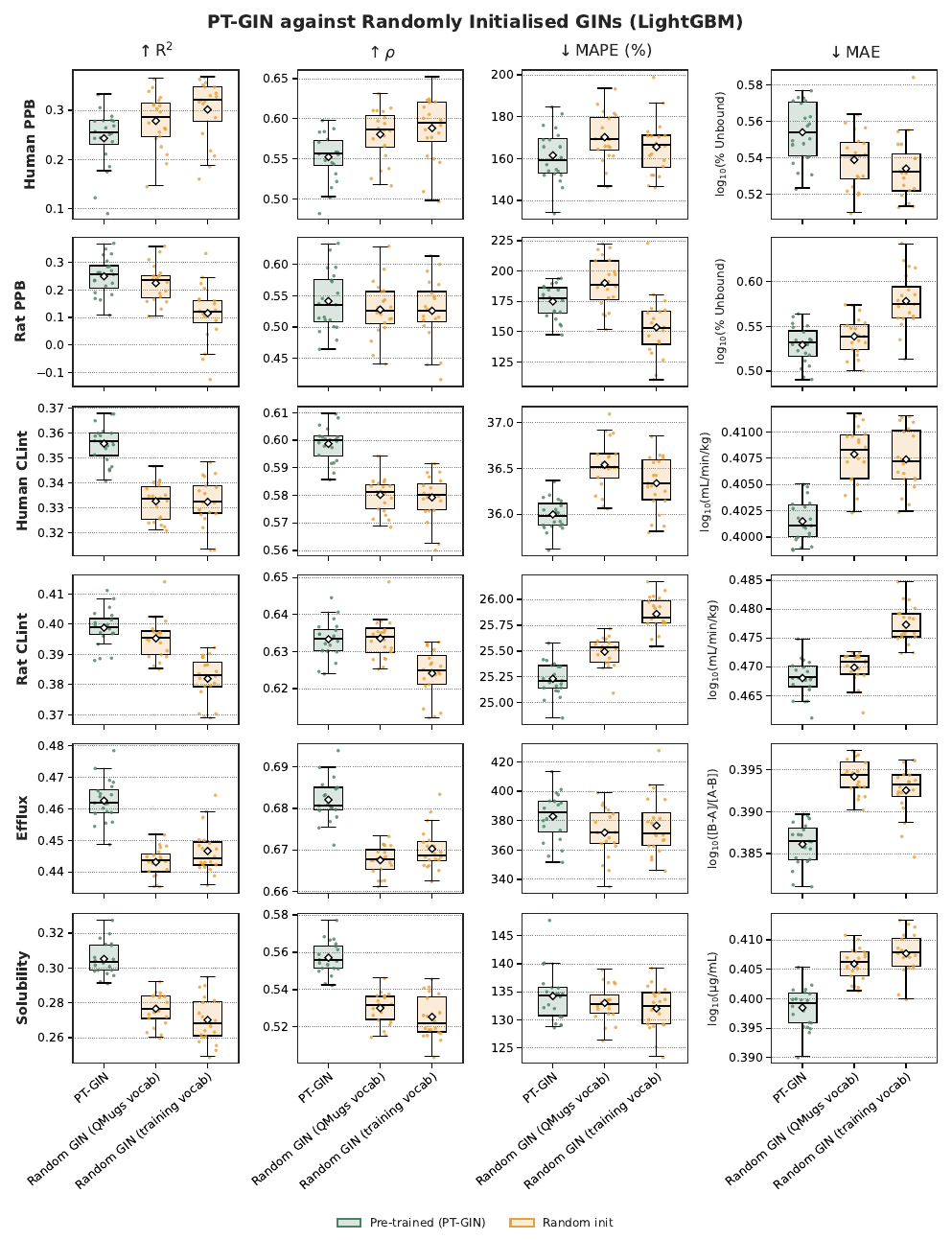}
    \caption{
        Performance metrics on Biogen tasks for GINs pre-trained to predict ECFP$4_{2048}$
        against a randomly initialised GIN with the same architecture.
        Metrics are calculated over 20 5-fold cross-validation repeats;
        each point represents the average performance across
        a single 5-fold CV repeat. PT-GIN is shown in green and the randomly initialised GINs in orange.
        Diamonds indicate the mean performance; the horizontal line
        within each box represents the median performance.
        Two versions of randomly initialised GINs are shown:
        one with the same Sort \& Slice vocabulary as the pre-trained GIN based on
        the QMugs dataset, and one with Sort \& Slice vocabularies based on
        benchmark training sets.
        Each row is a different task: Human Plasma Protein Binding (PPB), Rat PPB,
        Human Intrinsic Hepatic Clearance (CLint), Rat CLint, Efflux, and Solubility.
        Each column is a different metric: coefficient of determination ($\uparrow\mathrm{R}^2$),
        Pearson correlation coefficient ($\uparrow\rho$), mean absolute percentage error ($\downarrow$MAPE (\%)),
        and mean absolute error ($\downarrow$MAE).
    }\label{fig:random_init_gin_box_strip}
\end{figure}

\begin{figure}[!htbp]
    \centering
    \includegraphics[width=\textwidth, height=0.7\textheight, keepaspectratio]{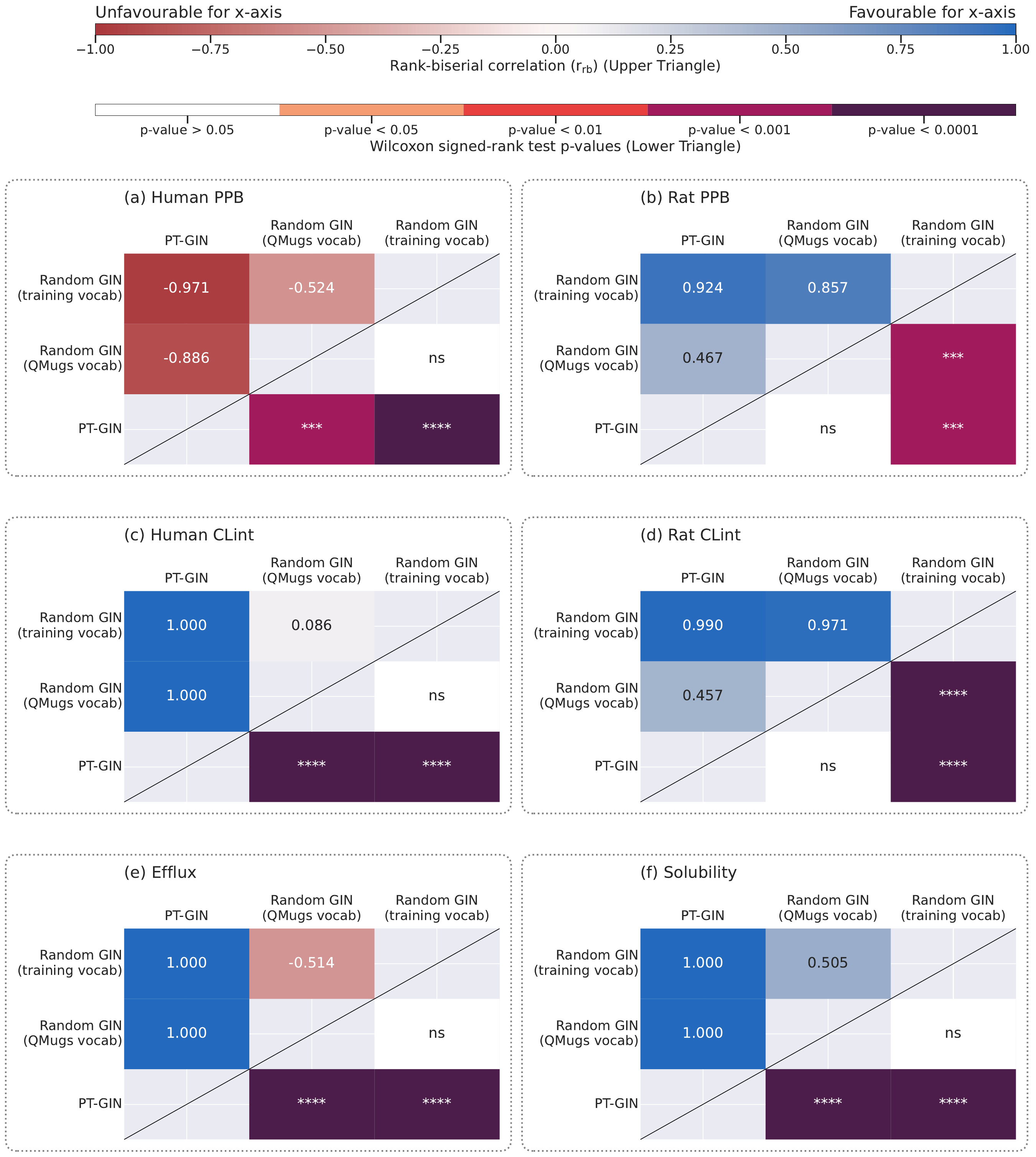}
    \caption{
        Rank-biserial correlation ($r_{rb}$) effect sizes (upper triangle) and 
        two-sided Wilcoxon signed-rank test \textit{p}-values with Bonferroni correction 
        (lower triangle) of the coefficient of determination 
        ($\uparrow \mathrm{R}^2$) between PT-GINs and randomly initialised GINs on Biogen tasks.
        Tasks are (a) Human Plasma Protein Binding (PPB), (b) Rat PPB, 
        (c) Human Intrinsic Hepatic Clearance (CLint), (d) Rat CLint, (e) Efflux, (f) Solubility.
        For $r_{rb}$ effect sizes (upper triangle): blue indicates the method on the \textit{x}-axis has a higher $\mathrm{R}^2$ than 
        the method on the \textit{y}-axis; red indicates the opposite. 
        For Wilcoxon signed-rank test \textit{p}-values (lower triangle): 
        white indicates no significant difference and darker indicates a smaller \textit{p}-value ($n=20$, $\alpha=0.05$).
    }
    \label{fig:random_init_gin_r2}
\end{figure}

\begin{figure}[!htbp]
    \centering
    \includegraphics[width=\textwidth, height=0.7\textheight, keepaspectratio]{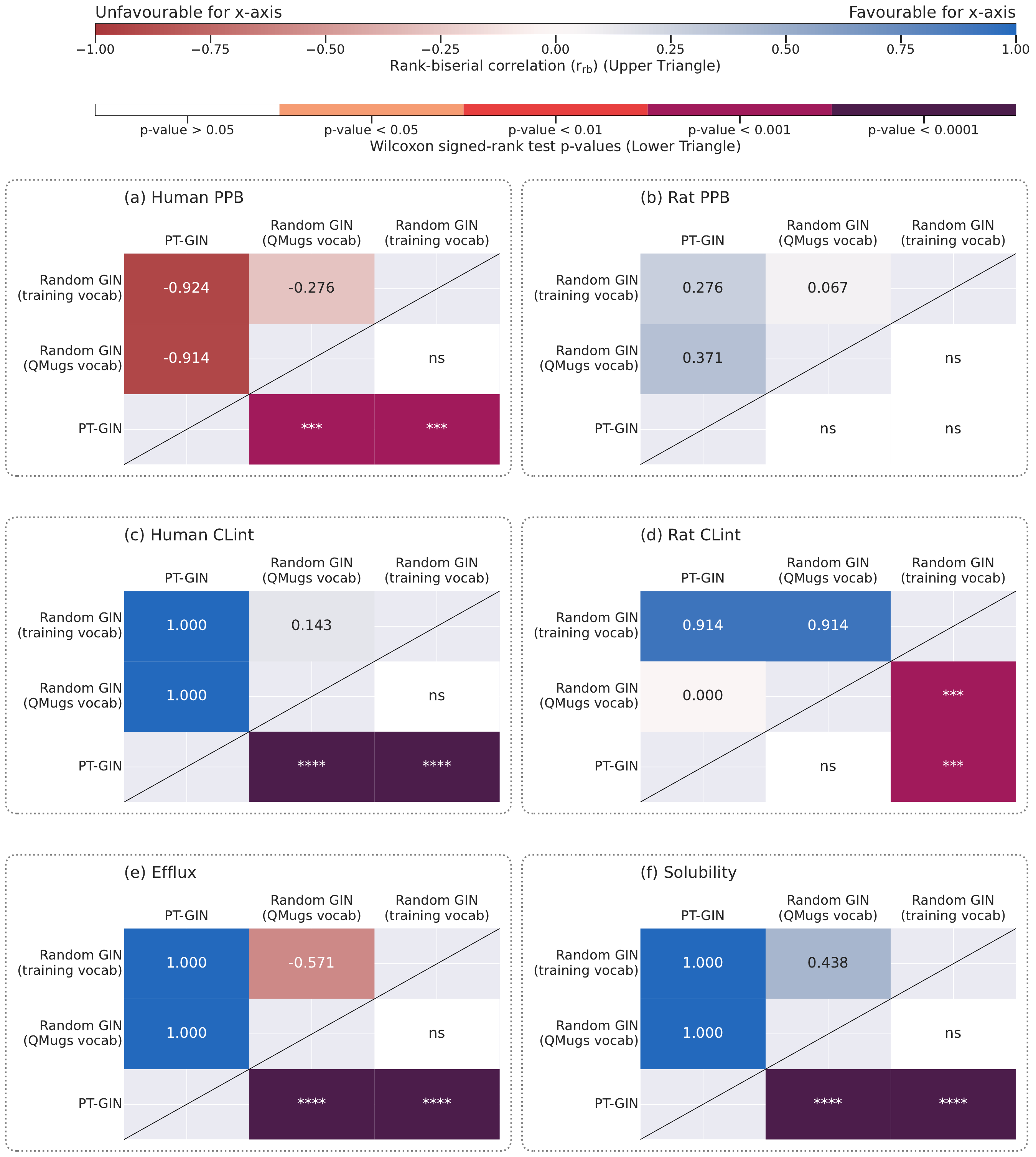}
    \caption{
        Rank-biserial correlation ($r_{rb}$) effect sizes (upper triangle) and 
        two-sided Wilcoxon signed-rank test \textit{p}-values with Bonferroni correction 
        (lower triangle) of the Pearson correlation coefficient
        ($\uparrow \rho$) between PT-GINs and randomly initialised GINs on Biogen tasks.
        Tasks are (a) Human Plasma Protein Binding (PPB), (b) Rat PPB, 
        (c) Human Intrinsic Hepatic Clearance (CLint), (d) Rat CLint, (e) Efflux, (f) Solubility.
        For $r_{rb}$ effect sizes (upper triangle): blue indicates the method on the \textit{x}-axis has a higher $\rho$ than 
        the method on the \textit{y}-axis; red indicates the opposite. 
        For Wilcoxon signed-rank test \textit{p}-values (lower triangle): 
        white indicates no significant difference and darker indicates a smaller \textit{p}-value ($n=20$, $\alpha=0.05$).
    }
    \label{fig:random_init_gin_pearson}
\end{figure}

\begin{figure}[!htbp]
    \centering
    \includegraphics[width=\textwidth, height=0.7\textheight, keepaspectratio]{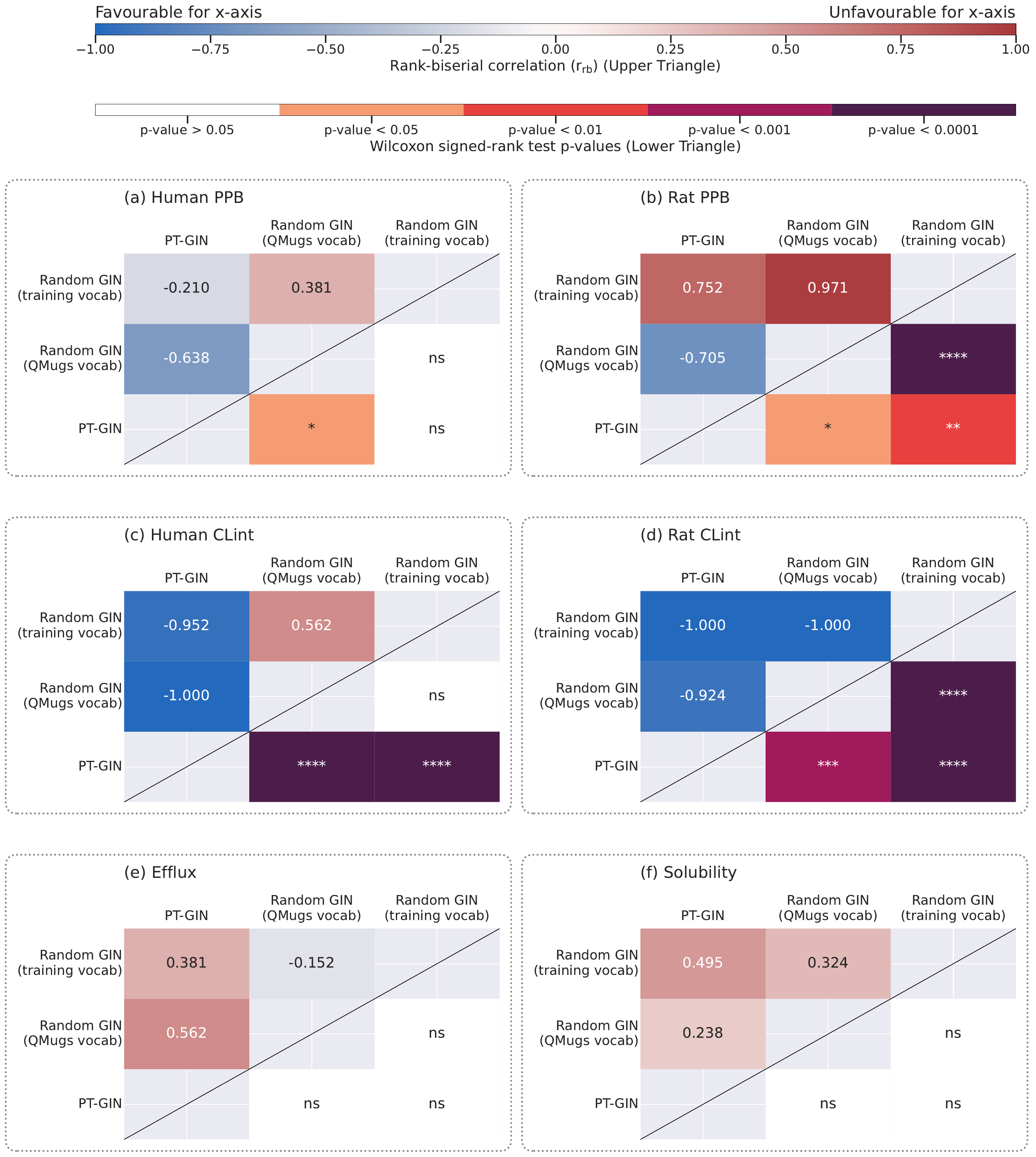}
    \caption{
        Rank-biserial correlation ($r_{rb}$) effect sizes (upper triangle) and 
        two-sided Wilcoxon signed-rank test \textit{p}-values with Bonferroni correction 
        (lower triangle) of the mean absolute percentage error 
        ($\downarrow\mathrm{MAPE}$) between PT-GINs and randomly initialised GINs on Biogen tasks.
        Tasks are (a) Human Plasma Protein Binding (PPB), (b) Rat PPB, 
        (c) Human Intrinsic Hepatic Clearance (CLint), (d) Rat CLint, (e) Efflux, (f) Solubility.
        For $r_{rb}$ effect sizes (upper triangle): blue indicates the method on the \textit{x}-axis has a lower $\mathrm{MAPE}$ than 
        the method on the \textit{y}-axis; red indicates the opposite. 
        For Wilcoxon signed-rank test \textit{p}-values (lower triangle): 
        white indicates no significant difference and darker indicates a smaller \textit{p}-value ($n=20$, $\alpha=0.05$).
    }
    \label{fig:random_init_gin_mape}
\end{figure}

\begin{figure}[!htbp]
    \centering
    \includegraphics[width=\textwidth, height=0.7\textheight, keepaspectratio]{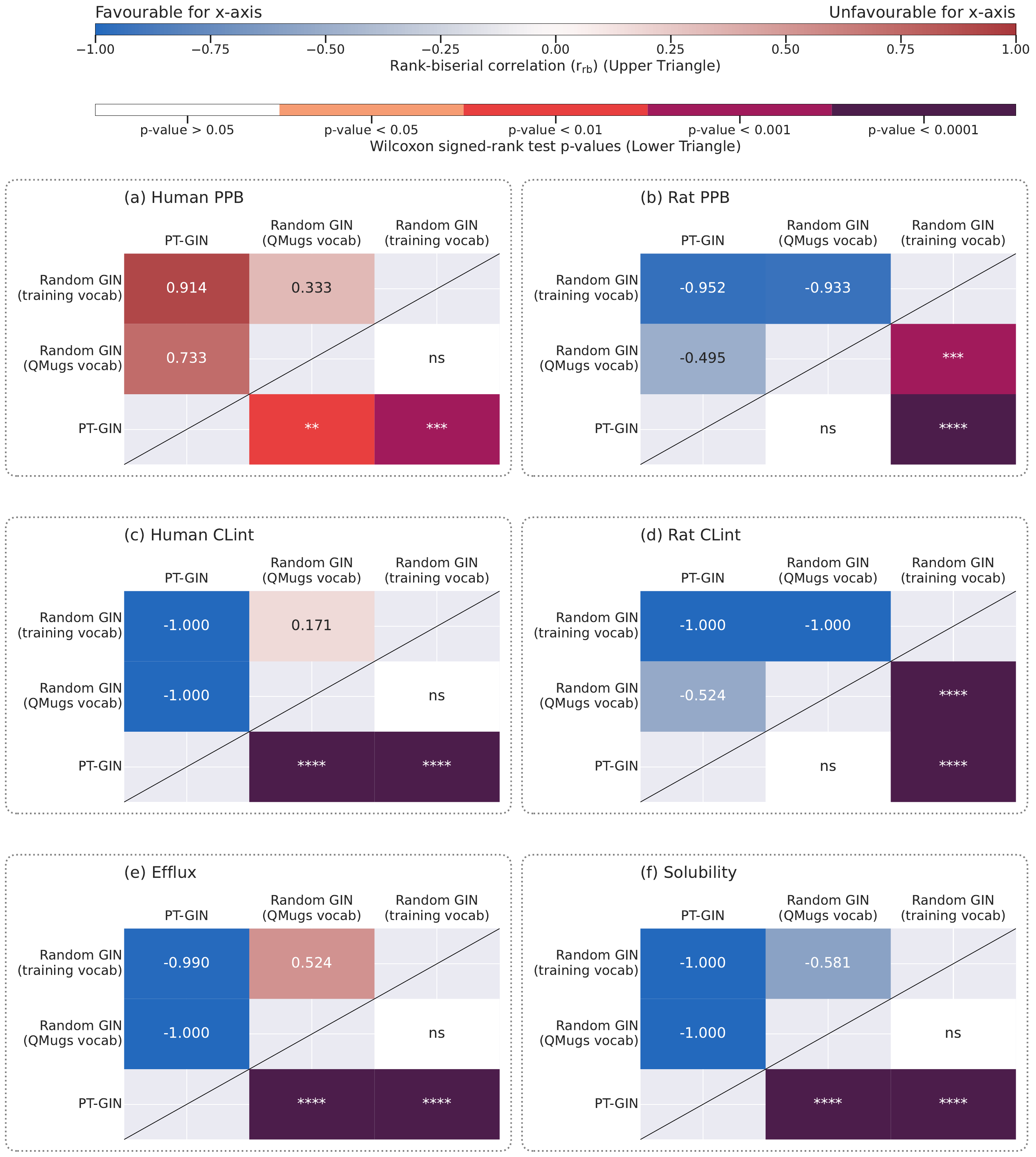}
    \caption{
        Rank-biserial correlation ($r_{rb}$) effect sizes (upper triangle) and 
        two-sided Wilcoxon signed-rank test \textit{p}-values with Bonferroni correction 
        (lower triangle) of the mean absolute error 
        ($\downarrow\mathrm{MAE}$) between PT-GINs and randomly initialised GINs on Biogen tasks.
        Tasks are (a) Human Plasma Protein Binding (PPB), (b) Rat PPB, 
        (c) Human Intrinsic Hepatic Clearance (CLint), (d) Rat CLint, (e) Efflux, (f) Solubility.
        For $r_{rb}$ effect sizes (upper triangle): blue indicates the method on the \textit{x}-axis has a lower $\mathrm{MAE}$ than 
        the method on the \textit{y}-axis; red indicates the opposite. 
        For Wilcoxon signed-rank test \textit{p}-values (lower triangle): 
        white indicates no significant difference and darker indicates a smaller \textit{p}-value ($n=20$, $\alpha=0.05$).
    }
    \label{fig:random_init_gin_mae}
\end{figure}

\clearpage
\subsection{Ablation study: impact of message passing on PT-GIN performance}\label{si:message_pass_ablation}

Although pre-trained GINs have been shown to outperform ECFPs on a range of molecular property prediction tasks,
it remains unclear how much of the performance gain is due to the message passing mechanism of GINs.
To validate the effectiveness of message passing in PT-GINs, 
we conducted an ablation study to assess the impact of message passing on model performance,
comparing the performance of pre-trained GINs (With MP) against a baseline model without message passing (No MP).
As with the ablation study on the impact of tokenisation (see \textit{\nameref{si:atom_feature_ablation}}),
we used LightGBM for all benchmarking tasks with hyperparameters optimised for each task,
and we used $20$ $5$-fold cross-validation repeats for evaluation.
The baseline model was implemented by removing the message passing layers from the GIN architecture,
and instead using two feedforward neural networks, an encoder and decoder, to project input graphs into a latent representation 
before decoding into the output target, ECFP$4_{2048}$.
The overall structure of the baseline model is analogous to an autoencoder, except that 
the input representation is a pooled graph representation of hashed Morgan substructures, 
rather than the target ECFP$4_{2048}$.

Both models were pre-trained on the same QMugs filtered data, with an ECFP$4_{2048}$ 
Tanimoto similarity threshold of $0.5$. Both models used the same vocabulary size for tokenisation,
and the same radius for the Morgan algorithm-based tokenisation. 
Both models had approximately the same number of trainable parameters, with the encoder in the 
baseline model having a similar number of parameters (3.9M) to the message passing layers within the
convolutional layers of the PT-GIN model.

\begin{figure}[!htbp]
    \centering
    \includegraphics[width=\textwidth, height=0.7\textheight, keepaspectratio]{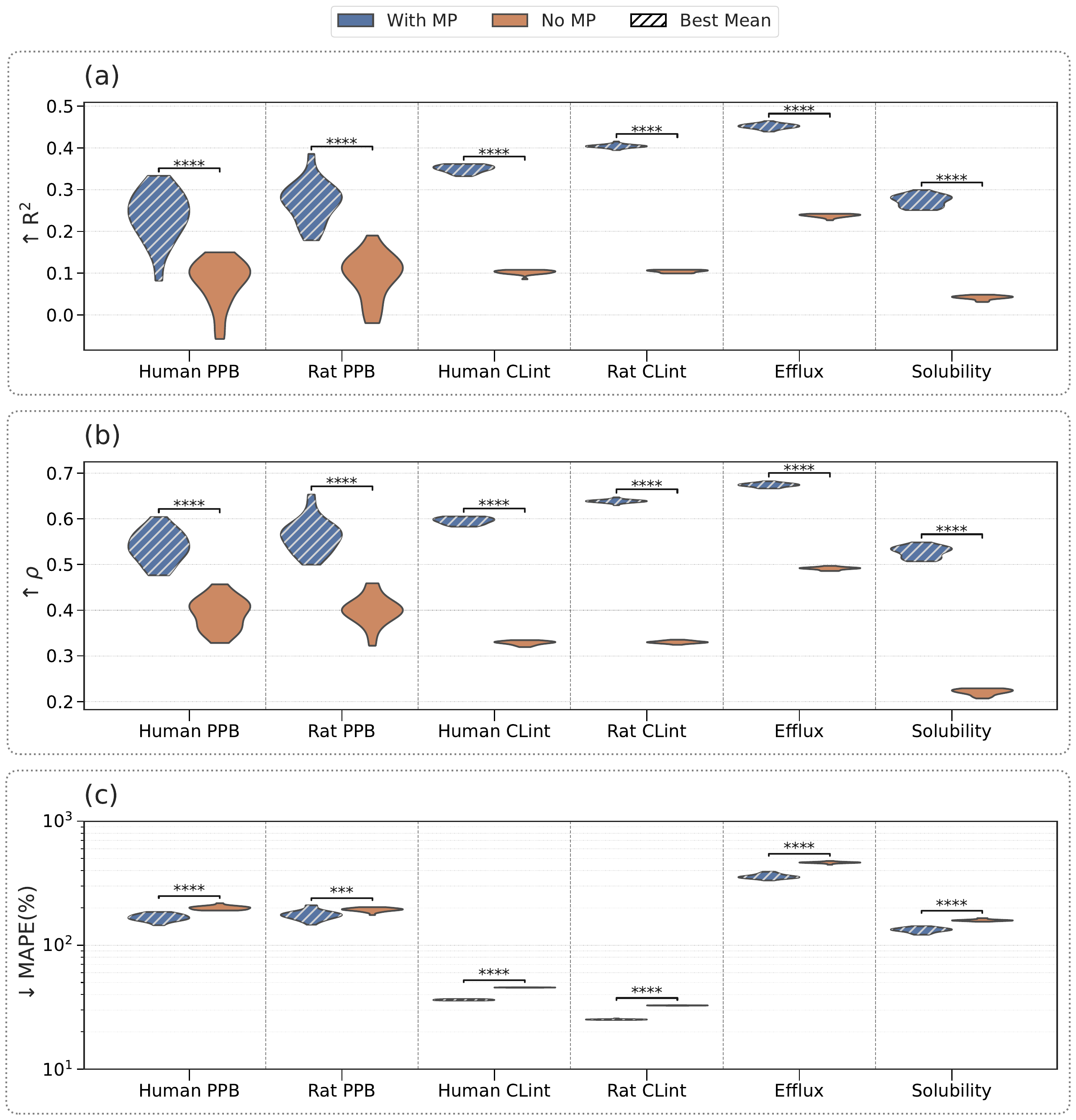}
    \caption{
        Performance metrics on Biogen regression tasks for pre-trained models 
        with (With MP) (blue) and without message passing (No MP) (orange).
        Tasks shown are Human Plasma Protein Binding (PPB), Rat PPB, 
        Human Intrinsic Hepatic Clearance (CLint), Rat CLint, Efflux, and Solubility.
        The metrics shown are (a) coefficient of determination ($\uparrow\mathrm{R}^2$), 
        (b) Pearson correlation coefficient ($\uparrow\rho$), and (c) mean absolute percentage error ($\downarrow$MAPE (\%)).
        For each task and metric, the approach with the best mean is hatched.
        Statistical significances, determined by the two-sided Wilcoxon signed-rank test ($n=20$, $\alpha=0.05$), 
        between the best method by mean and all other approaches are shown with asterisks; 
        ns (not significant) is $p \geq 0.05$;
        * is $0.01 \leq p < 0.05$;
        ** is $0.001 \leq p < 0.01$;
        *** is $0.0001 \leq p < 0.001$;
        and **** is $p < 0.0001$. 
        Violins are normalised such that each violin has the same maximum width.
    }
    \label{fig:message_passing_ablation_violin}
\end{figure}

The results of the ablation study suggest that message passing is a critical component of PT-GINs,
as the PT-GIN model outperforms the baseline model in all Biogen tasks across $\mathrm{R}^2$, $\rho$, and $\mathrm{MAPE}$
(Fig.~\ref{fig:message_passing_ablation_violin}). Without message passing, 
the baseline model is unable to effectively capture the local substructure-level 
information within the molecular graphs, resulting in a significant decrease in predictive performance.
Message passing may also enable PT-GIN to infer information about substructures that are not included
in the vocabulary, as the model can propagate information from known substructures to unknown substructures 
through convolutions.

\begin{figure}[!htbp]
    \centering
    \includegraphics[width=\textwidth, height=0.7\textheight, keepaspectratio]{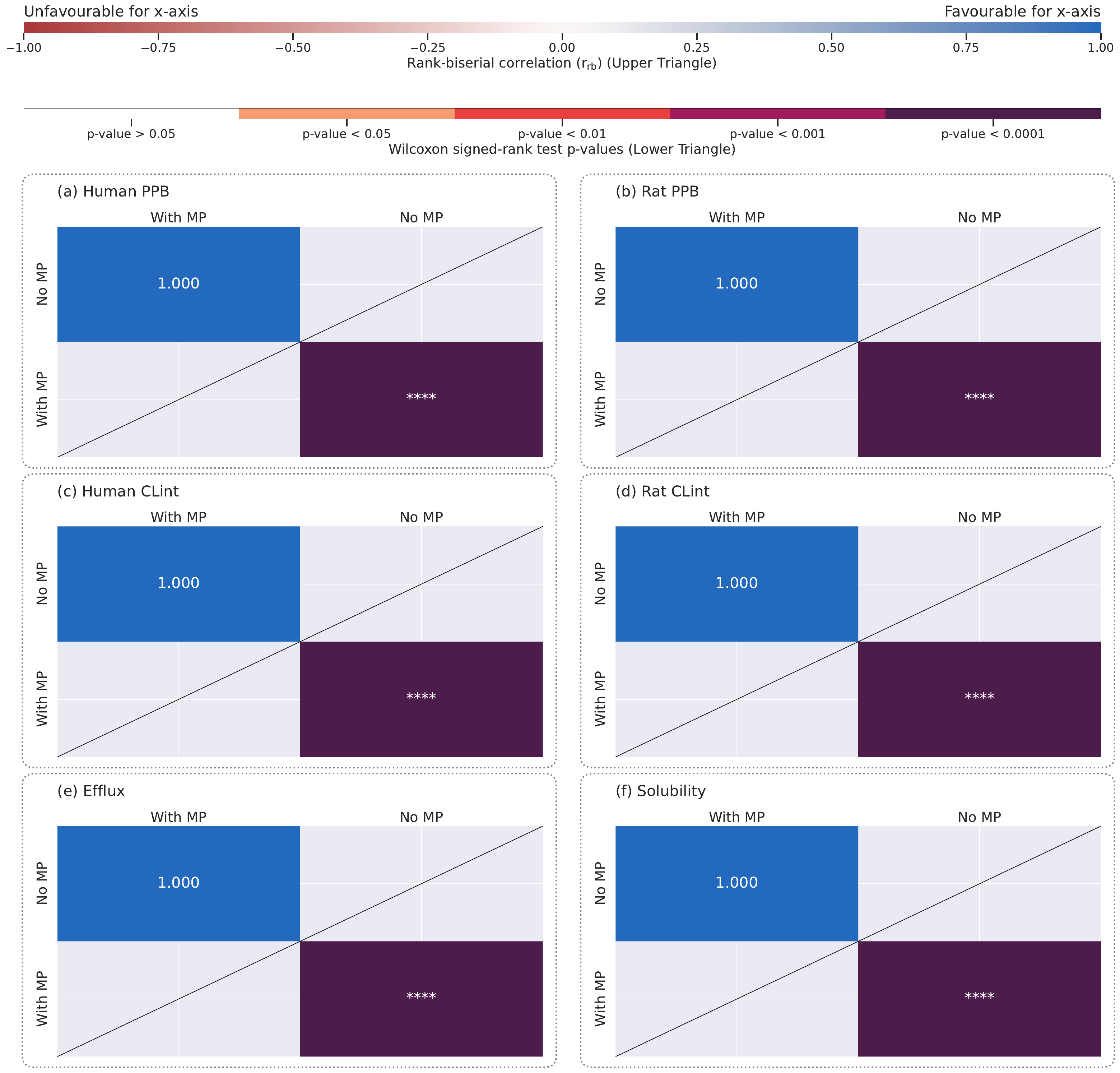}
    \caption{
        Rank-biserial correlation ($r_{rb}$) effect sizes (upper triangle) and two-sided Wilcoxon signed-rank test \textit{p}-values (lower triangle) of the coefficient of determination 
        ($\uparrow \mathrm{R}^2$) between pre-trained models with (With MP) and without message passing (No MP) on Biogen tasks.
        Tasks are (a) Human Plasma Protein Binding (PPB), (b) Rat PPB, 
        (c) Human Intrinsic Hepatic Clearance (CLint), (d) Rat CLint, (e) Efflux, (f) Solubility.
        For $r_{rb}$ effect sizes (upper triangle): blue indicates the method on the \textit{x}-axis has a higher $\mathrm{R}^2$ than 
        the method on the \textit{y}-axis; red indicates the opposite. 
        For Wilcoxon signed-rank test \textit{p}-values (lower triangle): 
        white indicates no significant difference and darker indicates a smaller \textit{p}-value ($n=20$, $\alpha=0.05$).
    }
    \label{fig:message_passing_ablation_r2}
\end{figure}

\begin{figure}[!htbp]
    \centering
    \includegraphics[width=\textwidth, height=0.7\textheight, keepaspectratio]{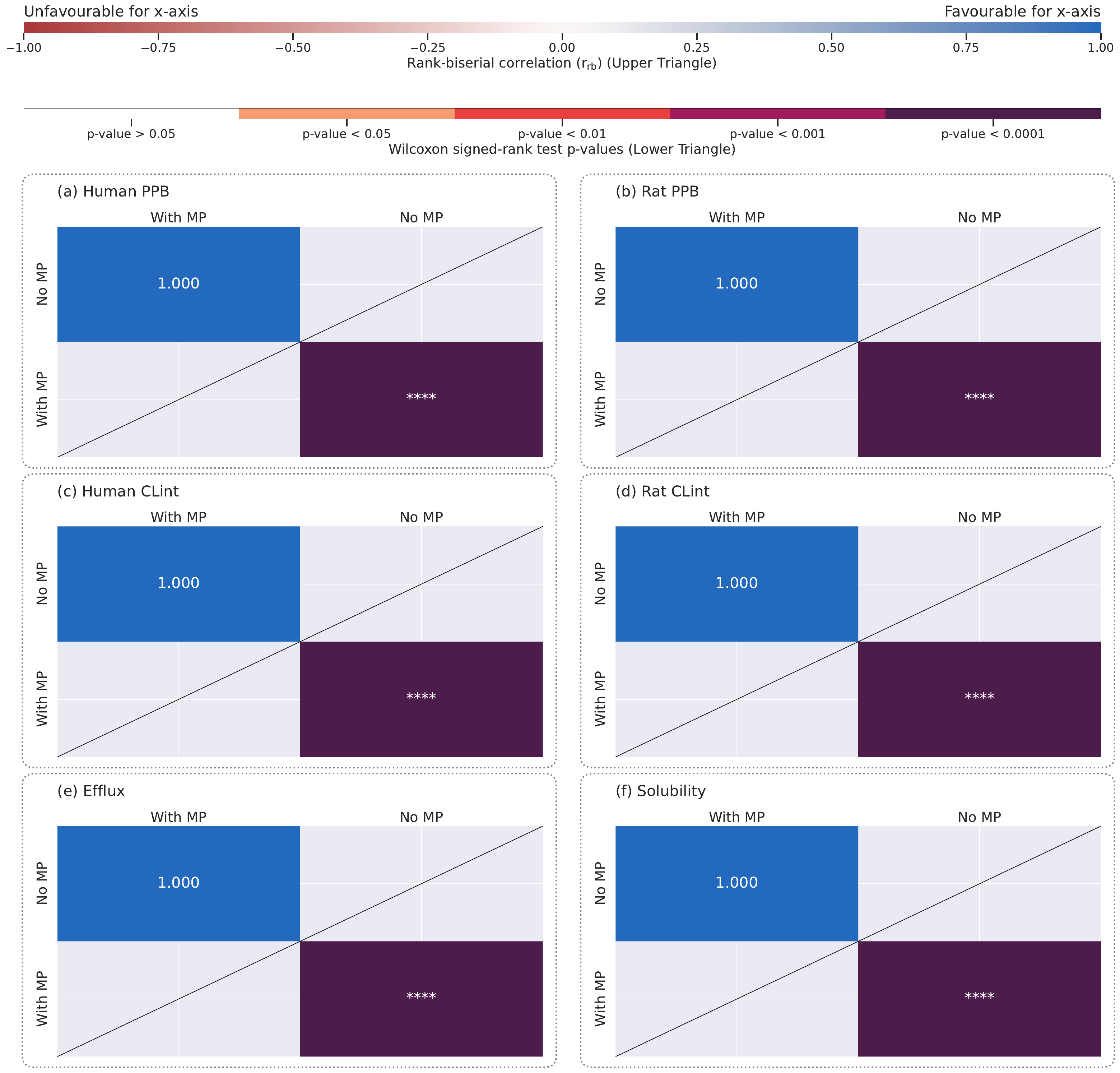}
    \caption{
        Rank-biserial correlation ($r_{rb}$) effect sizes (upper triangle) and two-sided Wilcoxon signed-rank test \textit{p}-values (lower triangle) of the Pearson correlation coefficient
        ($\uparrow \rho$) between pre-trained models with (With MP) and without message passing (No MP) on Biogen tasks. 
        Tasks are (a) Human Plasma Protein Binding (PPB), (b) Rat PPB, 
        (c) Human Intrinsic Hepatic Clearance (CLint), (d) Rat CLint, (e) Efflux, (f) Solubility.
        For $r_{rb}$ effect sizes (upper triangle): blue indicates the method on the \textit{x}-axis has a higher $\rho$ than 
        the method on the \textit{y}-axis; red indicates the opposite. 
        For Wilcoxon signed-rank test \textit{p}-values (lower triangle): 
        white indicates no significant difference and darker indicates a smaller \textit{p}-value ($n=20$, $\alpha=0.05$).
    }
    \label{fig:message_passing_ablation_pearson}
\end{figure}

\begin{figure}[!htbp]
    \centering
    \includegraphics[width=\textwidth, height=0.7\textheight, keepaspectratio]{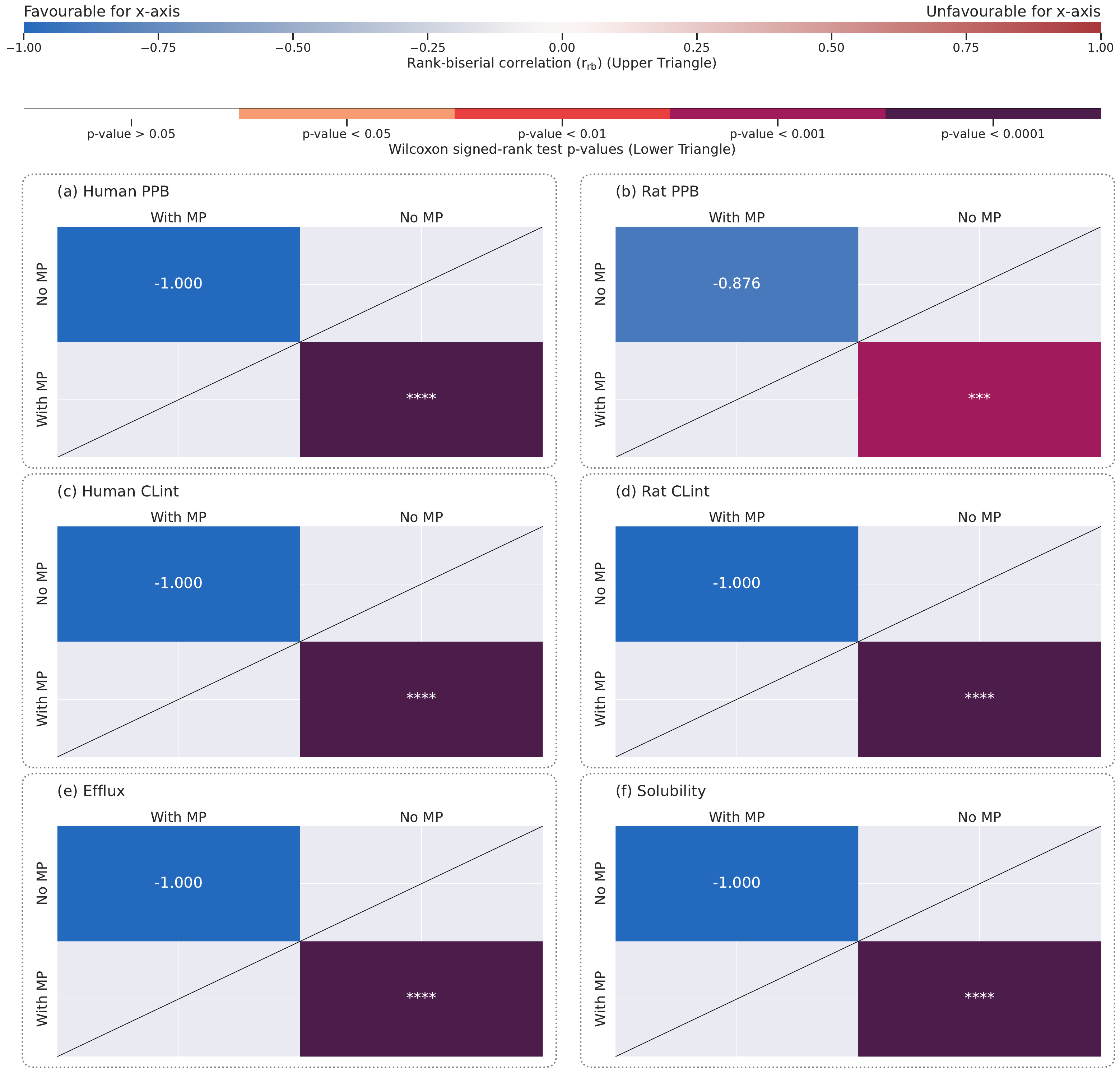}
    \caption{
        Rank-biserial correlation ($r_{rb}$) effect sizes (upper triangle) and two-sided Wilcoxon signed-rank test \textit{p}-values (lower triangle) of the mean absolute percentage error
        ($\downarrow\mathrm{MAPE}$) between pre-trained models with (With MP) and without message passing (No MP) on Biogen tasks.
        Tasks are (a) Human Plasma Protein Binding (PPB), (b) Rat PPB, 
        (c) Human Intrinsic Hepatic Clearance (CLint), (d) Rat CLint, (e) Efflux, (f) Solubility.
        For $r_{rb}$ effect sizes (upper triangle): blue indicates the method on the \textit{x}-axis has a lower $\mathrm{MAPE}$ than 
        the method on the \textit{y}-axis; red indicates the opposite. 
        For Wilcoxon signed-rank test \textit{p}-values (lower triangle): 
        white indicates no significant difference and darker indicates a smaller \textit{p}-value ($n=20$, $\alpha=0.05$).
    }
    \label{fig:message_passing_ablation_mape}
\end{figure}

\clearpage
\newpage

\subsection{Ablation study: impact of model size on Scratch GIN performance}\label{si:size_ablation}

Given the small amounts of data available in drug discovery benchmark datasets, we
constrained the size of Scratch GINs during hyperparameter optimisation such that the
number of trainable parameters was approximately between $10^0$ and $10^1$ times 
the size of the dataset. 
This allowed us to train models that were not overly complex for the available data,
and mitigate the risk of overfitting.

\begin{figure}[!htbp]
    \centering
    \includegraphics[width=\textwidth, height=0.7\textheight, keepaspectratio]{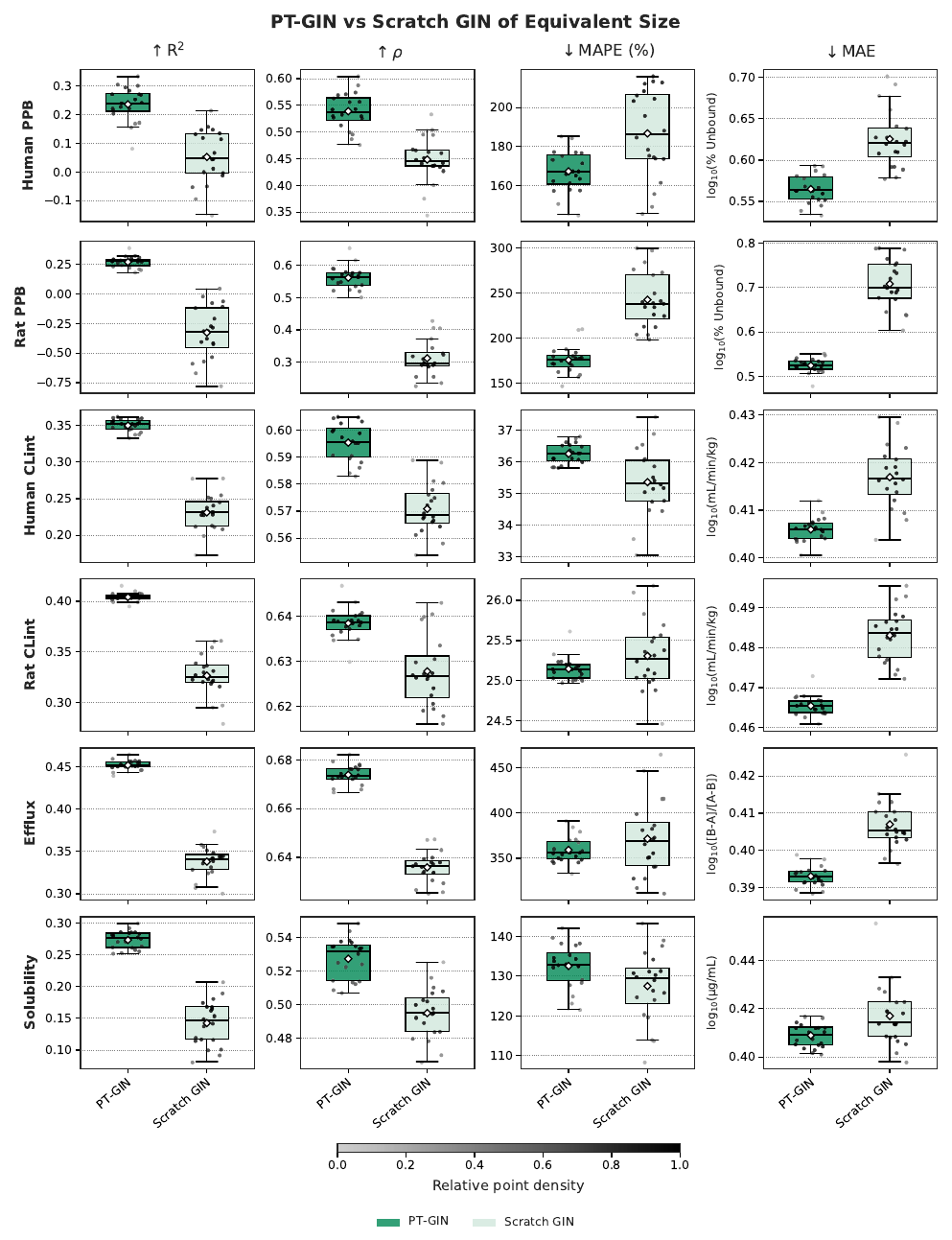}
    \caption{
        Performance metrics on Biogen regression tasks for PT-GIN (dark green)
        and Scratch GIN (light green)
        with the same model size, using the same number of 
        message passing convolutions ($3$), layers within each GIN layer ($3$),
        hidden dimensions ($512$), node dimensions ($512$),
        vocabulary sizes ($2048$), maximum substructure tokenisation radius ($2$), 
        prediction head layers ($2$), and prediction head hidden dimensions ($512$).
        Tasks shown are Human Plasma Protein Binding (PPB), Rat PPB,
        Human Intrinsic Hepatic Clearance (CLint), Rat CLint, Efflux, and Solubility.
        The metrics shown are coefficient of determination ($\uparrow\mathrm{R}^2$), 
        Pearson correlation coefficient ($\uparrow\rho$), mean absolute percentage error ($\downarrow$MAPE (\%)),
        and mean absolute error ($\downarrow$MAE).
    }
    \label{fig:scratch_gin_size_ablation}
\end{figure}

To validate whether this constraint was appropriate, we performed 
an ablation, comparing the performance 
of a Scratch GIN and PT-GIN model of equal size on Biogen tasks.
Our results show that PT-GINs outperform Scratch GINs of equal size
on all Biogen tasks across $\mathrm{R}^2$, $\rho$, and $\mathrm{MAE}$ (Supplementary Fig.~\ref{fig:scratch_gin_size_ablation}).
However, by $\mathrm{MAPE}$, Scratch GINs are more competitive with PT-GINs,
only clearly underperforming on both PPB tasks.
The discrepancy between $\mathrm{MAPE}$ and the other metrics likely
reflects the sensitivity of $\mathrm{MAPE}$ to errors from predicting smaller magnitude values.

These results suggest that the constraint on Scratch GIN size was appropriate,
as PT-GINs outperform Scratch GINs of equal size on all Biogen tasks across $\mathrm{R}^2$, $\rho$, and $\mathrm{MAE}$.

\clearpage
\newpage

\subsection{Example metric distribution plots}\label{si:distribution}

\subsubsection{Biogen}
\begin{figure}[!htbp]
    \centering
    \includegraphics[
        width=\textwidth, height=0.7\textheight, keepaspectratio
    ]{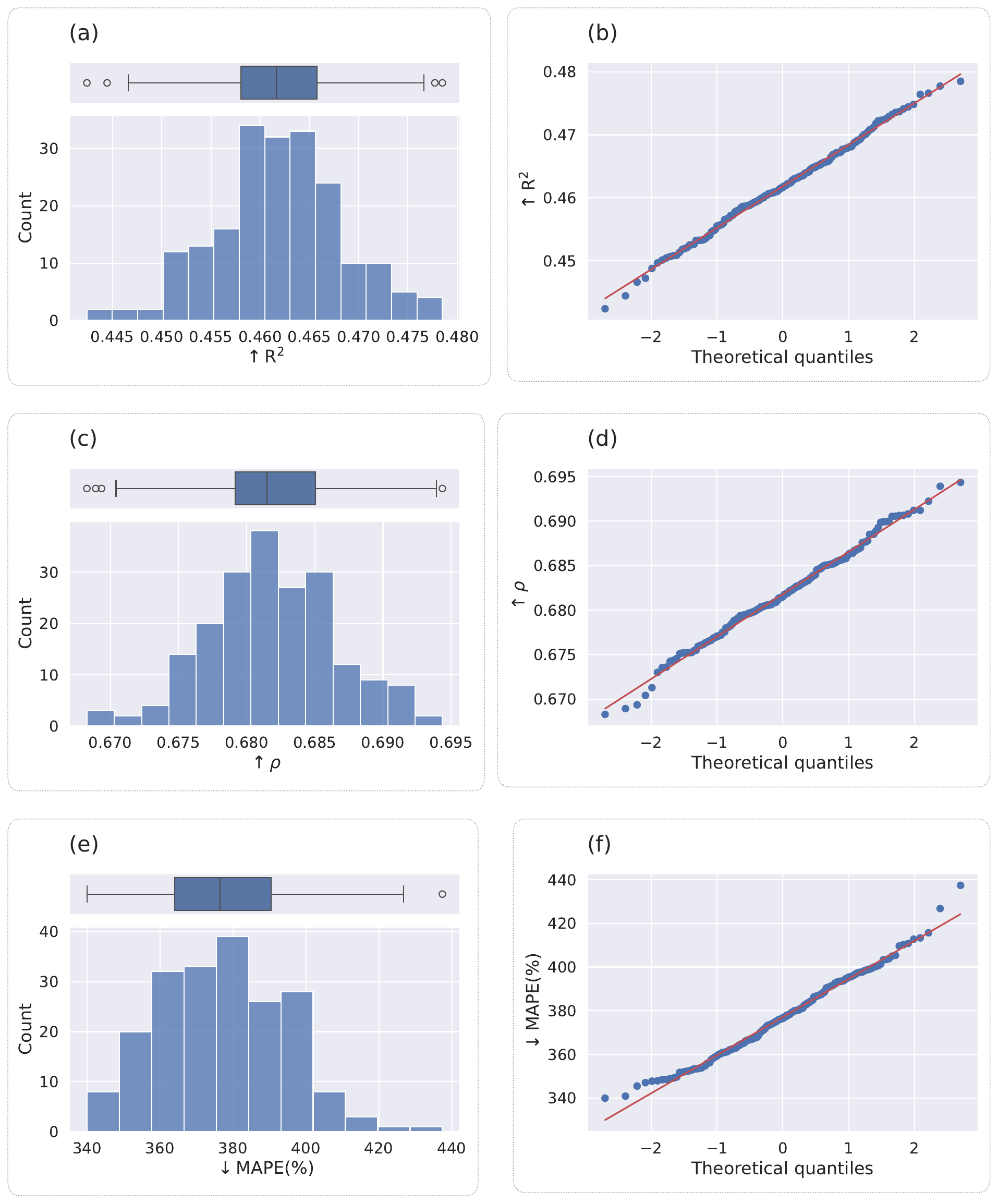}
    \caption{
        Metric distributions for PT-GIN on Efflux.
        The left column contains histograms and box plots; 
        the right column contains normal distribution Q-Q probability plots.
        Metrics shown are the (a, b) coefficient of determination ($\mathrm{R}^2$),
        (c, d) Pearson correlation coefficient ($\rho$), 
        and (e, f) mean absolute percentage error (MAPE (\%)).
    }
    \label{fig:dist_biogen_efflux}
\end{figure} 

\clearpage
\newpage
\subsubsection{MUV}

\begin{figure}[!htbp]
    \centering
    \includegraphics[
        width=\textwidth, height=0.7\textheight, keepaspectratio
    ]{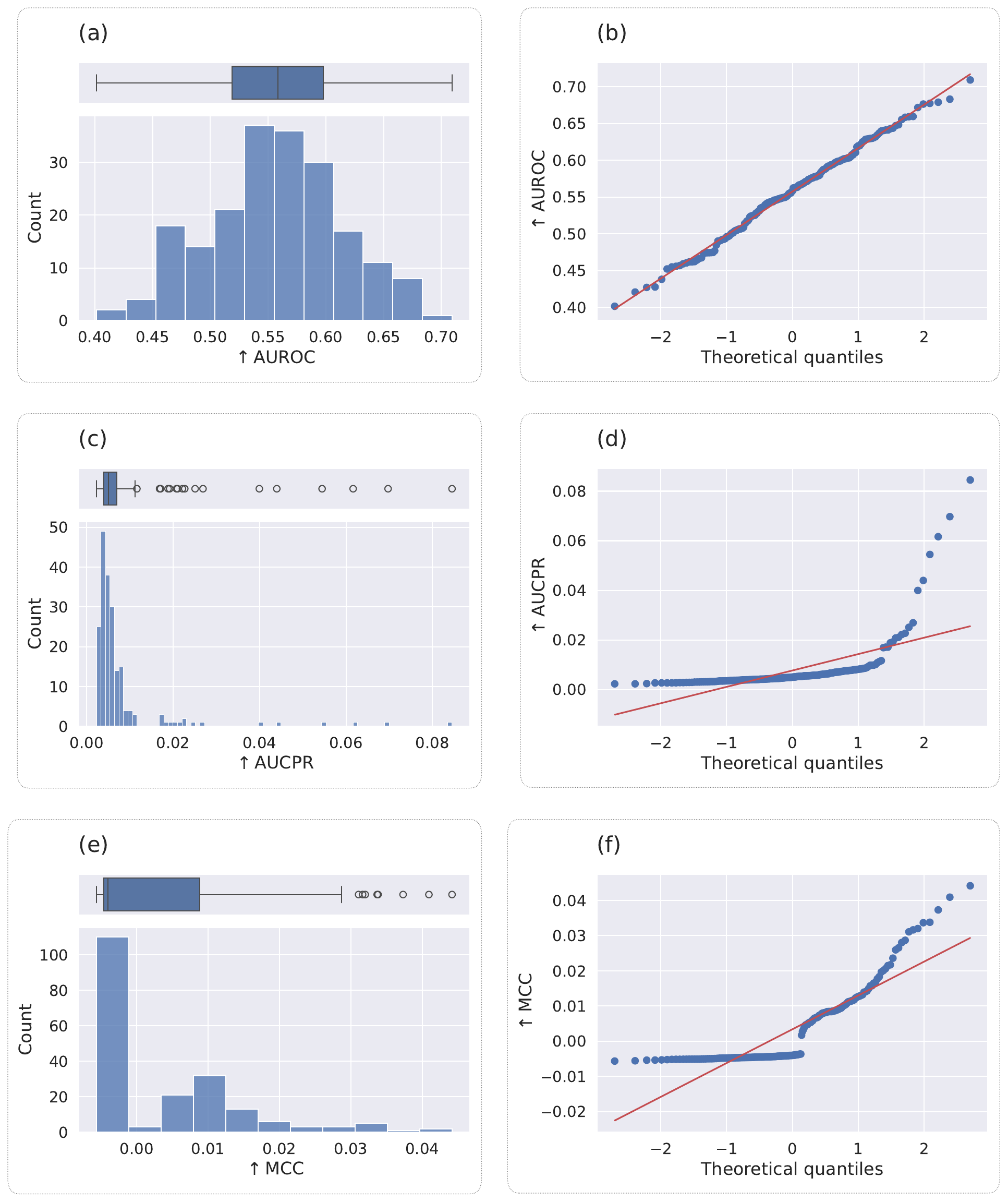}
    \caption{
        Metric distributions for PT-GIN on MUV466.
        The left column contains histograms and box plots; 
        the right column contains normal distribution Q-Q probability plots.
        Metrics shown are the 
        (a, b) area under the receiver operating characteristic curve ($\mathrm{AUROC}$),
        (c, d) area under the precision-recall curve ($\mathrm{AUCPR}$),
        and (e, f) Matthews correlation coefficient ($\mathrm{MCC}$).
    }
    \label{fig:dist_muv466}
\end{figure}

\begin{figure}[!htbp]
    \centering
    \includegraphics[
        width=\textwidth, height=0.7\textheight, keepaspectratio
    ]{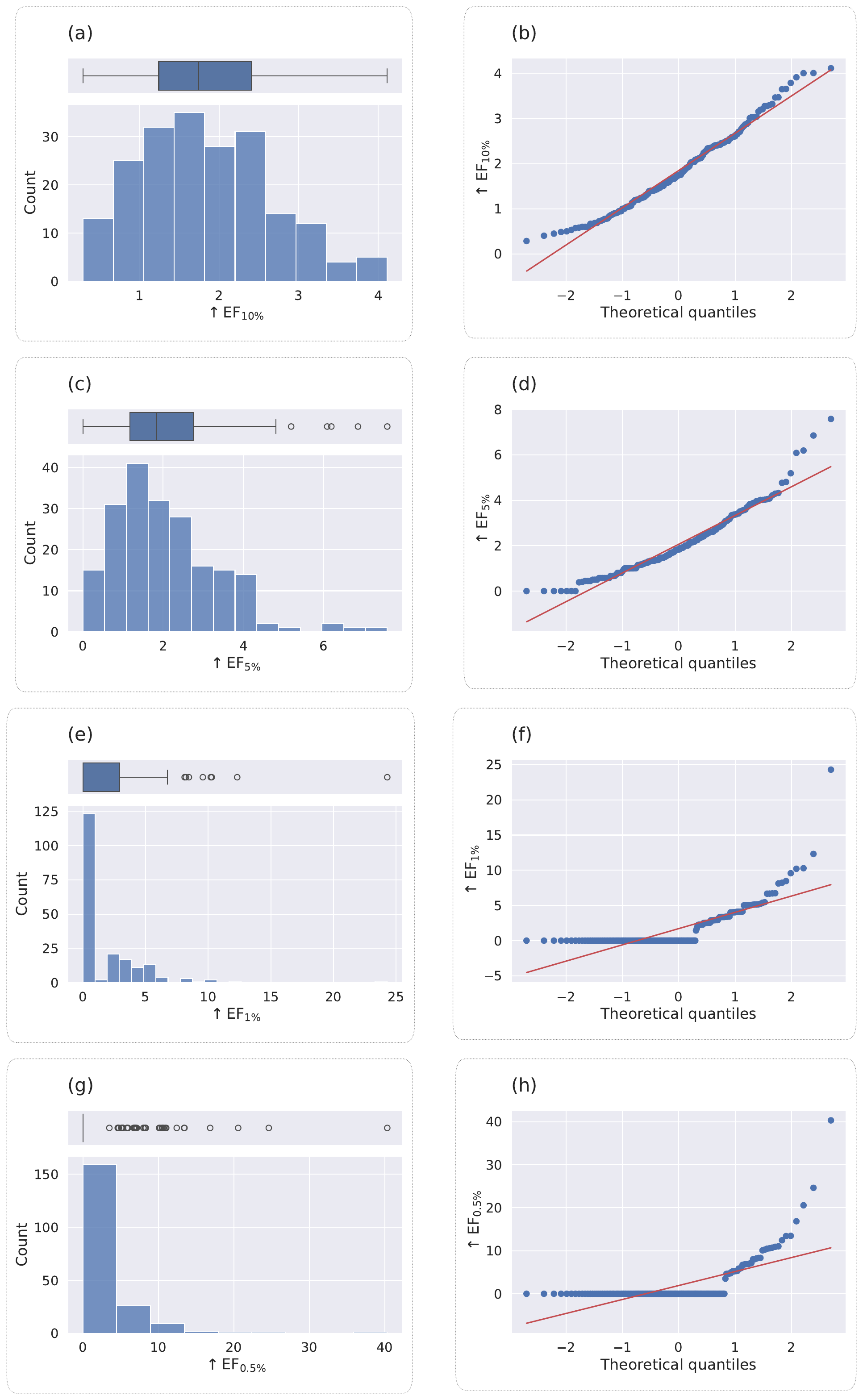}
    \caption{
        Enrichment Factor distributions for PT-GIN on MUV466.
        The left column contains histograms and box plots;
        the right column contains normal distribution Q-Q probability plots.
        Metrics shown are Enrichment Factor at
        (a, b) $10\%$,
        (c, d) $5\%$,
        (e, f) $1\%$,
        and (g, h) $0.5\%$.
    }
    \label{fig:dist_muv466_ef}
\end{figure}

\clearpage
\newpage

\section{Algorithms}

\begin{algorithm}
    \caption{Generating a molecular substructure (i.e., circular subgraph) vocabulary using Sort \& Slice}\label{alg:tokenizer}
    \begin{algorithmic}[1]
        \Require A molecular dataset, $\mathfrak{M}$
        \Require A Morgan algorithm-based circular substructure enumeration function, $\mathtt{morgan\_generator}$
        \Require The maximum radius for circular substructures, $r_{\max}$
        \Require The maximum substructure vocabulary size, $\mathtt{k}$
        \State $\mathtt{D} \gets \{\}$ \Comment{Initialise empty dictionary}
        \For{$\mathtt{molecule} \in \mathfrak{M}$} 
            \LineComment{Loop over molecules and tally the number of molecules each substructure appears in}
            \State $\mathfrak{S} \gets \mathtt{morgan\_generator}(\mathtt{molecule}, r_{\max})$ \Comment{Enumerate and hash circular substructures up to $r_{\max}$}
            \For{$(\mathtt{substructure}, \mathtt{radius}) \in \mathtt{set}(\mathfrak{S})$} \Comment{Loop over the set of substructure identifiers}
                \If{$\mathtt{substructure} \in \mathtt{D}$}
                    \State $\mathtt{D}[\mathtt{substructure}] \gets \mathtt{D}[\mathtt{substructure}] + 1$ \Comment{Add to count for substructure}
                \Else
                    \State $\mathtt{D}[\mathtt{substructure}] \gets 1$ \Comment{Initialise substructure count}
                \EndIf
            \EndFor
        \EndFor
        \State $\mathtt{D_{sorted}} \gets \mathtt{sort\_by\_value}(\mathtt{D}, \mathtt{descending})$ \Comment{Sort substructures by frequency}
        \State $\mathtt{D_{ranked}} \gets \mathtt{assign\_ranks}(\mathtt{D_{sorted}})$ \Comment{Assign integer ranks to substructures based on sorted positions}
        \State $\mathtt{vocab} \gets \mathtt{top\_k}(\mathtt{D_{ranked}}, \mathtt{k})$ \Comment{Slice dictionary to retain top-ranked substructures}
        \State $\mathtt{max\_id} \gets \max(\{\mathtt{vocab}[s] : s \in \text{keys}(\mathtt{vocab})\})$
        \State $\mathtt{vocab}[\mathtt{"UNK"}] \gets \mathtt{max\_id} + 1$ 
        \State \Return $\mathtt{vocab}$ \Comment{Return substructure vocabulary as a dictionary}
    \end{algorithmic}
\end{algorithm}

\begin{algorithm}
    \caption{Molecular substructure tokenisation}\label{alg:tokenisation}
    \begin{algorithmic}[1]
        \Require A $\mathtt{molecule}$
        \Require A dictionary output from Algorithm~\ref{alg:tokenizer}, $\mathtt{vocab}$
        \Require The same substructure enumeration function used for Algorithm~\ref{alg:tokenizer}, $\mathtt{morgan\_generator}$
        \Require The same maximum radius used for Algorithm~\ref{alg:tokenizer}, $r_{\max}$
        \State $\mathtt{graph} = (V, E)$ \textbf{where} $V \coloneq \mathtt{heavy\_atoms} \in \mathtt{molecule},\ E \coloneq \mathtt{bonds} \in \mathtt{molecule}$
        \State $\mathtt{graph.tokens} \gets \mathbf{0}^{|V| \times (r_{\max}+1)}$ \Comment{Initialise token tensor; rows for each atom and columns for each radius}
        \State $\mathfrak{S} \gets \mathtt{morgan\_generator}(\mathtt{molecule}, r_{\max})$ \Comment{Get circular substructures in the molecule}
        \For{$(\mathtt{substructure}, \mathtt{central\_atom}, \mathtt{radius}) \in \mathfrak{S}$} \Comment{Loop over all central atoms and their substructures}
            \If{$\mathtt{substructure} \in \mathtt{vocab}$}
                \State $\mathtt{token} \gets \mathtt{vocab}[\mathtt{substructure}]$ \Comment{Represent token using substructure rank}
            \Else
                \State $\mathtt{token} \gets \mathtt{vocab}[\mathtt{"UNK"}]$ \Comment{Get the unknown token}
            \EndIf
            \State $\mathtt{graph.tokens}[\mathtt{central\_atom}, \mathtt{radius}] \gets \mathtt{token}$ \Comment{Add substructure token to graph token tensor}
        \EndFor
        \State \Return $\mathtt{graph}$
    \end{algorithmic}
\end{algorithm}

\begin{algorithm}\small% arXiv: narrower sn-jnl text block
    \caption{Permute the embeddings of a graph for a particular token}\label{alg:permute}
    \begin{algorithmic}[1]
        \Require A $\mathtt{graph} = (V, E)$, \textbf{where} $n = |V|$
        \Require The token to permute $\mathtt{token} \in \{0, 1, \dots, v-1\}$, \textbf{where} $v$ is the token vocabulary size
        \Require A random index permutation for node embeddings $\mathtt{shuffled\_indices}\in \{0, 1, \dots, l-1\}^{l}$, \textbf{where} $l$ is the node embedding dimension
        \Require $l \geq |\mathtt{token\_locs}|$ \Comment{This assumption was always the case for our PT-GIN models}
        \Require Graph token tensor $\mathtt{graph.tokens} \in \{0, 1, \dots, v-1\}^{n \times m}$, \textbf{where} $m$ is the number of tokens per node
        \Require Graph embedding tensor $\mathtt{graph.F_v} \in \mathbb{R}^{n \times m \times l}$
        \LineComment{Find the locations of the token in the graph}
        \State $\mathtt{token\_locs} \gets \{(\mathtt{i}, \mathtt{j}) \mid \mathtt{graph.tokens}[\mathtt{i}, \mathtt{j}] = \mathtt{token},\ \mathtt{i} \in [0, \dots, n-1],\ \mathtt{j} \in [0, \dots, m-1]\}$
        \If{$\mathtt{token\_locs} = \emptyset$}
            \State \Return $\mathtt{graph}$ \Comment{If token is not present, return original graph}
        \EndIf
        \State $\mathtt{permuted\_graph} \gets \mathtt{copy}(\mathtt{graph})$
        \State $\mathtt{node\_embeddings} \gets \mathtt{permuted\_graph.F_v}$
        \State $\mathtt{start} \gets 0$
        \LineComment{Calculate the chunk size for permutations to distribute over the different locations of the token}
        \State $\mathtt{chunk\_size} \gets \left\lfloor \frac{l}{|\mathtt{token\_locs}|} \right\rfloor$
        \For{$(\mathtt{i}, \mathtt{j}) \in \mathtt{token\_locs}$} 
            \LineComment{Loop through instances of the token and distribute the permutations over the graph}
            \State $\mathtt{end} \gets \min(\mathtt{start} + \mathtt{chunk\_size}, l)$
            \State $\mathtt{shuffled\_subset} \gets \mathtt{shuffled\_indices}[\mathtt{start} : \mathtt{end}]$
            \State $\mathtt{node\_embeddings}[\mathtt{i}, \mathtt{j}, \mathtt{start} : \mathtt{end}] = \mathtt{node\_embeddings}[\mathtt{i}, \mathtt{j}, \mathtt{shuffled\_subset}]$
            \State $\mathtt{start} \gets \mathtt{end}$
        \EndFor
        \State $\mathtt{permuted\_graph.F_v} \gets \mathtt{node\_embeddings}$ 
        \State \Return $\mathtt{permuted\_graph}$
    \end{algorithmic}
\end{algorithm}

\begin{algorithm}\small% arXiv: narrower sn-jnl text block
    \caption{GNN substructure importance on one train-test split using embedding permutation}\label{alg:importance}
    \begin{algorithmic}[1]
        \Require A pre-trained graph neural network $\mathtt{GNN}$ with substructure tokenisation
        \Require Training data and labels, $\mathfrak{D}_{\text{train}}$ and $Y_{\text{train}}$
        \Require Testing data and labels, $\mathfrak{D}_{\text{test}}$ and $Y_{\text{test}}$
        \Require A $\mathtt{metric}$ function
        \Require The number of importance measurements to make, $\mathtt{iterations}$
        \Require Pre-trained $\mathtt{GNN}$ has a substructure token vocabulary, $\mathtt{vocab}$
        \Require Pre-trained $\mathtt{GNN}$ has learned node $\mathtt{embeddings} \in \mathbb{R}^{(|\mathtt{vocab}|) \times l}$, \textbf{where} $l$ is the embedding dimension
        \Require Pre-trained $\mathtt{GNN}$ has the substructure enumeration function used for tokenisation, $\mathtt{morgan\_generator}$
        \Require Pre-trained $\mathtt{GNN}$ has the maximum radius used for substructure tokenisation, $r_{\max}$
        \State $X_{\text{train}} \gets \mathtt{GNN}(\mathfrak{D}_{\mathtt{train}})$ \Comment{Compute frozen global embeddings of train data}
        \State $X_{\text{test}} \gets \mathtt{GNN}(\mathfrak{D}_{\mathtt{test}})$ \Comment{Compute frozen global embeddings of test data}
        \State $\mathtt{light\_gbm.fit}(X_{\text{train}}, Y_{\text{train}})$ \Comment{Fit a LightGBM model on training data}
        \State $\hat{Y}_{\text{test}} \gets \mathtt{light\_gbm}(X_{\text{test}})$
        \State $b_0 \gets \mathtt{metric}(\hat{Y}_{\text{test}}, Y_{\text{test}})$ \Comment{Calculate base metric score}
        \State $\mathtt{output} \gets \mathbf{0}^{|\mathtt{vocab}| \times \mathtt{iterations}}$ \Comment{Initialise output array}
        \Statex \vspace{0.25\baselineskip}
        \For{$\mathtt{iter} = 0, \dots, \mathtt{iterations} - 1$} \Comment{Loop over iterations}
            \For{$\mathtt{index} = 0, \dots, |\mathtt{vocab}| - 1$} \Comment{Loop over the vocabulary}
                \State $\mathtt{token} \gets \mathtt{index}$
                \State $\mathtt{shuffled\_indices} \gets \mathtt{shuffle\_array\_order([0, 1, \dots, l - 1])}$
                \State $X_{\text{permuted}} \gets [\quad]$
                \For{$\mathtt{molecule} \in \mathfrak{D}_{\text{test}}$}
                    \State $n \gets |\{\mathtt{heavy\_atoms} \in \mathtt{molecule}\}|$
                    \State $\mathtt{graph} \gets \mathtt{tokenize}(\mathtt{molecule}, \mathtt{vocab}, \mathtt{morgan\_generator}, r_{\max})$ \Comment{See Algorithm~\ref{alg:tokenisation}}
                    \State $\mathtt{graph} \gets \mathtt{permute\_graph}(\mathtt{graph}, \mathtt{token}, \mathtt{shuffled\_indices})$ \Comment{See Algorithm~\ref{alg:permute}}
                    \State $\mathtt{graph} \gets \mathtt{GNN.message\_passing(graph)}$
                    \State $\mathtt{x \in \mathbb{R}^h} \gets \mathtt{GNN.pooling(graph)}$, \textbf{where} $h$ is the dimension of the global embedding
                    \State $X_{\text{permuted}}.\mathtt{append}(x)$
                \EndFor
                \State $\hat{Y}_{\text{permuted}} \gets \mathtt{light\_gbm}(X_{\text{permuted}})$ \Comment{Predict label using permuted global embeddings}
                \State $b_{\text{permuted}} \gets \mathtt{metric}(\hat{Y}_{\text{permuted}}, Y_{\text{test}})$ \Comment{Calculate permuted metric}
                \State $\mathtt{score} \gets b_0 - b_{\text{permuted}}$ \Comment{Calculate metric difference}
                \State $\mathtt{output}[\mathtt{token}, \mathtt{iter}] \gets \mathtt{score}$ \Comment{Add score to output array}
            \EndFor
        \EndFor
        \State $\mathtt{output} \gets \mathtt{mean}(\mathtt{output}, \text{axis} = 1)$
        \State \Return $\mathtt{output}$
    \end{algorithmic}
\end{algorithm}

\clearpage
\newpage

\section{Package versions}

\begin{table}[!htbp]
    \centering
    \caption{
        Python package versions used in this work. Note: PyTorch was bumped to version 2.8.0 for security reasons 
        (along with PyTorch Geometric for compatibility) after the completion of this work.
    }
    \label{tab:package_versions}
    \begin{tabular}{lr}
        \toprule
        \textbf{Package} & \textbf{Version} \\
        \midrule
        LightGBM & 4.6.0\\
        NumPy & 2.2.3 \\
        Optuna & 4.2.1 \\
        Pandas & 2.2.3 \\
        PyTorch & 2.5.1 \\
        PyTorch Geometric & 2.6.1 \\
        RDKit & 2024.09.03 \\
        Scikit-learn & 1.6.0 \\
        Scikit-posthocs & 0.11.4 \\
        SciPy & 1.15.2 \\
        TorchEval & 0.0.7 \\
        \bottomrule
    \end{tabular}
\end{table}

\clearpage
\newpage
\section{Datasets}\label{si:datasets}
\subsection{Dataset summaries}

\begin{tableorg}[!htbp]% arXiv: plain float so the tabular can be scaled
    \centering
    \caption{Summary of the benchmark datasets used in this study. The number of molecules shown is the total number of SMILES per-task that passed RDKit standardisation.}
    \label{tab:datasets}
    \resizebox{\textwidth}{!}{% arXiv: narrower sn-jnl text block
    \begin{tabular}{llr}
        \toprule
        \textbf{Dataset} & \textbf{Source} & \textbf{Number of Molecules} \\
        \midrule
        Efflux & Biogen~\cite{fangC-2023-ProspectiveValidationMachineLearningAlgorithmsAbsorptionDistributionMetabolismExcretionPrediction} 
        & 2642\\
        Human Plasma Protein Binding & Biogen~\cite{fangC-2023-ProspectiveValidationMachineLearningAlgorithmsAbsorptionDistributionMetabolismExcretionPrediction} 
        & 194\\
        Human Intrinsic Hepatic Clearance & Biogen~\cite{fangC-2023-ProspectiveValidationMachineLearningAlgorithmsAbsorptionDistributionMetabolismExcretionPrediction} 
        & 3087\\
        Rat Plasma Protein Binding & Biogen~\cite{fangC-2023-ProspectiveValidationMachineLearningAlgorithmsAbsorptionDistributionMetabolismExcretionPrediction} 
        & 168\\
        Rat Intrinsic Hepatic Clearance & Biogen~\cite{fangC-2023-ProspectiveValidationMachineLearningAlgorithmsAbsorptionDistributionMetabolismExcretionPrediction} 
        & 3054 \\
        Solubility & Biogen~\cite{fangC-2023-ProspectiveValidationMachineLearningAlgorithmsAbsorptionDistributionMetabolismExcretionPrediction} 
        & 2173\\
        ESOL & MoleculeNet~\cite{wuZ-2018-MoleculeNet}, 
        ESOL~\cite{delaneyJS-2004-ESOL} 
        & 1069\\
        FreeSolv & MoleculeNet~\cite{wuZ-2018-MoleculeNet}, 
        FreeSolv~\cite{mobleyDL-2014-FreeSolv} 
        & 640\\
        Lipophilicity & MoleculeNet~\cite{wuZ-2018-MoleculeNet}, ChEMBL~\cite{zdrazilB-2024-ChEMBLDatabase2023} & 4200\\
        MUV466& MoleculeNet~\cite{wuZ-2018-MoleculeNet}, Rohrer et al.~\cite{rohrerSG-2009-MaximumUnbiasedValidationMUVData} & 14844\\
        MUV548& MoleculeNet~\cite{wuZ-2018-MoleculeNet}, Rohrer et al.~\cite{rohrerSG-2009-MaximumUnbiasedValidationMUVData} & 14737\\
        MUV600& MoleculeNet~\cite{wuZ-2018-MoleculeNet}, Rohrer et al.~\cite{rohrerSG-2009-MaximumUnbiasedValidationMUVData} & 14734\\
        MUV644& MoleculeNet~\cite{wuZ-2018-MoleculeNet}, Rohrer et al.~\cite{rohrerSG-2009-MaximumUnbiasedValidationMUVData} & 14633\\
        MUV652& MoleculeNet~\cite{wuZ-2018-MoleculeNet}, Rohrer et al.~\cite{rohrerSG-2009-MaximumUnbiasedValidationMUVData} & 14903\\
        MUV689& MoleculeNet~\cite{wuZ-2018-MoleculeNet}, Rohrer et al.~\cite{rohrerSG-2009-MaximumUnbiasedValidationMUVData} & 14606\\
        MUV692& MoleculeNet~\cite{wuZ-2018-MoleculeNet}, Rohrer et al.~\cite{rohrerSG-2009-MaximumUnbiasedValidationMUVData} & 14647\\
        MUV712& MoleculeNet~\cite{wuZ-2018-MoleculeNet}, Rohrer et al.~\cite{rohrerSG-2009-MaximumUnbiasedValidationMUVData} & 14415\\
        MUV713& MoleculeNet~\cite{wuZ-2018-MoleculeNet}, Rohrer et al.~\cite{rohrerSG-2009-MaximumUnbiasedValidationMUVData} & 14841\\
        MUV733& MoleculeNet~\cite{wuZ-2018-MoleculeNet}, Rohrer et al.~\cite{rohrerSG-2009-MaximumUnbiasedValidationMUVData} & 14691\\
        MUV737& MoleculeNet~\cite{wuZ-2018-MoleculeNet}, Rohrer et al.~\cite{rohrerSG-2009-MaximumUnbiasedValidationMUVData} & 14696\\
        MUV810& MoleculeNet~\cite{wuZ-2018-MoleculeNet}, Rohrer et al.~\cite{rohrerSG-2009-MaximumUnbiasedValidationMUVData} & 14646\\
        MUV832& MoleculeNet~\cite{wuZ-2018-MoleculeNet}, Rohrer et al.~\cite{rohrerSG-2009-MaximumUnbiasedValidationMUVData} & 14676\\
        MUV846& MoleculeNet~\cite{wuZ-2018-MoleculeNet}, Rohrer et al.~\cite{rohrerSG-2009-MaximumUnbiasedValidationMUVData} & 14714\\
        MUV852& MoleculeNet~\cite{wuZ-2018-MoleculeNet}, Rohrer et al.~\cite{rohrerSG-2009-MaximumUnbiasedValidationMUVData} & 14658\\
        MUV858& MoleculeNet~\cite{wuZ-2018-MoleculeNet}, Rohrer et al.~\cite{rohrerSG-2009-MaximumUnbiasedValidationMUVData} & 14775\\
        MUV859& MoleculeNet~\cite{wuZ-2018-MoleculeNet}, Rohrer et al.~\cite{rohrerSG-2009-MaximumUnbiasedValidationMUVData} & 14751\\
        NR AR & MoleculeNet~\cite{wuZ-2018-MoleculeNet}, Tox21~\cite{richardAM-2021-Tox2110KCompoundLibrary} &7257\\
        NR AR LBD & MoleculeNet~\cite{wuZ-2018-MoleculeNet}, Tox21~\cite{richardAM-2021-Tox2110KCompoundLibrary} &6750\\
        NR AhR & MoleculeNet~\cite{wuZ-2018-MoleculeNet}, Tox21~\cite{richardAM-2021-Tox2110KCompoundLibrary} &6541\\
        NR Aromatase & MoleculeNet~\cite{wuZ-2018-MoleculeNet}, Tox21~\cite{richardAM-2021-Tox2110KCompoundLibrary} &5814\\
        NR ER& MoleculeNet~\cite{wuZ-2018-MoleculeNet}, Tox21~\cite{richardAM-2021-Tox2110KCompoundLibrary} & 6185\\
        NR ER LBD& MoleculeNet~\cite{wuZ-2018-MoleculeNet}, Tox21~\cite{richardAM-2021-Tox2110KCompoundLibrary} & 6947\\
        NR PPAR $\gamma$ & MoleculeNet~\cite{wuZ-2018-MoleculeNet}, Tox21~\cite{richardAM-2021-Tox2110KCompoundLibrary} &6442\\
        SR ARE & MoleculeNet~\cite{wuZ-2018-MoleculeNet}, Tox21~\cite{richardAM-2021-Tox2110KCompoundLibrary} &5824\\
        SR ATAD5 & MoleculeNet~\cite{wuZ-2018-MoleculeNet}, Tox21~\cite{richardAM-2021-Tox2110KCompoundLibrary} &7064\\
        SR HSE & MoleculeNet~\cite{wuZ-2018-MoleculeNet}, Tox21~\cite{richardAM-2021-Tox2110KCompoundLibrary} &6459\\
        SR MMP & MoleculeNet~\cite{wuZ-2018-MoleculeNet}, Tox21~\cite{richardAM-2021-Tox2110KCompoundLibrary} &5804\\
        SR p53 & MoleculeNet~\cite{wuZ-2018-MoleculeNet}, Tox21~\cite{richardAM-2021-Tox2110KCompoundLibrary} &6766\\
        DRD2 & ChEMBL~\cite{zdrazilB-2024-ChEMBLDatabase2023}, Dablander et al.~\cite{dablanderM-2023-ExploringQSARModelsActivitycliffPrediction} & 6333 \\
        Factor Xa & ChEMBL~\cite{zdrazilB-2024-ChEMBLDatabase2023}, Dablander et al.~\cite{dablanderM-2023-ExploringQSARModelsActivitycliffPrediction} & 3605\\
        \bottomrule
    \end{tabular}
    }%
\end{tableorg}

\begin{table}[!htbp]
    \centering
    \caption{
        Total number of test sets used for metric evaluation for MUV tasks.
        As MUV tasks are highly imbalanced, 
        StratifiedGroupKFold splitting occasionally produces one fold with no positive samples.
        Folds without positive samples were excluded as test sets for metric evaluation;
        cross-validation averages were determined using the remaining available folds 
        for each repeat.
    }
    \label{tab:muv_n_train_test_splits}
    \begin{tabular}{lr}
        \toprule
        \textbf{Dataset} & \textbf{Number of Train-Test Splits with} \\
                         & \textbf{Positive Samples in the Test Set} \\
        \midrule
        MUV466 & 991\\
        MUV548 & 990\\
        MUV600 & 994\\
        MUV644 & 989\\
        MUV652 & 993\\
        MUV689 & 992\\
        MUV692 & 994\\
        MUV712 & 991\\
        MUV713 & 990\\
        MUV733 & 992\\
        MUV737 & 986\\
        MUV810 & 994\\
        MUV832 & 992\\
        MUV846 & 989\\
        MUV852 & 994\\
        MUV858 & 992\\
        MUV859 & 987\\
        \bottomrule
    \end{tabular}
\end{table}

\begin{tableorg}[!htbp]% arXiv: plain float so the tabular can be scaled
    \centering
    \caption{
        Tox21 toxicity tasks~\cite{richardAM-2021-Tox2110KCompoundLibrary}.
        See \url{https://tripod.nih.gov/tox21/challenge/data.jsp}.
        NR = nuclear receptor signalling pathway, SR = stress response pathway.
        All twelve tasks correspond to quantitative high-throughput screening (qHTS) assay endpoints.
    }
    \label{tab:tox21_tasks}
    \resizebox{\textwidth}{!}{% arXiv: narrower sn-jnl text block
    \begin{tabular}{llr}
        \toprule
        Abbreviation & Label & PubChem Assay ID \\
        \midrule
        NR AR & androgen receptor activation & $743040$ \\
        NR AR LBD & androgen receptor ligand binding domain activation & $743053$ \\
        NR AhR & aryl hydrocarbon receptor activation & $743122$ \\
        NR Aromatase & aromatase enzyme inhibition & $743139$ \\
        NR ER & estrogen receptor alpha activation & $743079$ \\
        NR ER LBD & estrogen receptor alpha ligand binding domain activation & $743077$ \\
        NR PPAR $\gamma$ & peroxisome proliferator-activated receptor $\gamma$ activation & $743140$ \\
        SR ARE & antioxidant responsive element signalling pathway activation & $743219$ \\
        SR ATAD5 & ATPase family AAA domain containing 5 expression & $720516$ \\
        SR HSE & heat shock element signalling pathway activation & $743228$\\
        SR MMP & mitochondrial membrane potential disruption& $720637$ \\
        SR p53 & p53 signalling pathway activation & $720552$ \\
        \bottomrule
    \end{tabular}
    }%
\end{tableorg}

\begin{figure}[!htbp]
    \centering
    \includegraphics[width=\textwidth, height=0.7\textheight, keepaspectratio]{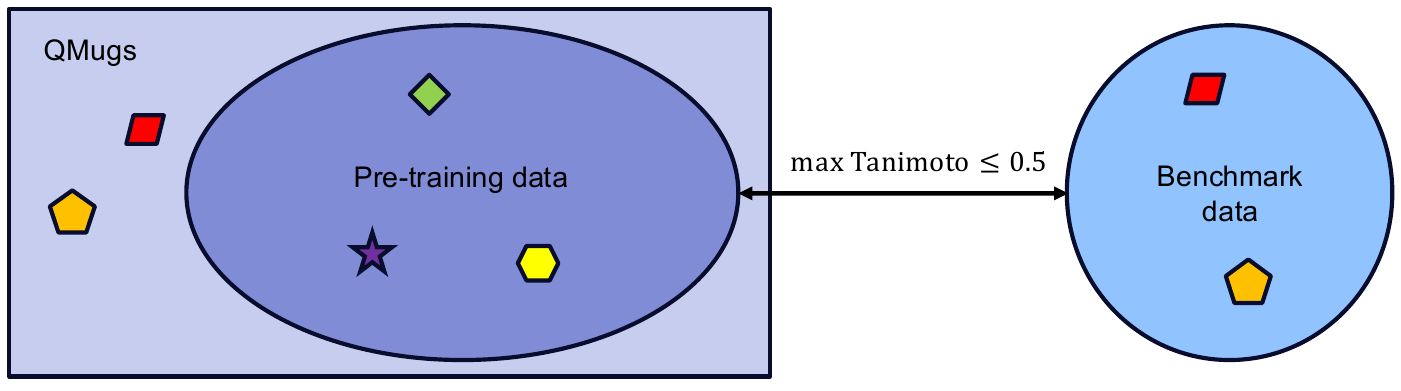}
    \caption{
        Tanimoto similarity filtering of the pre-training dataset.
    }
    \label{fig:tanimoto_filtering_sets}
\end{figure}

\clearpage
\newpage
\subsection{SMILES failures}\label{si:smiles_failures}

SMILES that failed RDKit standardisation and were excluded from pre-training, training, and testing are listed below.

% QMugs: Hypervalent P with a valence of 7 (1 SMILES)
\begin{flushleft}
\noindent\textbf{QMugs: Hypervalent P with a valence of 7 ($n = 1$)}
\begin{itemize}[nosep]
	\item \texttt{F[PH]\allowbreak{}(F)\allowbreak{}(F)\allowbreak{}(F)\allowbreak{}(F)\allowbreak{}F.\allowbreak{}N=\allowbreak{}c1ccc2c\allowbreak{}(-c3ccccc3)\allowbreak{}c3ccc\allowbreak{}(N)\allowbreak{}cc3sc-2c1}
\end{itemize}
\end{flushleft}

% ESOL: Hypervalent N with a valence of 5 (58 SMILES)
\begin{flushleft}
\begin{multicols}{2}[\noindent\textbf{ESOL: Hypervalent N with a valence of 5 ($n = 58$)}]
\begin{itemize}[nosep]
	\item \texttt{CN\allowbreak{}(=\allowbreak{}O)\allowbreak{}=\allowbreak{}O}
	\item \texttt{CCN\allowbreak{}(=\allowbreak{}O)\allowbreak{}=\allowbreak{}O}
	\item \texttt{CCCN\allowbreak{}(=\allowbreak{}O)\allowbreak{}=\allowbreak{}O}
	\item \texttt{CC\allowbreak{}(C)\allowbreak{}N\allowbreak{}(=\allowbreak{}O)\allowbreak{}=\allowbreak{}O}
	\item \texttt{O=\allowbreak{}N\allowbreak{}(=\allowbreak{}O)\allowbreak{}c1ccccc1}
	\item \texttt{Cc1ccccc1N\allowbreak{}(=\allowbreak{}O)\allowbreak{}=\allowbreak{}O}
	\item \texttt{Nc1ccccc1N\allowbreak{}(=\allowbreak{}O)\allowbreak{}=\allowbreak{}O}
	\item \texttt{Oc1ccccc1N\allowbreak{}(=\allowbreak{}O)\allowbreak{}=\allowbreak{}O}
	\item \texttt{COc1ccccc1N\allowbreak{}(=\allowbreak{}O)\allowbreak{}=\allowbreak{}O}
	\item \texttt{Clc1ccccc1N\allowbreak{}(=\allowbreak{}O)\allowbreak{}=\allowbreak{}O}
	\item \texttt{Cc1ccc\allowbreak{}(cc1)\allowbreak{}N\allowbreak{}(=\allowbreak{}O)\allowbreak{}=\allowbreak{}O}
	\item \texttt{Cc1cccc\allowbreak{}(c1)\allowbreak{}N\allowbreak{}(=\allowbreak{}O)\allowbreak{}=\allowbreak{}O}
	\item \texttt{ClC\allowbreak{}(Cl)\allowbreak{}(Cl)\allowbreak{}N\allowbreak{}(=\allowbreak{}O)\allowbreak{}=\allowbreak{}O}
	\item \texttt{Nc1ccc\allowbreak{}(cc1)\allowbreak{}N\allowbreak{}(=\allowbreak{}O)\allowbreak{}=\allowbreak{}O}
	\item \texttt{Nc1cccc\allowbreak{}(c1)\allowbreak{}N\allowbreak{}(=\allowbreak{}O)\allowbreak{}=\allowbreak{}O}
	\item \texttt{Oc1ccc\allowbreak{}(cc1)\allowbreak{}N\allowbreak{}(=\allowbreak{}O)\allowbreak{}=\allowbreak{}O}
	\item \texttt{Oc1cccc\allowbreak{}(c1)\allowbreak{}N\allowbreak{}(=\allowbreak{}O)\allowbreak{}=\allowbreak{}O}
	\item \texttt{COc1ccc\allowbreak{}(cc1)\allowbreak{}N\allowbreak{}(=\allowbreak{}O)\allowbreak{}=\allowbreak{}O}
	\item \texttt{Clc1ccc\allowbreak{}(cc1)\allowbreak{}N\allowbreak{}(=\allowbreak{}O)\allowbreak{}=\allowbreak{}O}
	\item \texttt{Clc1cccc\allowbreak{}(c1)\allowbreak{}N\allowbreak{}(=\allowbreak{}O)\allowbreak{}=\allowbreak{}O}
	\item \texttt{Cc1ncc\allowbreak{}(N\allowbreak{}(=\allowbreak{}O)\allowbreak{}=\allowbreak{}O)\allowbreak{}n1CCO}
	\item \texttt{ON=\allowbreak{}Cc1ccc\allowbreak{}(o1)\allowbreak{}N\allowbreak{}(=\allowbreak{}O)\allowbreak{}=\allowbreak{}O}
	\item \texttt{O=\allowbreak{}N\allowbreak{}(=\allowbreak{}O)\allowbreak{}c1cccc2ccccc12}
	\item \texttt{O=\allowbreak{}N\allowbreak{}(=\allowbreak{}O)\allowbreak{}c1ccccc1N\allowbreak{}(=\allowbreak{}O)\allowbreak{}=\allowbreak{}O}
	\item \texttt{O=\allowbreak{}N\allowbreak{}(=\allowbreak{}O)\allowbreak{}c1c\allowbreak{}(Cl)\allowbreak{}c\allowbreak{}(Cl)\allowbreak{}ccc1}
	\item \texttt{O=\allowbreak{}N\allowbreak{}(=\allowbreak{}O)\allowbreak{}c1cc\allowbreak{}(Cl)\allowbreak{}c\allowbreak{}(Cl)\allowbreak{}cc1}
	\item \texttt{CC\allowbreak{}(=\allowbreak{}O)\allowbreak{}Nc1ccc\allowbreak{}(cc1)\allowbreak{}N\allowbreak{}(=\allowbreak{}O)\allowbreak{}=\allowbreak{}O}
	\item \texttt{Nc1cc\allowbreak{}(nc\allowbreak{}(N)\allowbreak{}n1=\allowbreak{}O)\allowbreak{}N2CCCCC2}
	\item \texttt{O=\allowbreak{}N\allowbreak{}(=\allowbreak{}O)\allowbreak{}c1ccc\allowbreak{}(cc1)\allowbreak{}N\allowbreak{}(=\allowbreak{}O)\allowbreak{}=\allowbreak{}O}
	\item \texttt{O=\allowbreak{}N\allowbreak{}(=\allowbreak{}O)\allowbreak{}c1cccc\allowbreak{}(c1)\allowbreak{}N\allowbreak{}(=\allowbreak{}O)\allowbreak{}=\allowbreak{}O}
	\item \texttt{Cc1c\allowbreak{}(cccc1N\allowbreak{}(=\allowbreak{}O)\allowbreak{}=\allowbreak{}O)\allowbreak{}N\allowbreak{}(=\allowbreak{}O)\allowbreak{}=\allowbreak{}O}
	\item \texttt{Cc1ccc\allowbreak{}(cc1N\allowbreak{}(=\allowbreak{}O)\allowbreak{}=\allowbreak{}O)\allowbreak{}N\allowbreak{}(=\allowbreak{}O)\allowbreak{}=\allowbreak{}O}
	\item \texttt{O=\allowbreak{}C1NCCN1c2ncc\allowbreak{}(s2)\allowbreak{}N\allowbreak{}(=\allowbreak{}O)\allowbreak{}=\allowbreak{}O}
	\item \texttt{Cc1cc\allowbreak{}(cc\allowbreak{}(N\allowbreak{}(=\allowbreak{}O)\allowbreak{}=\allowbreak{}O)\allowbreak{}c1O)\allowbreak{}N\allowbreak{}(=\allowbreak{}O)\allowbreak{}=\allowbreak{}O}
	\item \texttt{O=\allowbreak{}N\allowbreak{}(=\allowbreak{}O)\allowbreak{}OCC\allowbreak{}(CON\allowbreak{}(=\allowbreak{}O)\allowbreak{}=\allowbreak{}O)\allowbreak{}ON\allowbreak{}(=\allowbreak{}O)\allowbreak{}=\allowbreak{}O}
	\item \texttt{O=\allowbreak{}C\allowbreak{}(Cn1ccnc1N\allowbreak{}(=\allowbreak{}O)\allowbreak{}=\allowbreak{}O)\allowbreak{}NCc2ccccc2}
	\item \texttt{CCOP\allowbreak{}(=\allowbreak{}S)\allowbreak{}(OCC)\allowbreak{}Oc1ccc\allowbreak{}(cc1)\allowbreak{}N\allowbreak{}(=\allowbreak{}O)\allowbreak{}=\allowbreak{}O}
	\item \texttt{COP\allowbreak{}(=\allowbreak{}S)\allowbreak{}(OC)\allowbreak{}Oc1ccc\allowbreak{}(cc1Cl)\allowbreak{}N\allowbreak{}(=\allowbreak{}O)\allowbreak{}=\allowbreak{}O}
	\item \texttt{COP\allowbreak{}(=\allowbreak{}S)\allowbreak{}(OC)\allowbreak{}Oc1ccc\allowbreak{}(N\allowbreak{}(=\allowbreak{}O)\allowbreak{}=\allowbreak{}O)\allowbreak{}c\allowbreak{}(C)\allowbreak{}c1}
	\item \texttt{CCC\allowbreak{}(C)\allowbreak{}c1cc\allowbreak{}(cc\allowbreak{}(N\allowbreak{}(=\allowbreak{}O)\allowbreak{}=\allowbreak{}O)\allowbreak{}c1O)\allowbreak{}N\allowbreak{}(=\allowbreak{}O)\allowbreak{}=\allowbreak{}O}
	\item \texttt{Clc1c\allowbreak{}(Cl)\allowbreak{}c\allowbreak{}(Cl)\allowbreak{}c\allowbreak{}(N\allowbreak{}(=\allowbreak{}O)\allowbreak{}=\allowbreak{}O)\allowbreak{}c\allowbreak{}(Cl)\allowbreak{}c1Cl}
	\item \texttt{O=\allowbreak{}N\allowbreak{}(=\allowbreak{}O)\allowbreak{}c1cc\allowbreak{}(cc\allowbreak{}(c1)\allowbreak{}N\allowbreak{}(=\allowbreak{}O)\allowbreak{}=\allowbreak{}O)\allowbreak{}N\allowbreak{}(=\allowbreak{}O)\allowbreak{}=\allowbreak{}O}
	\item \texttt{Cc1c\allowbreak{}(cc\allowbreak{}(cc1N\allowbreak{}(=\allowbreak{}O)\allowbreak{}=\allowbreak{}O)\allowbreak{}N\allowbreak{}(=\allowbreak{}O)\allowbreak{}=\allowbreak{}O)\allowbreak{}N\allowbreak{}(=\allowbreak{}O)\allowbreak{}=\allowbreak{}O}
	\item \texttt{Clc2ccc\allowbreak{}(Oc1ccc\allowbreak{}(cc1)\allowbreak{}N\allowbreak{}(=\allowbreak{}O)\allowbreak{}=\allowbreak{}O)\allowbreak{}c\allowbreak{}(Cl)\allowbreak{}c2}
	\item \texttt{O=\allowbreak{}C2CN\allowbreak{}(N=\allowbreak{}Cc1ccc\allowbreak{}(o1)\allowbreak{}N\allowbreak{}(=\allowbreak{}O)\allowbreak{}=\allowbreak{}O)\allowbreak{}C\allowbreak{}(=\allowbreak{}O)\allowbreak{}N2}
	\item \texttt{O=\allowbreak{}C3CN=\allowbreak{}C\allowbreak{}(c1ccccc1)\allowbreak{}c2cc\allowbreak{}(ccc2N3)\allowbreak{}N\allowbreak{}(=\allowbreak{}O)\allowbreak{}=\allowbreak{}O}
	\item \texttt{Oc1ccc\allowbreak{}(Cl)\allowbreak{}cc1C\allowbreak{}(=\allowbreak{}O)\allowbreak{}Nc2ccc\allowbreak{}(cc2Cl)\allowbreak{}N\allowbreak{}(=\allowbreak{}O)\allowbreak{}=\allowbreak{}O}
	\item \texttt{Clc1ccccc1C2=\allowbreak{}NCC\allowbreak{}(=\allowbreak{}O)\allowbreak{}Nc3ccc\allowbreak{}(cc23)\allowbreak{}N\allowbreak{}(=\allowbreak{}O)\allowbreak{}=\allowbreak{}O}
	\item \texttt{OCC\allowbreak{}(NC\allowbreak{}(=\allowbreak{}O)\allowbreak{}C\allowbreak{}(Cl)\allowbreak{}Cl)\allowbreak{}C\allowbreak{}(O)\allowbreak{}c1ccc\allowbreak{}(cc1)\allowbreak{}N\allowbreak{}(=\allowbreak{}O)\allowbreak{}=\allowbreak{}O}
	\item \texttt{CCCN\allowbreak{}(CCC)\allowbreak{}c1c\allowbreak{}(cc\allowbreak{}(cc1N\allowbreak{}(=\allowbreak{}O)\allowbreak{}=\allowbreak{}O)\allowbreak{}C\allowbreak{}(C)\allowbreak{}C)\allowbreak{}N\allowbreak{}(=\allowbreak{}O)\allowbreak{}=\allowbreak{}O}
	\item \texttt{CN2C\allowbreak{}(=\allowbreak{}O)\allowbreak{}CN=\allowbreak{}C\allowbreak{}(c1ccccc1)\allowbreak{}c3cc\allowbreak{}(ccc23)\allowbreak{}N\allowbreak{}(=\allowbreak{}O)\allowbreak{}=\allowbreak{}O}
	\item \texttt{CCCCN\allowbreak{}(CC)\allowbreak{}c1c\allowbreak{}(cc\allowbreak{}(cc1N\allowbreak{}(=\allowbreak{}O)\allowbreak{}=\allowbreak{}O)\allowbreak{}C\allowbreak{}(F)\allowbreak{}(F)\allowbreak{}F)\allowbreak{}N\allowbreak{}(=\allowbreak{}O)\allowbreak{}=\allowbreak{}O}
	\item \texttt{CCCN\allowbreak{}(CCC)\allowbreak{}c1c\allowbreak{}(cc\allowbreak{}(cc1N\allowbreak{}(=\allowbreak{}O)\allowbreak{}=\allowbreak{}O)\allowbreak{}C\allowbreak{}(F)\allowbreak{}(F)\allowbreak{}F)\allowbreak{}N\allowbreak{}(=\allowbreak{}O)\allowbreak{}=\allowbreak{}O}
	\item \texttt{CCN\allowbreak{}(CC)\allowbreak{}c1c\allowbreak{}(cc\allowbreak{}(c\allowbreak{}(N)\allowbreak{}c1N\allowbreak{}(=\allowbreak{}O)\allowbreak{}=\allowbreak{}O)\allowbreak{}C\allowbreak{}(F)\allowbreak{}(F)\allowbreak{}F)\allowbreak{}N\allowbreak{}(=\allowbreak{}O)\allowbreak{}=\allowbreak{}O}
	\item \texttt{CCCN\allowbreak{}(CCC)\allowbreak{}c1c\allowbreak{}(cc\allowbreak{}(cc1N\allowbreak{}(=\allowbreak{}O)\allowbreak{}=\allowbreak{}O)\allowbreak{}S\allowbreak{}(N)\allowbreak{}(=\allowbreak{}O)\allowbreak{}=\allowbreak{}O)\allowbreak{}N\allowbreak{}(=\allowbreak{}O)\allowbreak{}=\allowbreak{}O}
	\item \texttt{CCN\allowbreak{}(CC\allowbreak{}(C)\allowbreak{}=\allowbreak{}C)\allowbreak{}c1c\allowbreak{}(cc\allowbreak{}(cc1N\allowbreak{}(=\allowbreak{}O)\allowbreak{}=\allowbreak{}O)\allowbreak{}C\allowbreak{}(F)\allowbreak{}(F)\allowbreak{}F)\allowbreak{}N\allowbreak{}(=\allowbreak{}O)\allowbreak{}=\allowbreak{}O}
	\item \texttt{COC\allowbreak{}(=\allowbreak{}O)\allowbreak{}C1=\allowbreak{}C\allowbreak{}(C)\allowbreak{}NC\allowbreak{}(=\allowbreak{}C\allowbreak{}(C1c2\allowbreak{}ccccc2N\allowbreak{}(=\allowbreak{}O)\allowbreak{}=\allowbreak{}O)\allowbreak{}C\allowbreak{}(=\allowbreak{}O)\allowbreak{}OC)\allowbreak{}C}
	\item \texttt{CCN\allowbreak{}(Cc1c\allowbreak{}(F)\allowbreak{}cccc1Cl)\allowbreak{}c2c\allowbreak{}(cc\allowbreak{}(cc2N\allowbreak{}(=\allowbreak{}O)\allowbreak{}=\allowbreak{}O)\allowbreak{}C\allowbreak{}(F)\allowbreak{}(F)\allowbreak{}F)\allowbreak{}N\allowbreak{}(=\allowbreak{}O)\allowbreak{}=\allowbreak{}O}
\end{itemize}
\end{multicols}
\end{flushleft}

% FreeSolv: Hypervalent N with a valence of 5 (2 SMILES)
\begin{flushleft}
\noindent\textbf{FreeSolv: Hypervalent N with a valence of 5 ($n = 2$)}
\begin{itemize}[nosep]
	\item \texttt{c1cc\allowbreak{}(ccc1N)\allowbreak{}N\allowbreak{}(=\allowbreak{}O)\allowbreak{}=\allowbreak{}O}
	\item \texttt{c1\allowbreak{}(c\allowbreak{}(c\allowbreak{}(c\allowbreak{}(c\allowbreak{}(c1Cl)\allowbreak{}Cl)\allowbreak{}Cl)\allowbreak{}Cl)\allowbreak{}Cl)\allowbreak{}N\allowbreak{}(=\allowbreak{}O)\allowbreak{}=\allowbreak{}O}
\end{itemize}
\end{flushleft}

% Tox21: Hypervalent Sb with a valence of 9 (1 SMILES)
\begin{flushleft}
\noindent\textbf{Tox21: Hypervalent Sb with a valence of 9 ($n = 1$)}
\begin{itemize}[nosep]
	\item \texttt{[SbH6+3]\allowbreak{}}
\end{itemize}
\end{flushleft}

\clearpage
\newpage

% Tox21: Hypervalent Al with a valence of 6 (8 SMILES)
\begin{flushleft}
\noindent\textbf{Tox21: Hypervalent Al with a valence of 6 ($n = 8$)}
\begin{itemize}[nosep]
	\item \texttt{CC\allowbreak{}(=\allowbreak{}O)\allowbreak{}O[AlH3]\allowbreak{}(O)\allowbreak{}O}
	\item \texttt{O=\allowbreak{}CO[AlH3]\allowbreak{}(OC=\allowbreak{}O)\allowbreak{}OC=\allowbreak{}O}
	\item \texttt{CC\allowbreak{}(=\allowbreak{}O)\allowbreak{}O[AlH3]\allowbreak{}(O)\allowbreak{}OC\allowbreak{}(C)\allowbreak{}=\allowbreak{}O}
	\item \texttt{CCCCO[AlH3]\allowbreak{}(OCCCC)\allowbreak{}OCCCC}
	\item \texttt{NC\allowbreak{}(=\allowbreak{}O)\allowbreak{}NC1N=\allowbreak{}C\allowbreak{}(O[AlH3]\allowbreak{}(O)\allowbreak{}O)\allowbreak{}NC1=\allowbreak{}O}
	\item \texttt{CCCCCCCCCCCCCCCCCC\allowbreak{}(=\allowbreak{}O)\allowbreak{}O[AlH3]\allowbreak{}(O)\allowbreak{}O}
	\item \texttt{CCOC\allowbreak{}(=\allowbreak{}O)\allowbreak{}/C=\allowbreak{}C\allowbreak{}(/C)\allowbreak{}O[AlH3]\allowbreak{}(OC\allowbreak{}(C)\allowbreak{}CC)\allowbreak{}OC\allowbreak{}(C)\allowbreak{}CC}
	\item \texttt{O=\allowbreak{}S\allowbreak{}(=\allowbreak{}O)\allowbreak{}(OC[C@H]\allowbreak{}1O[C@H]\allowbreak{}(O[C@]\allowbreak{}2\allowbreak{}(COS\allowbreak{}(=\allowbreak{}O)\allowbreak{}(=\allowbreak{}O)\allowbreak{}O[AlH3]\allowbreak{}(O)\allowbreak{}O)\allowbreak{}O[C@H]\allowbreak{}(COS\allowbreak{}(=\allowbreak{}O)\allowbreak{}(=\allowbreak{}O)\allowbreak{}O[AlH3]\allowbreak{}(O)\allowbreak{}O)\allowbreak{}[C@@H]\allowbreak{}(OS\allowbreak{}(=\allowbreak{}O)\allowbreak{}(=\allowbreak{}O)\allowbreak{}O[AlH3]\allowbreak{}(O)\allowbreak{}O)\allowbreak{}[C@@H]\allowbreak{}2OS\allowbreak{}(=\allowbreak{}O)\allowbreak{}(=\allowbreak{}O)\allowbreak{}O[AlH3]\allowbreak{}(O)\allowbreak{}O)\allowbreak{}[C@H]\allowbreak{}(OS\allowbreak{}(=\allowbreak{}O)\allowbreak{}(=\allowbreak{}O)\allowbreak{}O[AlH3]\allowbreak{}(O)\allowbreak{}O)\allowbreak{}[C@@H]\allowbreak{}(OS\allowbreak{}(=\allowbreak{}O)\allowbreak{}(=\allowbreak{}O)\allowbreak{}O[AlH3]\allowbreak{}(O)\allowbreak{}O)\allowbreak{}[C@@H]\allowbreak{}1OS\allowbreak{}(=\allowbreak{}O)\allowbreak{}(=\allowbreak{}O)\allowbreak{}O[AlH3]\allowbreak{}(O)\allowbreak{}O)\allowbreak{}O[AlH3]\allowbreak{}(O)\allowbreak{}O.\allowbreak{}O[AlH3]\allowbreak{}(O)\allowbreak{}[AlH3]\allowbreak{}(O)\allowbreak{}O.\allowbreak{}O[AlH3]\allowbreak{}(O)\allowbreak{}[AlH3]\allowbreak{}(O)\allowbreak{}O.\allowbreak{}O[AlH3]\allowbreak{}(O)\allowbreak{}[AlH3]\allowbreak{}(O)\allowbreak{}O.\allowbreak{}O[AlH3]\allowbreak{}(O)\allowbreak{}[AlH3]\allowbreak{}(O)\allowbreak{}O}
\end{itemize}
\end{flushleft}

\clearpage
\newpage
\subsection{Maximum train-test similarities across splits}

To assess the degree of similarity between training and test sets for each task, 
we computed the maximum Tanimoto similarity to the training set for each molecule every time it 
was included in the test set. We then averaged these maximum similarities across 
all test splits for each molecule, and plotted the resulting distributions for each task.
For comparison, we computed the same distributions for random splits, 
and quantified the difference between the Butina and Random distributions 
using the Wasserstein distance ($\mathrm{W}_{1}$), with sign-flip permutation testing.

\begin{figure}[!htbp]
    \centering
    \includegraphics[scale=0.32]{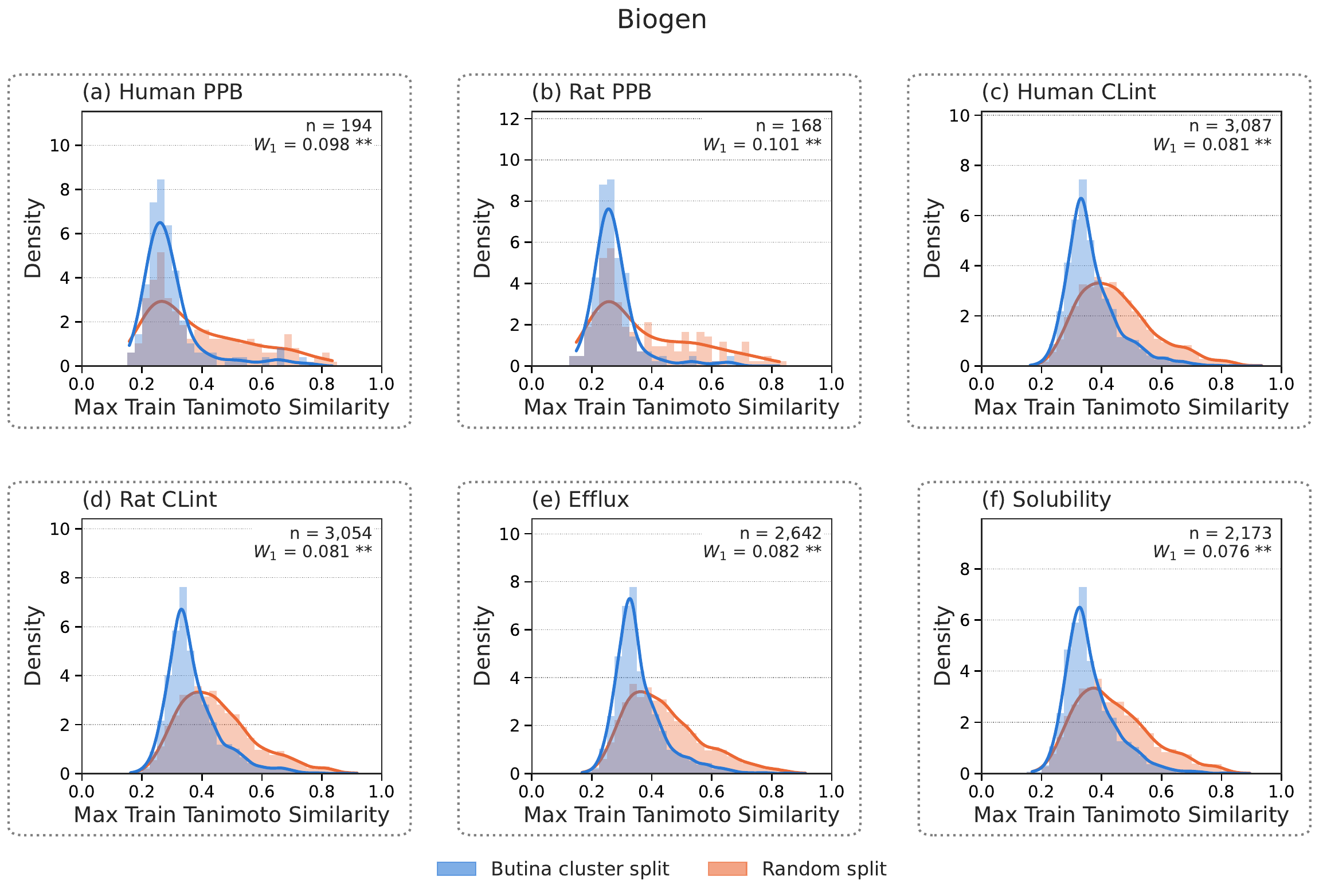}
    \caption{
        Maximum train-test Tanimoto similarities for Biogen tasks. 
        (a) Human Plasma Protein Binding (PPB); (b) Rat PPB; 
        (c) Human Intrinsic Hepatic Clearance (CLint); (d) Rat CLint; (e) Efflux; (f) Solubility.
        Blue is Butina GroupKFold splitting, and orange is Random splitting.
        $\mathrm{W}_{1}$ is the Wasserstein distance between the Butina and Random distributions; asterisks indicate statistical significance ($p < 0.05$)
        using sign-flip permutation testing with 9999 permutations.
    }
    \label{fig:train_test_similarities_biogen}
\end{figure}

\begin{figure}[!htbp]
    \centering
    \includegraphics[scale=0.32]{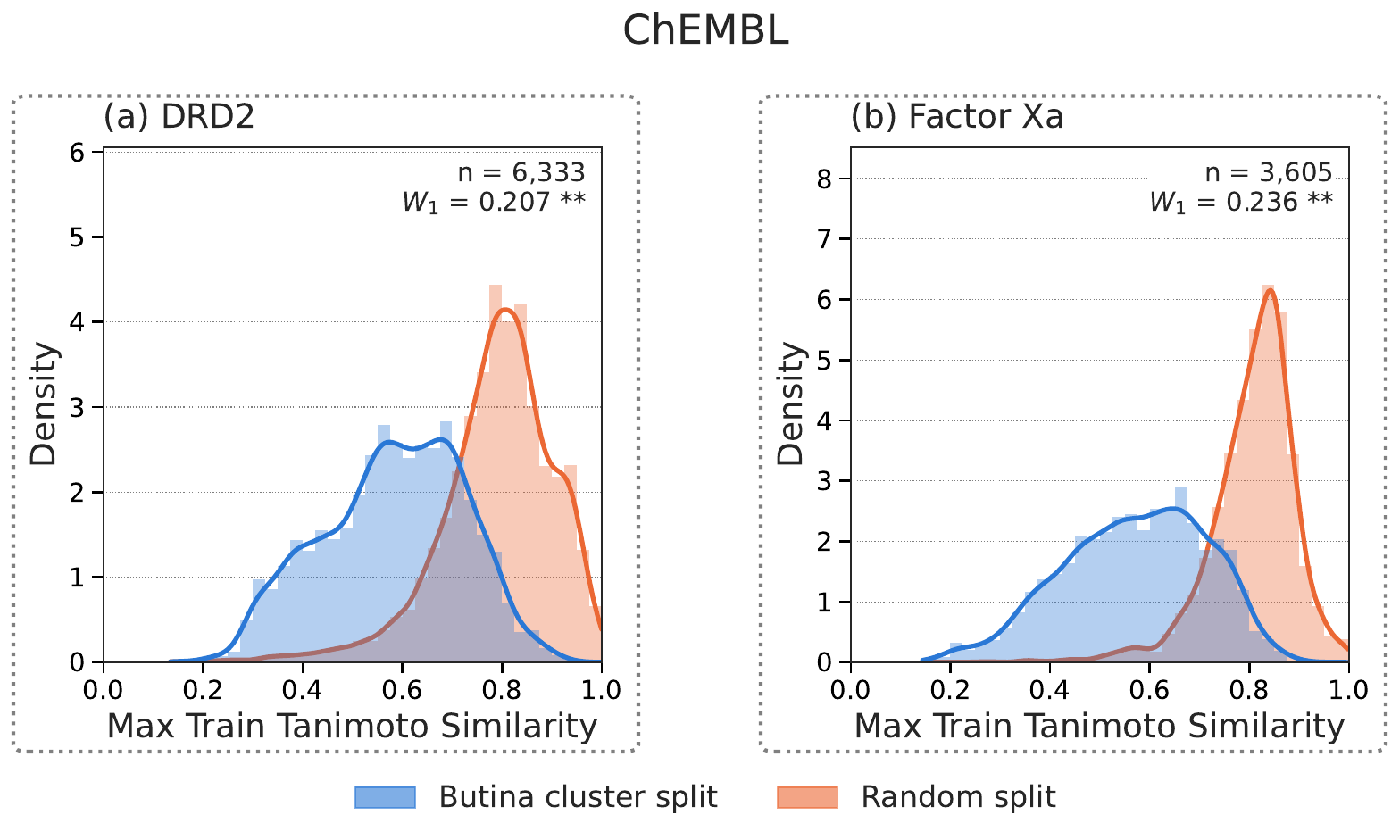}
    \caption{
        Maximum train-test Tanimoto similarities for ChEMBL binding affinity tasks.
        (a) DRD2; (b) Factor Xa.
        Blue is Butina GroupKFold splitting, and orange is Random splitting.
        $\mathrm{W}_{1}$ is the Wasserstein distance between the Butina and Random distributions; asterisks indicate statistical significance ($p < 0.05$)
        using sign-flip permutation testing with 9999 permutations.
    }
    \label{fig:train_test_similarities_chembl}
\end{figure}

\begin{figure}[!htbp]
    \centering
    \includegraphics[width=\textwidth, height=0.7\textheight, keepaspectratio]{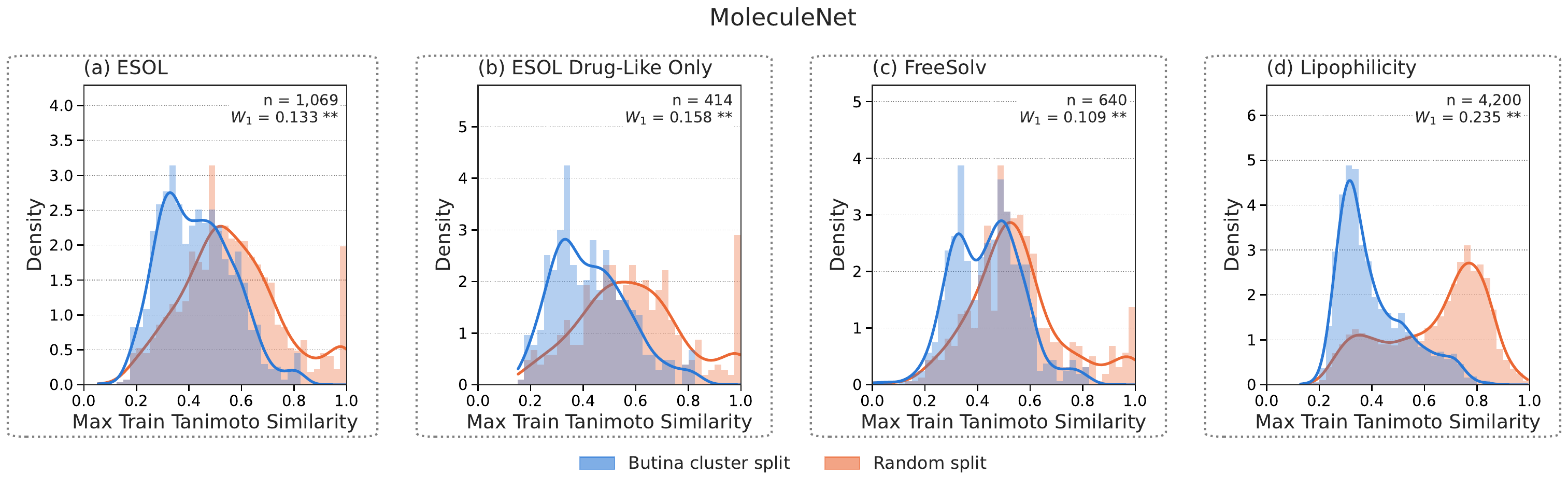}
    \caption{
        Maximum train-test Tanimoto similarities for MoleculeNet regression tasks.
        (a) ESOL; (b) ESOL restricted to drug-like molecules; (c) FreeSolv; (d) Lipophilicity.
        Blue is Butina GroupKFold splitting, and orange is Random splitting.
        $\mathrm{W}_{1}$ is the Wasserstein distance between the Butina and Random distributions; asterisks indicate statistical significance ($p < 0.05$)
        using sign-flip permutation testing with 9999 permutations.
    }
    \label{fig:train_test_similarities_molnet}
\end{figure}

\begin{figure}[!htbp]
    \centering
    \includegraphics[width=\textwidth, height=0.8\textheight, keepaspectratio]{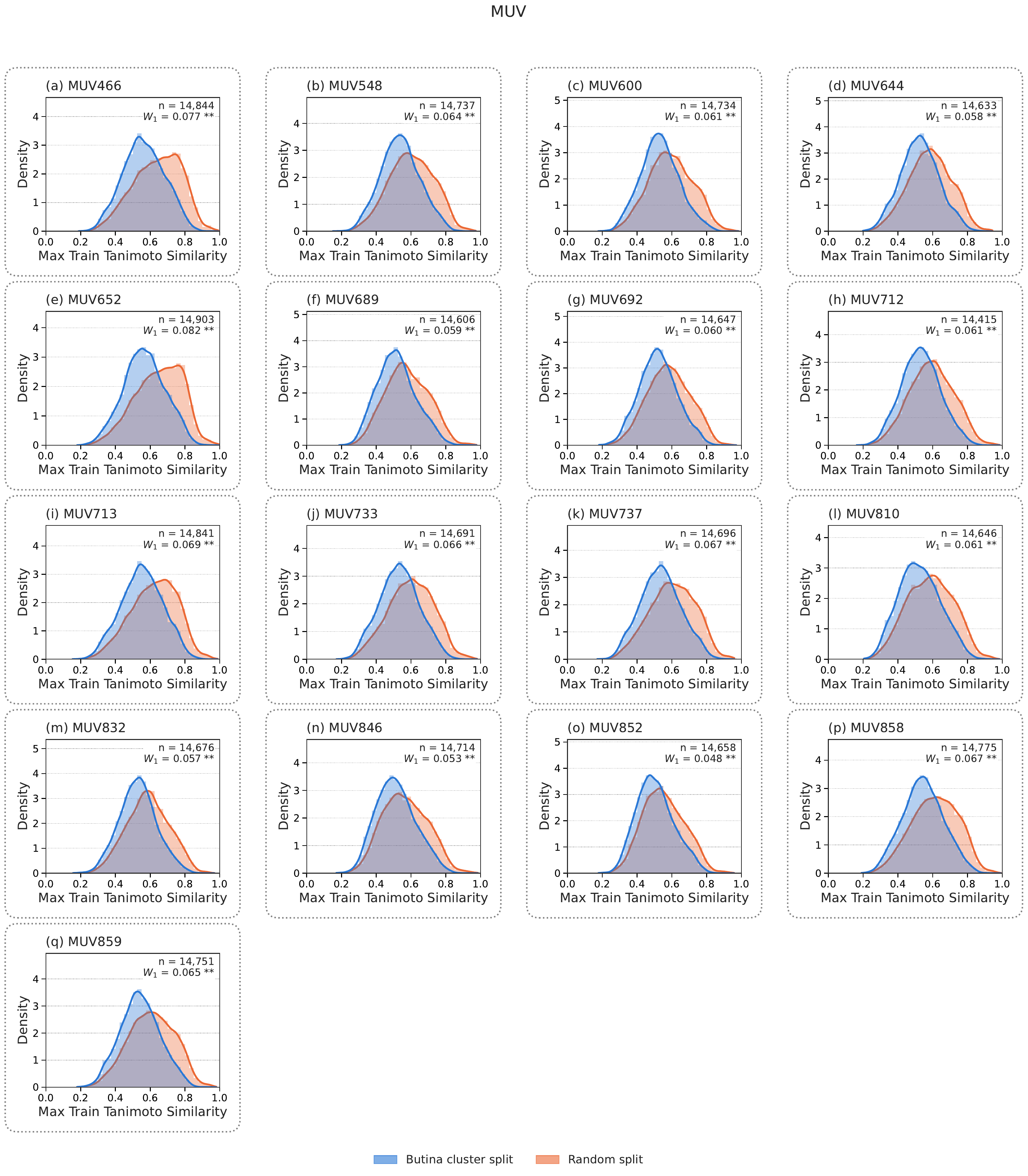}
    \caption{
        Maximum train-test Tanimoto similarities for MUV virtual screening tasks.
        Blue is Butina StratifiedGroupKFold splitting, and orange is Random splitting.
        $\mathrm{W}_{1}$ is the Wasserstein distance between the Butina and Random distributions; asterisks indicate statistical significance ($p < 0.05$)
        using sign-flip permutation testing with 9999 permutations.
    }
    \label{fig:train_test_similarities_muv}
\end{figure}

\begin{figure}[!htbp]
    \centering
    \includegraphics[width=\textwidth, height=0.7\textheight, keepaspectratio]{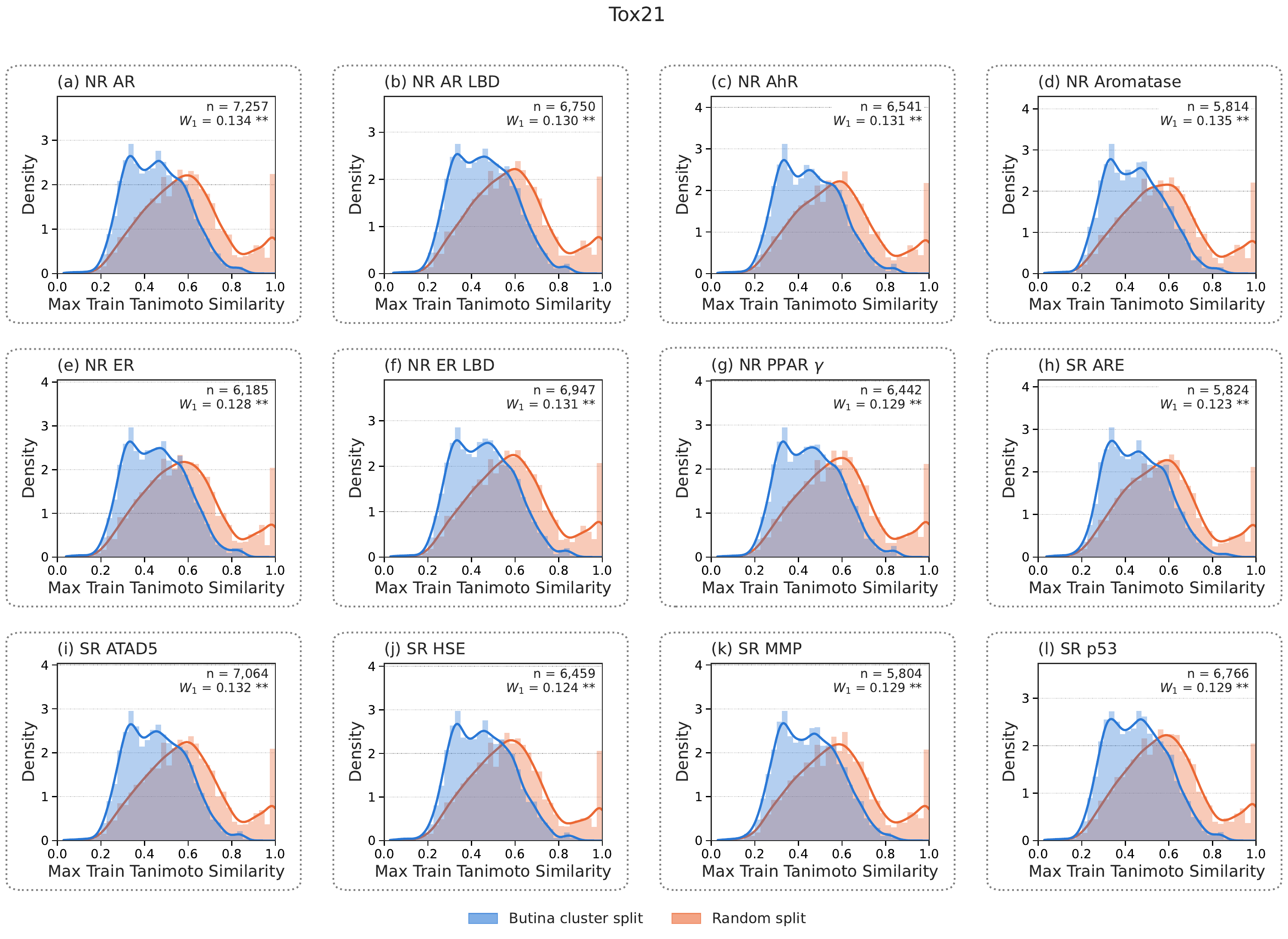}
    \caption{
        Maximum train-test Tanimoto similarities for Tox21 toxicity classification tasks.
        Blue is Butina StratifiedGroupKFold splitting, and orange is Random splitting.
        $\mathrm{W}_{1}$ is the Wasserstein distance between the Butina and Random distributions; asterisks indicate statistical significance ($p < 0.05$)
        using sign-flip permutation testing with 9999 permutations.
    }
    \label{fig:train_test_similarities_tox21}
\end{figure}

\clearpage
\newpage

\subsection{Label distributions}\label{si:label_distributions}

\begin{figure}[!htbp]
    \centering
    \includegraphics[scale=0.32]{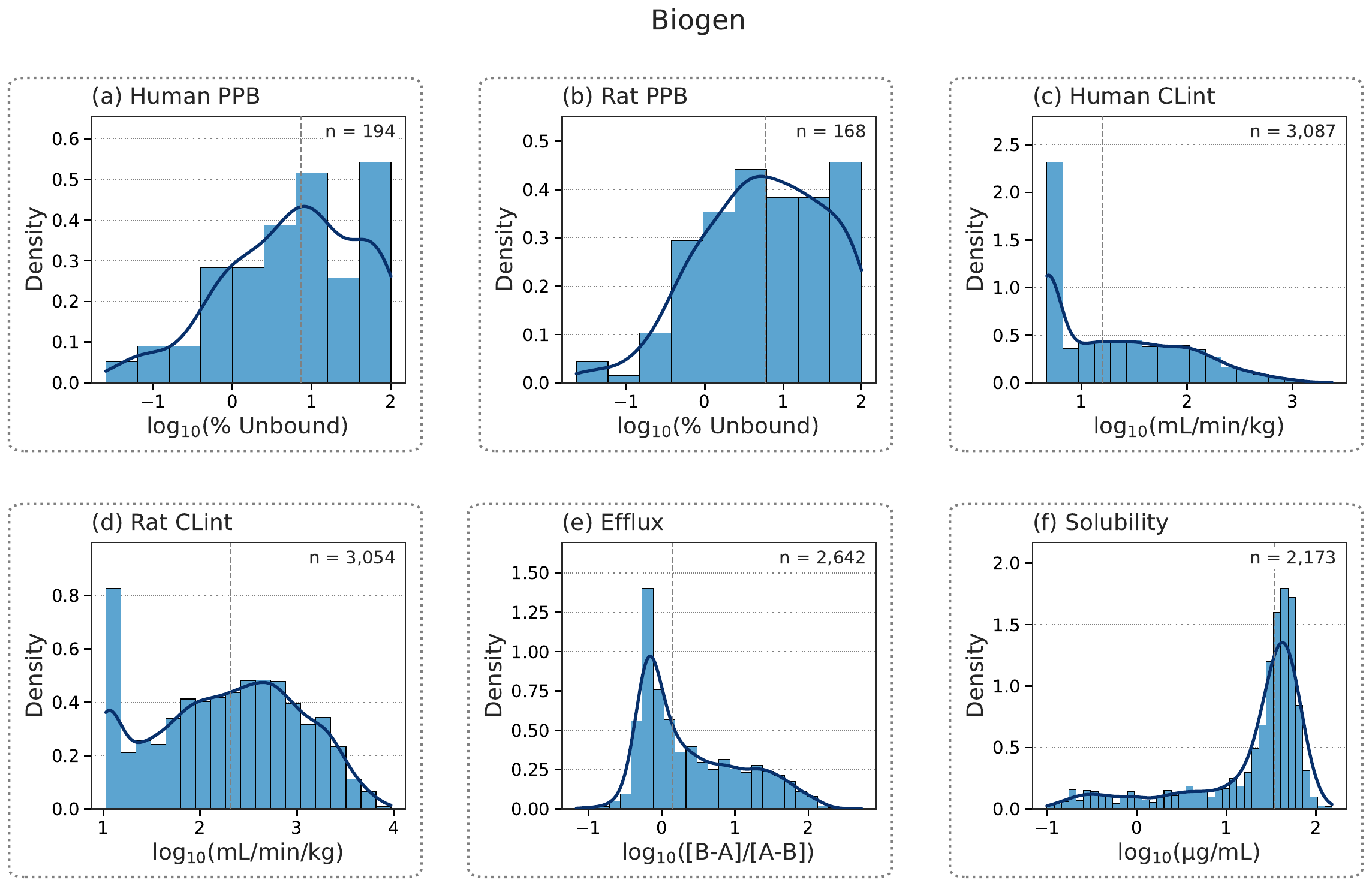}
    \caption{
        Distributions of labels in the Biogen tasks.
        (a) Human Plasma Protein Binding (PPB) ($\mathrm{log_{10}(\% \ Unbound)}$);
        (b) Rat PPB ($\mathrm{log_{10}(\% \ Unbound)}$);
        (c) Human Intrinsic Hepatic Clearance (CLint) ($\mathrm{log_{10}(mL/min/kg)}$);
        (d) Rat CLint ($\mathrm{log_{10}(mL/min/kg)}$);
        (e) Efflux ($\mathrm{log_{10}([B\text{-}A]/[A\text{-}B])}$);
        (f) Solubility ($\mathrm{log_{10}(\mu g/mL)}$).
        Histograms are shown with a kernel density estimate overlay;
        dashed vertical lines indicate the median label value,
        and $n$ is the number of molecules in each dataset.
    }
    \label{fig:label_dist_biogen}
\end{figure}

\begin{figure}[!htbp]
    \centering
    \includegraphics[scale=0.32]{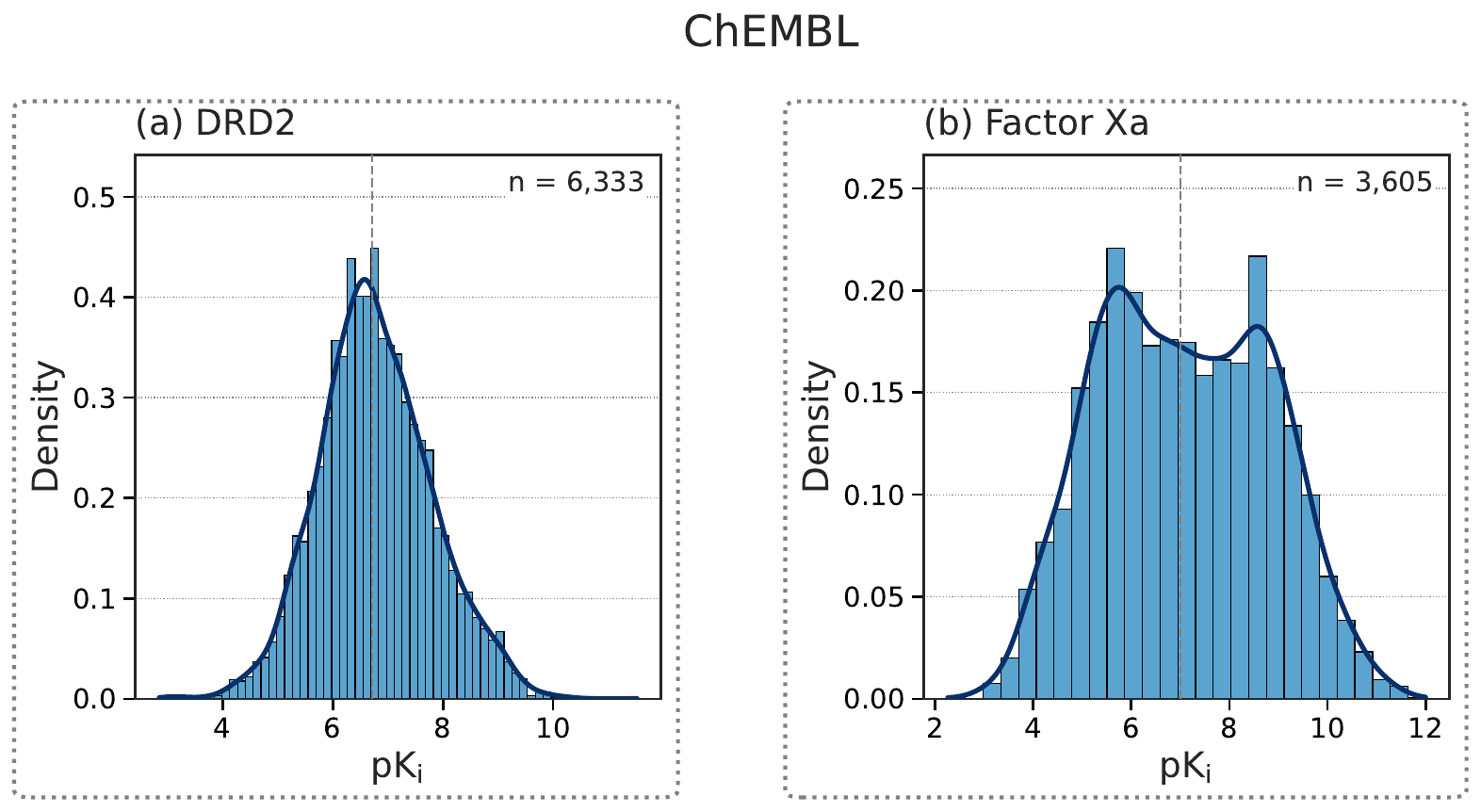}
    \caption{
        Distributions of labels in the ChEMBL binding affinity tasks.
        (a) DRD2 ($\mathrm{pK_{i}}$);
        (b) Factor Xa ($\mathrm{pK_{i}}$).
        Histograms are shown with a kernel density estimate overlay;
        dashed vertical lines indicate the median label value,
        and $n$ is the number of molecules in each dataset.
    }
    \label{fig:label_dist_chembl}
\end{figure}

\begin{figure}[!htbp]
    \centering
    \includegraphics[width=\textwidth, height=0.7\textheight, keepaspectratio]{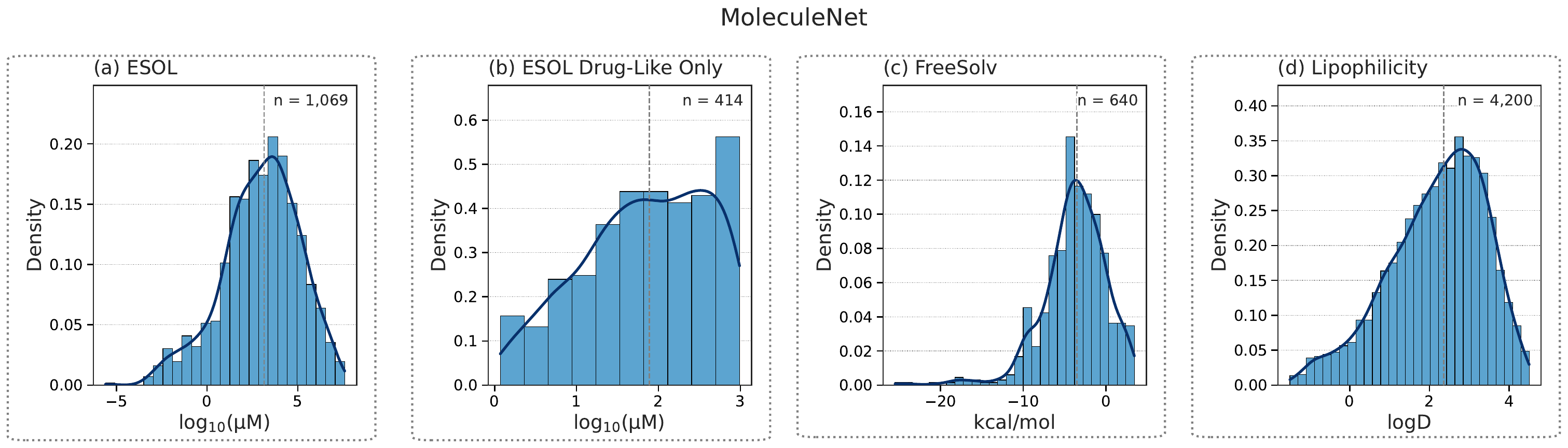}
    \caption{
        Distributions of labels in the MoleculeNet regression tasks.
        (a) ESOL aqueous solubility ($\mathrm{log_{10}(\mu M)}$);
        (b) ESOL restricted to drug-like molecules with $0 < \mathrm{log_{10}(S (\mu M))} < 3$ ($\mathrm{log_{10}(\mu M)}$);
        (c) FreeSolv free energy of solvation ($\mathrm{kcal/mol}$);
        (d) Lipophilicity ($\mathrm{logD}$).
        Histograms are shown with a kernel density estimate overlay;
        dashed vertical lines indicate the median label value,
        and $n$ is the number of molecules in each dataset.
    }
    \label{fig:label_dist_molnet}
\end{figure}

\clearpage
\newpage
\section{Hyperparameters}\label{sec:si_hps}

\subsection{Search spaces and values}

\begin{table}[!htbp]
    \centering
    \caption{GIN hyperparameters explored for pre-training.}
    \label{tab:pt_hp_opt}
    \begin{tabular}{llr}
        \toprule
        \textbf{Hyperparameter} & \textbf{Search Type} & \textbf{Search Space}\\
        \midrule
        Activation & categorical & LeakyReLU, GELU, Hardswish \\
        Batch Normalization & categorical & True, False \\
        Dropout & float & 0.0--0.5 \\
        Embedding dimensions & categorical & 128, 256, 512 \\
        GIN MLP Layers & int & 1--3 \\
        Graph Pooling & categorical & mean, sum, max \\
        Hidden dimensions & categorical & 128, 256, 512 \\
        Message Passing Layers & categorical & 1, 2, 3 \\
        Prediction Head Layers & categorical & 2, 3, 4 \\
        Weight Sharing & categorical & True, False \\
        \bottomrule
    \end{tabular}
\end{table}

\begin{table}[!htbp]
    \centering
    \caption{
        PT-GIN hyperparameters. 
        For pre-training, layer aggregation was set to $\mathtt{"last"}$ to use the output of the
        final GIN layer as the global embedding. 
        This was done to reduce the number of parameters in the prediction head and speed up pre-training.
        For benchmarking, layer aggregation was set to $\mathtt{"concat"}$ 
        to concatenate the outputs of all GIN layers as the global embedding.
        This was done to maximise the information available in the global embedding for benchmark tasks;
        the number of features was not a concern for benchmarking, 
        as LightGBM models are capable of handling high-dimensional
        input relatively well.
    }
    \label{tab:pt_hp}
    \begin{tabular}{lr}
        \toprule
        \textbf{Hyperparameter} & \textbf{Value} \\
        \midrule
        Activation & Hardswish \\
        Batch Normalization & False \\
        Batch Size & 128 \\
        Bias initialization & Zeros \\
        Dropout & 0.125 \\
        Embedding Dimensions & 512 \\
        Epochs & 50 \\
        GIN MLP Layers & 3 \\
        Graph pooling & Sum \\
        Hidden Dimensions & 512 \\
        Layer aggregation (readout) & last (pre-training); concat (benchmarking)\\
        Learning Rate Half-life (Epochs) & 5 \\
        Learning Rate Scaling Factor & 0.5 \\
        Message Passing Layers & 3 \\
        Optimizer & Adam \\
        Prediction Head Layers & 2 \\
        Share Weights & True \\
        Train eps & True \\
        Warm-up Epochs & 2 \\
        Weight initialization  & Xavier Normal \\
        \bottomrule
    \end{tabular}
\end{table}

\begin{tableorg}[!htbp]% arXiv: plain float so the tabular can be scaled
    \centering
    \small
    \caption{PT-GIN embedding dimensions before and after convolutions.}
    \label{tab:pt_emb_dim}
    \resizebox{\textwidth}{!}{% arXiv: narrower sn-jnl text block
    \begin{tabular}{rrrrrr}
        \toprule
        \textbf{Substructure} & \textbf{Token Embedding} & \textbf{Node} & \textbf{Hidden} & \textbf{\# of} & \textbf{Final} \\
        \textbf{Max. Radius} & \textbf{Dimension} & \textbf{Dimension} & \textbf{Dimension} & \textbf{Convolutions} & \textbf{Dimension} \\
        \midrule
        $0$ & $512$ & $512$ & $512$ & $3$ & $2048$ \\
        $1$ & $512$ & $1024$ & $512$ & $3$ & $2560$ \\
        $2$ & $512$ & $1536$ & $512$ & $3$ & $3072$ \\
        \bottomrule
    \end{tabular}
    }%
\end{tableorg}

\begin{table}[!htbp]
    \centering
    \caption{LightGBM hyperparameters explored in benchmarking.}
    \label{tab:lgbm_hp_opt}
    \begin{tabular}{llr}
        \toprule
        \textbf{Hyperparameter} & \textbf{Search Type} & \textbf{Search Space}\\
        \midrule
        num\_leaves & int & 10--100 \\
        learning\_rate & float with log & 0.001--0.3 \\
        n\_estimators & int & 10--120 \\
        subsample\_for\_bin & int & 20000--200000\\
        min\_data\_in\_leaf & int with log & 1--1000 \\
        bagging\_fraction & float & 0.25--1.0 \\
        feature\_fraction & float & 0.25--1.0 \\
        reg\_alpha & float with log & 0.01--10.0 \\
        reg\_lambda & float with log & 0.01--10.0 \\
        \bottomrule
    \end{tabular}
\end{table}

\begin{table}[!htbp]
    \centering
    \caption{Hyperparameters explored for Scratch GIN models in benchmarking, regardless of dataset size. The initial learning rate was calculated as $lr = c\cdot p_{model}^{-0.5}$, where $c$ is the learning rate scaling factor and $p_{model}$ is the number of model parameters.}
    \label{tab:scratch_hp_opt}
    \begin{tabular}{llr}
        \toprule
        \textbf{Hyperparameter} & \textbf{Search Type} & \textbf{Search Space}\\
        \midrule
        Activation & categorical & Hardswish \\
        Dropout & float & 0.0--0.5 \\
        Epochs & fixed & 50 \\
        Graph pooling & categorical & Max, Sum, Mean \\
        Layer aggregation & categorical & concatenate \\
        Learning Rate Half-life (Epochs) & int & 5--50 \\
        Learning Rate Scaling Factor & float & 0.1--1.0 \\
        Optimizer & fixed & Adam \\
        Message Passing Layers & int & 1--5 \\
        Prediction Head Layers & int & 1--5 \\
        Radius & categorical & 0, 1 \\
        Train eps & categorical & True, False \\
        Weight Sharing & categorical & True, False \\
        Vocabulary Size & categorical & 16, 32, 64, 128, 256 \\
        \bottomrule
    \end{tabular}
\end{table}

\begin{table}[!htbp]
    \centering
    \caption{Hyperparameters explored for Scratch GIN models on datasets with fewer than 500 samples.}
    \label{tab:scratch_hp_opt_500}
    \begin{tabular}{llr}
        \toprule
        \textbf{Hyperparameter} & \textbf{Search Type} & \textbf{Search Space}\\
        \midrule
        Batch Size & int & 16 \\
        Embedding dimensions & categorical & 2, 4, 8 \\
        GIN MLP Layers & int & 1--3 \\
        Hidden dimensions & categorical & 4, 8 \\
        \bottomrule
    \end{tabular}
\end{table}

\begin{table}[!htbp]
    \centering
    \caption{Hyperparameters explored for Scratch GIN models on datasets with $500 \leq n < 1000$ samples.}
    \label{tab:scratch_hp_opt_1000}
    \begin{tabular}{llr}
        \toprule
        \textbf{Hyperparameter} & \textbf{Search Type} & \textbf{Search Space}\\
        \midrule
        Batch Size & int & 32 \\
        Embedding dimensions & categorical & 4, 8, 16 \\
        GIN MLP Layers & int & 1--5 \\
        Hidden dimensions & categorical & 4, 8, 16 \\
        \bottomrule
    \end{tabular}
\end{table}

\begin{table}[!htbp]
    \centering
    \caption{Hyperparameters explored for Scratch GIN models on datasets with $1000 \leq n < 5000$ samples.}
    \label{tab:scratch_hp_opt_5000}
    \begin{tabular}{llr}
        \toprule
        \textbf{Hyperparameter} & \textbf{Search Type} & \textbf{Search Space}\\
        \midrule
        Batch Size & int & 64 \\
        Embedding dimensions & categorical & 8, 16, 32 \\
        GIN MLP Layers & int & 1--5 \\
        Hidden dimensions & categorical & 8, 16, 32 \\
        \bottomrule
    \end{tabular}
\end{table}

\begin{table}[!htbp]
    \centering
    \caption{Hyperparameters explored for Scratch GIN models on datasets with $n \geq 5000$ samples.}
    \label{tab:scratch_hp_opt_5000_plus}
    \begin{tabular}{llr}
        \toprule
        \textbf{Hyperparameter} & \textbf{Search Type} & \textbf{Search Space}\\
        \midrule
        Batch Size & int & 128 \\
        Embedding dimensions & categorical & 16, 32, 64 \\
        GIN MLP Layers & int & 1--5 \\
        Hidden dimensions & categorical & 16, 32, 64 \\
        \bottomrule
    \end{tabular}
\end{table}

\subsection{Hyperparameter tuning results for Morgan algorithm parameters}

\subsubsection{PT-GIN}

\begin{figure}[!htbp]
    \centering
    \includegraphics[width=\textwidth, height=0.7\textheight, keepaspectratio]{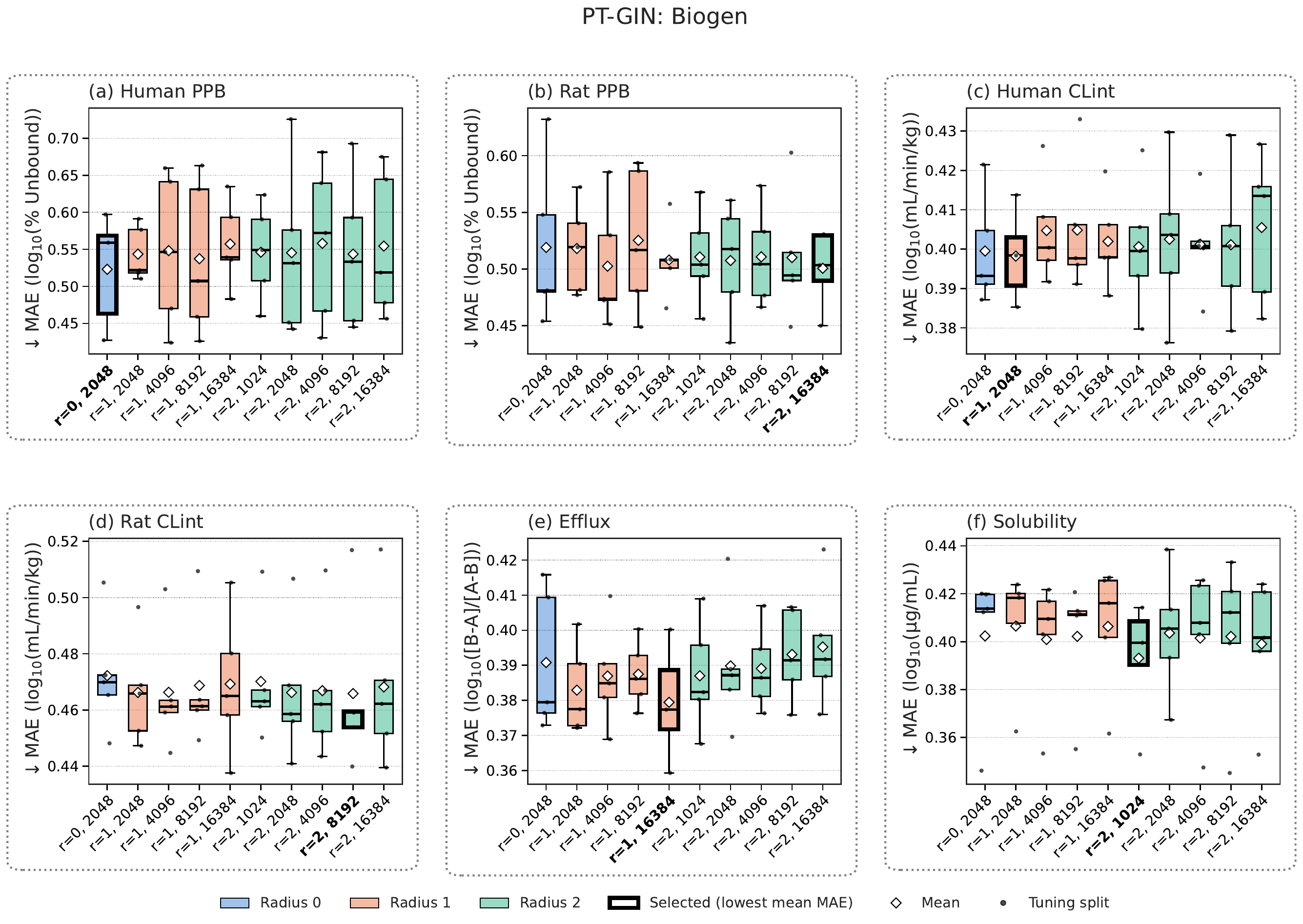}
    \caption{
        PT-GIN hyperparameter tuning results on Biogen tasks.
    }
    \label{fig:tuning_PT_GIN_biogen}
\end{figure}

\begin{figure}[!htbp]
    \centering
    \includegraphics[scale=0.365]{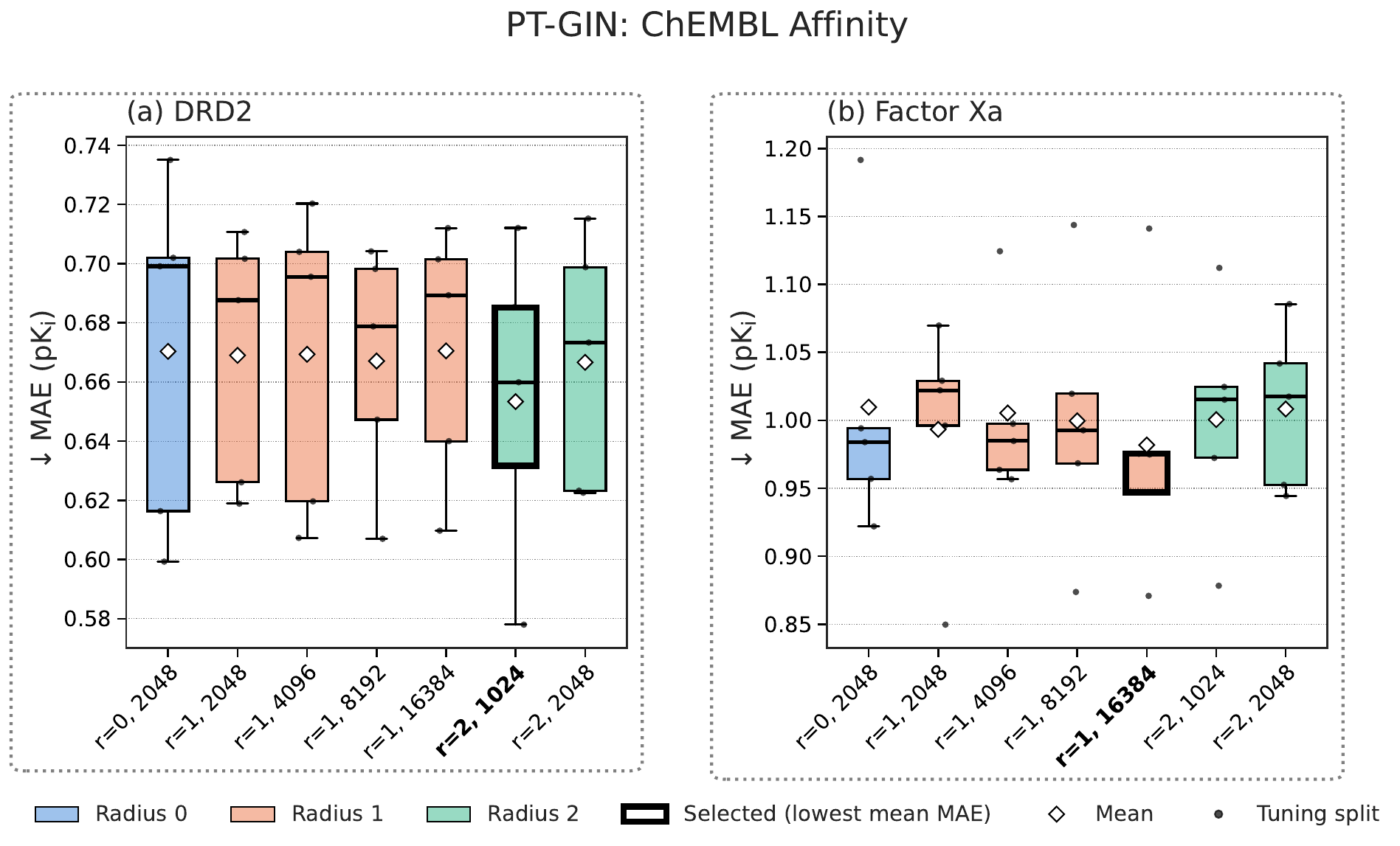}
    \caption{
        PT-GIN hyperparameter tuning results on ChEMBL binding affinity tasks.
    }
    \label{fig:tuning_PT_GIN_chembl}
\end{figure}

\begin{figure}[!htbp]
    \centering
    \includegraphics[width=\textwidth, height=0.7\textheight, keepaspectratio]{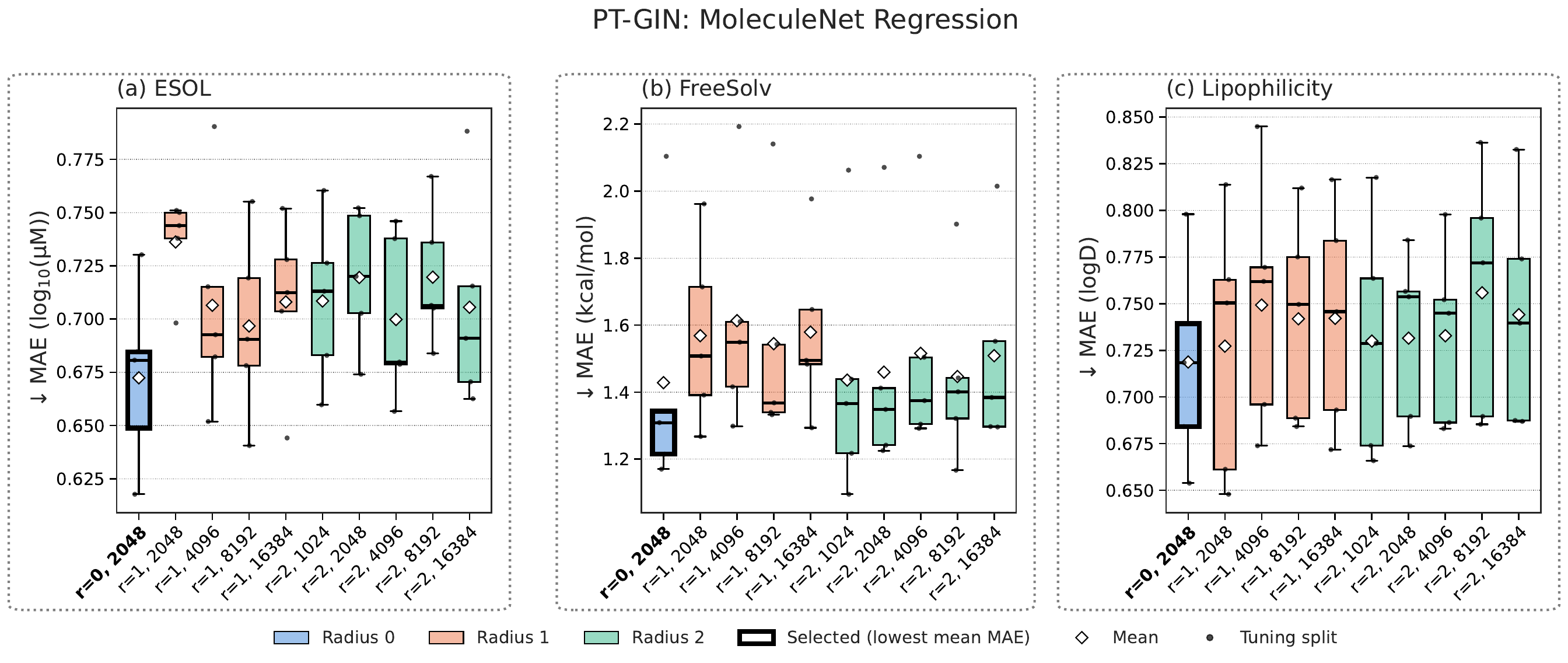}
    \caption{
        PT-GIN hyperparameter tuning results on MoleculeNet regression tasks.
    }
    \label{fig:tuning_PT_GIN_molnet}
\end{figure}

\begin{figure}[!htbp]
    \centering
    \includegraphics[width=\textwidth, height=0.7\textheight, keepaspectratio]{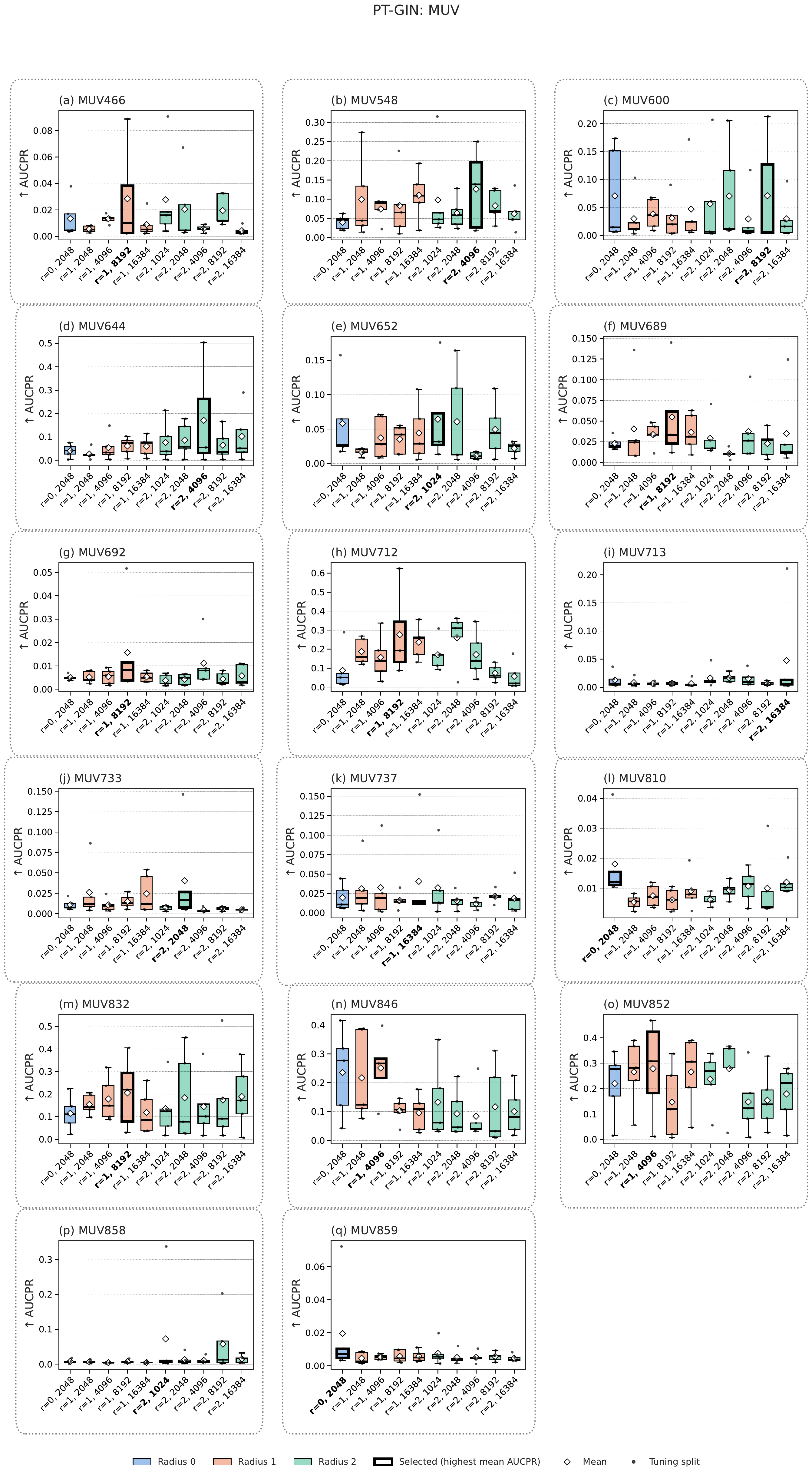}
    \caption{
        PT-GIN hyperparameter tuning results on MUV virtual screening tasks.
    }
    \label{fig:tuning_PT_GIN_muv}
\end{figure}

\begin{figure}[!htbp]
    \centering
    \includegraphics[width=\textwidth, height=0.7\textheight, keepaspectratio]{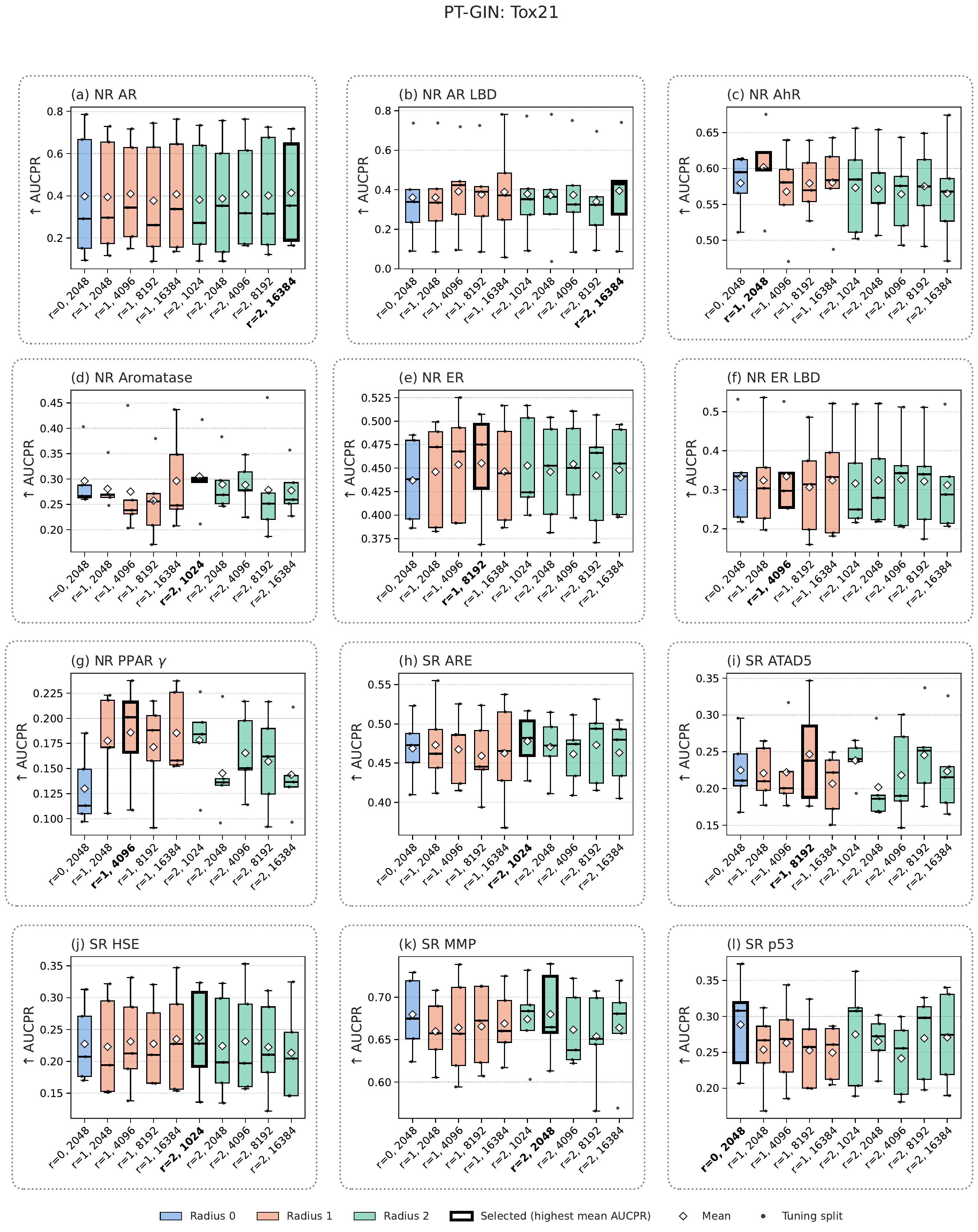}
    \caption{
        PT-GIN hyperparameter tuning results on Tox21 tasks.
    }
    \label{fig:tuning_PT_GIN_tox21}
\end{figure}

\clearpage
\newpage

\subsubsection{Scratch GIN}

\begin{figure}[!htbp]
    \centering
    \includegraphics[width=\textwidth, height=0.7\textheight, keepaspectratio]{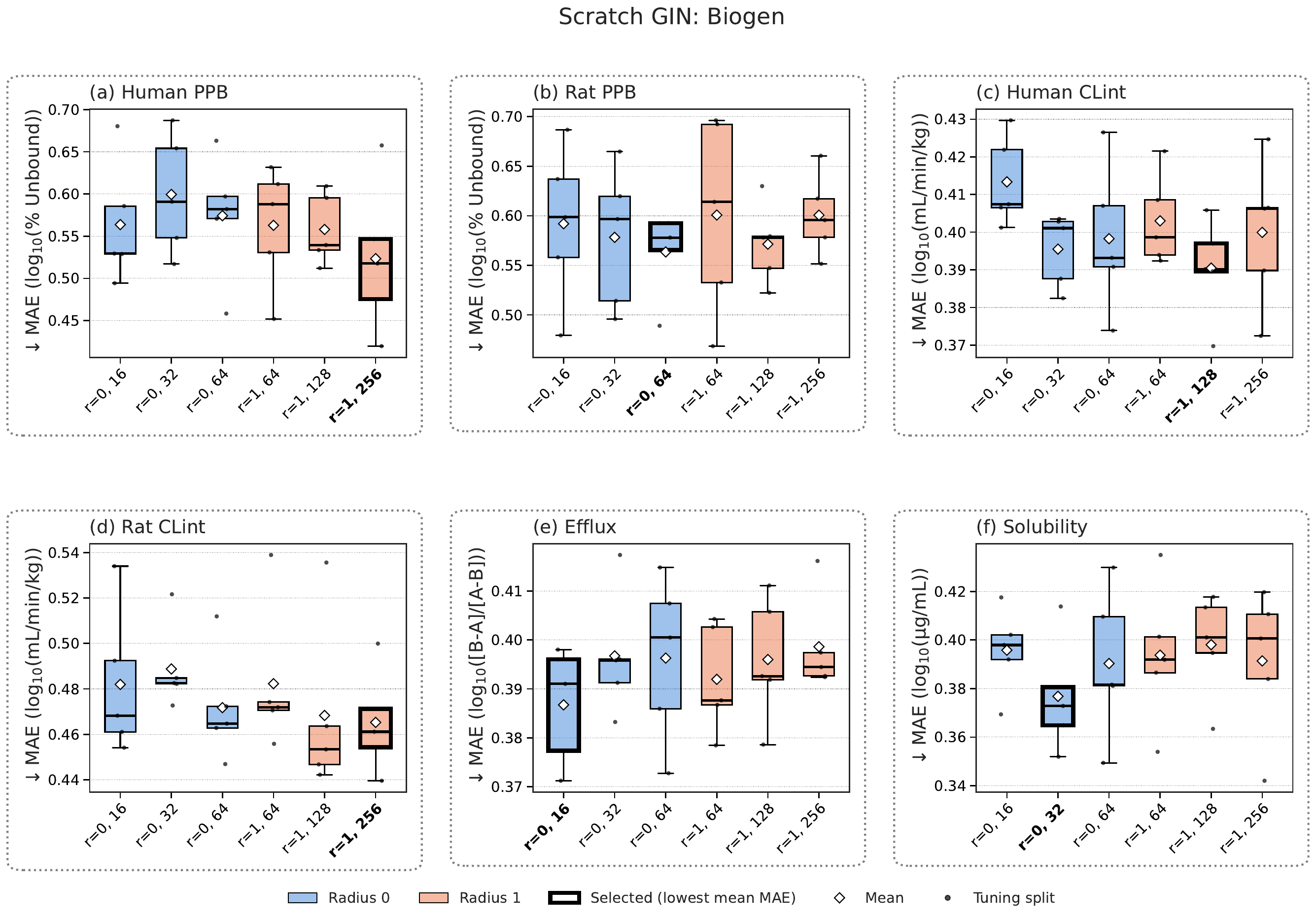}
    \caption{
        Scratch GIN hyperparameter tuning results on Biogen tasks.
    }
    \label{fig:tuning_Scratch_GIN_biogen}
\end{figure}

\clearpage
\newpage

\subsubsection{ECFP Sort \& Slice}

\begin{figure}[!htbp]
    \centering
    \includegraphics[width=\textwidth, height=0.7\textheight, keepaspectratio]{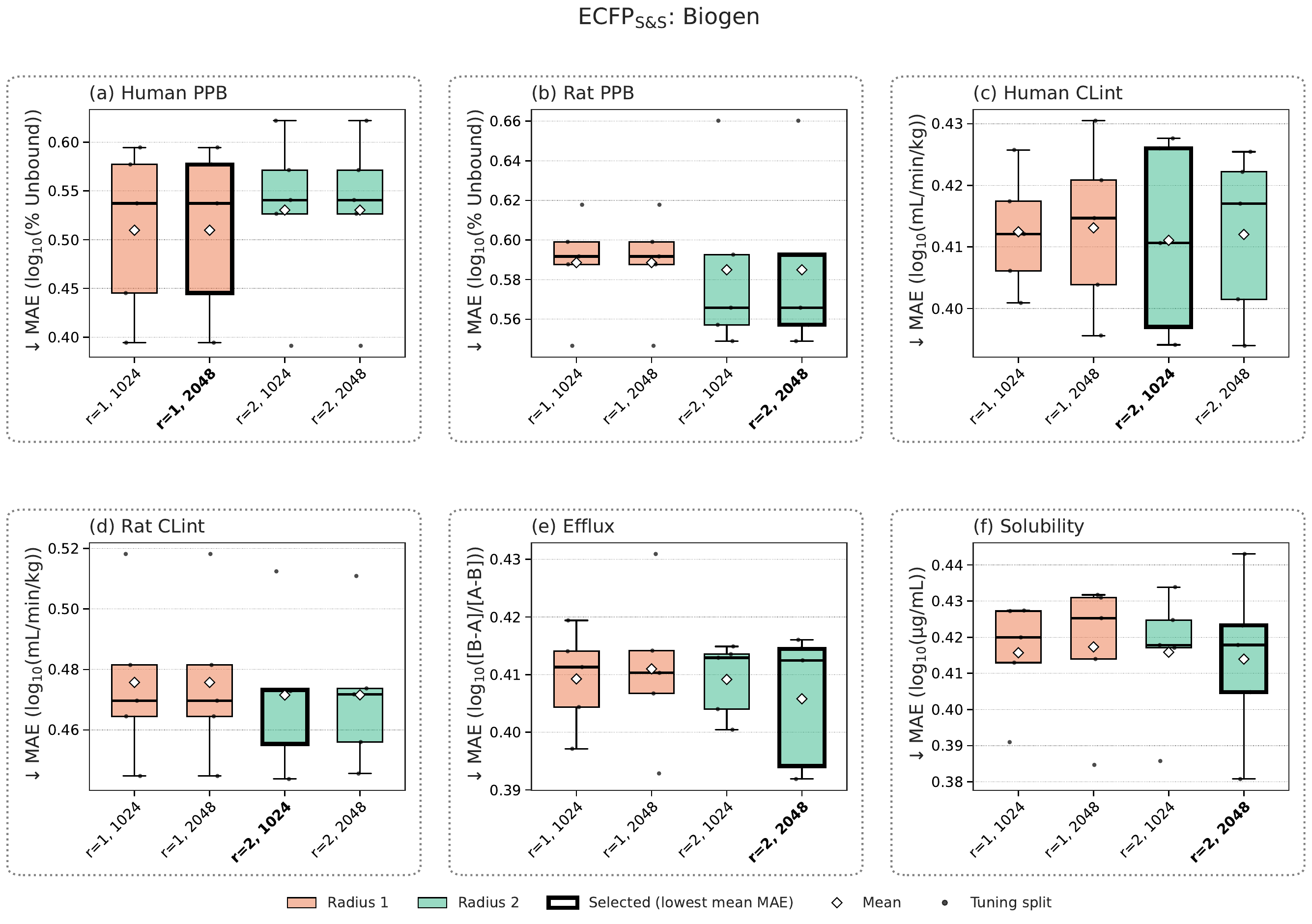}
    \caption{
        ECFP$_{\mathrm{S\&S}}$ hyperparameter tuning results on Biogen tasks.
    }
    \label{fig:tuning_ECFP_SNS_biogen}
\end{figure}

\begin{figure}[!htbp]
    \centering
    \includegraphics[scale=0.365]{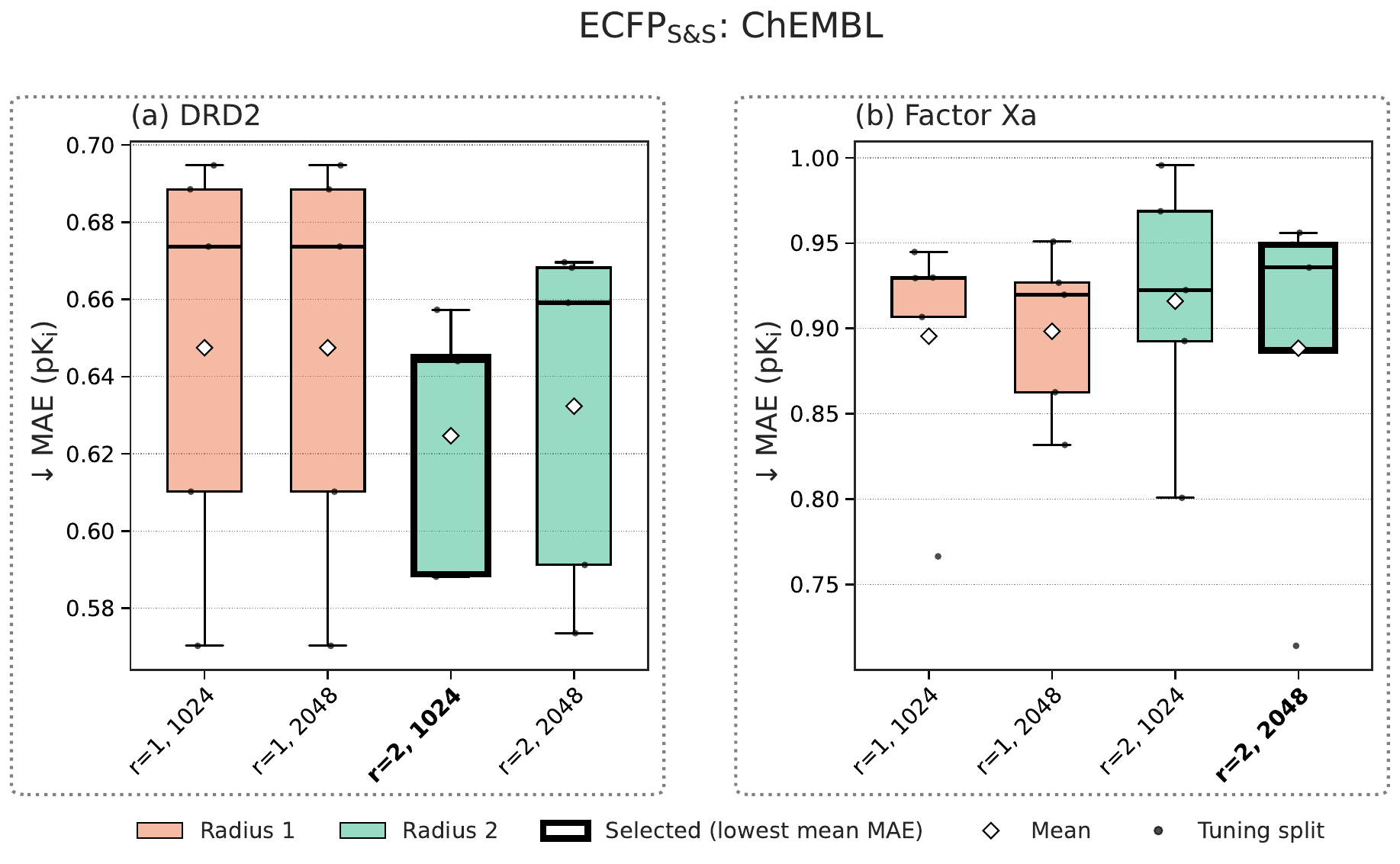}
    \caption{
        ECFP$_{\mathrm{S\&S}}$ hyperparameter tuning results on ChEMBL binding affinity tasks.
    }
    \label{fig:tuning_ECFP_SNS_chembl}
\end{figure}

\begin{figure}[!htbp]
    \centering
    \includegraphics[width=\textwidth, height=0.7\textheight, keepaspectratio]{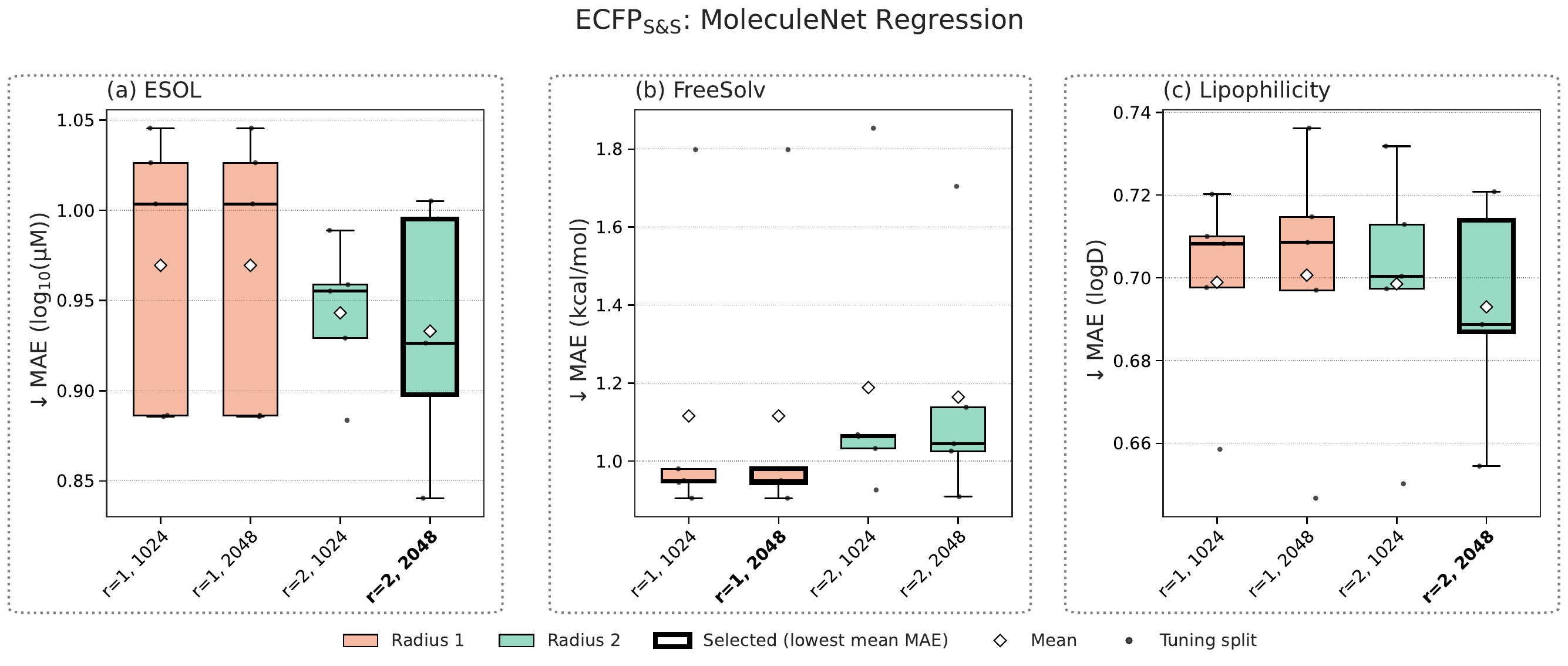}
    \caption{
        ECFP$_{\mathrm{S\&S}}$ hyperparameter tuning results on MoleculeNet regression tasks.
    }
    \label{fig:tuning_ECFP_SNS_molnet}
\end{figure}

\begin{figure}[!htbp]
    \centering
    \includegraphics[width=\textwidth, height=0.7\textheight, keepaspectratio]{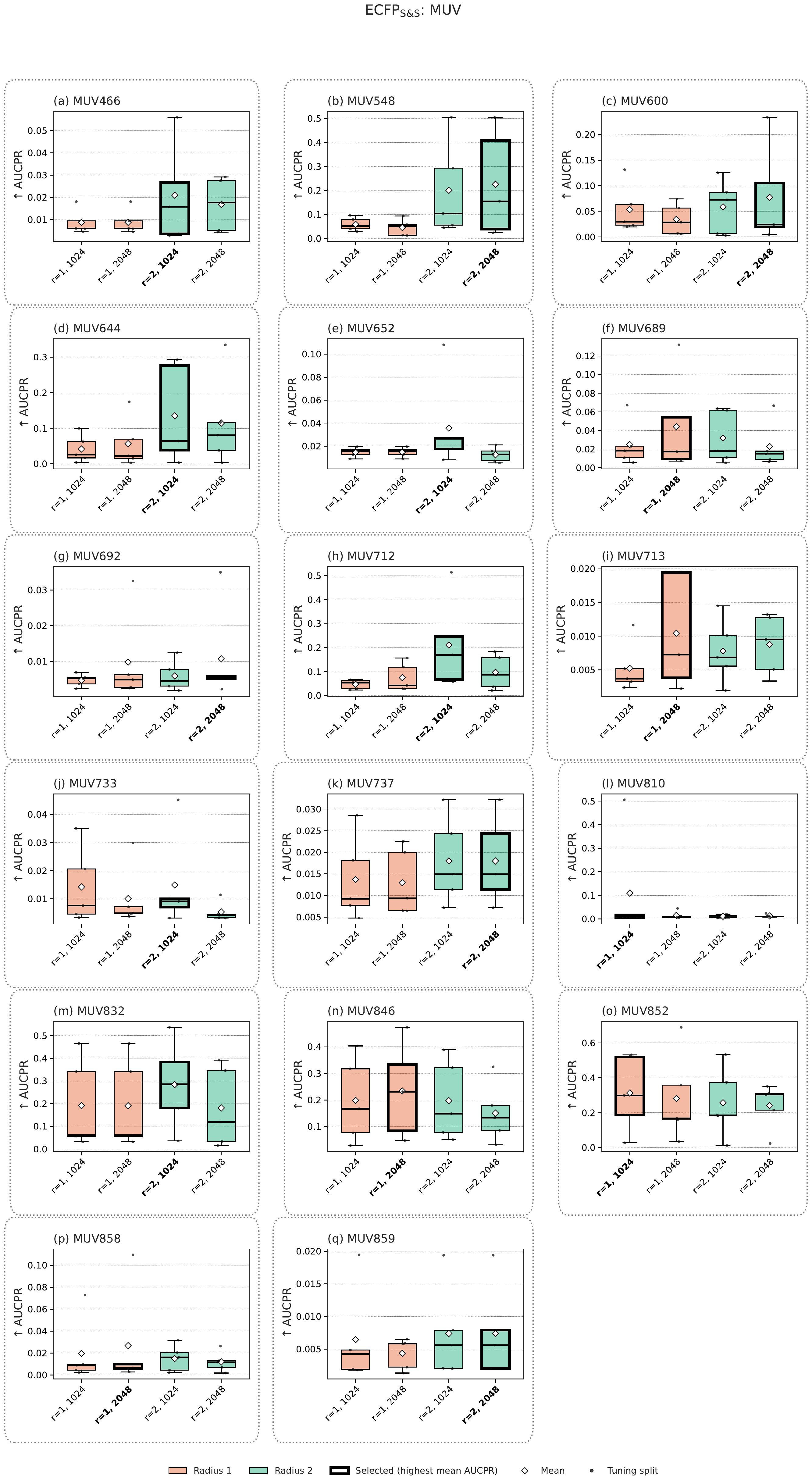}
    \caption{
        ECFP$_{\mathrm{S\&S}}$ hyperparameter tuning results on MUV virtual screening tasks.
    }
    \label{fig:tuning_ECFP_SNS_muv}
\end{figure}

\begin{figure}[!htbp]
    \centering
    \includegraphics[width=\textwidth, height=0.7\textheight, keepaspectratio]{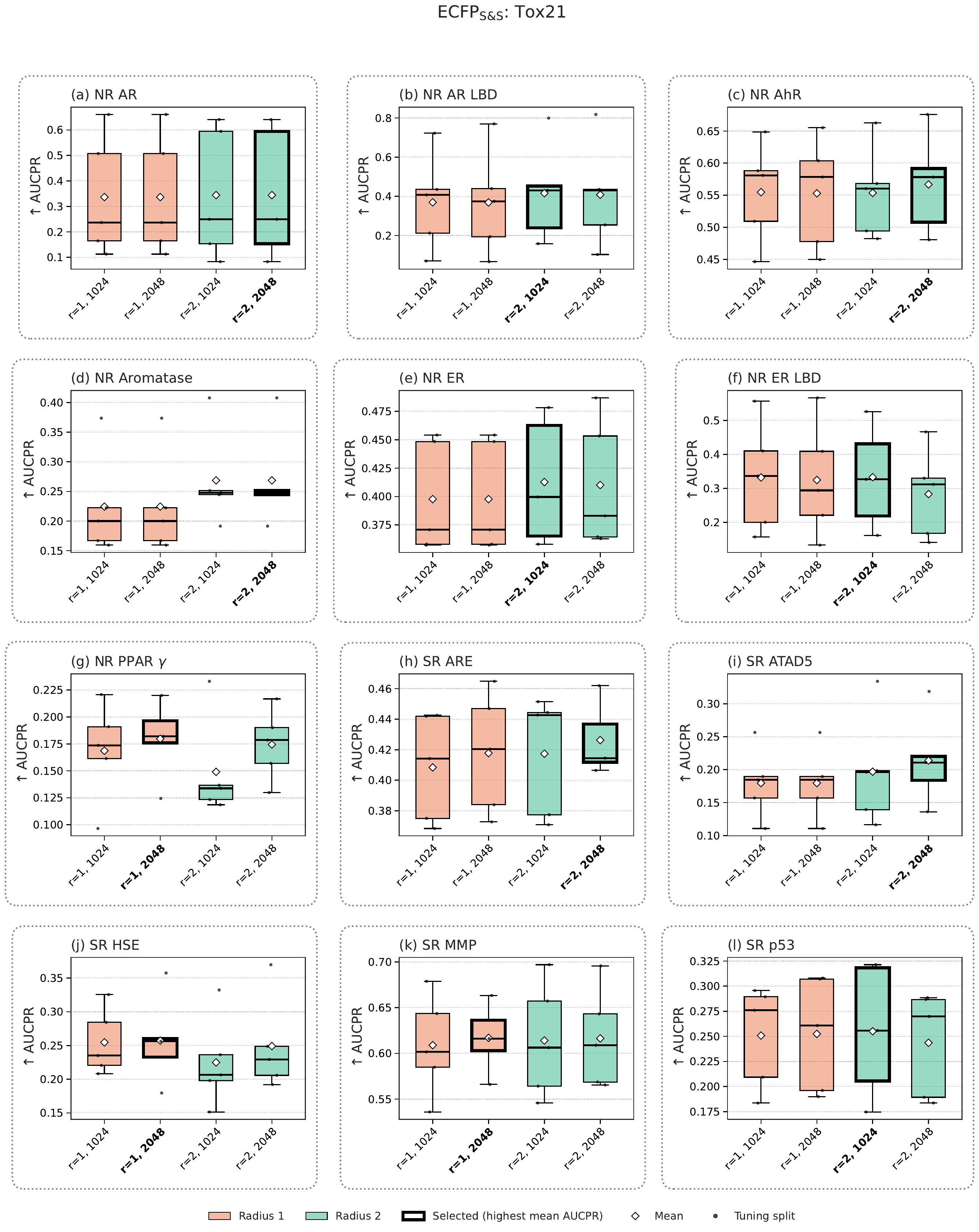}
    \caption{
        ECFP$_{\mathrm{S\&S}}$ hyperparameter tuning results on Tox21 tasks.
    }
    \label{fig:tuning_ECFP_SNS_tox21}
\end{figure}
\clearpage
\newpage
\subsubsection{ECFP hashed}

\begin{figure}[!htbp]
    \centering
    \includegraphics[width=\textwidth, height=0.7\textheight, keepaspectratio]{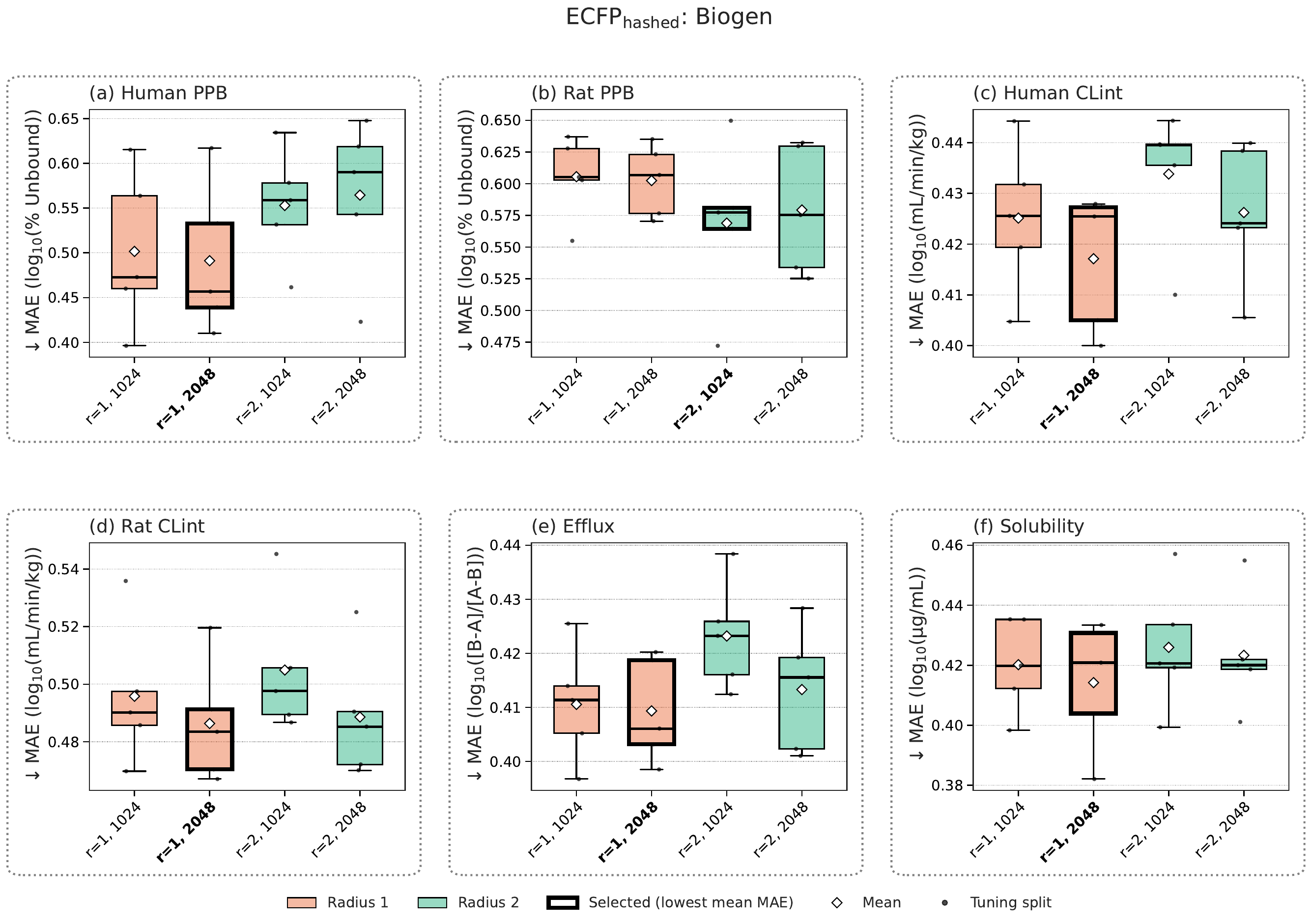}
    \caption{
        ECFP$_{\mathrm{hashed}}$ hyperparameter tuning results on Biogen tasks.
    }
    \label{fig:tuning_ECFP_hashed_biogen}
\end{figure}

\begin{figure}[!htbp]
    \centering
    \includegraphics[scale=0.365]{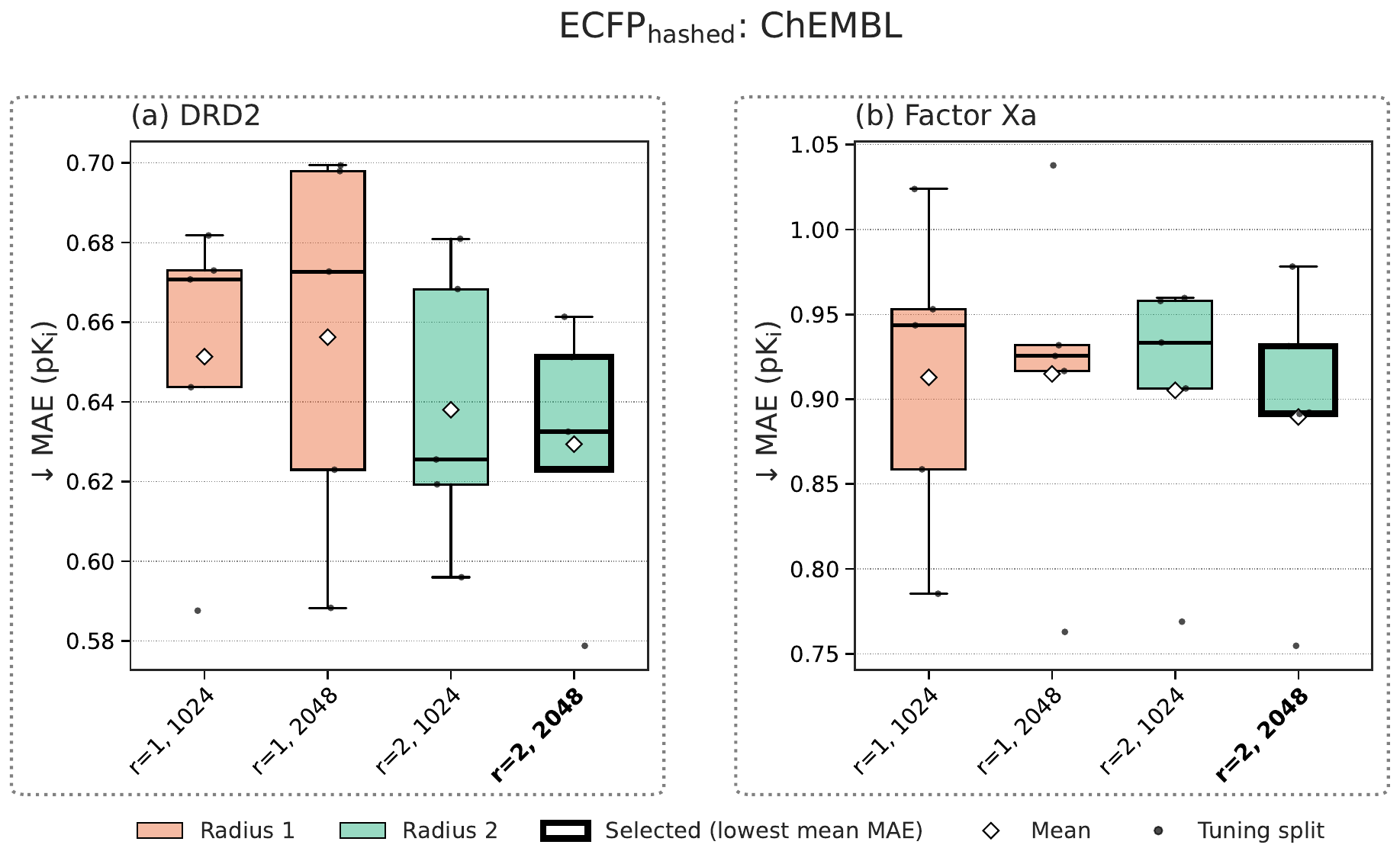}
    \caption{
        ECFP$_{\mathrm{hashed}}$ hyperparameter tuning results on ChEMBL binding affinity tasks.
    }
    \label{fig:tuning_ECFP_hashed_chembl}
\end{figure}

\begin{figure}[!htbp]
    \centering
    \includegraphics[width=\textwidth, height=0.7\textheight, keepaspectratio]{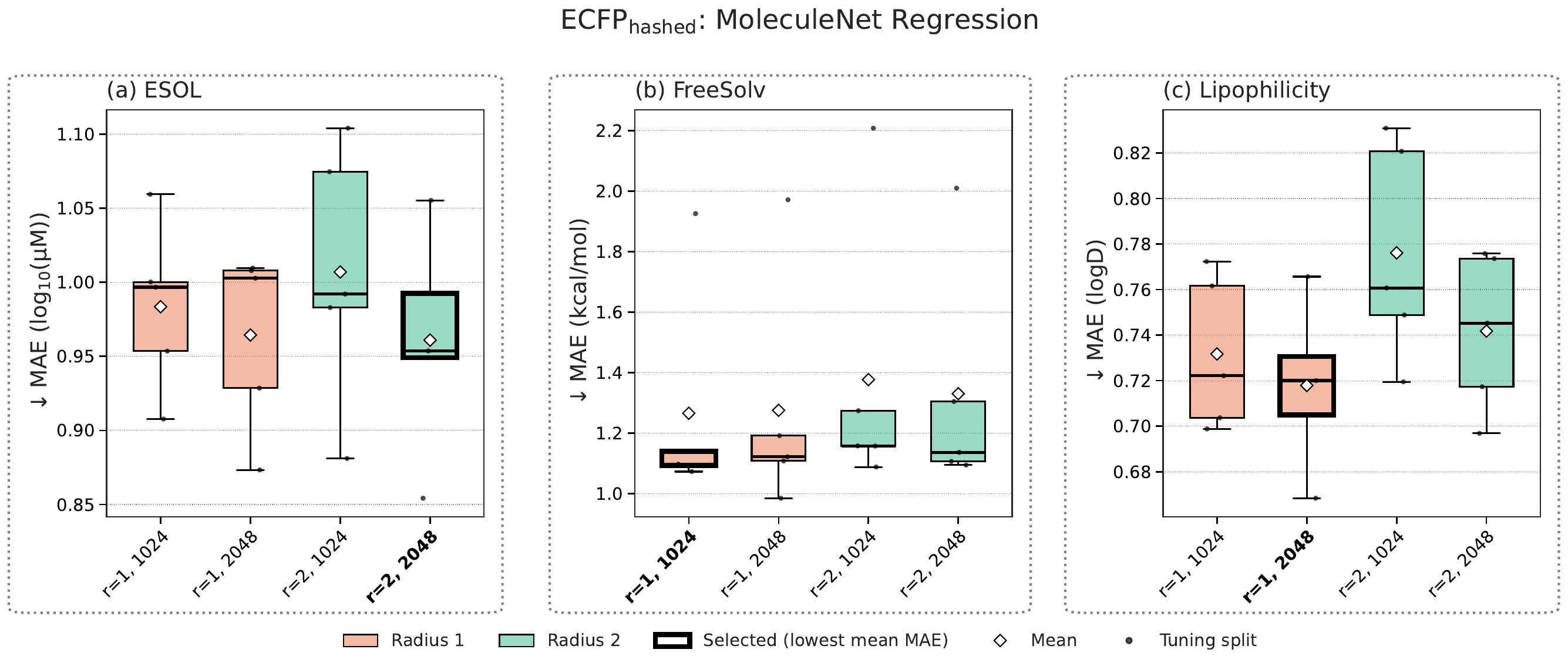}
    \caption{
        ECFP$_{\mathrm{hashed}}$ hyperparameter tuning results on MoleculeNet regression tasks.
    }
    \label{fig:tuning_ECFP_hashed_molnet}
\end{figure}

\begin{figure}[!htbp]
    \centering
    \includegraphics[width=\textwidth, height=0.7\textheight, keepaspectratio]{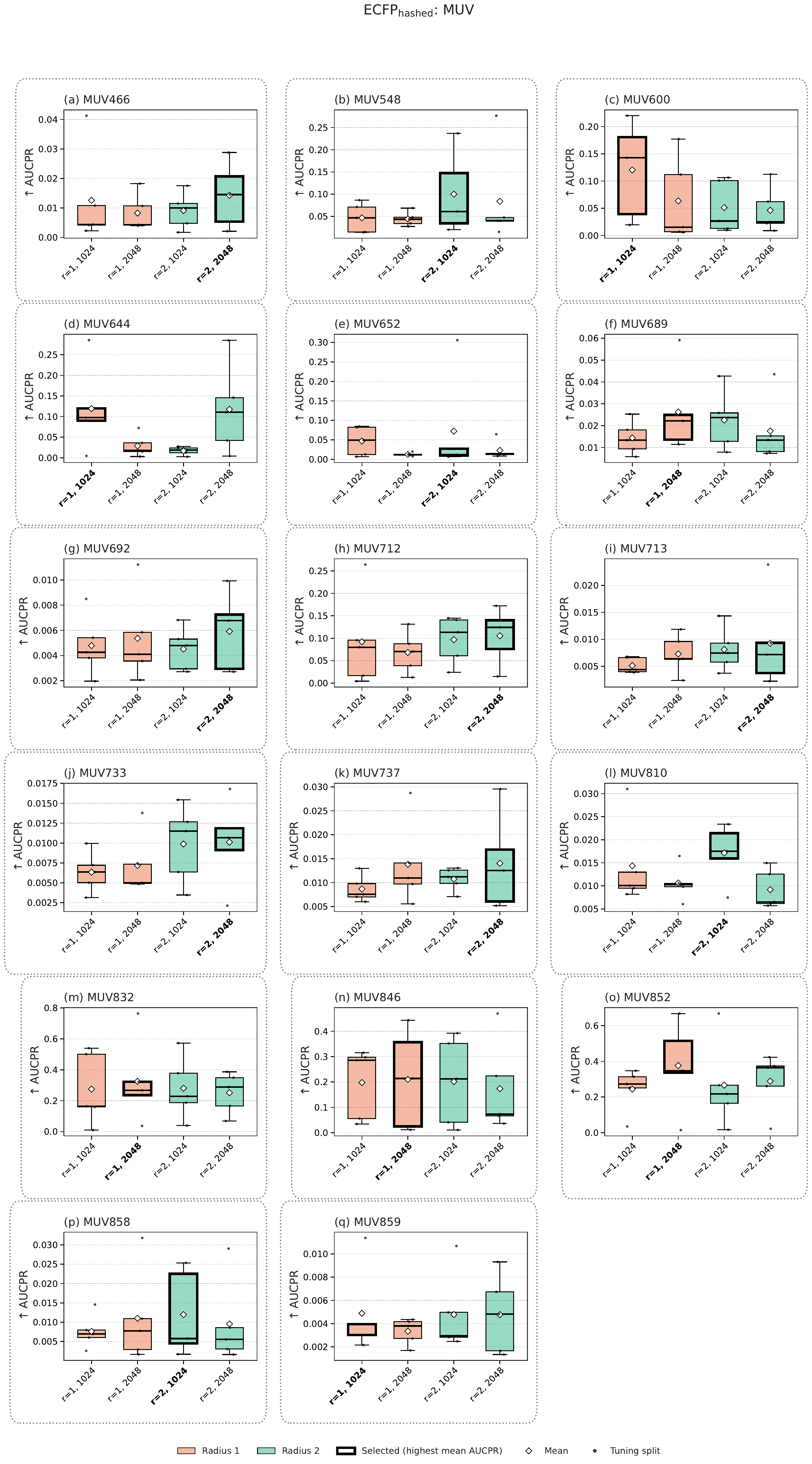}
    \caption{
        ECFP$_{\mathrm{hashed}}$ hyperparameter tuning results on MUV virtual screening tasks.
    }
    \label{fig:tuning_ECFP_hashed_muv}
\end{figure}

\begin{figure}[!htbp]
    \centering
    \includegraphics[width=\textwidth, height=0.7\textheight, keepaspectratio]{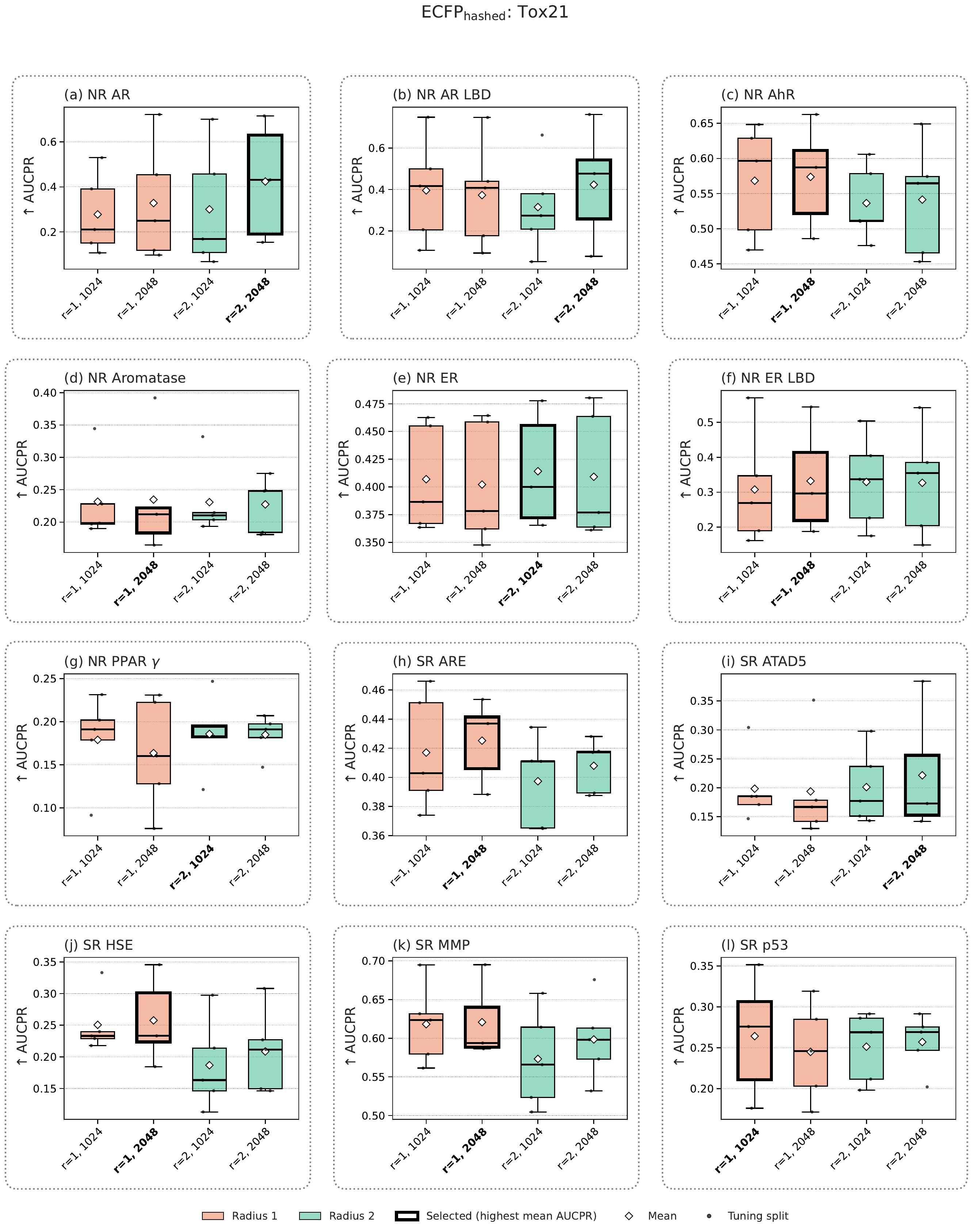}
    \caption{
        ECFP$_{\mathrm{hashed}}$ hyperparameter tuning results on Tox21 tasks.
    }
    \label{fig:tuning_ECFP_hashed_tox21}
\end{figure}
\clearpage
\newpage
\subsubsection{FCFP hashed}

\begin{figure}[!htbp]
    \centering
    \includegraphics[width=\textwidth, height=0.7\textheight, keepaspectratio]{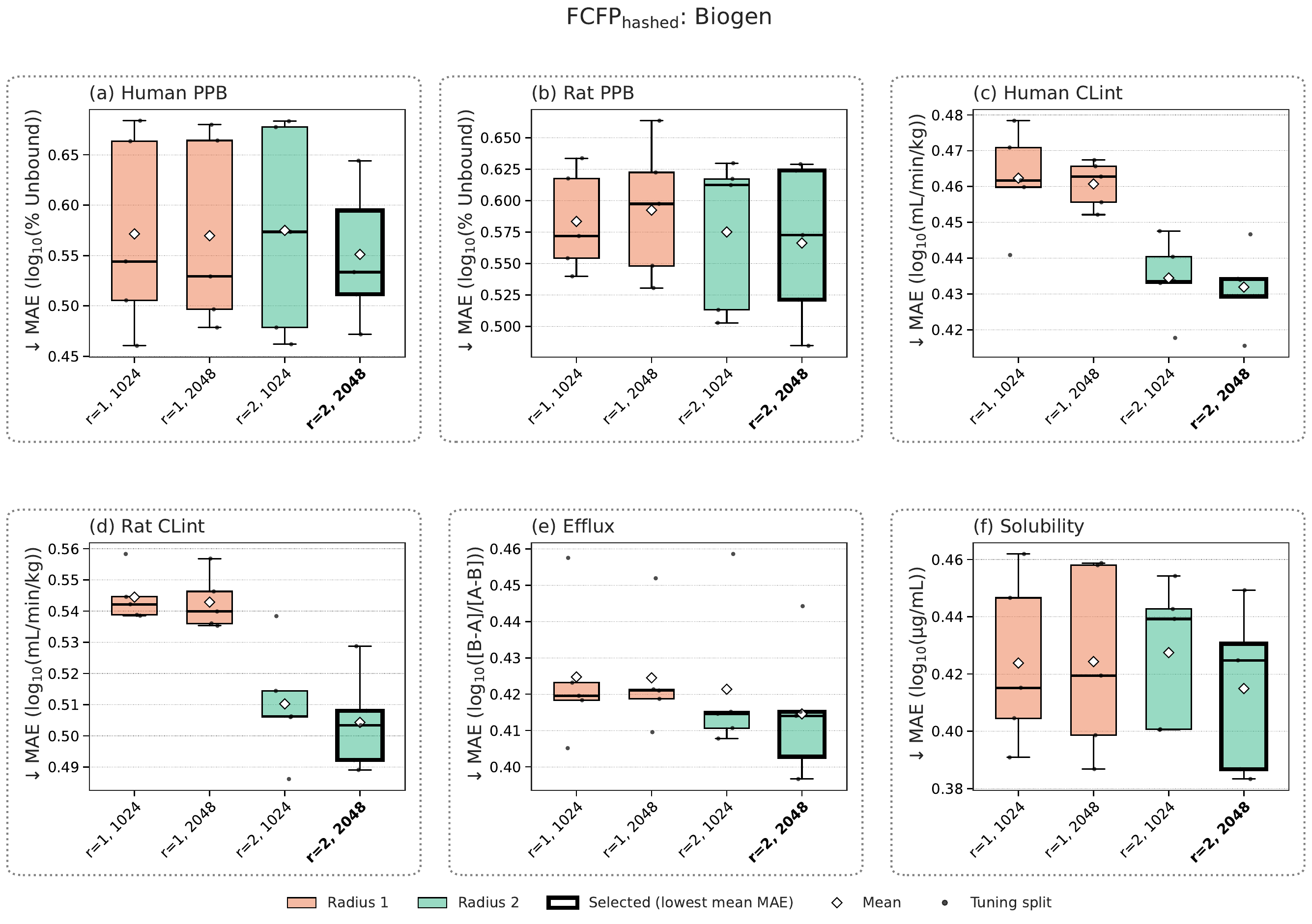}
    \caption{
        FCFP$_{\mathrm{hashed}}$ hyperparameter tuning results on Biogen tasks.
    }
    \label{fig:tuning_FCFP_hashed_biogen}
\end{figure}

\begin{figure}[!htbp]
    \centering
    \includegraphics[scale=0.365]{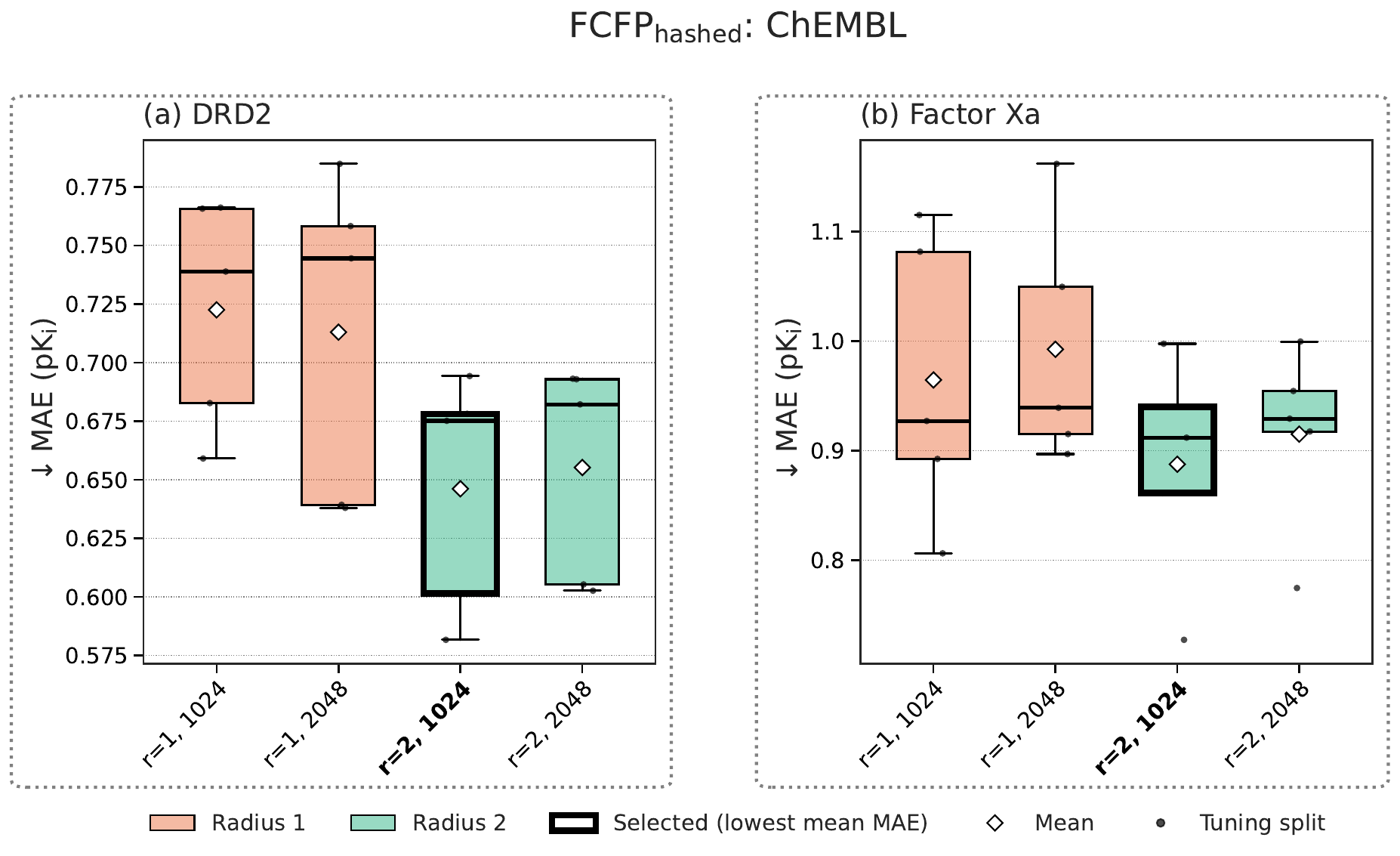}
    \caption{
        FCFP$_{\mathrm{hashed}}$ hyperparameter tuning results on ChEMBL binding affinity tasks.
    }
    \label{fig:tuning_FCFP_hashed_chembl}
\end{figure}

\begin{figure}[!htbp]
    \centering
    \includegraphics[width=\textwidth, height=0.7\textheight, keepaspectratio]{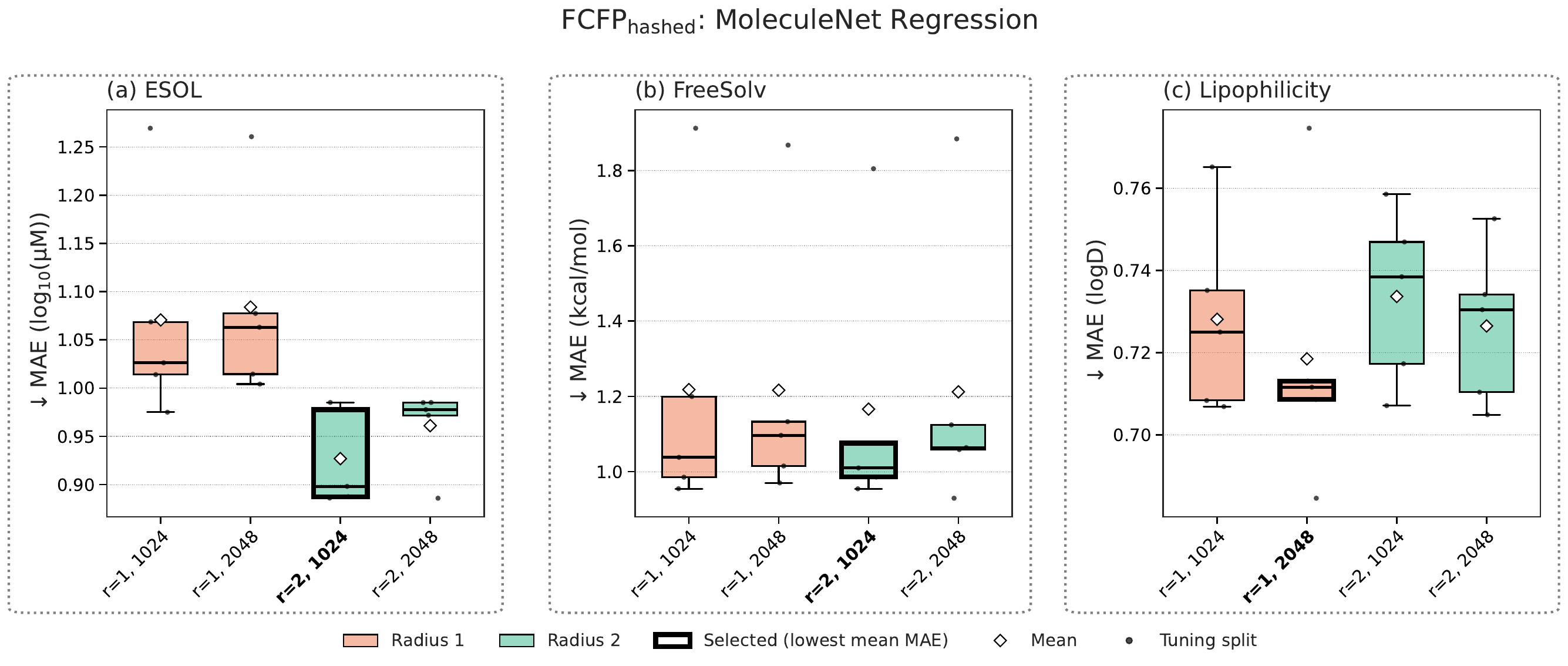}
    \caption{
        FCFP$_{\mathrm{hashed}}$ hyperparameter tuning results on MoleculeNet regression tasks.
    }
    \label{fig:tuning_FCFP_hashed_molnet}
\end{figure}

\begin{figure}[!htbp]
    \centering
    \includegraphics[width=\textwidth, height=0.7\textheight, keepaspectratio]{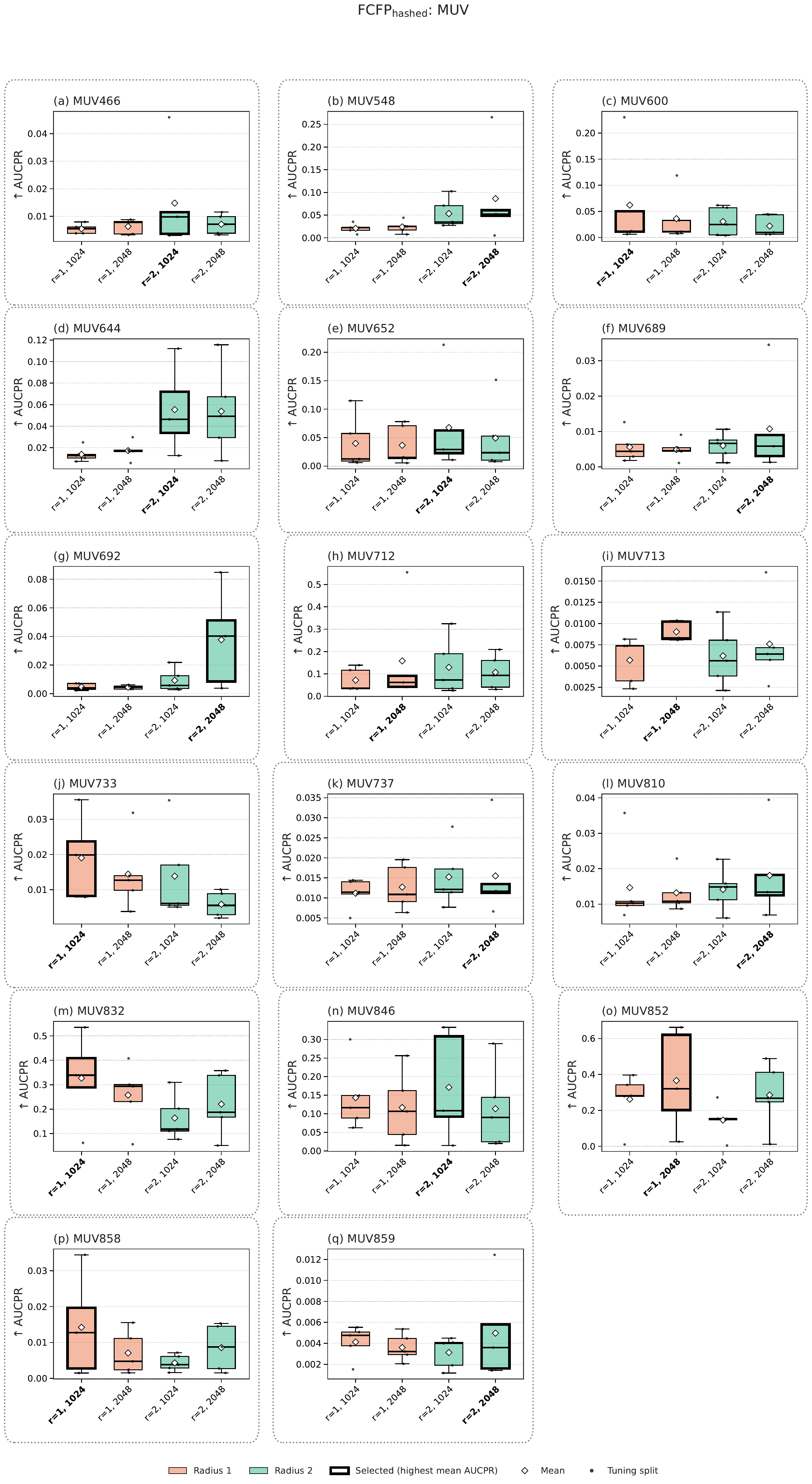}
    \caption{
        FCFP$_{\mathrm{hashed}}$ hyperparameter tuning results on MUV virtual screening tasks.
    }
    \label{fig:tuning_FCFP_hashed_muv}
\end{figure}

\begin{figure}[!htbp]
    \centering
    \includegraphics[width=\textwidth, height=0.7\textheight, keepaspectratio]{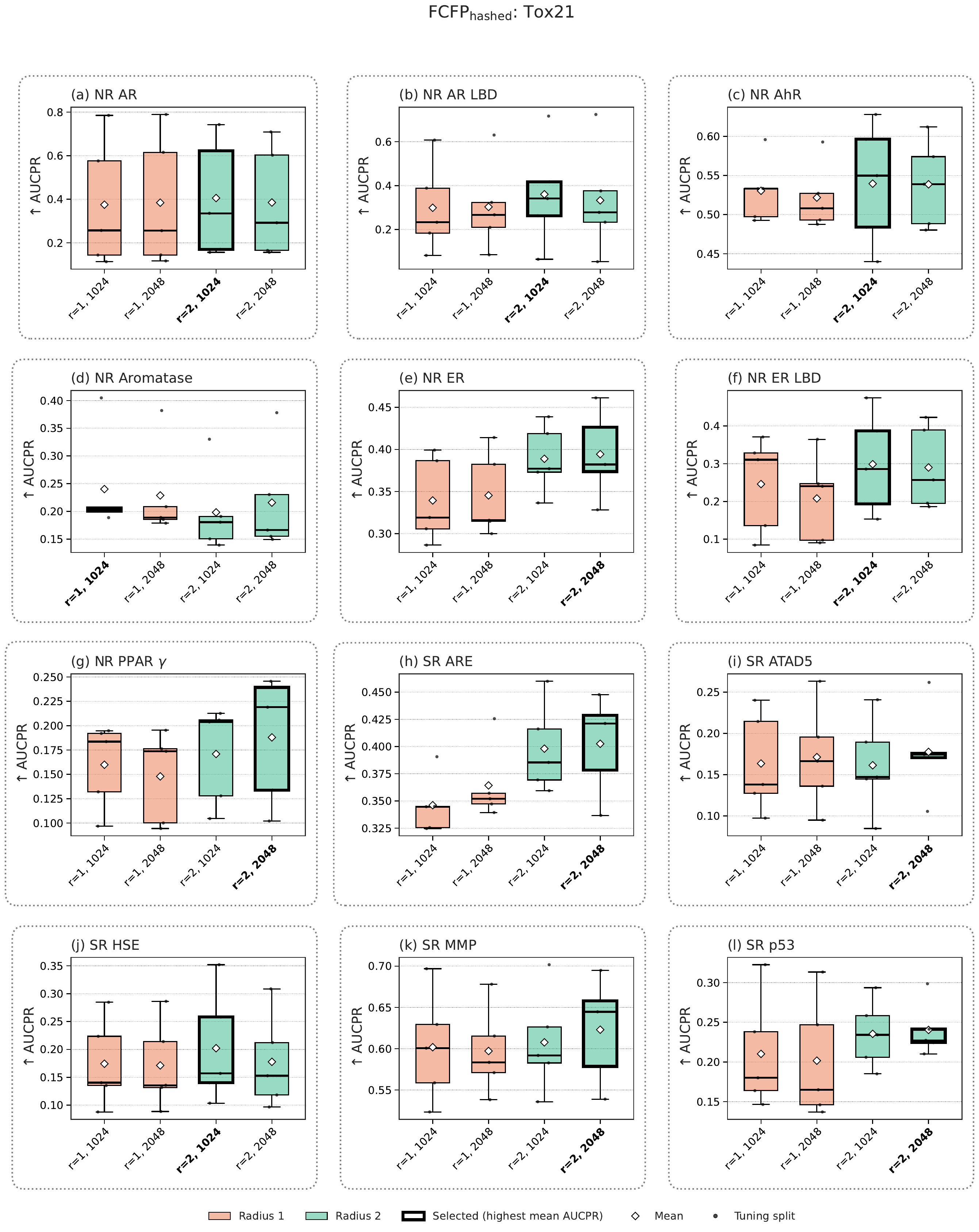}
    \caption{
        FCFP$_{\mathrm{hashed}}$ hyperparameter tuning results on Tox21 tasks.
    }
    \label{fig:tuning_FCFP_hashed_tox21}
\end{figure}

\clearpage
\newpage

\subsubsection{Selected Morgan algorithm configurations}

\begin{table}[!htbp]
\centering
    \caption{
        Morgan algorithm hyperparameters selected on each benchmark dataset for pre-trained GINs and fixed fingerprints,
        chosen by the best mean performance on the five tuning splits (lowest MAE for regression, highest AUCPR for classification). 
        $r$: radius of the substructure environments (for GINs and fingerprints);
        $|\mathcal{V}|$: substructure vocabulary size (for GINs);
        Bits: fingerprint length (for folded topological fingerprints).
        ESOL Drug-Like Only uses the same configurations selected on ESOL.
    }
    \label{tab:tuning_selected_configs}
    \small
    \setlength{\tabcolsep}{4pt}
    \renewcommand{\arraystretch}{0.8}% arXiv: shorter sn-jnl text block
    \begin{tabular}{l rr rr rr rr rr}
    \toprule
    & \multicolumn{2}{c}{PT-GIN} & \multicolumn{2}{c}{Scratch GIN} & \multicolumn{2}{c}{ECFP$_{\mathrm{S\&S}}$} & \multicolumn{2}{c}{ECFP$_{\mathrm{hashed}}$} & \multicolumn{2}{c}{FCFP$_{\mathrm{hashed}}$} \\
    \cmidrule(lr){2-3} \cmidrule(lr){4-5} \cmidrule(lr){6-7} \cmidrule(lr){8-9} \cmidrule(lr){10-11}
    Dataset & $r$ & $|\mathcal{V}|$ & $r$ & $|\mathcal{V}|$ & $r$ & Bits & $r$ & Bits & $r$ & Bits \\
    \midrule
    \multicolumn{11}{l}{\textit{Biogen (MAE)}} \\
    Human PPB & 0 & 2048 & 1 & 256 & 1 & 2048 & 1 & 2048 & 2 & 2048 \\
    Rat PPB & 2 & 16384 & 0 & 64 & 2 & 2048 & 2 & 1024 & 2 & 2048 \\
    Human CLint & 1 & 2048 & 1 & 128 & 2 & 1024 & 1 & 2048 & 2 & 2048 \\
    Rat CLint & 2 & 8192 & 1 & 256 & 2 & 1024 & 1 & 2048 & 2 & 2048 \\
    Efflux & 1 & 16384 & 0 & 16 & 2 & 2048 & 1 & 2048 & 2 & 2048 \\
    Solubility & 2 & 1024 & 0 & 32 & 2 & 2048 & 1 & 2048 & 2 & 2048 \\
    \midrule
    \multicolumn{11}{l}{\textit{ChEMBL Affinity (MAE)}} \\
    DRD2 & 2 & 1024 & -- & -- & 2 & 1024 & 2 & 2048 & 2 & 1024 \\
    Factor Xa & 1 & 16384 & -- & -- & 2 & 2048 & 2 & 2048 & 2 & 1024 \\
    \midrule
    \multicolumn{11}{l}{\textit{MoleculeNet Regression (MAE)}} \\
    ESOL & 0 & 2048 & -- & -- & 2 & 2048 & 2 & 2048 & 2 & 1024 \\
    FreeSolv & 0 & 2048 & -- & -- & 1 & 2048 & 1 & 1024 & 2 & 1024 \\
    Lipophilicity & 0 & 2048 & -- & -- & 2 & 2048 & 1 & 2048 & 1 & 2048 \\
    \midrule
    \multicolumn{11}{l}{\textit{MUV (AUCPR)}} \\
    MUV466 & 1 & 8192 & -- & -- & 2 & 1024 & 2 & 2048 & 2 & 1024 \\
    MUV548 & 2 & 4096 & -- & -- & 2 & 2048 & 2 & 1024 & 2 & 2048 \\
    MUV600 & 2 & 8192 & -- & -- & 2 & 2048 & 1 & 1024 & 1 & 1024 \\
    MUV644 & 2 & 4096 & -- & -- & 2 & 1024 & 1 & 1024 & 2 & 1024 \\
    MUV652 & 2 & 1024 & -- & -- & 2 & 1024 & 2 & 1024 & 2 & 1024 \\
    MUV689 & 1 & 8192 & -- & -- & 1 & 2048 & 1 & 2048 & 2 & 2048 \\
    MUV692 & 1 & 8192 & -- & -- & 2 & 2048 & 2 & 2048 & 2 & 2048 \\
    MUV712 & 1 & 8192 & -- & -- & 2 & 1024 & 2 & 2048 & 1 & 2048 \\
    MUV713 & 2 & 16384 & -- & -- & 1 & 2048 & 2 & 2048 & 1 & 2048 \\
    MUV733 & 2 & 2048 & -- & -- & 2 & 1024 & 2 & 2048 & 1 & 1024 \\
    MUV737 & 1 & 16384 & -- & -- & 2 & 2048 & 2 & 2048 & 2 & 2048 \\
    MUV810 & 0 & 2048 & -- & -- & 1 & 1024 & 2 & 1024 & 2 & 2048 \\
    MUV832 & 1 & 8192 & -- & -- & 2 & 1024 & 1 & 2048 & 1 & 1024 \\
    MUV846 & 1 & 4096 & -- & -- & 1 & 2048 & 1 & 2048 & 2 & 1024 \\
    MUV852 & 1 & 4096 & -- & -- & 1 & 1024 & 1 & 2048 & 1 & 2048 \\
    MUV858 & 2 & 1024 & -- & -- & 1 & 2048 & 2 & 1024 & 1 & 1024 \\
    MUV859 & 0 & 2048 & -- & -- & 2 & 2048 & 1 & 1024 & 2 & 2048 \\
    \midrule
    \multicolumn{11}{l}{\textit{Tox21 (AUCPR)}} \\
    NR AR & 2 & 16384 & -- & -- & 2 & 2048 & 2 & 2048 & 2 & 1024 \\
    NR AR LBD & 2 & 16384 & -- & -- & 2 & 1024 & 2 & 2048 & 2 & 1024 \\
    NR AhR & 1 & 2048 & -- & -- & 2 & 2048 & 1 & 2048 & 2 & 1024 \\
    NR Aromatase & 2 & 1024 & -- & -- & 2 & 2048 & 1 & 2048 & 1 & 1024 \\
    NR ER & 1 & 8192 & -- & -- & 2 & 1024 & 2 & 1024 & 2 & 2048 \\
    NR ER LBD & 1 & 4096 & -- & -- & 2 & 1024 & 1 & 2048 & 2 & 1024 \\
    NR PPAR $\gamma$ & 1 & 4096 & -- & -- & 1 & 2048 & 2 & 1024 & 2 & 2048 \\
    SR ARE & 2 & 1024 & -- & -- & 2 & 2048 & 1 & 2048 & 2 & 2048 \\
    SR ATAD5 & 1 & 8192 & -- & -- & 2 & 2048 & 2 & 2048 & 2 & 2048 \\
    SR HSE & 2 & 1024 & -- & -- & 1 & 2048 & 1 & 2048 & 2 & 1024 \\
    SR MMP & 2 & 2048 & -- & -- & 1 & 2048 & 1 & 2048 & 2 & 2048 \\
    SR p53 & 0 & 2048 & -- & -- & 2 & 1024 & 1 & 1024 & 2 & 2048 \\
    \bottomrule
\end{tabular}
\end{table}
\clearpage